\documentclass{article}

\usepackage[nonatbib,final]{neurips_2022}

\usepackage[utf8]{inputenc} %
\usepackage[T1]{fontenc}    %
\usepackage[pagebackref=true,breaklinks=true,colorlinks,bookmarks=true]{hyperref}
\usepackage{url}            %
\usepackage{booktabs}       %
\usepackage{amsfonts}       %
\usepackage{nicefrac}       %
\usepackage{microtype}      %
\usepackage{xcolor}         %
\usepackage{pifont}         %
\usepackage{makecell}
\usepackage{colortbl}
\usepackage{graphicx}
\usepackage{amsmath,amsfonts,amssymb,amsthm}
\usepackage{color}
\usepackage{caption}
\usepackage{subcaption}
\usepackage{xspace}
\usepackage{array}
\usepackage{cite}
\usepackage{tikzducks}
\usepackage{wrapfig}
\usepackage[titletoc,title]{appendix}
\usepackage{lscape}
\usepackage{enumitem}
\usepackage{multirow}

\usepackage[scaled=0.85]{beramono}

\title{MonoSDF: Exploring Monocular Geometric Cues \\for Neural Implicit Surface Reconstruction}

\author{
Zehao Yu$^{1}$\quad Songyou Peng$^{2,3}$\quad Michael Niemeyer$^{1,3}$\quad Torsten Sattler$^{4}$\quad Andreas Geiger$^{1,3}$\vspace{8pt}\\
$^{1}$University of T\"ubingen\qquad $^{2}$ETH Zurich\qquad $^{3}$MPI for Intelligent Systems, T\"ubingen
\\
$^{4}$Czech Technical University in Prague\vspace{8pt}\\
\url{https://niujinshuchong.github.io/monosdf}
}

\providecommand{\impath}[1]{}
\providecommand{\impatha}[1]{}
\providecommand{\impathb}[1]{}
\providecommand{\impathc}[1]{}
\providecommand{\impathd}[1]{}
\providecommand{\impathe}[1]{}

\newcommand{\myheight}{3.1cm}
\newcommand{\teaser}{
\begin{figure*}[t]
    \centering
    \setlength{\tabcolsep}{0.1em}
    \renewcommand{\arraystretch}{0.8}
    \hfill{}\hspace*{-0.5em}
    \footnotesize
    \begin{tabular}{cccc}
            \rot{\qquad VolSDF~\cite{Yariv2021NEURIPS}}&
            \includegraphics[height=\myheight]{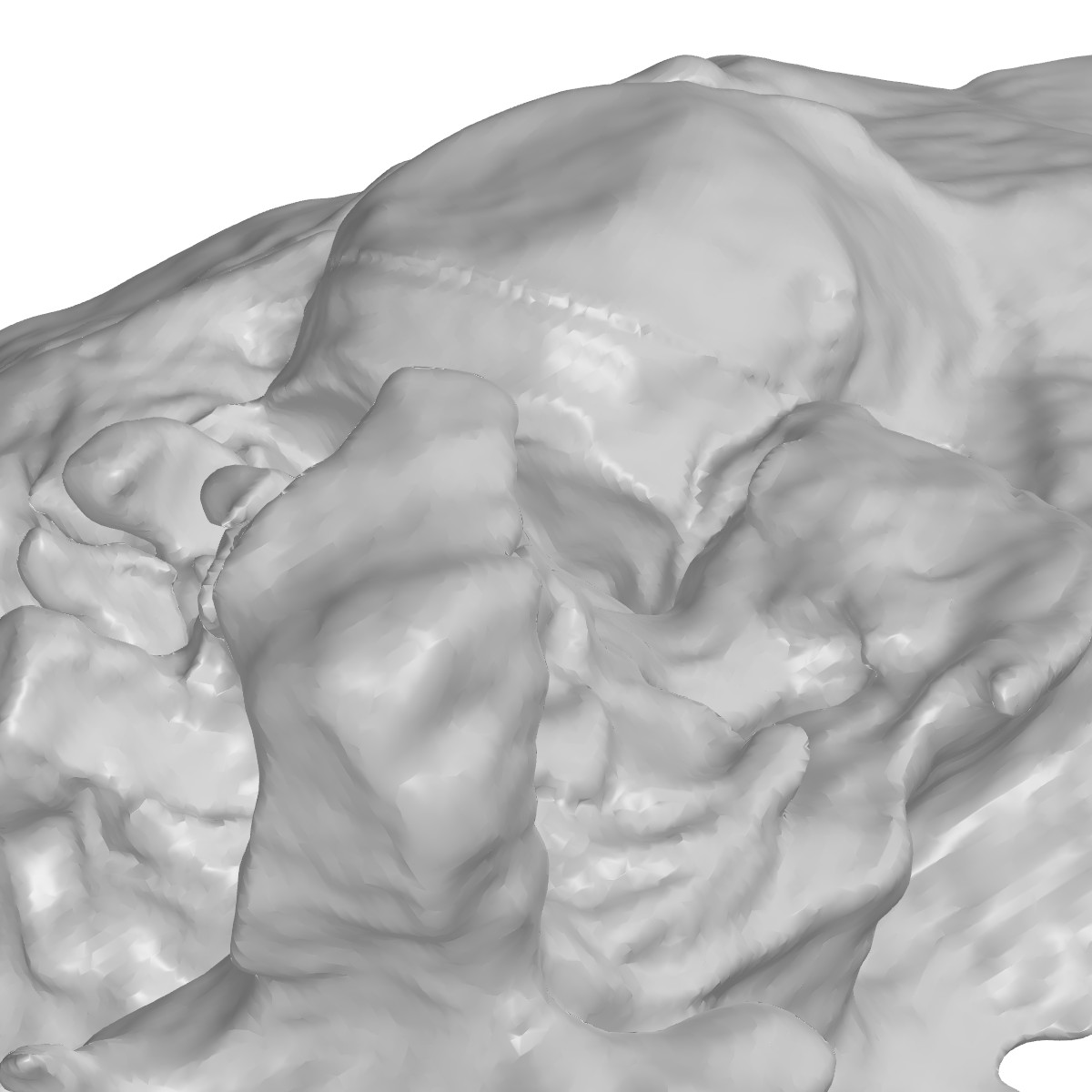}&
            \includegraphics[height=\myheight]{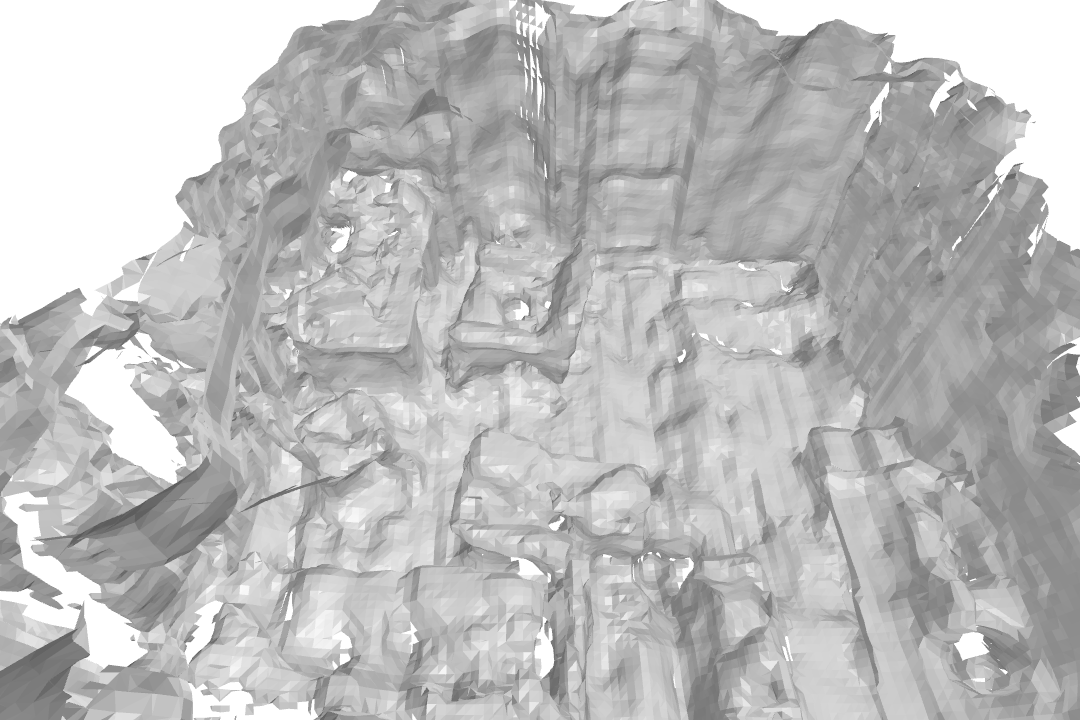}&
            \includegraphics[height=\myheight]{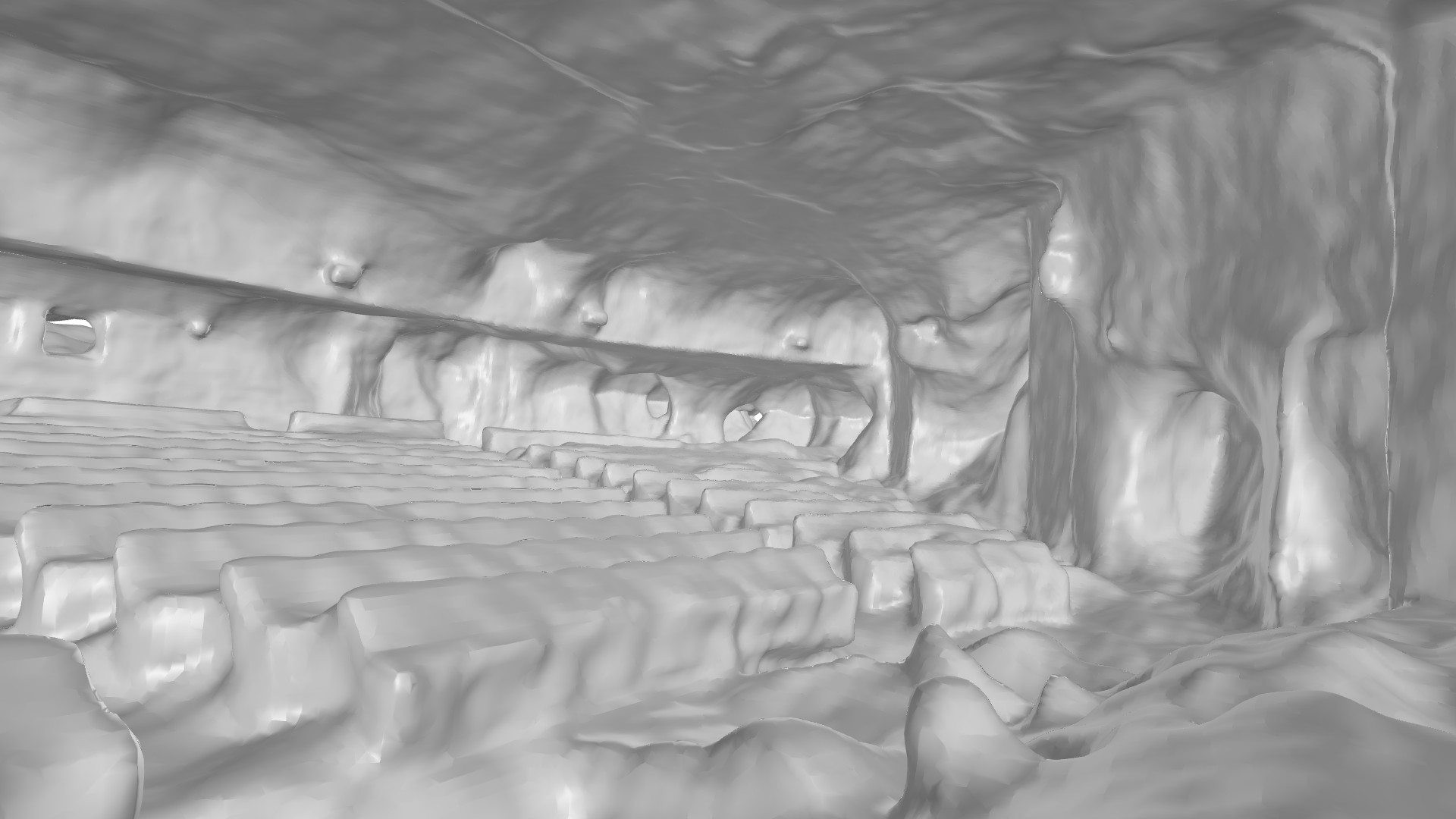}
            \\
            \rot{\;+ Monocular Cues}&
            \includegraphics[height=\myheight]{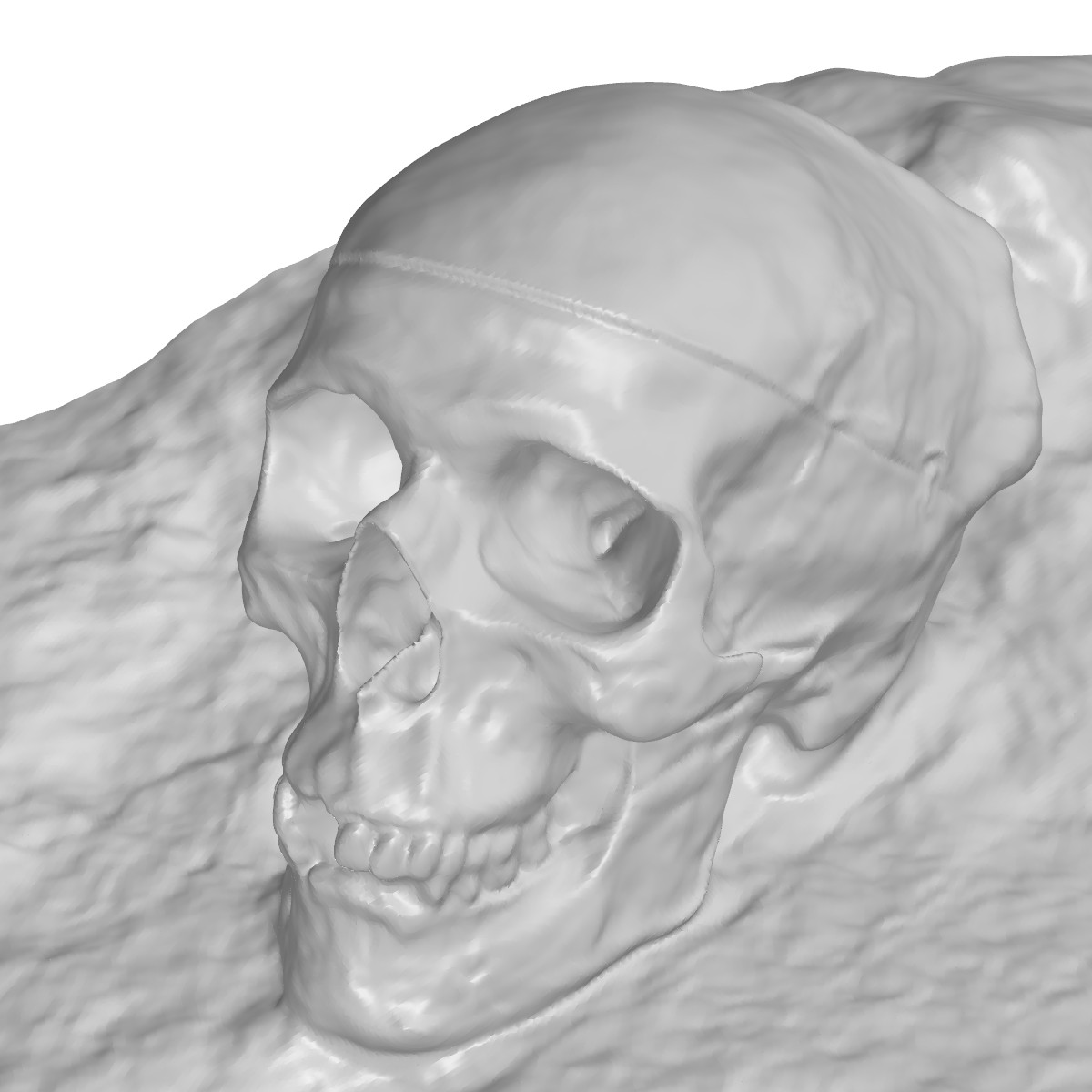}&
            \includegraphics[height=\myheight]{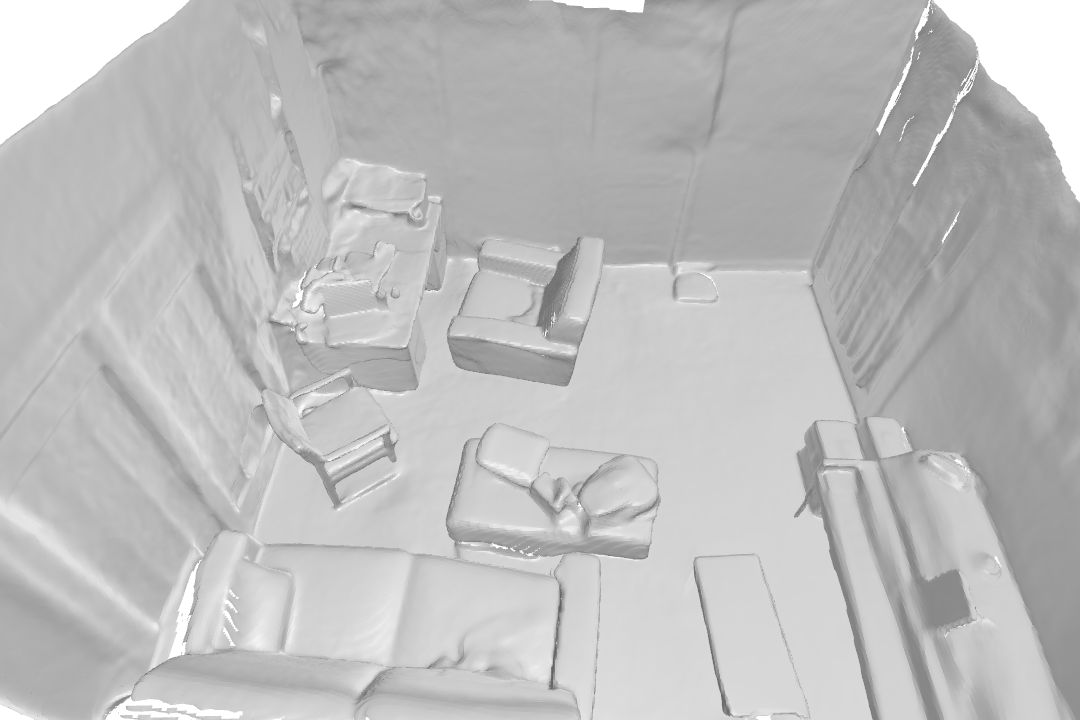}&
            \includegraphics[height=\myheight]{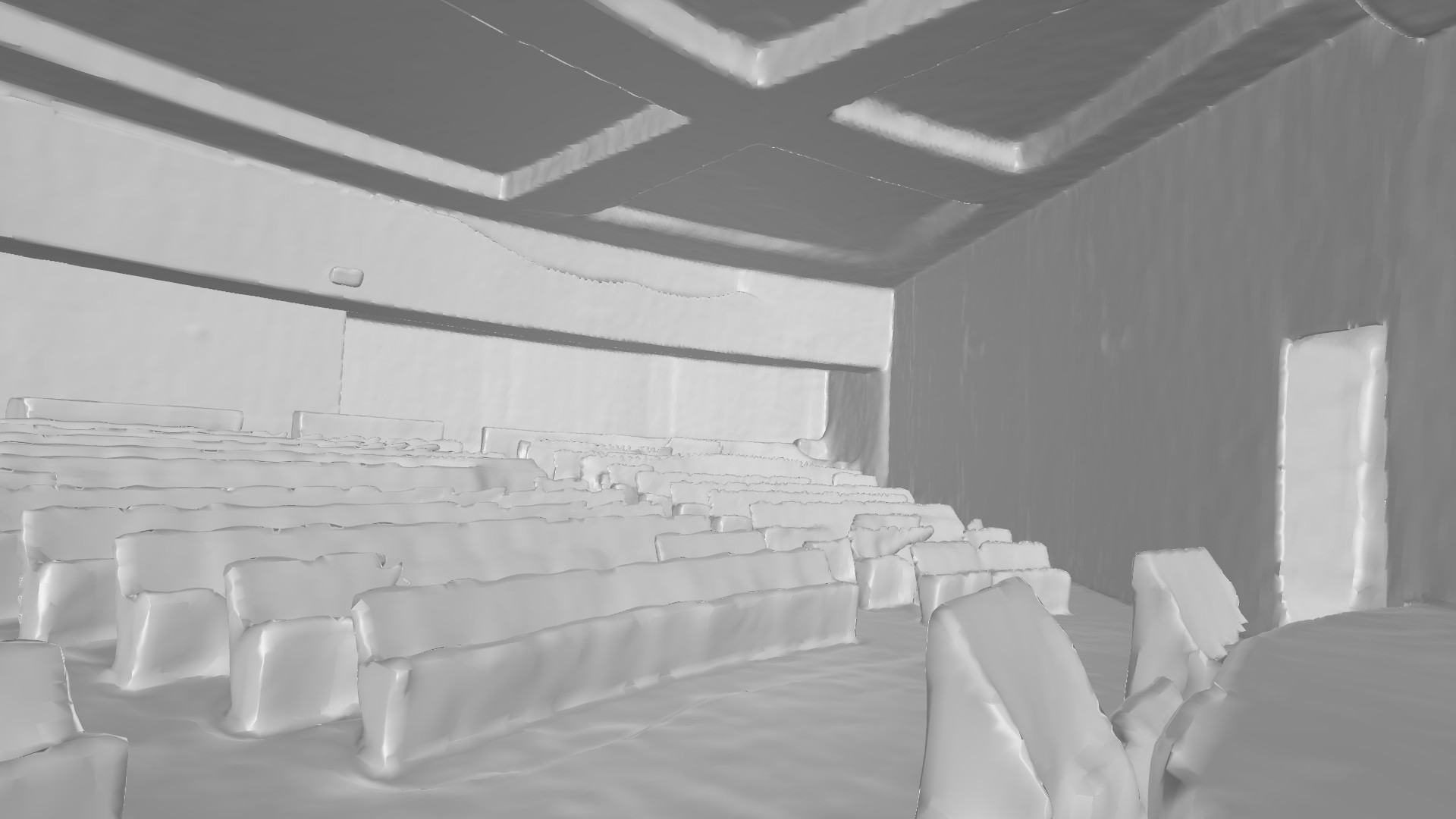}\\
            {}&DTU (3 views) & ScanNet (464 views) & Tanks \& Temples (298 views)
    \end{tabular}
  \caption{\textbf{MonoSDF.} 
  Top: State-of-the-art neural implicit surface reconstruction methods fail in the presence of limited input views or when applied to complex multi-object scenes. Bottom: We demonstrate
  that incorporating geometric cues from general-purpose monocular predictors enables scaling to larger scenes while yielding more accurate reconstructions and speeding up optimization.
  An image resolution of $384\times384$ pixels was %
  used for all results shown above.
}
  \label{fig:teaser}
\end{figure*}
}

\newcommand{\method}{
\begin{figure*}[t]
  \includegraphics[width=\textwidth]{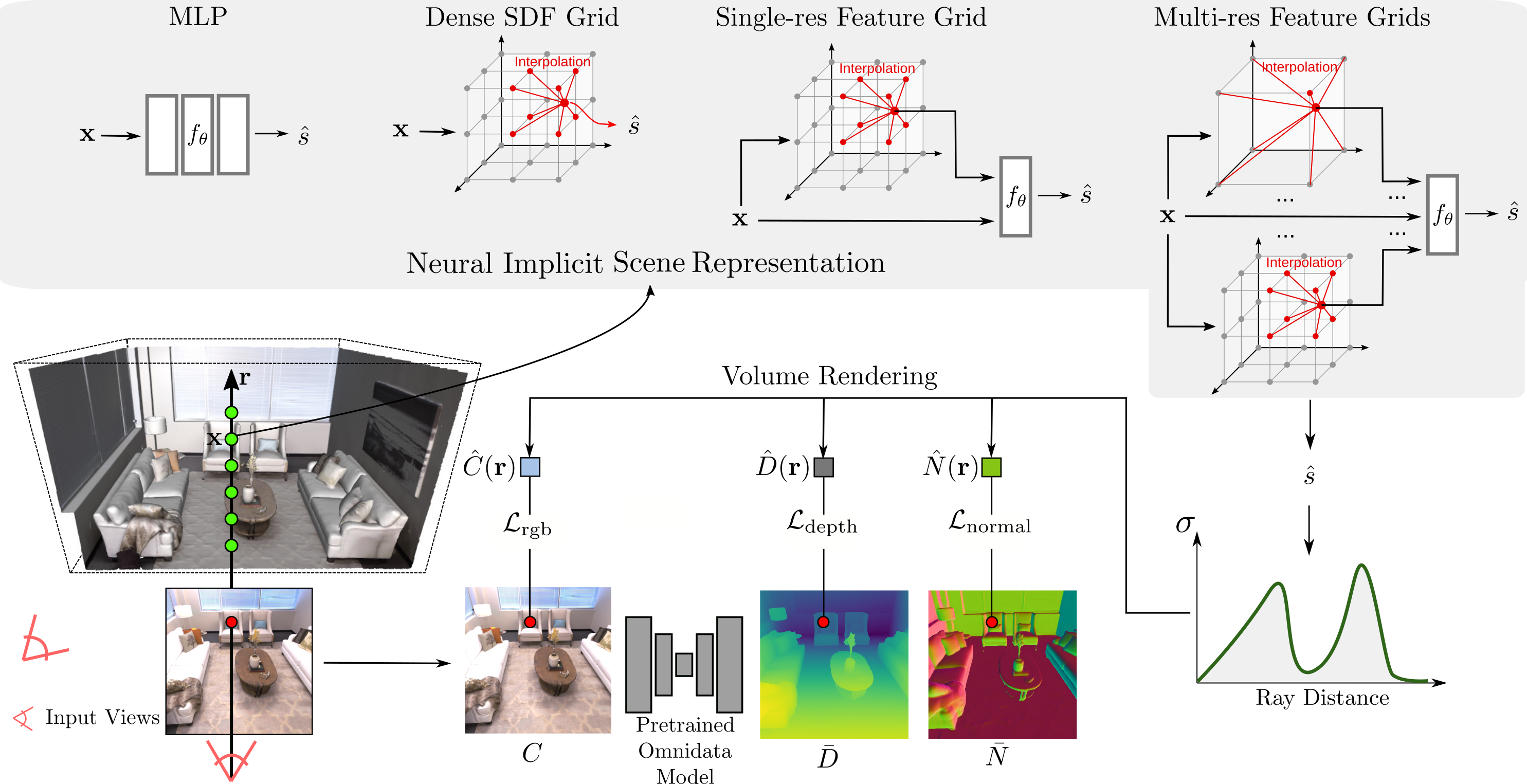}
  \caption{\textbf{Overview.} In this work we use monocular geometric cues predicted by a general-purpose pretrained network to guide the optimization of neural implicit surface models. %
  More specifically, for a batch of rays, we volume render predicted RGB colors, depth, and normals, and optimize \wrt the input RGB images and monocular geometric cues. Further, we investigate different design choices for neural implicit architectures and provide an in-depth analysis. For clarity, we only show the SDF and not the color prediction branch above. %
  }
  \label{fig:overview}
\end{figure*}
}

\newcommand{\rwidth}{0.242\textwidth}

\newcommand{\fivewidth}{0.196\textwidth}
\newcommand{\figureablationgrid}{
\begin{figure*}[t]
        \centering
        \setlength{\tabcolsep}{0.1em}
        \renewcommand{\arraystretch}{0.5}
        \hfill{}\hspace*{-0.5em}
        \begin{tabular}{ccccc}
            \includegraphics[width=\fivewidth]{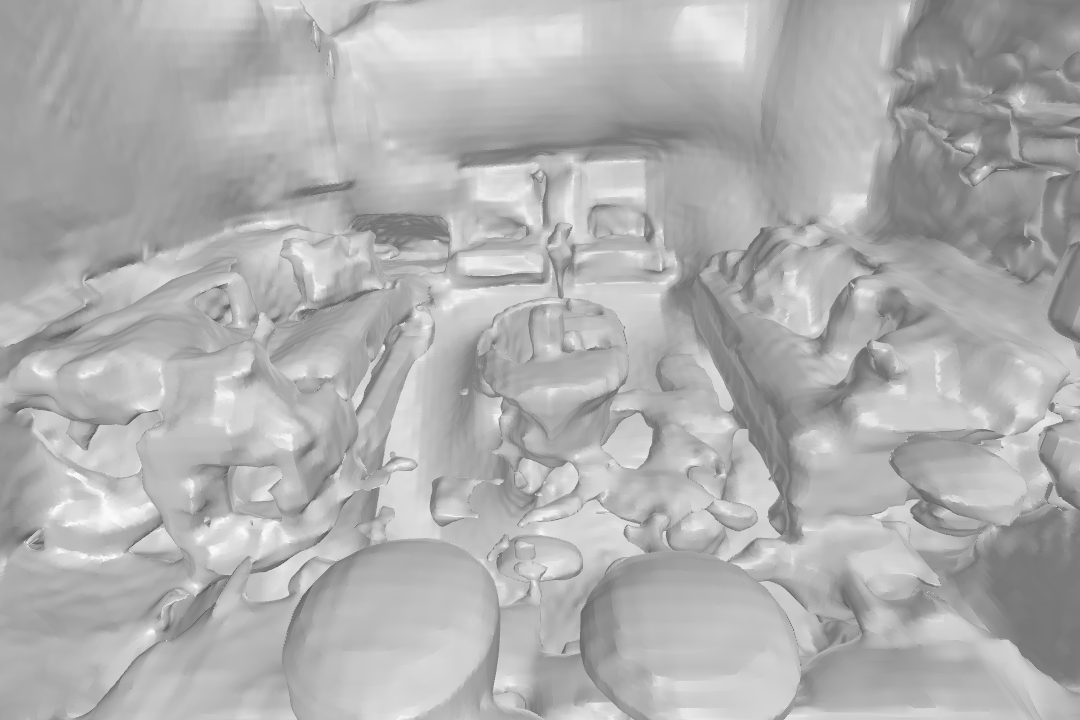}&
            \includegraphics[width=\fivewidth]{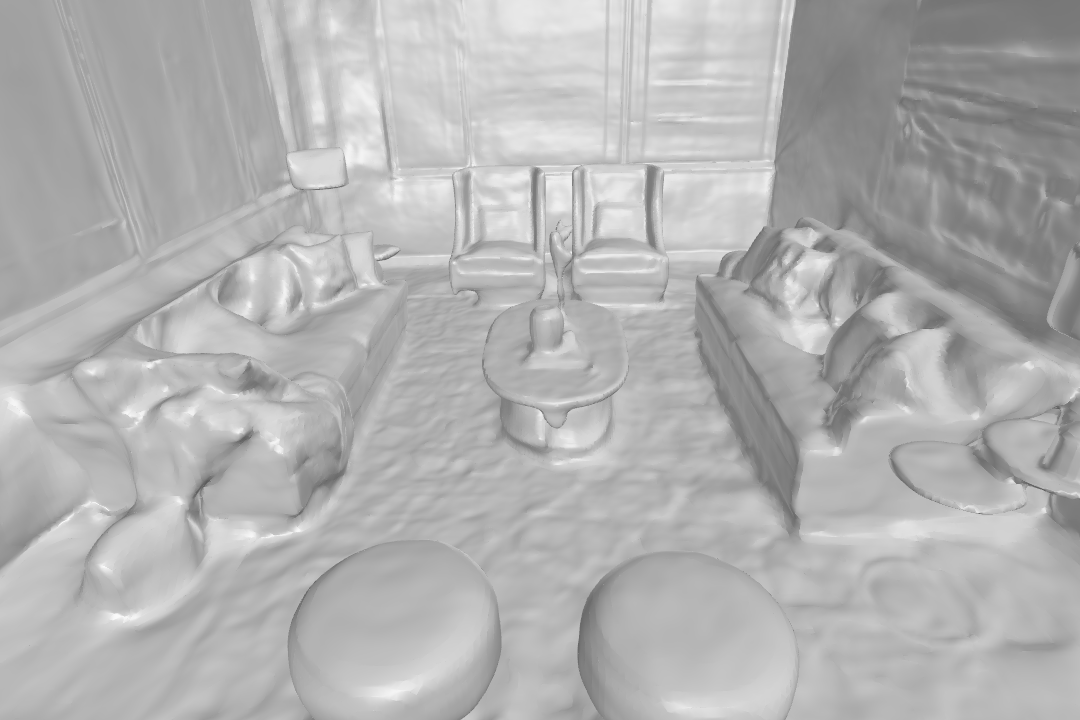}&
            \includegraphics[width=\fivewidth]{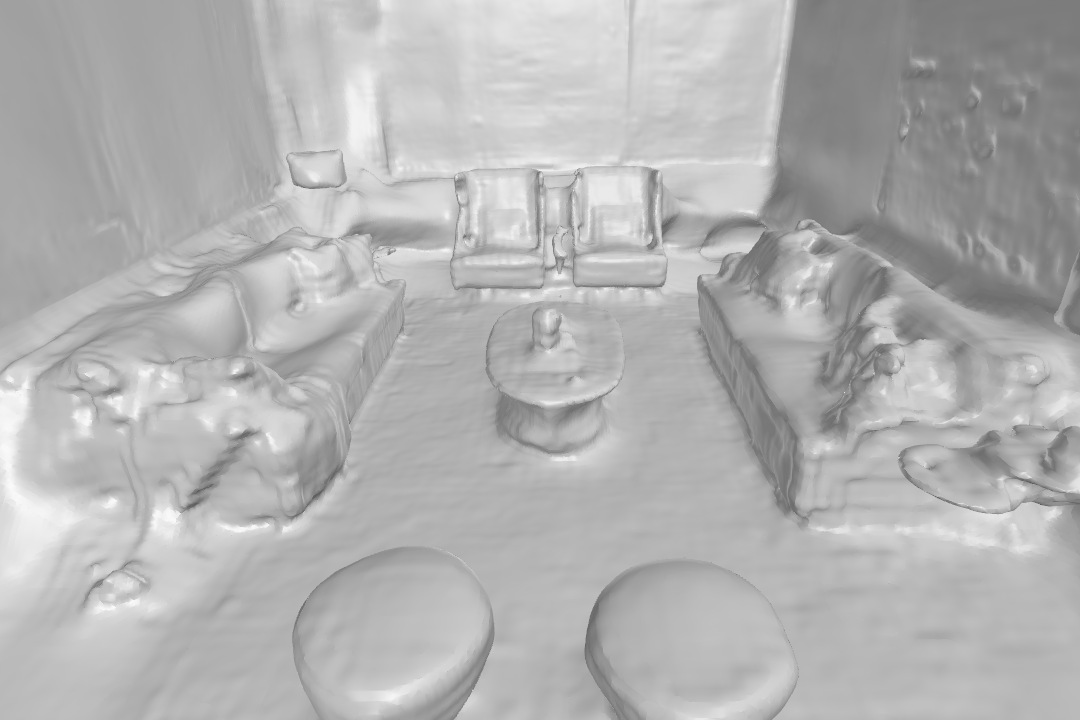}&
            \includegraphics[width=\fivewidth]{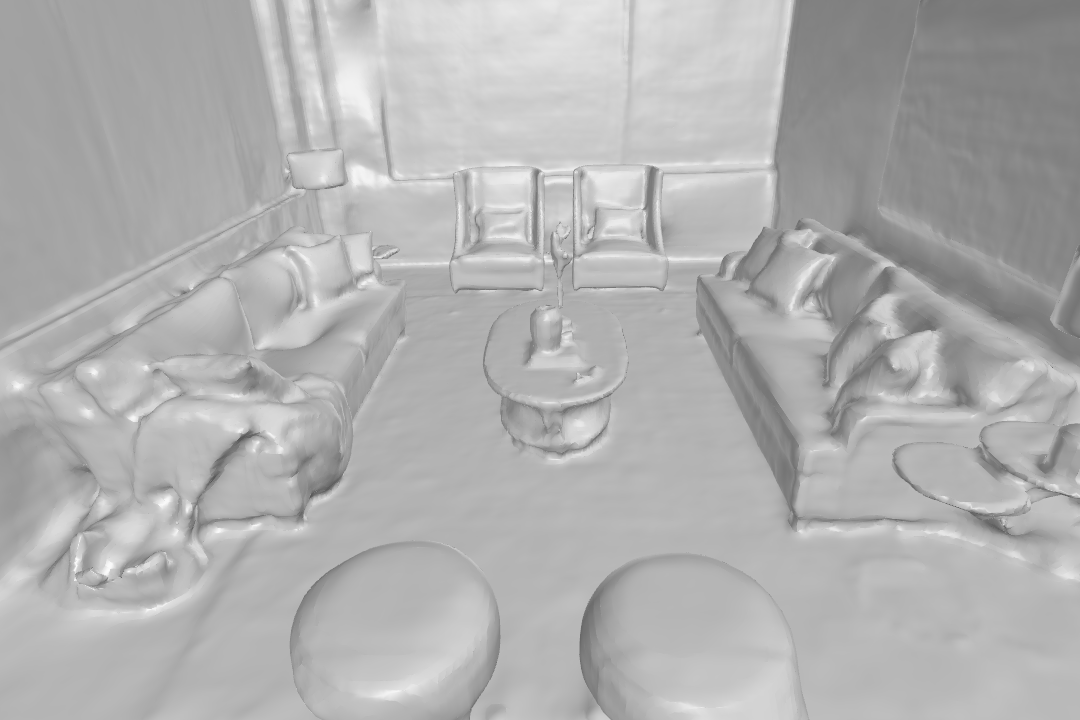}&
            \includegraphics[width=\fivewidth]{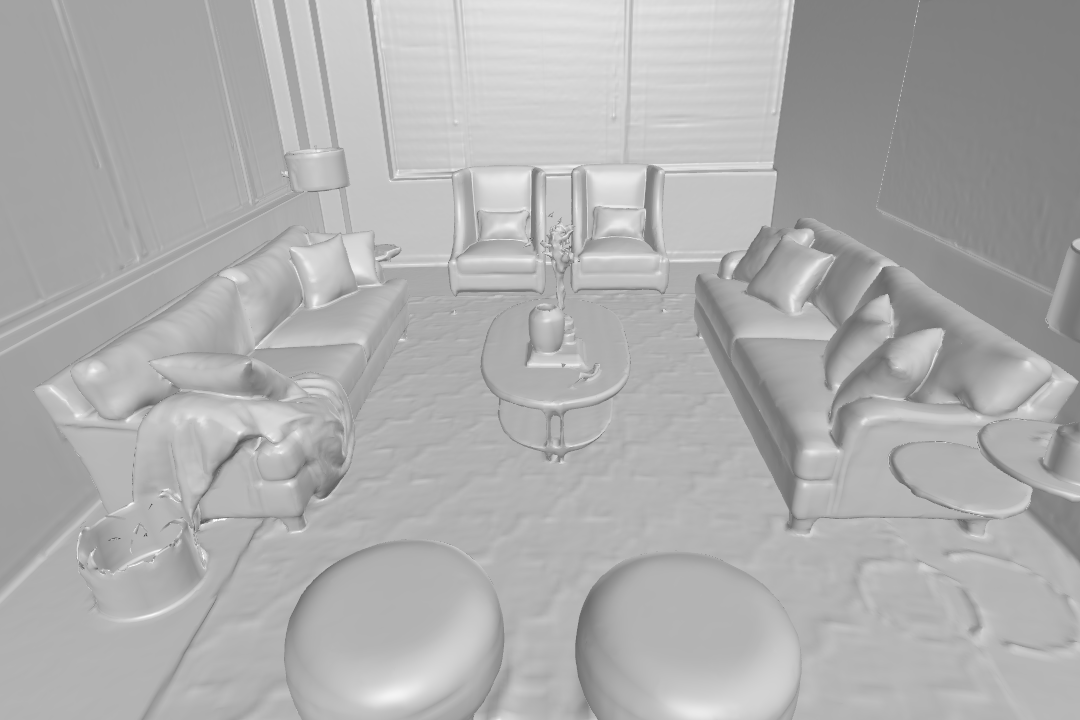}\\
              \small Dense SDF Grid&
              \small MLP  &
              \small Single-Res.&
              \small Multi-Res. &
              \small Ground Truth\\
              &
              &
              \small Fea. Grid&
              \small Fea. Grids&
              \\
        \end{tabular}
        \caption{
        \textbf{Architectural Ablation Study.}
        Comparing different design choices for neural implicit surface representations, %
        we observe that a dense SDF grid leads to noisy reconstructions due to a missing smoothness bias. 
        The MLP and the Single-Res. Fea. Grid improve %
        results, but geometry tends to be overly smooth with missing details. 
        The best results are obtained using Multi-Res. Fea. Grids. 
        }
        \label{fig:ablation_architecture}
\end{figure*}
}

\newcommand{\figureablationcues}{
\begin{figure*}[t]
        \centering
        \setlength{\tabcolsep}{0.1em}
        \renewcommand{\arraystretch}{0.7}
        \hfill{}\hspace*{-0.5em}
        \begin{tabular}{ccccc}
            \includegraphics[width=\fivewidth]{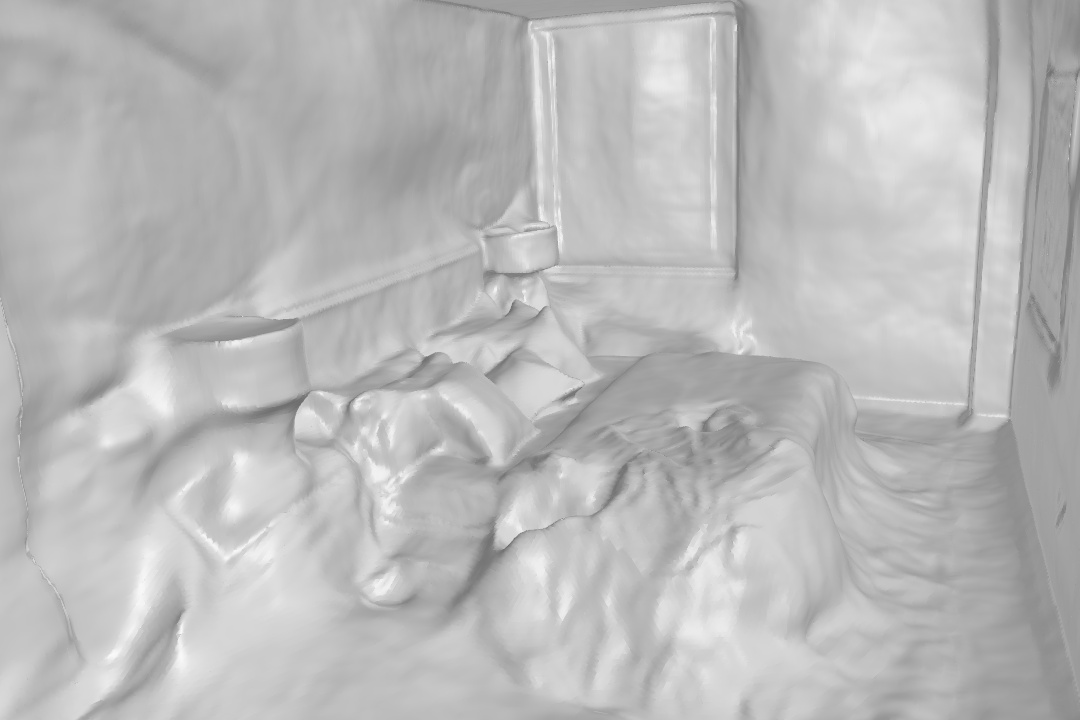}&
            \includegraphics[width=\fivewidth]{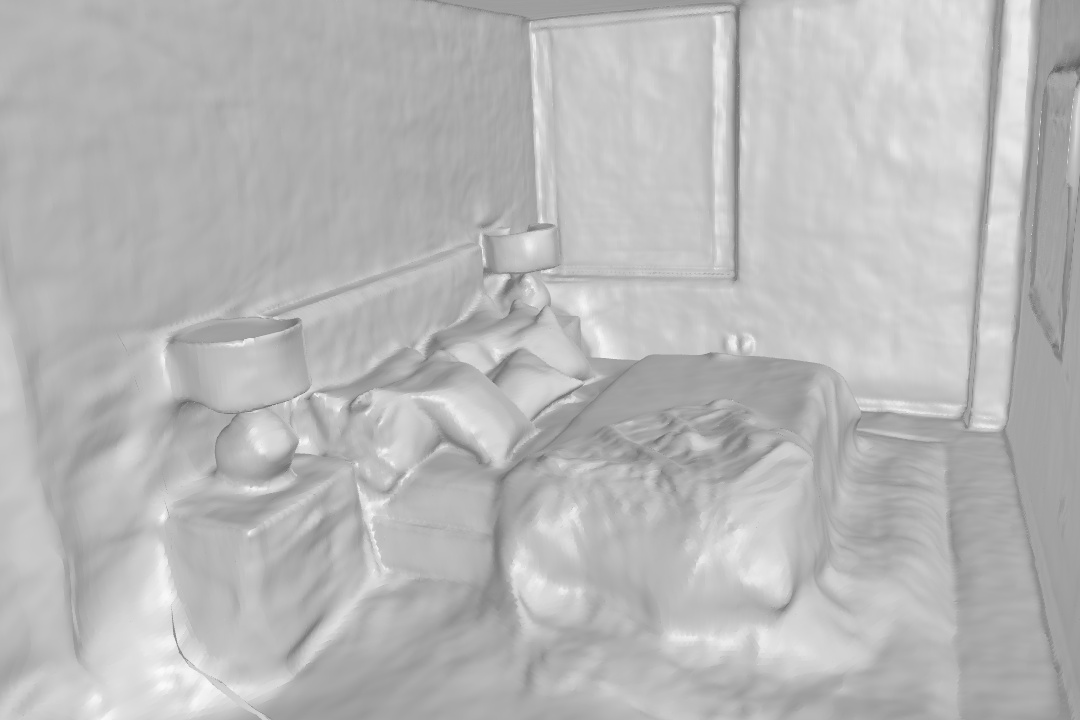}&
            \includegraphics[width=\fivewidth]{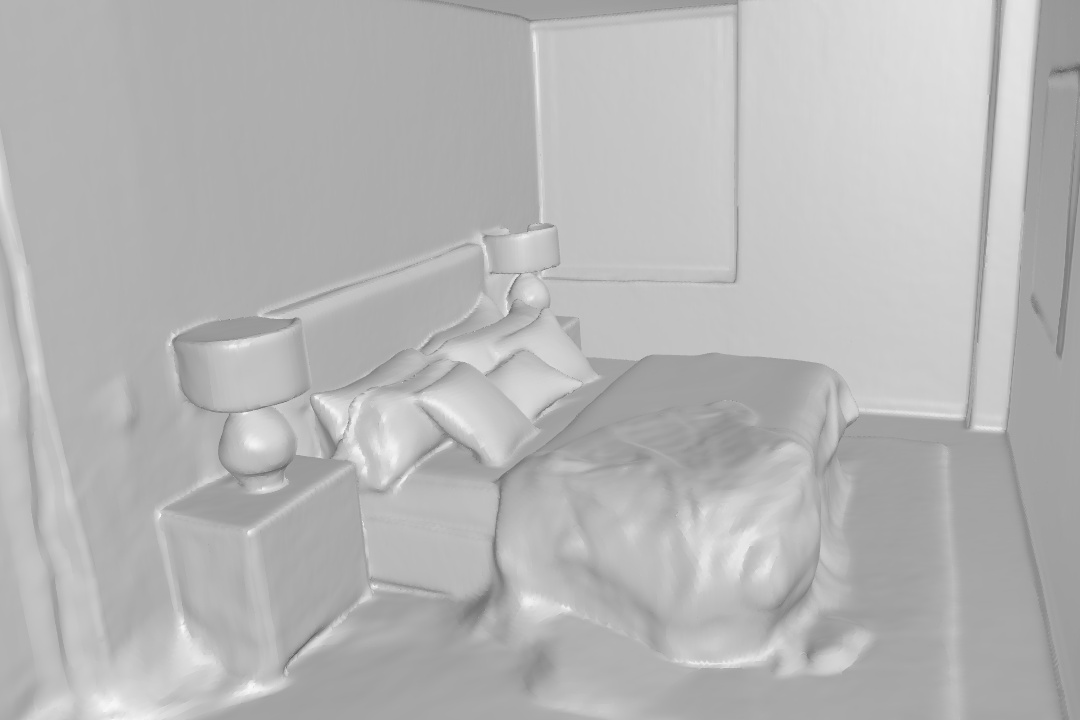}&
            \includegraphics[width=\fivewidth]{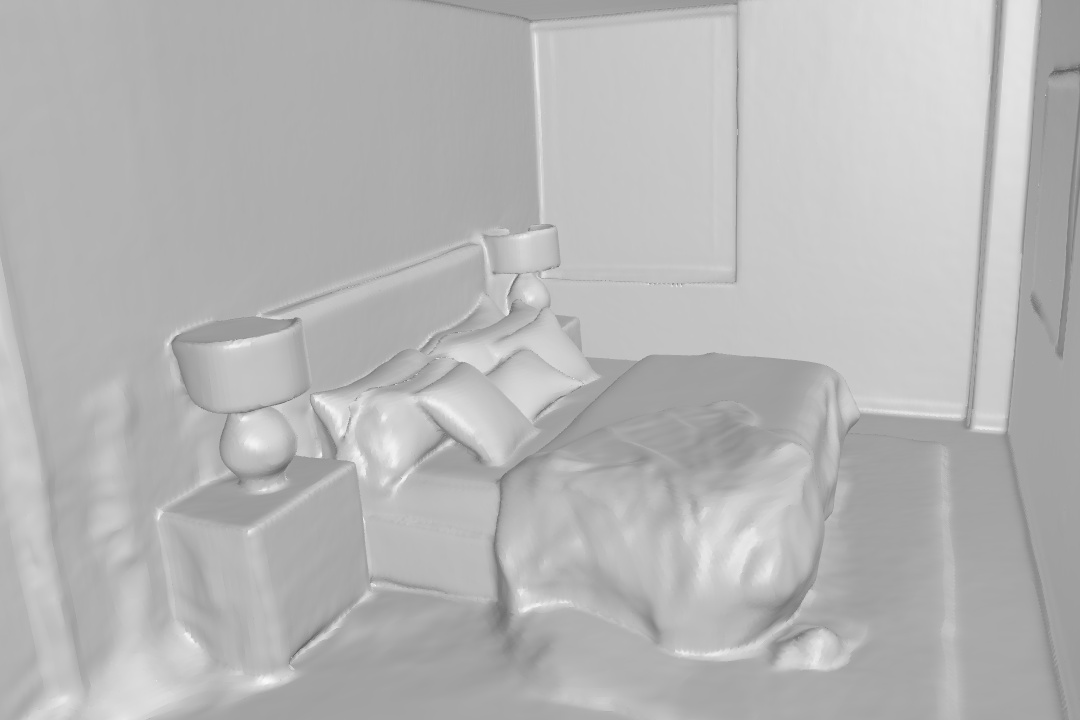}&
            \includegraphics[width=\fivewidth]{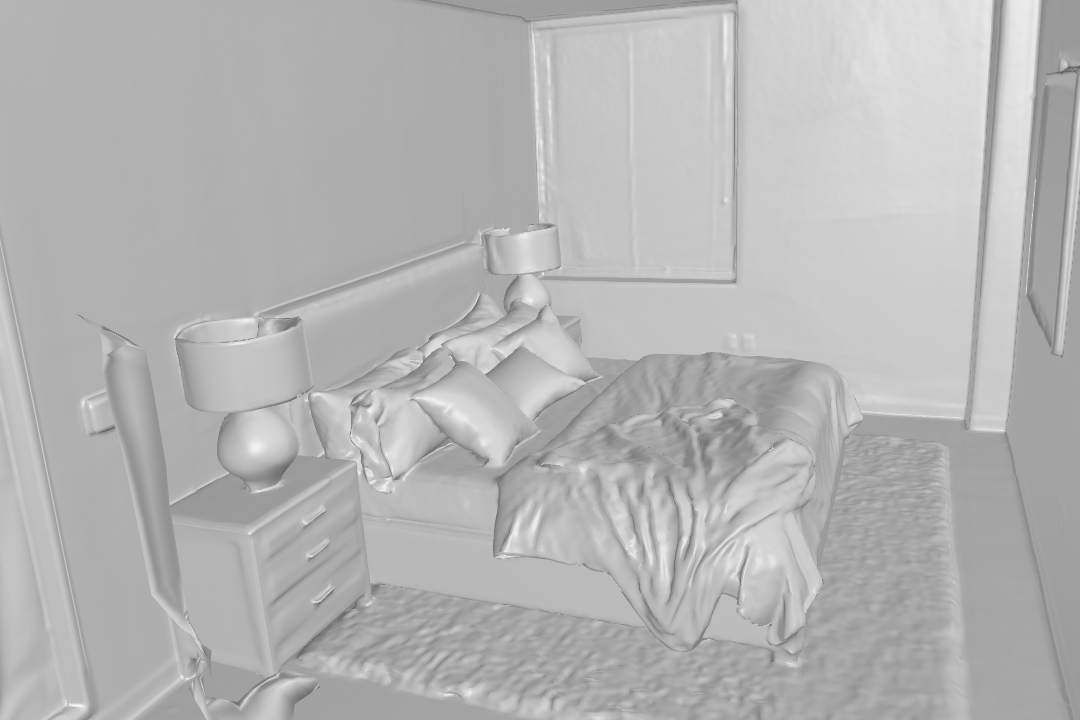}
            \\
             \small No Cue &
             \small + Depth  &
             \small + Normal&
             \small + Both &
             \small Ground Truth\\
        \end{tabular}
        \caption{
        \textbf{Ablation of Monocular Geometric Cues.}
        Monocular geometric cues significantly improve reconstruction quality for both architectures (we show our MLP variant). %
        With %
        monocular depth cues, the recovered geometry contains more details and a better overall structure. %
        With %
        normal cues, missing details are added and the results become smoother. Using both cues leads to the best performance.
        }
        \label{fig:ablation_cues}
\end{figure*}
}

\newcommand{\replicatable}{
\begin{wraptable}[7]{r}{0.44\linewidth}
    \centering
    \vspace{-1em}
    \resizebox{\linewidth}{!}{
    
    \setlength{\tabcolsep}{1pt} %
    \footnotesize
    \begin{tabular}{lccc}
    \toprule
    {} &Normal C.$\uparrow$& Chamfer-$L_1$~$\downarrow$ & F-score~$\uparrow$\\
    \midrule%
    MLP~\cite{Yariv2021NEURIPS}      & 86.48 & 6.75 & 66.88\\
    \hline
    Dense SDF Grid                   & 57.30 & 26.68& 15.50\\
    Single-res. Fea. Grid             & 86.41 & 6.28 & 64.22\\
    Multi-res. Fea. Grids             & \textbf{87.95} & \textbf{5.03} & \textbf{78.38}\\ 
    \bottomrule
    \end{tabular}
    }
    \caption{\small\textbf{Architectural Ablation on Replica.}
    }
    \label{tab:replica_architecture}
\end{wraptable}
}

\newcommand{\replicaablation}{
\begin{table}[!t]
	\centering
	\vspace{-2em}
	\setlength{\tabcolsep}{2pt}
    \begin{tabular}{cc}
    	\resizebox{0.58\textwidth}{!}{
    	    \begin{tabular}{clccc}
                \toprule
                \multicolumn{2}{c}{} &Normal C.$\uparrow$& Chamfer-$L_1$~$\downarrow$ & F-score~$\uparrow$\\
                \hline
                \multirow{4}{*}{\textbf{MLP}} & No Cues     & 86.48& 6.75& 66.88\\
                {}                            & Only Depth  & 90.56& 4.26& 76.42\\
                {}                            & Only Normal & 91.35& 3.19& 85.84\\
                {}                            & Both Cues   & \textbf{92.11}& \textbf{2.94}& \textbf{86.18}\\
                \hline
                {}                            & No Cues     & 87.95& 5.03&78.38\\
                \textbf{Multi-Res.}            & Only Depth  & 90.87& 3.75&80.32\\
                \textbf{Grids}                & Only Normal & 89.90& 3.61&81.28\\
                {}                            & Both Cues   & \textbf{90.93}& \textbf{3.23} &\textbf{85.91}\\
                \bottomrule
            \end{tabular}
    	}
    	&
    	\begin{minipage}{0.42\textwidth}
    	    \vspace{1em}
    		\includegraphics[width=0.9\textwidth]{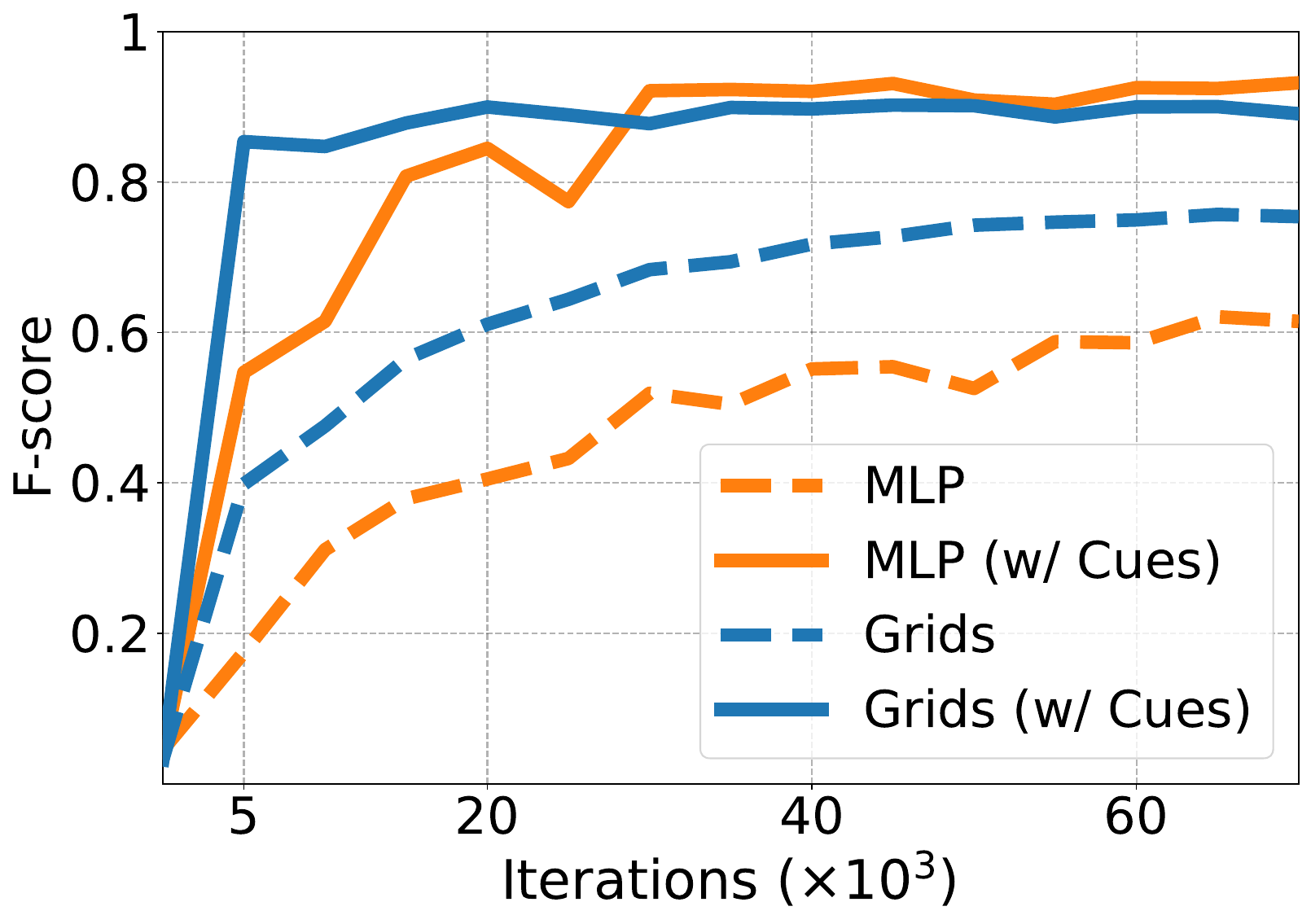}
    	\end{minipage}\\
    (a) \small Different Cues & \small (b) Optimization Time	\\
    \end{tabular}
    \caption{%
    \textbf{Ablation of Monocular Geometric Cues.}
    a.) We report reconstruction results on Replica for MLP and Multi-Res.\ Grids with and without the monocular geometric cues. We observe that monocular cues improve reconstruction quality for both architectures, and using both cues in combination leads to the best performance.
    b.) The optimization speed becomes significantly faster when incorporating monocular cues. Comparing the two architectures, we observe that the grid approach yields faster convergences while the MLP with both cues leads to the best results.
    }
    \vspace{-0.3cm}
    \label{tab:ablation_cues}
\end{table}
}

\newcommand{\dtufused}{
\begin{wraptable}[13]{r}{0.3086\linewidth}
    \centering
    \vspace{-1em}
    \setlength{\tabcolsep}{2pt}
    \resizebox{\linewidth}{!}{

    \begin{tabular}{lc}
            \toprule
             & Chamfer-$L_1$ $\downarrow$\\
            \midrule
            TSDF-Fusion~\cite{Curless1996SIGGRAPH} & 4.80\\  COLMAP~\cite{Schoenberger2016ECCV} & 2.56\\
            RealityCapture & 2.84\\
            Grids & 6.47\\
            Grids w/ cues& 3.68 \\
            MLP~\cite{Yariv2021NEURIPS} & 4.21\\
            MLP w/ cues & \textbf{1.86}\\
            \bottomrule
    \end{tabular}
    }
    \caption{\small \textbf{Reconstruction on DTU (3 Views).}
    We report the average over the test split from~\cite{Yariv2021NEURIPS} (see supplementary for per-object results).
    }
    \vspace{-1.0cm}
    \label{tab:dtu_fused}
\end{wraptable}
}

\newcommand{\fivew}{0.195\textwidth}
\newcommand{\scannetfused}{
\begin{table}[!tb]
	\centering
	\setlength{\tabcolsep}{2pt}
    \begin{tabular}{c}
    	\begin{minipage}{1.0\textwidth}
        	\centering
            \setlength{\tabcolsep}{0.1em}
            \renewcommand{\arraystretch}{0.5}
            \hfill{}\hspace*{-0.5em}
            \footnotesize
    	    \begin{tabular}{ccccc}
    	    \small COLMAP \cite{Schoenberger2016ECCV}&
             \small VolSDF \cite{Mildenhall2020ECCV} &
             \small Manhattan-SDF \cite{guo2022manhattan}&
             \small  \textbf{Ours} (MLP) & Ground Truth\\
            \includegraphics[width=\fivew]{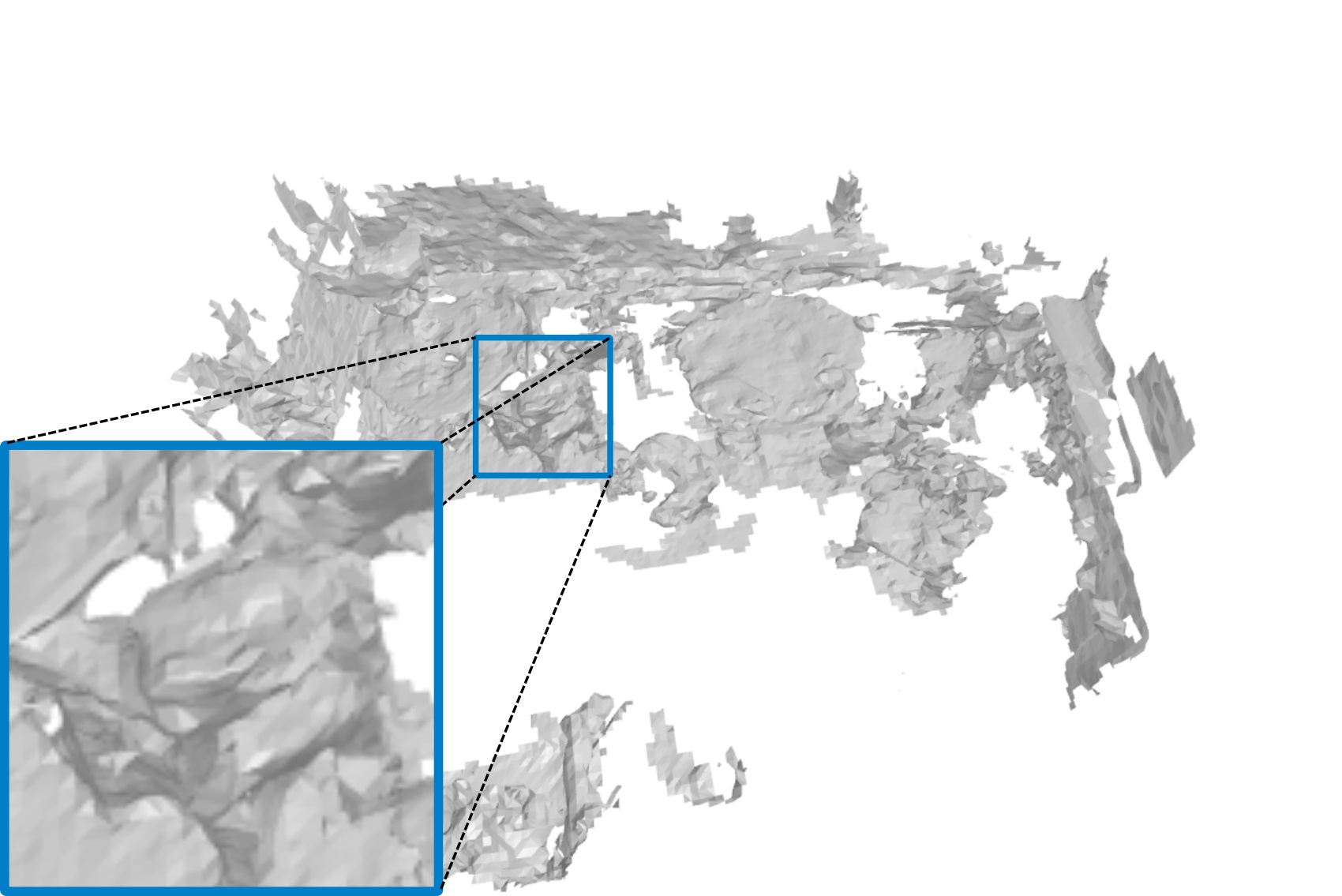}&
            \includegraphics[width=\fivew]{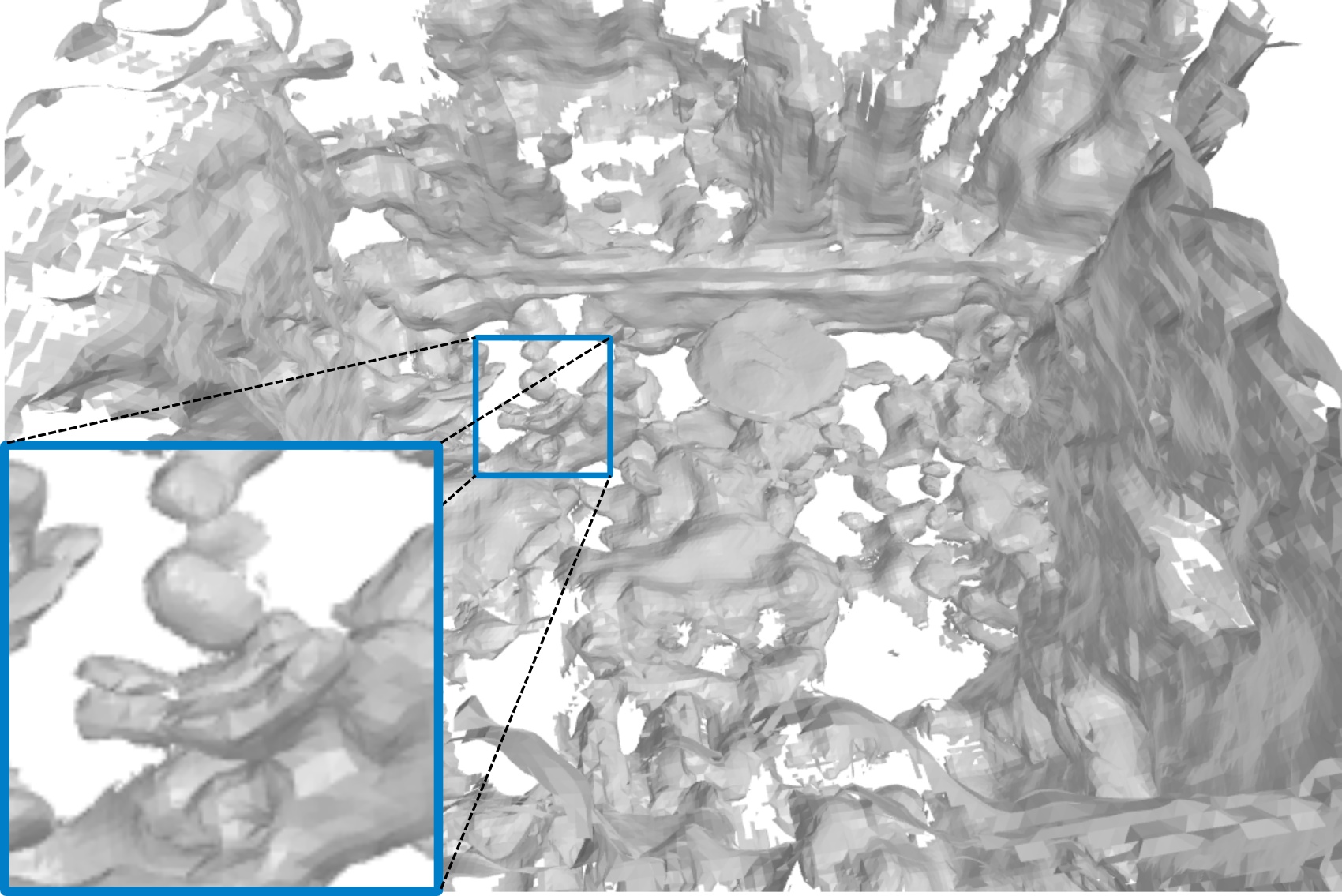}&
            \includegraphics[width=\fivew]{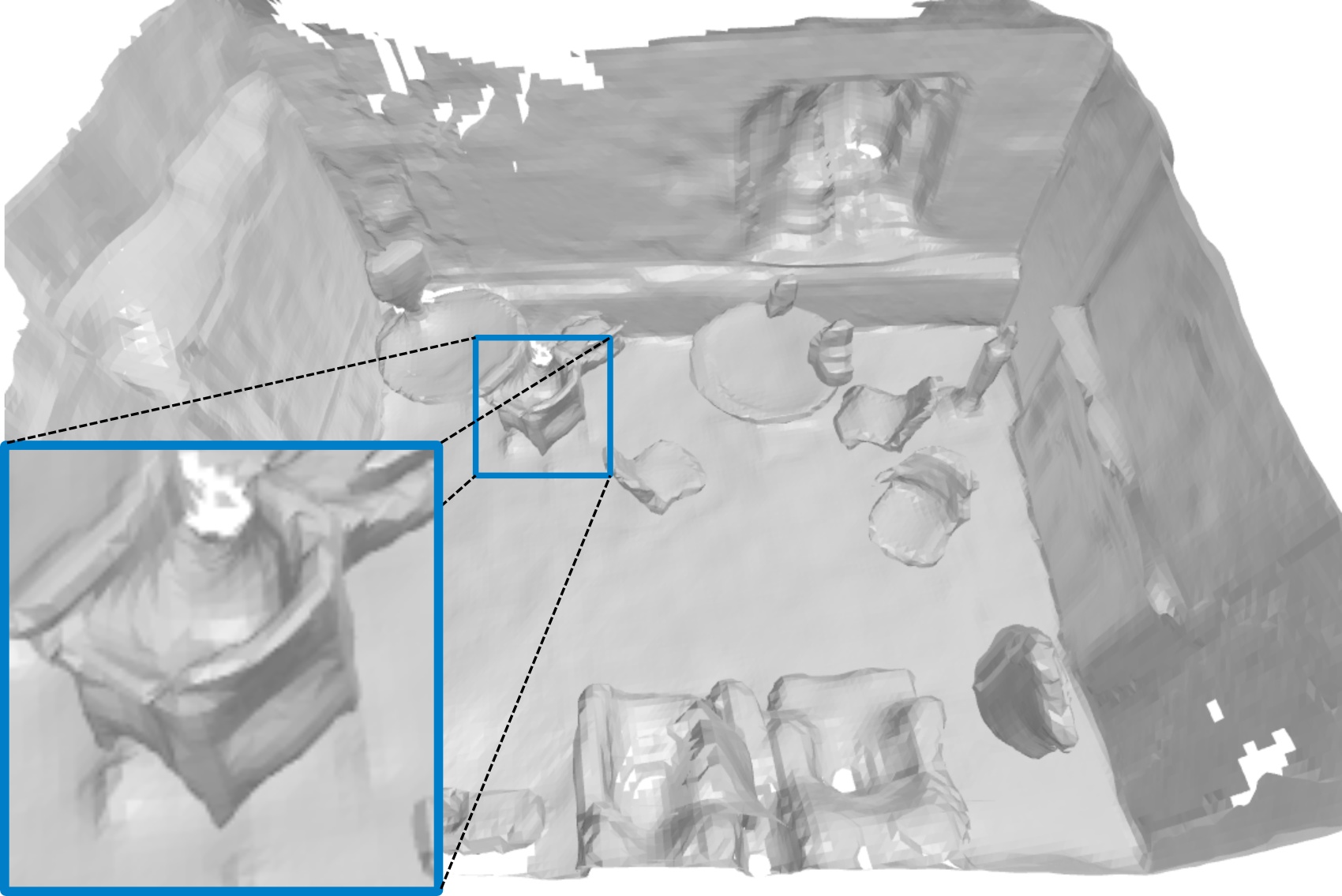}&
            \includegraphics[width=\fivew]{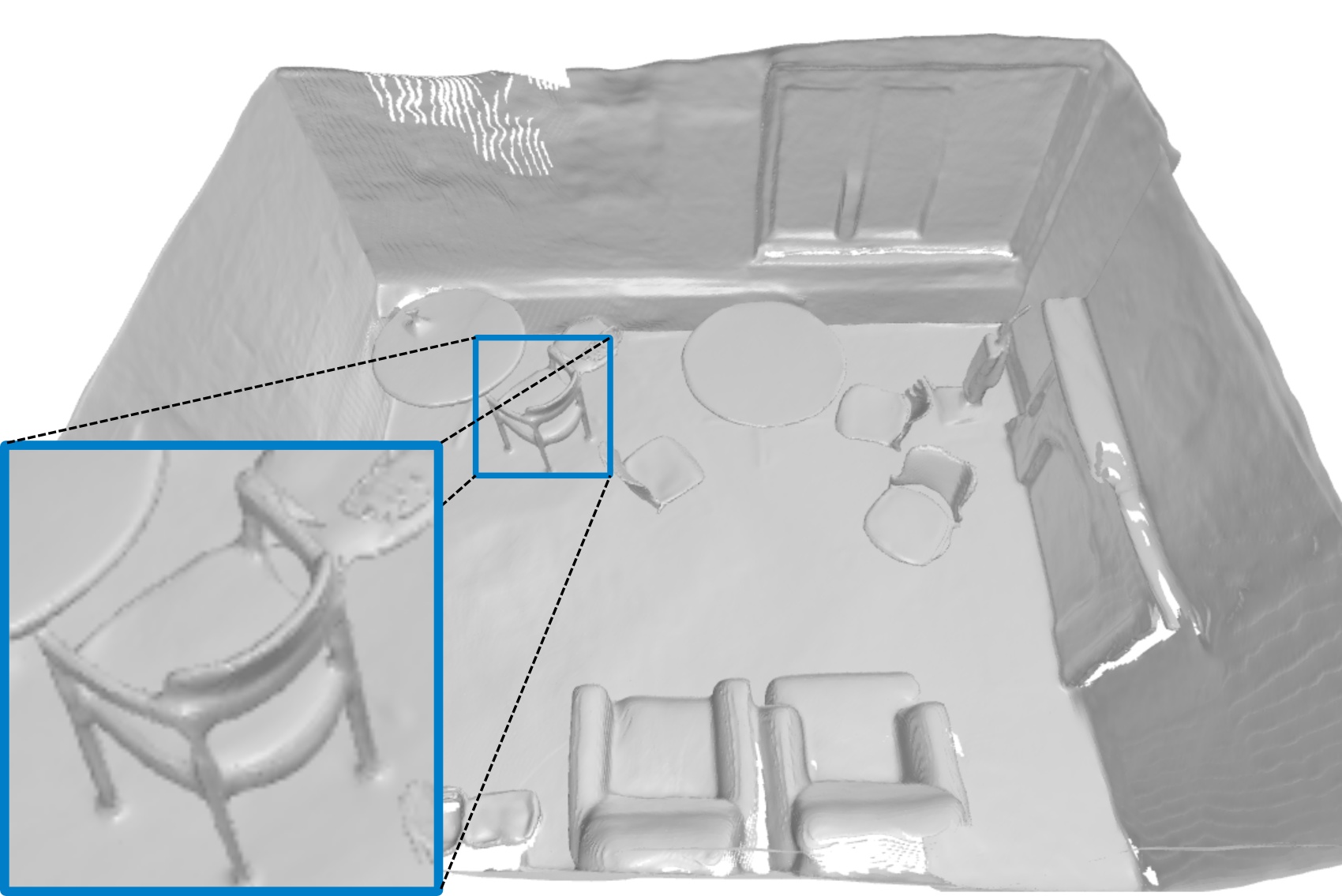}&
            \includegraphics[width=\fivew]{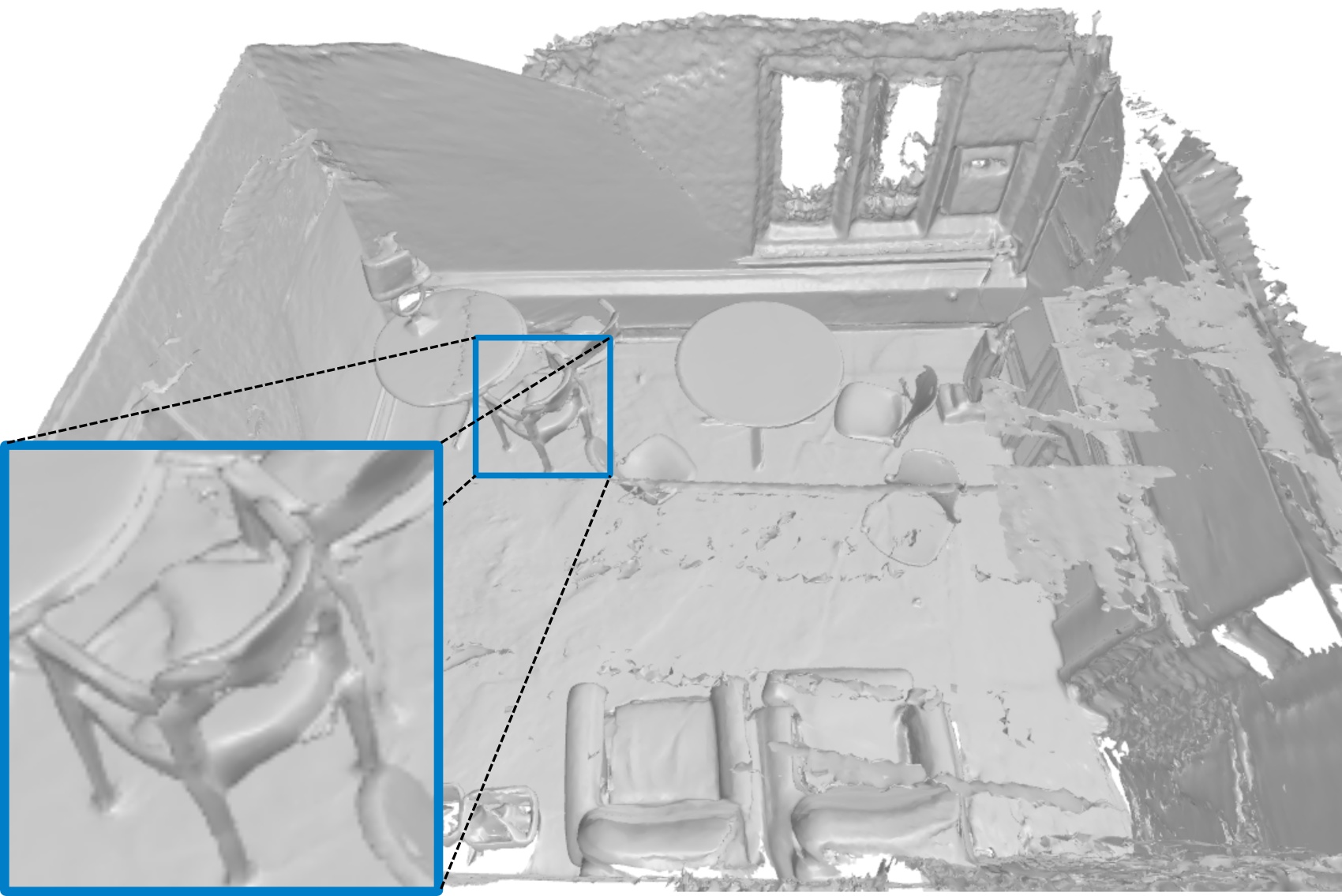}\\
        \end{tabular}
    	\end{minipage}
    	\\
    	\rule{0pt}{3ex}
    	\vspace{2em}
    	\begin{minipage}{1.0\textwidth}
    	    \centering
    	    \footnotesize
    	    \resizebox{\textwidth}{!}{%
    	    \begin{tabular}{ccccccccc}
            \toprule
            {} & COLMAP~\cite{Schoenberger2016ECCV} & UNISURF~\cite{Oechsle2021ICCV} & NeuS~\cite{Wang2021ARXIVb} & VolSDF~\cite{Yariv2021NEURIPS}  & M-SDF~\cite{guo2022manhattan} &NeuRIS~\cite{wang2022neuris} & Ours (Grids) & Ours (MLP) \\
            \midrule
            Chamfer-$L_1$ $\downarrow$ & 0.141 & 0.359 & 0.194 & 0.267 &0.070 & 0.050& 0.064 & \textbf{0.042}  \\
            F-score$\uparrow$          & 0.537 & 0.267&0.291& 0.364&0.602 &0.692 &0.626&\textbf{0.733}\\ 
            \bottomrule
            \end{tabular}
            }
    	\end{minipage}
    \end{tabular}
    \caption{ 
    \textbf{Scene-level Reconstruction on ScanNet.} 
    Colmap and VolSDF do not lead to competitive reconstructions. Manhatten-SDF achieves compelling results, but less-observed areas are noisier and details are missing. 
    In contrast, our approaches reconstruct smooth and details surfaces, achieving the best results. %
    Further, MLPs are more robust to the motion blur and noise in camera poses. %
    }
    \label{tab:scannet_fused}
    \vspace{-0.2cm}
\end{table}
}

\newcommand{\dtuallviewfused}{
\begin{table}[!tb]
	\centering
	\setlength{\tabcolsep}{2pt}
    \begin{tabular}{c}
    	\begin{minipage}{1.0\textwidth}
        	\centering
            \setlength{\tabcolsep}{0.1em}
            \renewcommand{\arraystretch}{0.5}
            \hfill{}\hspace*{-0.5em}
            \footnotesize
    	    \begin{tabular}{ccccc}
    	    \small NeuS \cite{Wang2021NEURIPS}&
             \small VolSDF \cite{Yariv2021NEURIPS} &
             \small \textbf{Ours} (MLP)&
             \small  \textbf{Ours} (Grids) & Ground Truth View\\
            \includegraphics[width=\fivew]{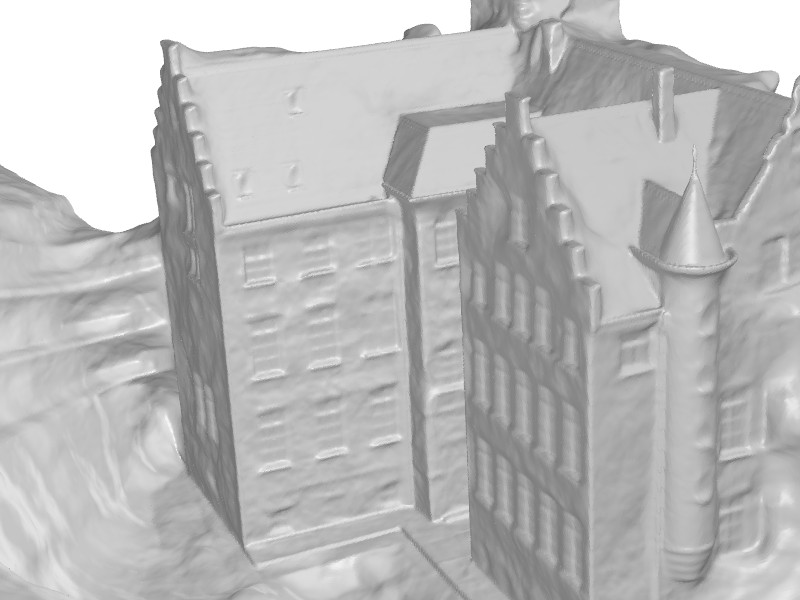}&
            \includegraphics[width=\fivew]{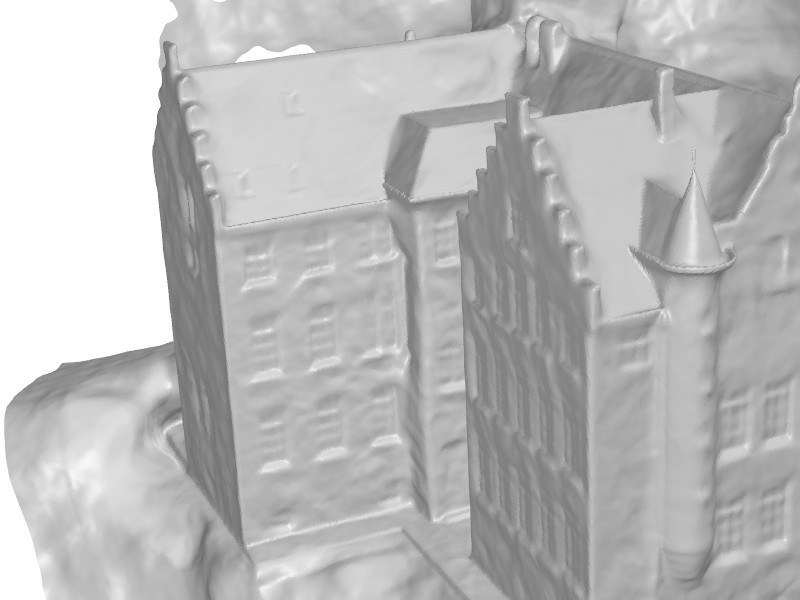}&
            \includegraphics[width=\fivew]{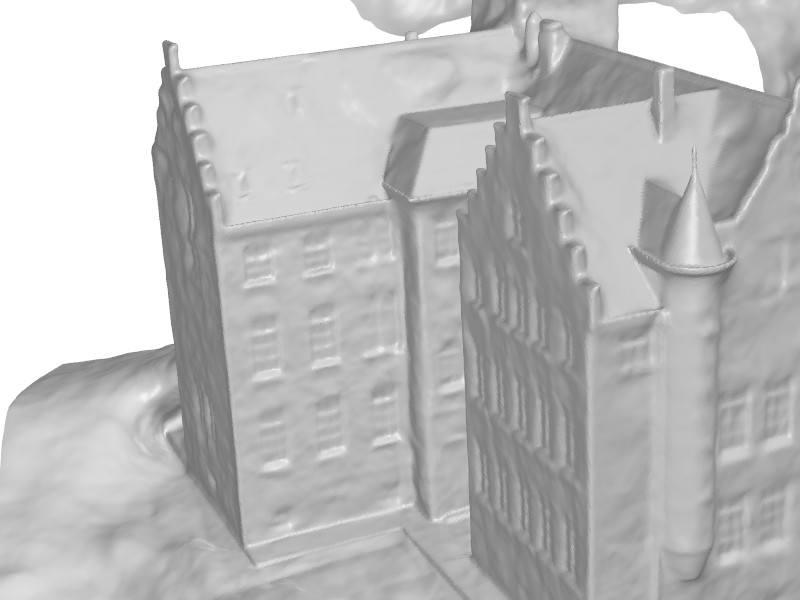}&
            \includegraphics[width=\fivew]{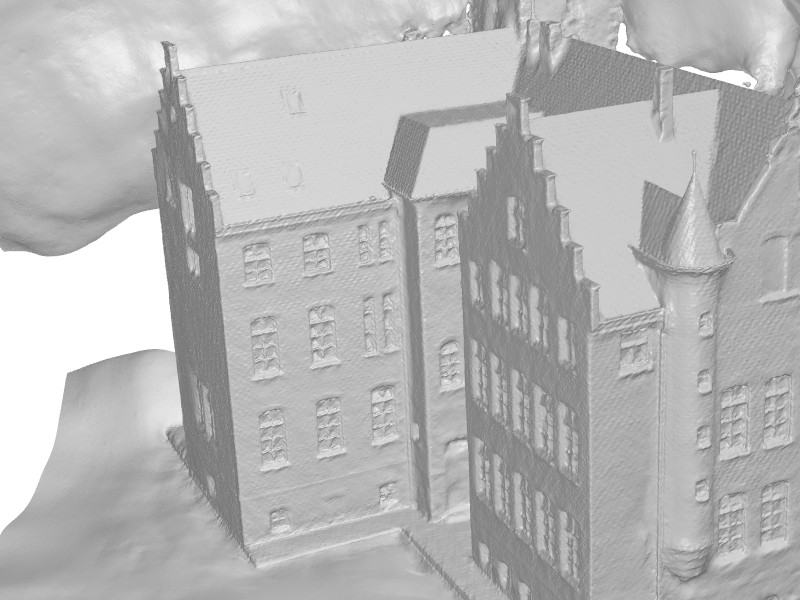}&
            \includegraphics[width=\fivew]{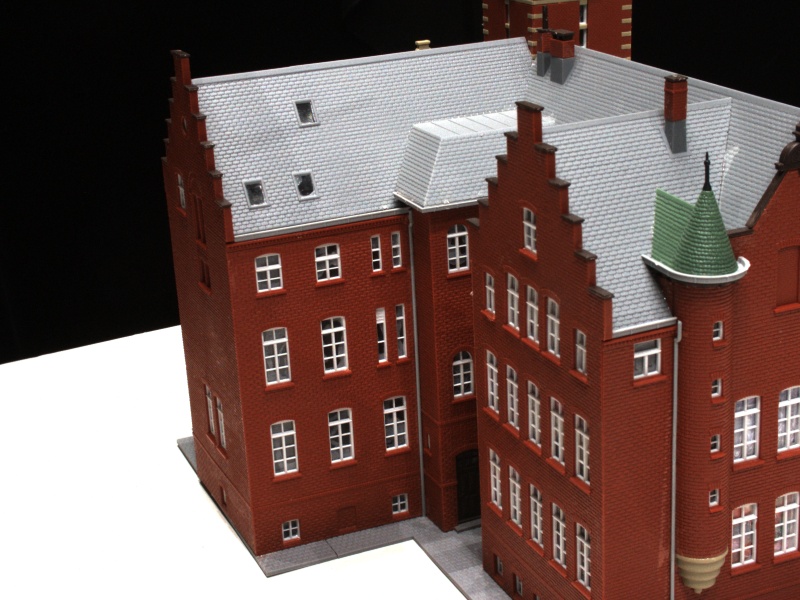}\\
        \end{tabular}
    	\end{minipage}
    	\\
    	\rule{0pt}{3ex}
    	\vspace{2em}
    	\begin{minipage}{1.0\textwidth}
    	    \centering
    	    \footnotesize
    	    \resizebox{\textwidth}{!}{%
    	    \centering
            \begin{tabular}{ccccccccccccccccc}
                \toprule
                Scan & 24 & 37 & 40 & 55 & 63 & 65 & 69 &83 & 97 &105 & 106 & 110 & 114 & 118 & 122 & \textbf{Mean}   \\
                \midrule
                COLMAP                         & 0.81 & 2.05 & 0.73 & 1.22 & 1.79 & 1.58 & 1.02 & 3.05 & 1.40 & 2.05 & 1.00 & 1.32 & 0.49 & 0.78 & 1.17 & 1.36 \\
                NeRF~\cite{Mildenhall2020ECCV} & 1.90 & 1.60 & 1.85 & 0.58 & 2.28 & 1.27 & 1.47 & 1.67 & 2.05 & 1.07 & 0.88 & 2.53 & 1.06 & 1.15 & 0.96 & 1.49 \\
                UniSurf~\cite{Oechsle2021ICCV} & 1.32 & 1.36 & 1.72 & 0.44 & 1.35 & 0.79 & 0.80 & 1.49 & 1.37 & 0.89 & 0.59 & 1.47 & 0.46 & 0.59 & 0.62 & 1.02 \\
                NeuS~\cite{Wang2021NEURIPS}    & 1.00 & 1.37 & 0.93 & 0.43 & 1.10 & \textbf{0.65} & \textbf{0.57} & 1.48 & \textbf{1.09} & 0.83 & \textbf{0.52} & 1.20 & 0.35 & \textbf{0.49} & 0.54 & 0.84 \\
                VolSDF~\cite{Yariv2021NEURIPS} & 1.14 & 1.26 & 0.81 & 0.49 & 1.25 & 0.70 & 0.72 & 1.29 & 1.18 & 0.70 & 0.66 & 1.08 & 0.42 & 0.61 & 0.55 & 0.86 \\
                \midrule
                Ours (MLP)                     & 0.83 & 1.61 & 0.65 & 0.47 & 0.92 & 0.87 & 0.87 & 1.30 & 1.25 & 0.68 & 0.65 & \textbf{0.96} & \textbf{0.41} & 0.62 & 0.58 & 0.84 \\
                Ours(Grids)                    & \textbf{0.66} & \textbf{0.88} & \textbf{0.43} & \textbf{0.40} & \textbf{0.87} & 0.78 & 0.81 & \textbf{1.23} & 1.18 & \textbf{0.66} & 0.66 & \textbf{0.96} & \textbf{0.41} & 0.57 & \textbf{0.51} & \textbf{0.73} \\
                \bottomrule
                \end{tabular}
            }
    	\end{minipage}
    \end{tabular}
    \caption{ 
    \red{\textbf{Object-level Reconstruction on DTU Dataset will All Input Views.} We compare Chamfer distance with state-of-the-art methods. Our approach with MLP achieves similar results to previous methods, while our method with multi-resolution feature grids leads to more detailed surfaces and outperforms previous work by a large margin.
    }}
    \label{tab:dtuallview_fused}
    \vspace{-0.4cm}
\end{table}
}

\newcommand{\bc}{\mathbf{c}}
\newcommand{\bd}{\mathbf{d}}

\newcommand{\bh}{\mathbf{h}}

\newcommand{\bn}{\mathbf{n}}
\newcommand{\bo}{\mathbf{o}}
\newcommand{\bp}{\mathbf{p}}

\newcommand{\br}{\mathbf{r}}

\newcommand{\bv}{\mathbf{v}}

\newcommand{\bx}{\mathbf{x}}

\newcommand{\bz}{\mathbf{z}}

\newcommand{\nR}{\mathbb{R}}

\newcommand{\cR}{\mathcal{R}}

\newcommand{\cX}{\mathcal{X}}

\newcommand{\figref}[1]{Fig.~\ref{#1}}
\newcommand{\secref}[1]{Section~\ref{#1}}

\newcommand{\eqnref}[1]{Eq.~\eqref{#1}}
\newcommand{\tabref}[1]{Table~\ref{#1}}

\DeclareMathOperator*{\argmin}{argmin~}

\makeatletter
\DeclareRobustCommand\onedot{\futurelet\@let@token\@onedot}
\def\@onedot{\ifx\@let@token.\else.\null\fi\xspace}
\def\eg{e.g\onedot} 
 
\def\cf{cf\onedot}

\def\wrt{wrt\onedot}

\makeatother

\newcommand*\rot{\rotatebox{90}}

\newcommand{\boldparagraph}[1]{\vspace{-0.1cm}\noindent{\bf #1} }

\definecolor{darkgreen}{rgb}{0,0.7,0}
\definecolor{darkblue}{RGB}{31,119,180}
\definecolor{darkred}{RGB}{214,39,40}

\newcommand{\red}[1]{#1}

\usepackage{appendix}

\providecommand{\impath}[1]{}
\providecommand{\impatha}[1]{}
\providecommand{\impathb}[1]{}
\providecommand{\impathc}[1]{}
\providecommand{\impathd}[1]{}
\providecommand{\impathe}[1]{}

\newcommand{\architecture}{
\begin{figure*}[t]
  \includegraphics[width=\textwidth]{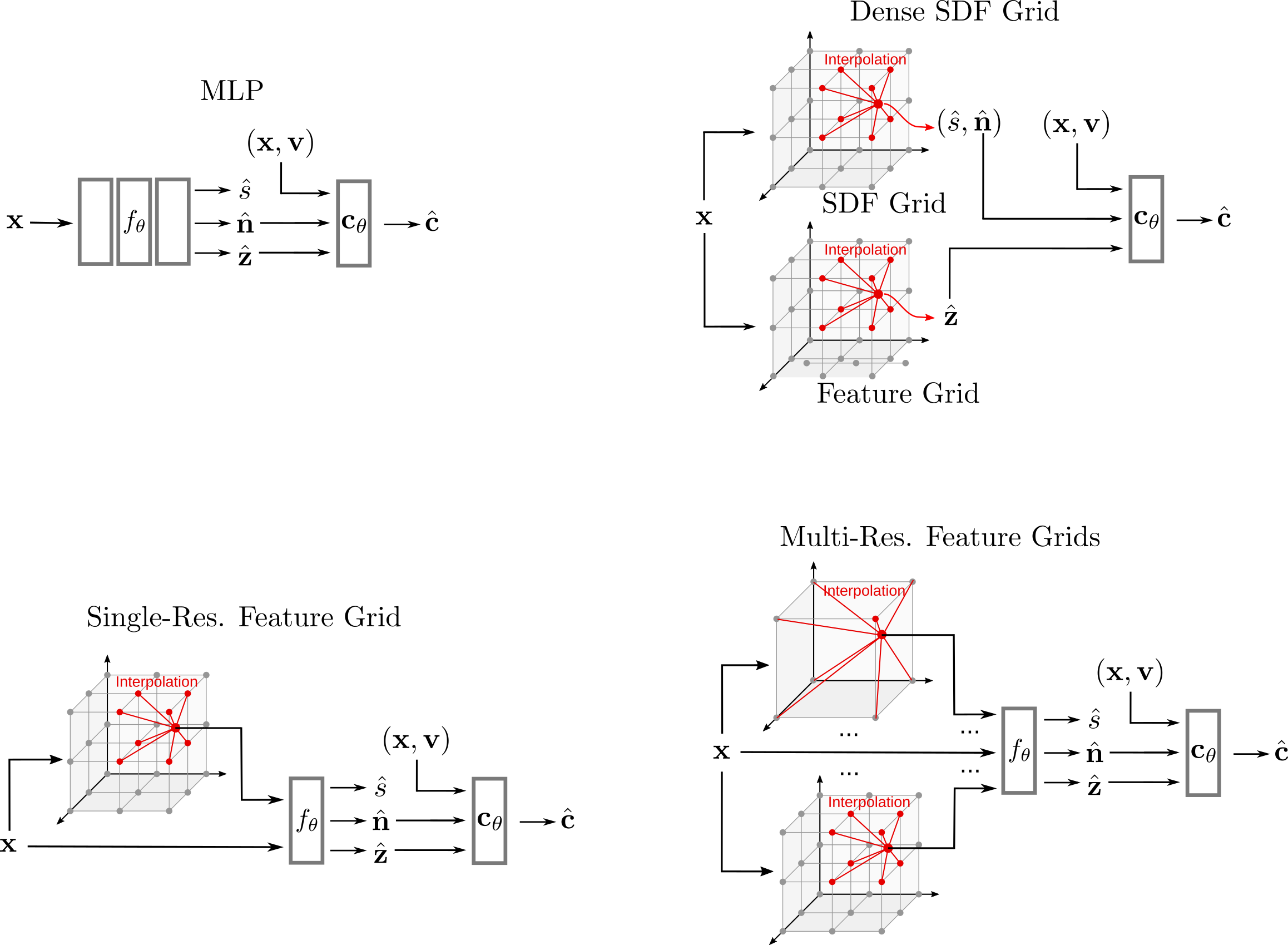}
  \caption{\textbf{Architectures.} We show an overview over four different scene representations considered in this paper.
  }
  \label{fig:architecture}
\end{figure*}
}

\newcommand{\newwidth}{0.142\textwidth}
\newcommand{\figuredturgb}{
\begin{figure*}[t]
        \centering
        \setlength{\tabcolsep}{0.1em}
        \renewcommand{\arraystretch}{0.7}
        \footnotesize
        \begin{tabular}{ccccccc}
            \includegraphics[width=\newwidth]{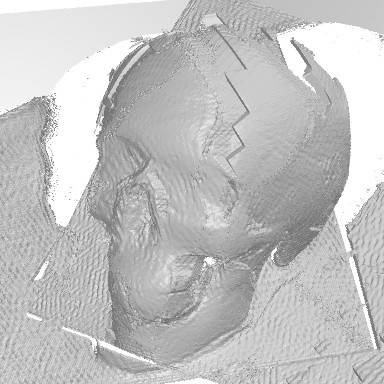}&
            \includegraphics[width=\newwidth]{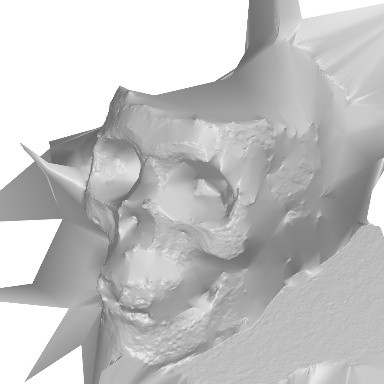}&
            \includegraphics[width=\newwidth]{gfx/DTU_vis/scan65MLP_noprior_render_23.jpg}&
            \includegraphics[width=\newwidth]{gfx/DTU_vis/scan65MLP_wprior_decay_render_23.jpg}&
            \includegraphics[width=\newwidth]{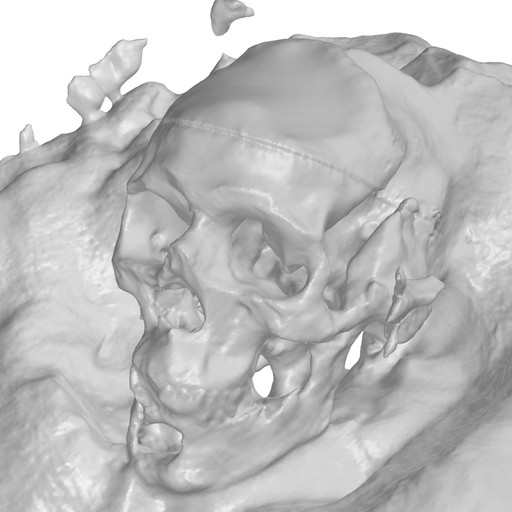}&
            \includegraphics[width=\newwidth]{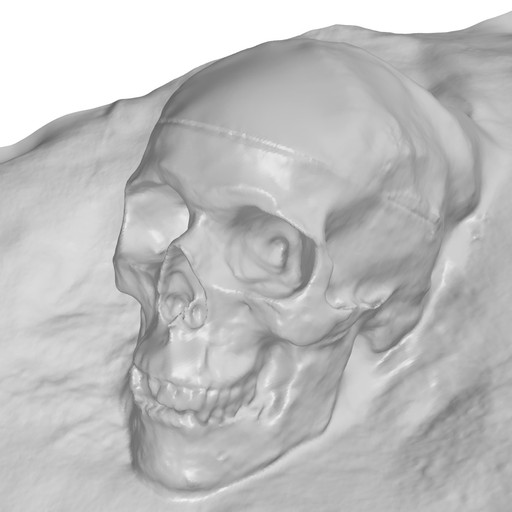}&
            \includegraphics[width=\newwidth]{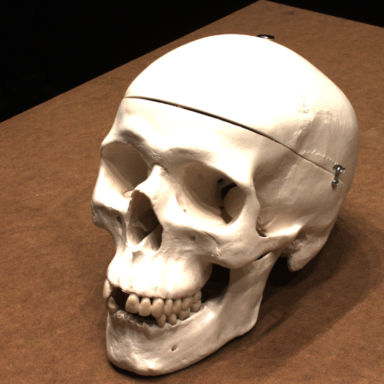}\\
            \includegraphics[width=\newwidth]{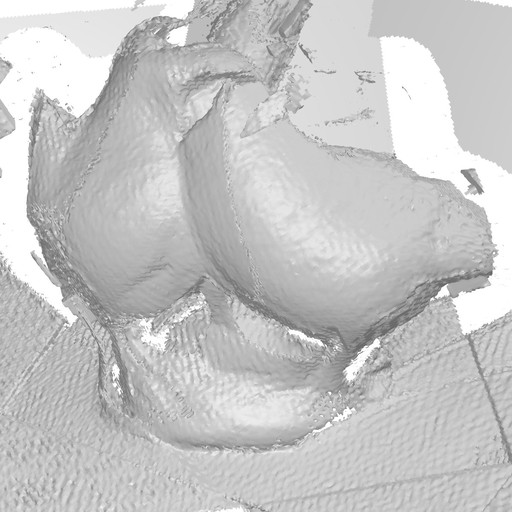}&
            \includegraphics[width=\newwidth]{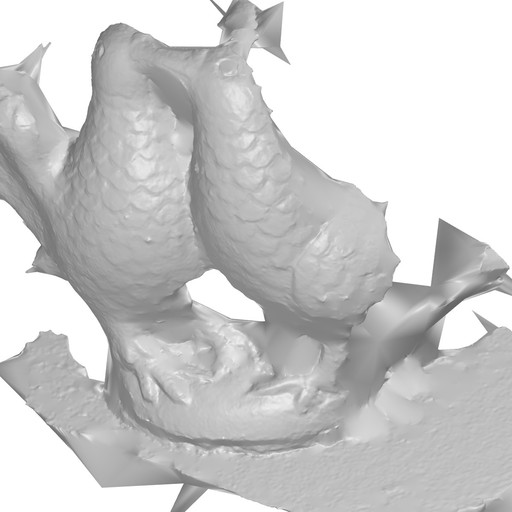}&
            \includegraphics[width=\newwidth]{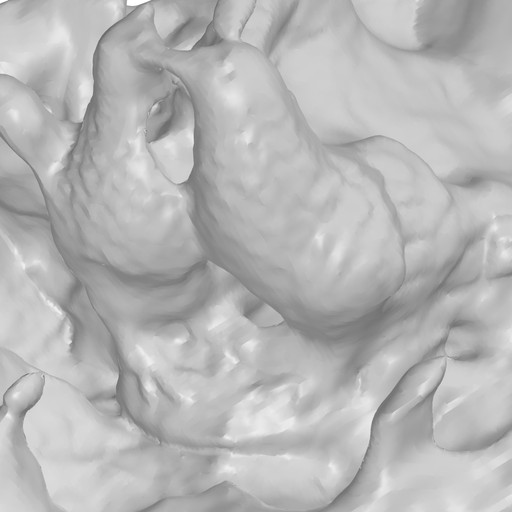}&
            \includegraphics[width=\newwidth]{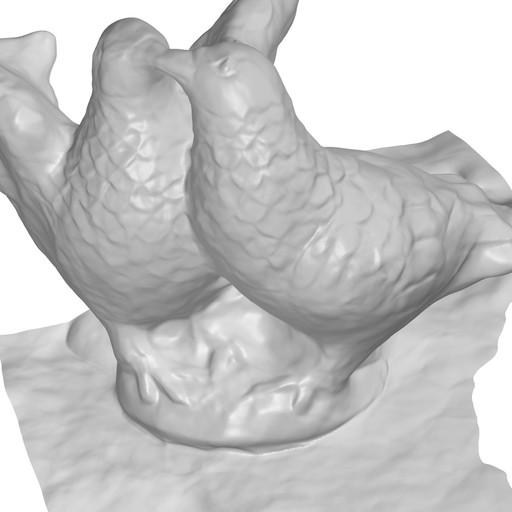}&
            \includegraphics[width=\newwidth]{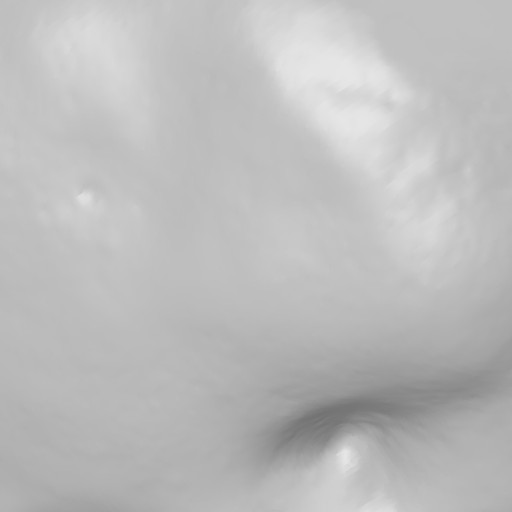}&
            \includegraphics[width=\newwidth]{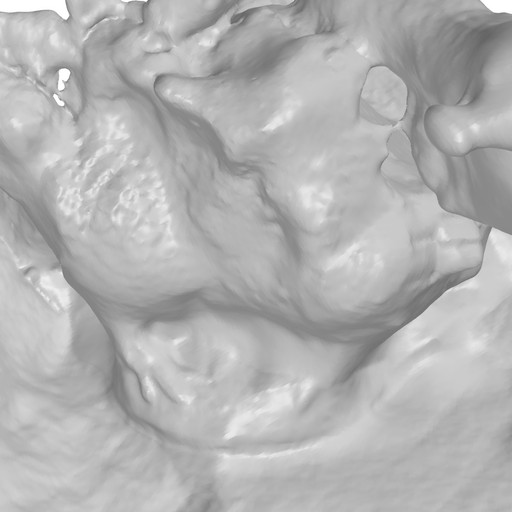}&
            \includegraphics[width=\newwidth]{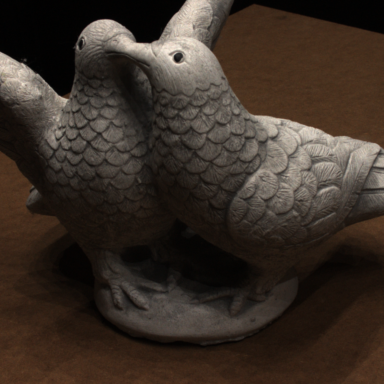}\\
            \includegraphics[width=\newwidth]{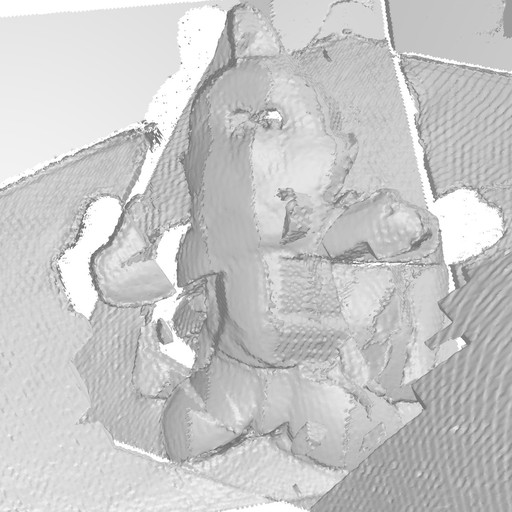}&
            \includegraphics[width=\newwidth]{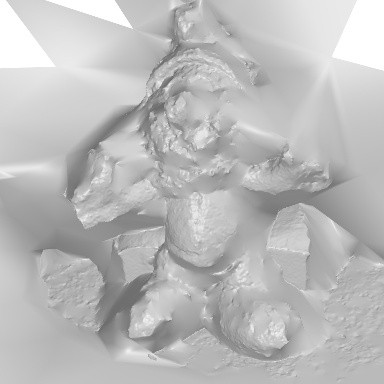}&
            \includegraphics[width=\newwidth]{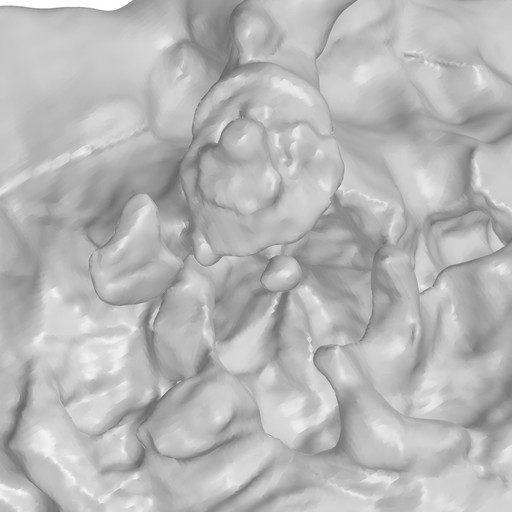}&
            \includegraphics[width=\newwidth]{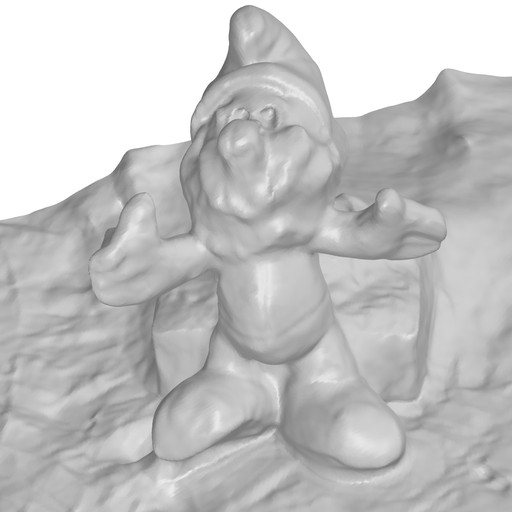}&
            \includegraphics[width=\newwidth]{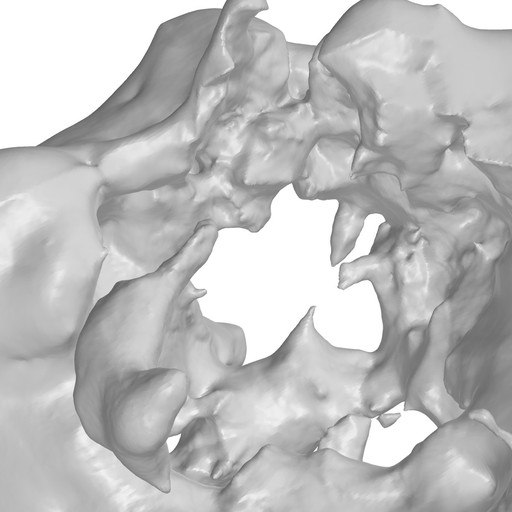}&
            \includegraphics[width=\newwidth]{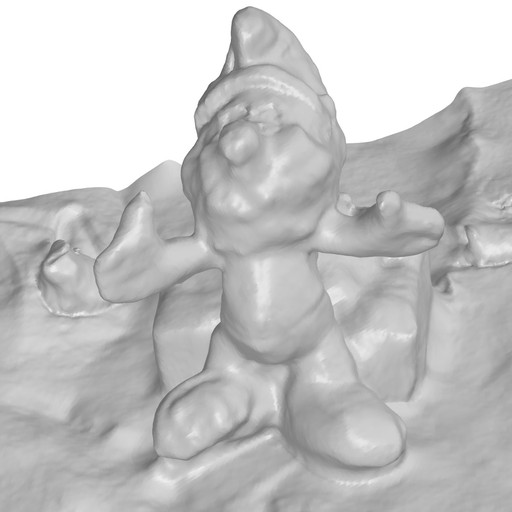}&
            \includegraphics[width=\newwidth]{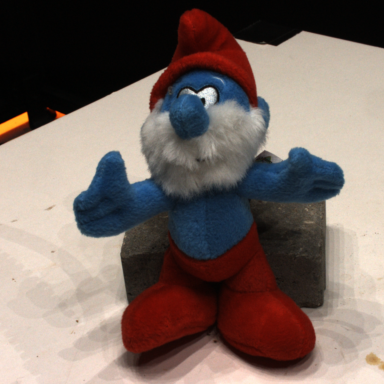}\\
            \includegraphics[width=\newwidth]{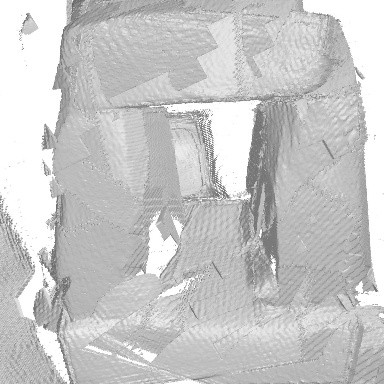}&
            \includegraphics[width=\newwidth]{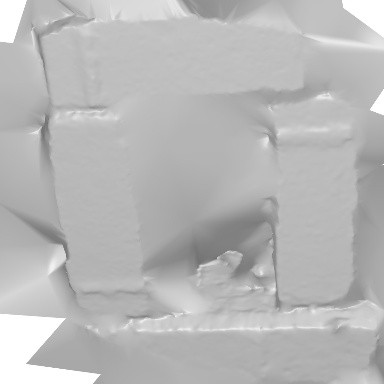}&
            \includegraphics[width=\newwidth]{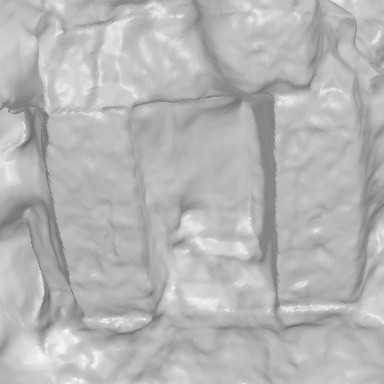}&
            \includegraphics[width=\newwidth]{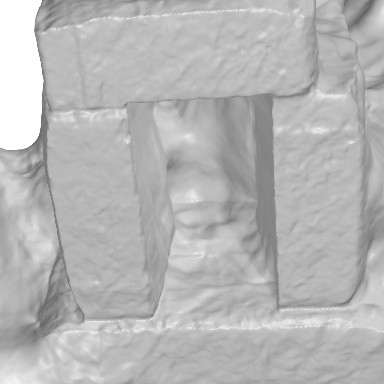}&
            \includegraphics[width=\newwidth]{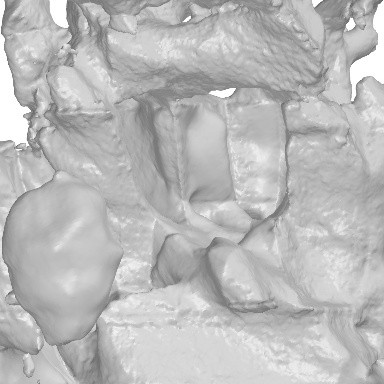}&
            \includegraphics[width=\newwidth]{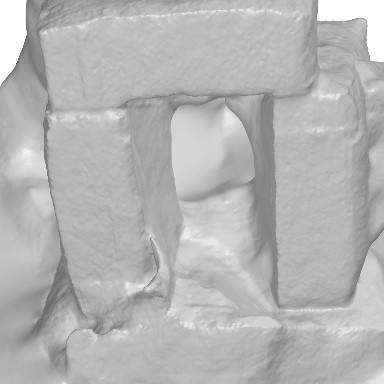}&
            \includegraphics[width=\newwidth]{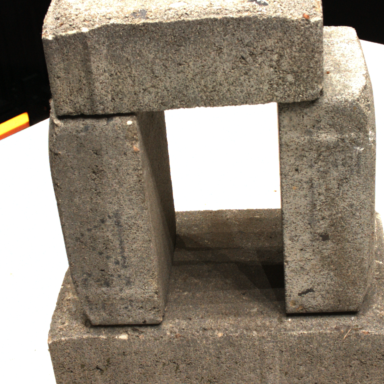}\\
            \includegraphics[width=\newwidth]{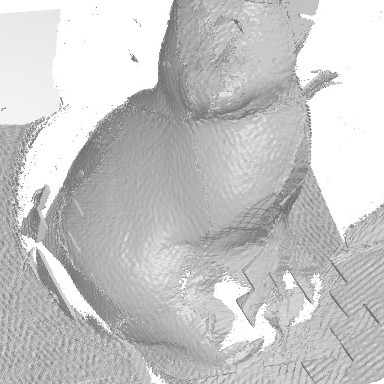}&
            \includegraphics[width=\newwidth]{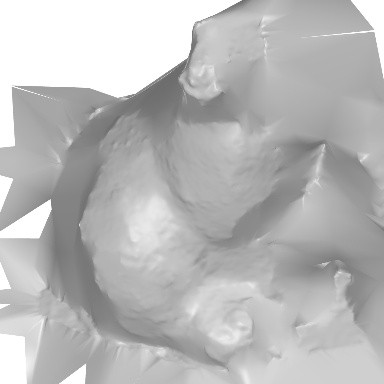}&
            \includegraphics[width=\newwidth]{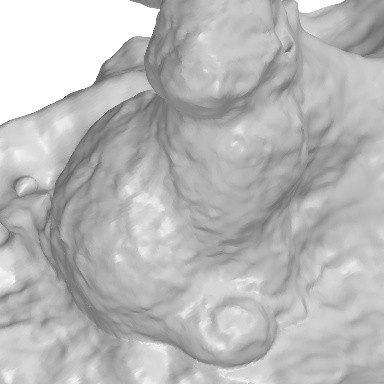}&
            \includegraphics[width=\newwidth]{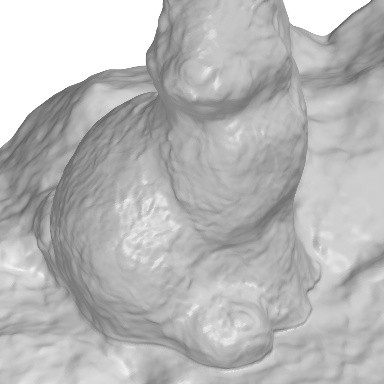}&
            \includegraphics[width=\newwidth]{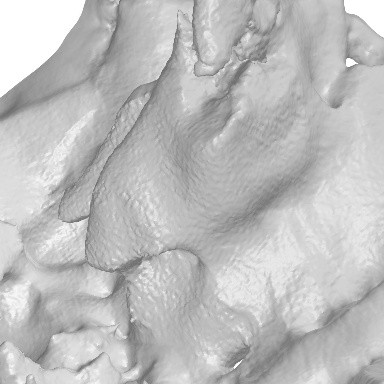}&
            \includegraphics[width=\newwidth]{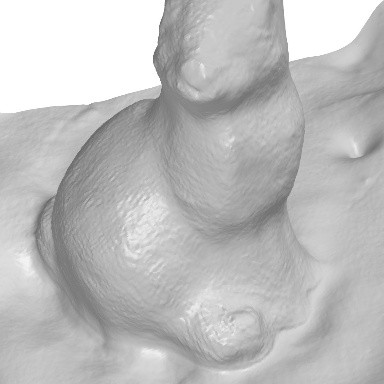}&
            \includegraphics[width=\newwidth]{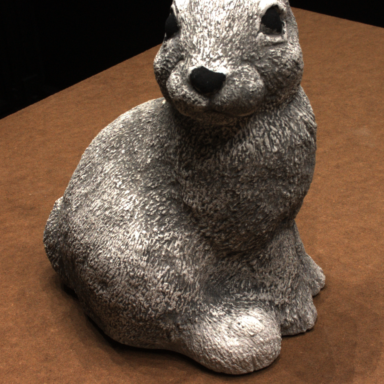}\\
            \includegraphics[width=\newwidth]{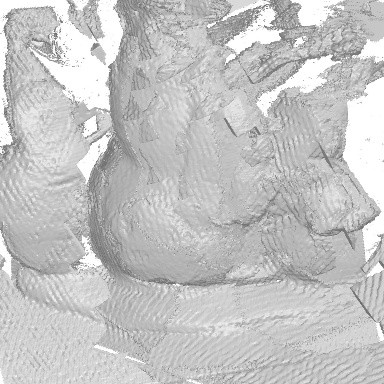}&
            \includegraphics[width=\newwidth]{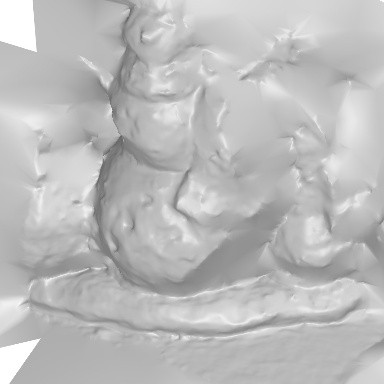}&
            \includegraphics[width=\newwidth]{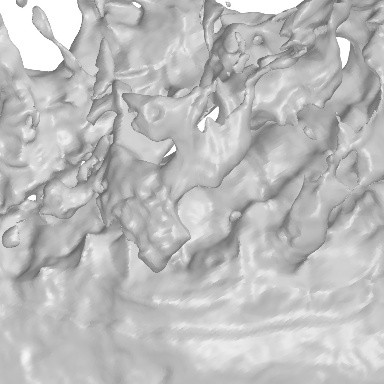}&
            \includegraphics[width=\newwidth]{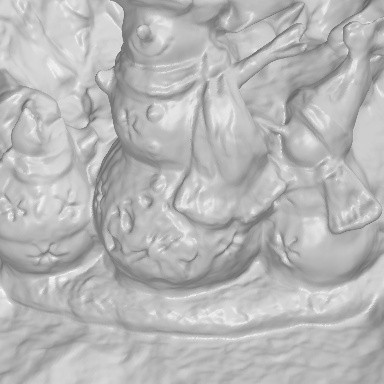}&
            \includegraphics[width=\newwidth]{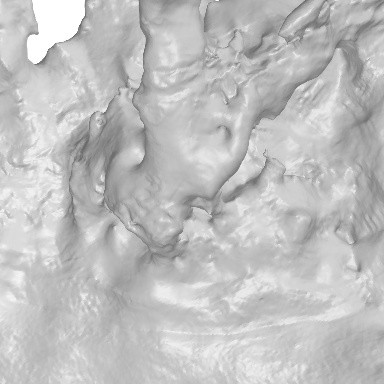}&
            \includegraphics[width=\newwidth]{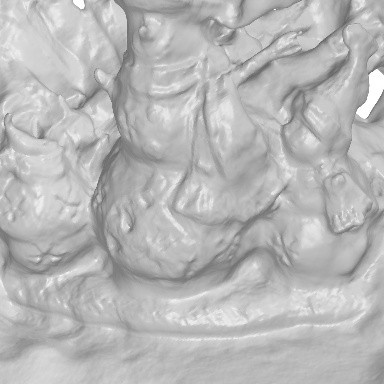}&
            \includegraphics[width=\newwidth]{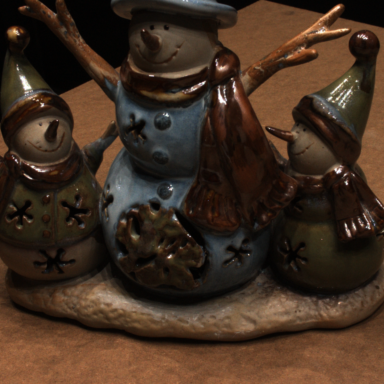}\\
            \includegraphics[width=\newwidth]{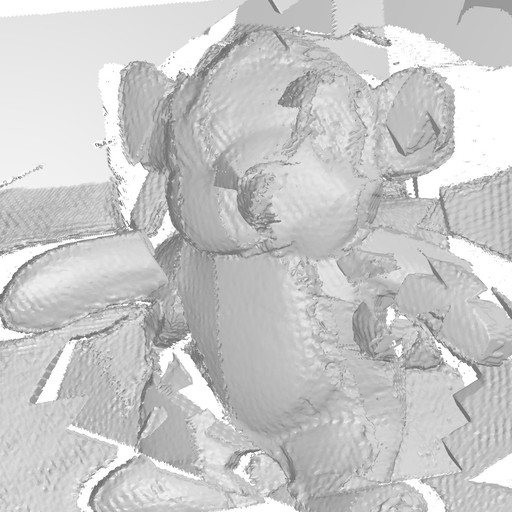}&
            \includegraphics[width=\newwidth]{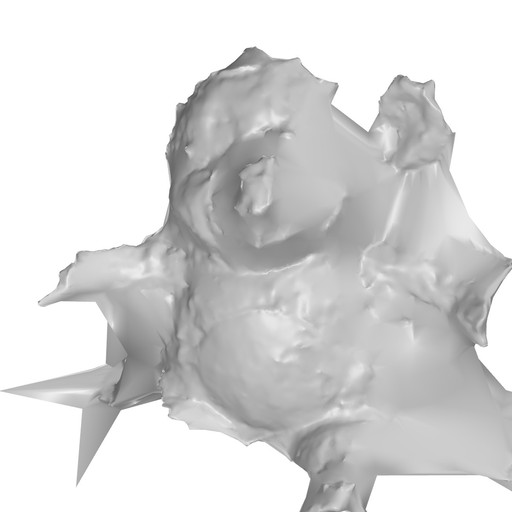}&
            \includegraphics[width=\newwidth]{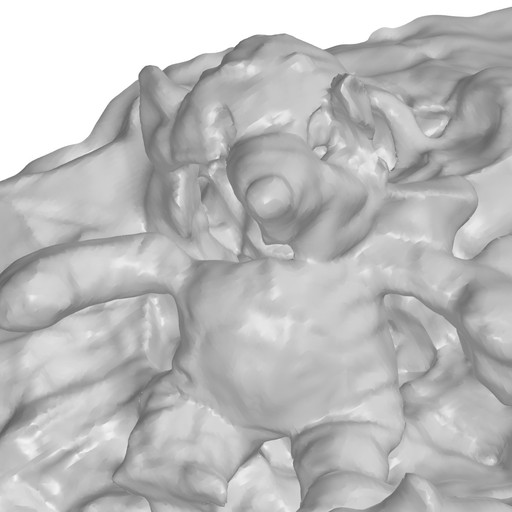}&
            \includegraphics[width=\newwidth]{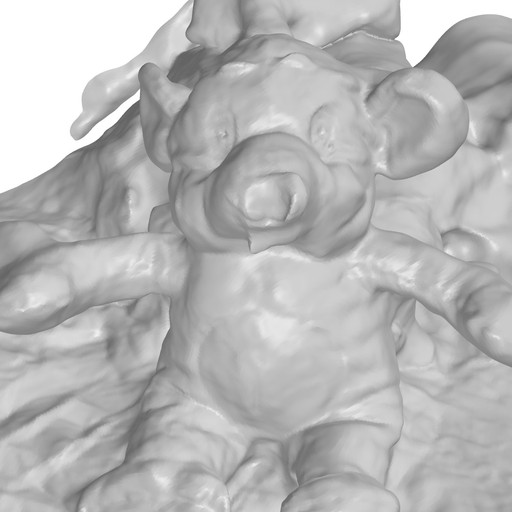}&
            \includegraphics[width=\newwidth]{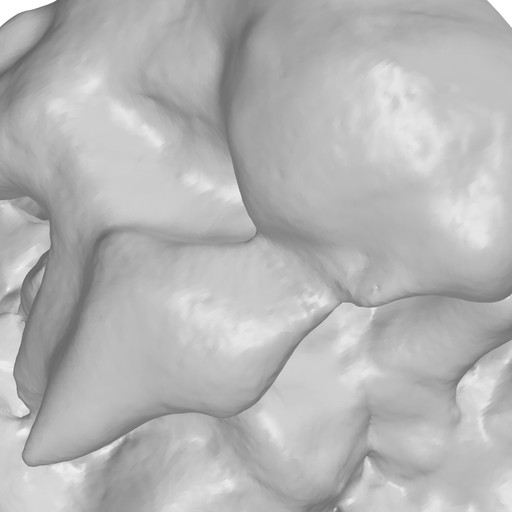}&
            \includegraphics[width=\newwidth]{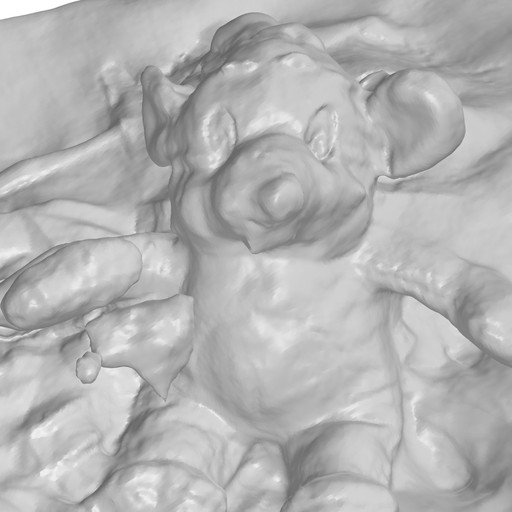}&
            \includegraphics[width=\newwidth]{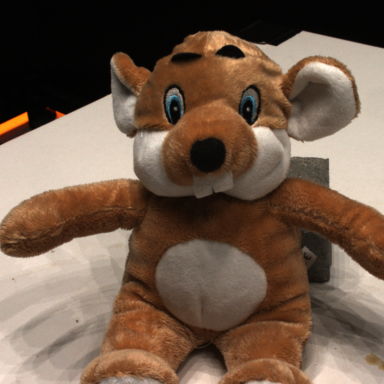}\\
            \includegraphics[width=\newwidth]{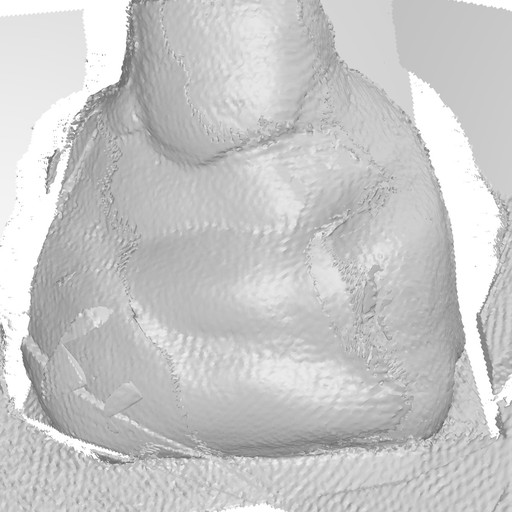}&
            \includegraphics[width=\newwidth]{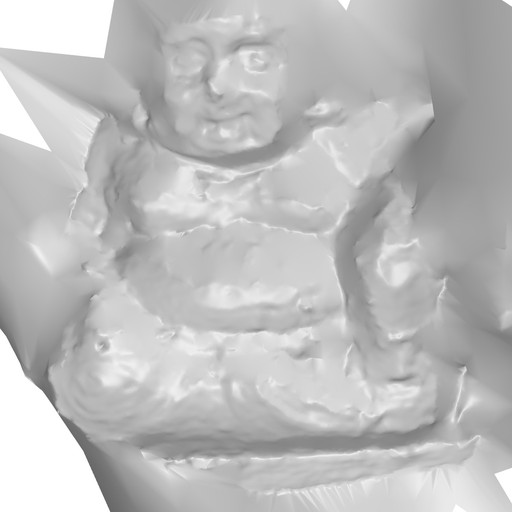}&
            \includegraphics[width=\newwidth]{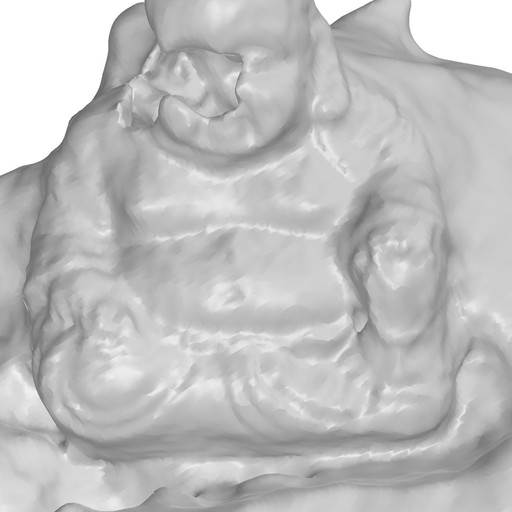}&
            \includegraphics[width=\newwidth]{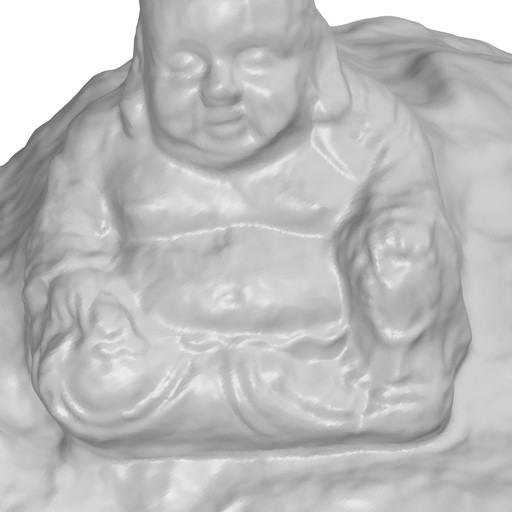}&
            \includegraphics[width=\newwidth]{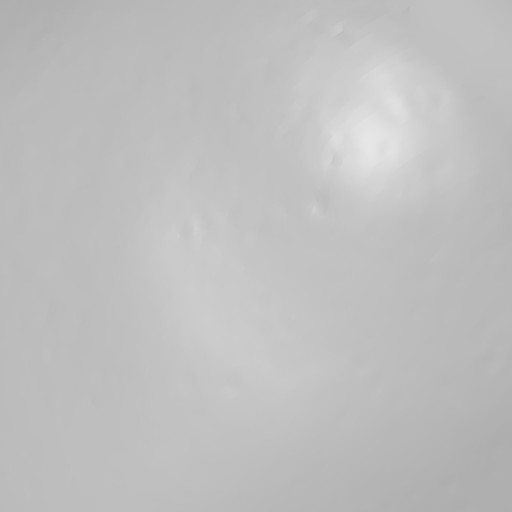}&
            \includegraphics[width=\newwidth]{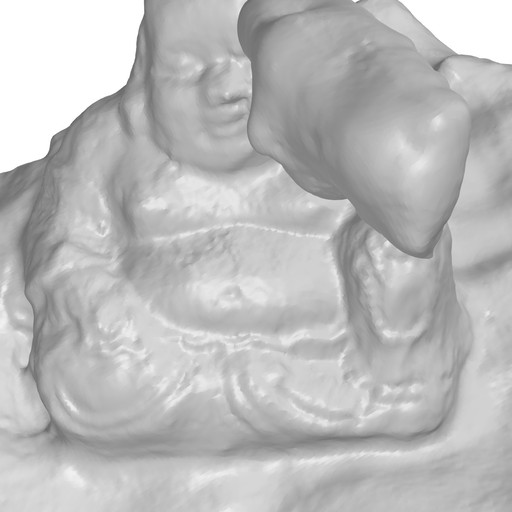}&
            \includegraphics[width=\newwidth]{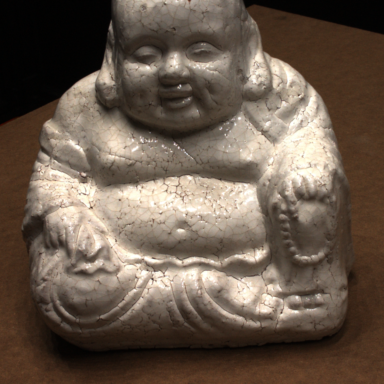}\\
            \includegraphics[width=\newwidth]{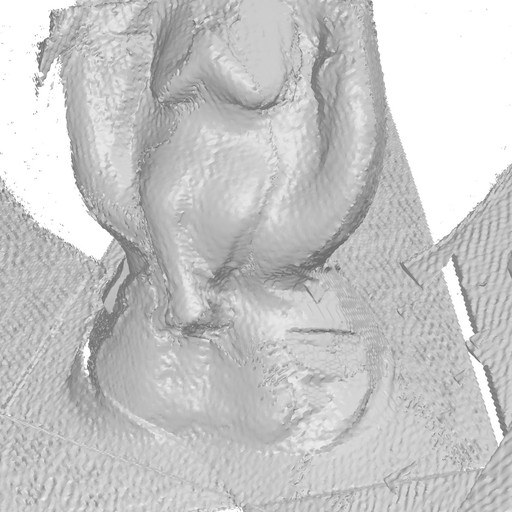}&
            \includegraphics[width=\newwidth]{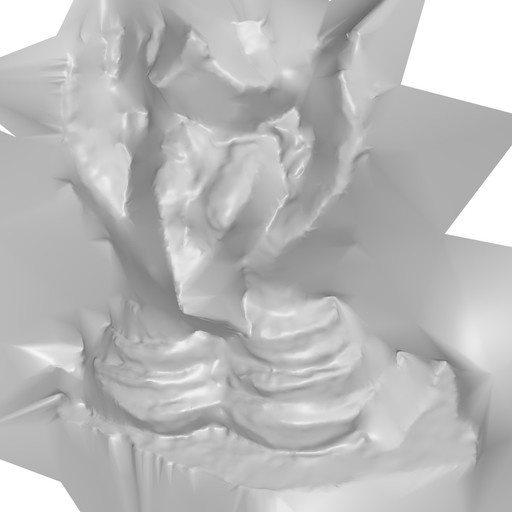}&
            \includegraphics[width=\newwidth]{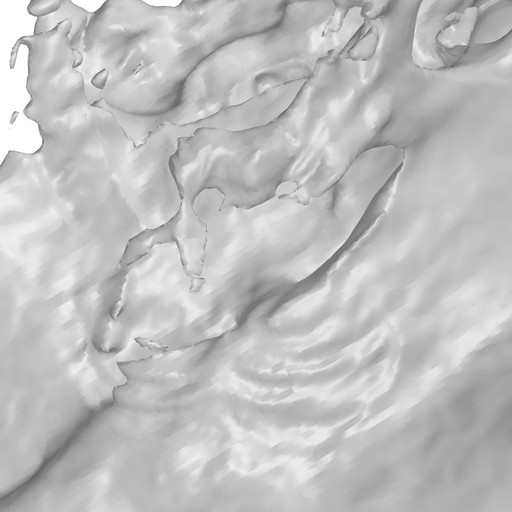}&
            \includegraphics[width=\newwidth]{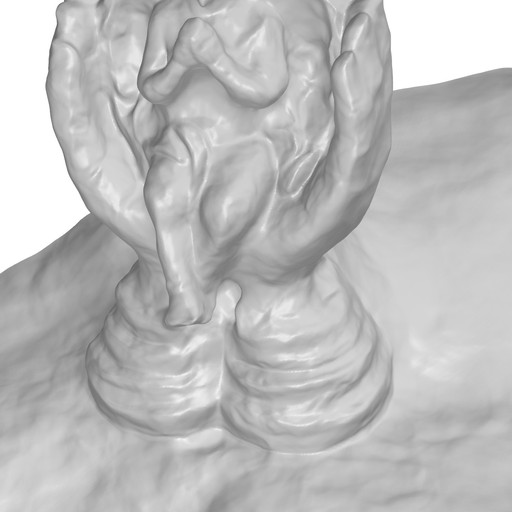}&
            \includegraphics[width=\newwidth]{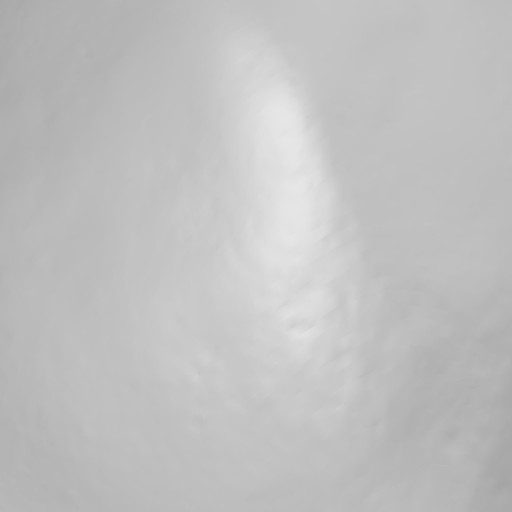}&
            \includegraphics[width=\newwidth]{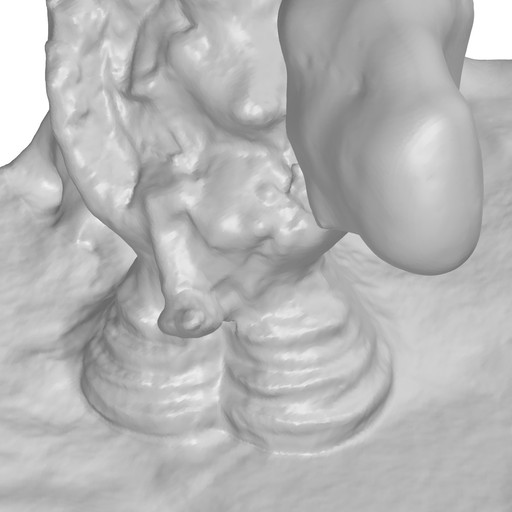}&
            \includegraphics[width=\newwidth]{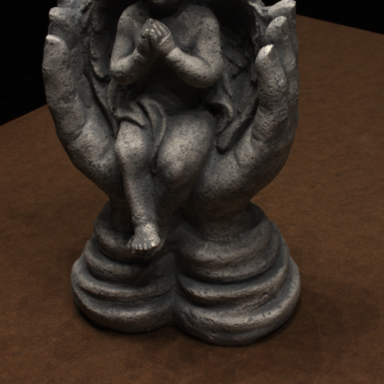}\\
            
            \includegraphics[width=\newwidth]{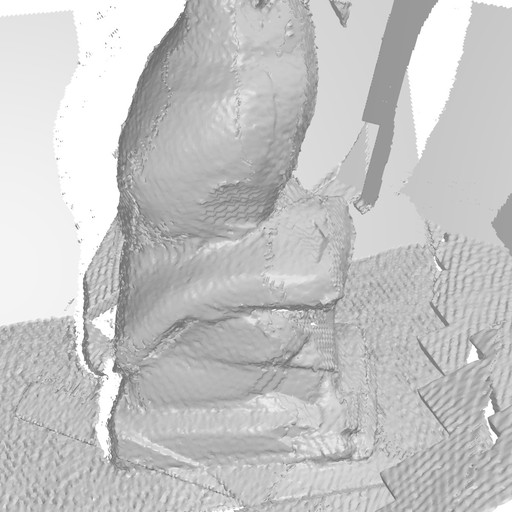}&
            \includegraphics[width=\newwidth]{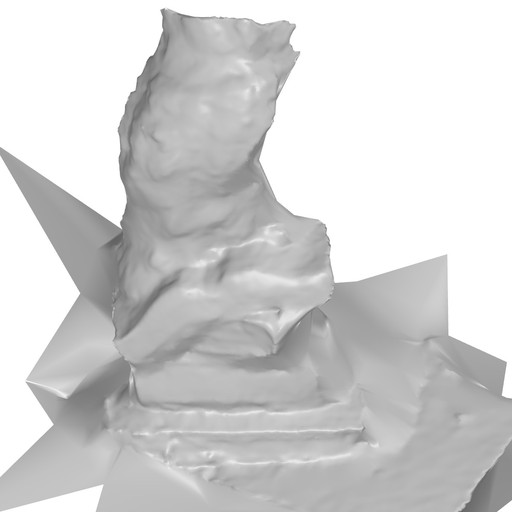}&
            \includegraphics[width=\newwidth]{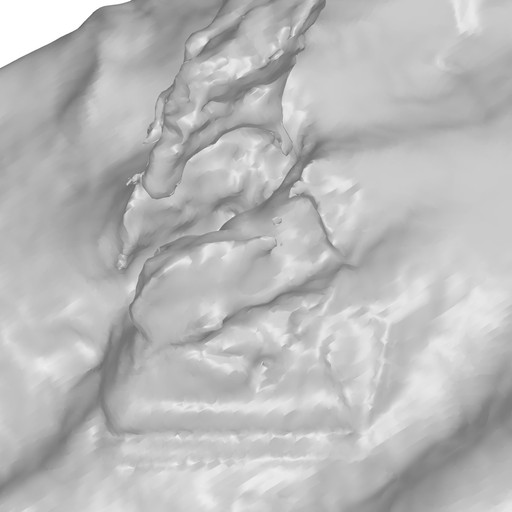}&
            \includegraphics[width=\newwidth]{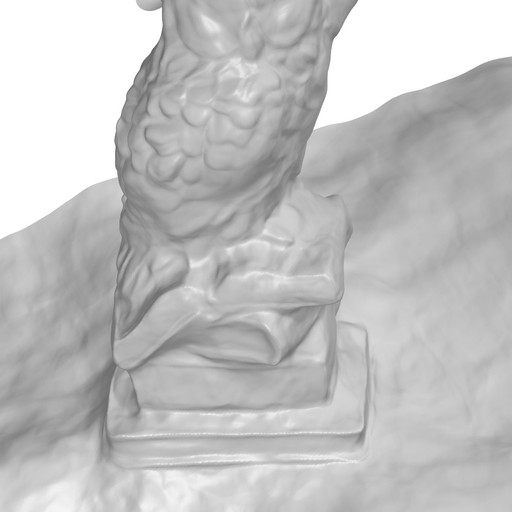}&
            \includegraphics[width=\newwidth]{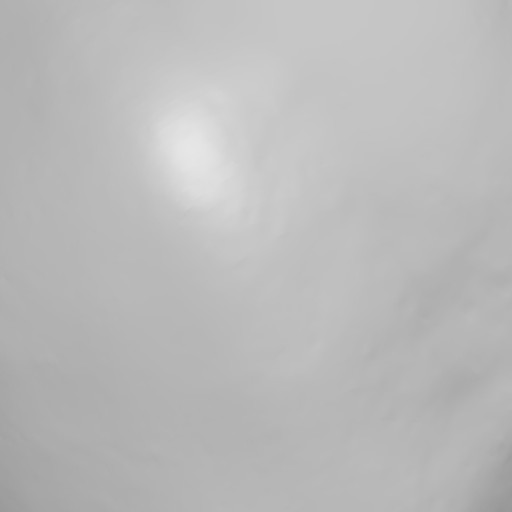}&
            \includegraphics[width=\newwidth]{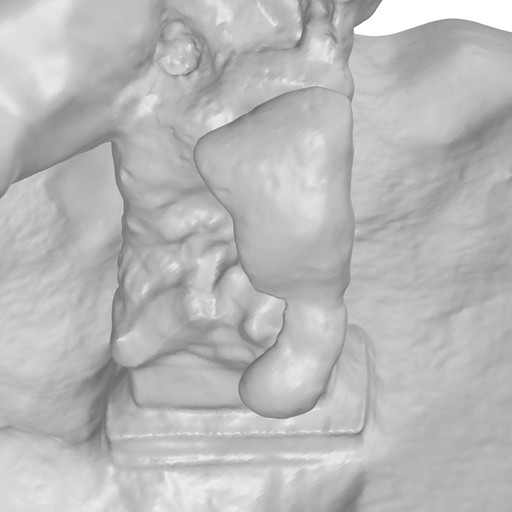}&
            \includegraphics[width=\newwidth]{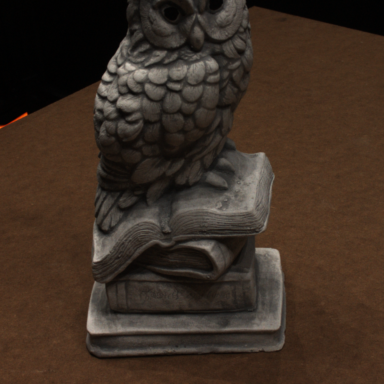}\\

             TSDF-Fusion &
             RealityCapture &
             MLP~\cite{Yariv2021NEURIPS}&
             MLP & 
             Multi-Res.& 
             Multi-Res.& 
             GT View \\
             {}&
             {}&
             {}&
             w/ cues& 
             Grids& 
             Grids w/ cues& 
             {}\\
        \end{tabular}
    \caption{\textbf{Qualitative Comparison on the DTU Dataset with 3 Input Views.} Adding monocular geometric cues improves 3D reconstruction quality for both MLP and Multi-Res. Grids. We show a failure case on the last row. }
    \label{fig:dtu}
\end{figure*}
}

\newcommand{\mywidth}{0.196\textwidth}
\newcommand{\figurescannetjpg}{
\begin{figure*}[t]
        \centering
        \setlength{\tabcolsep}{0.1em}
        \renewcommand{\arraystretch}{0.5}
        \hfill{}\hspace*{-0.5em}
        \footnotesize
        \begin{tabular}{ccccc}
            \includegraphics[width=\mywidth]{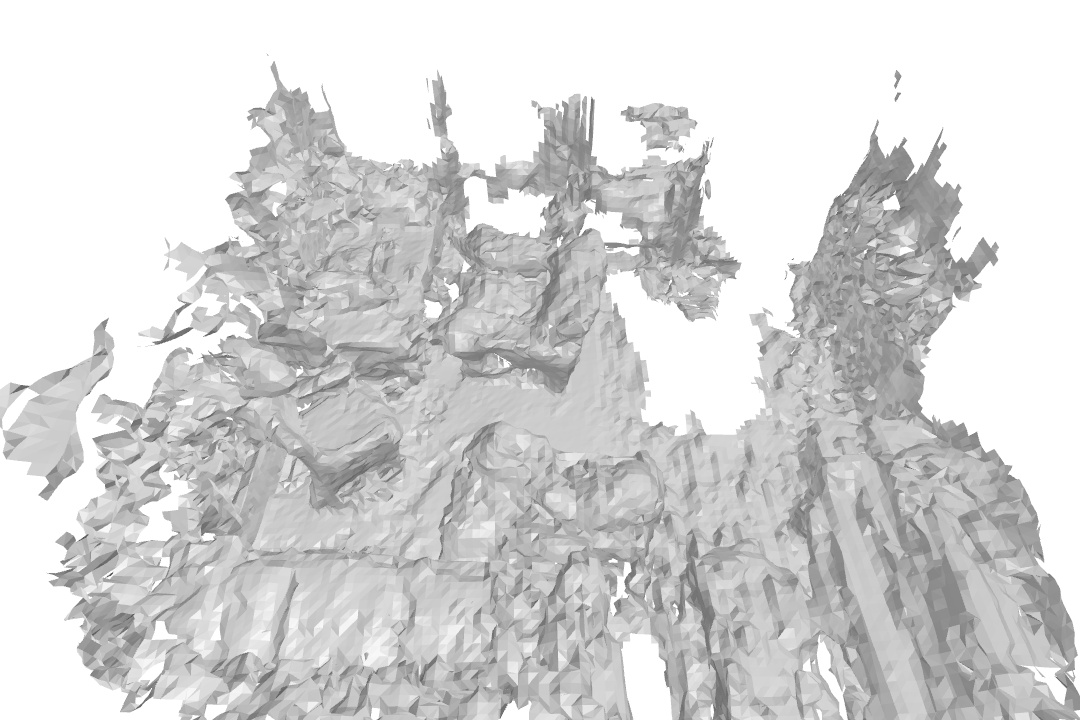}&
            \includegraphics[width=\mywidth]{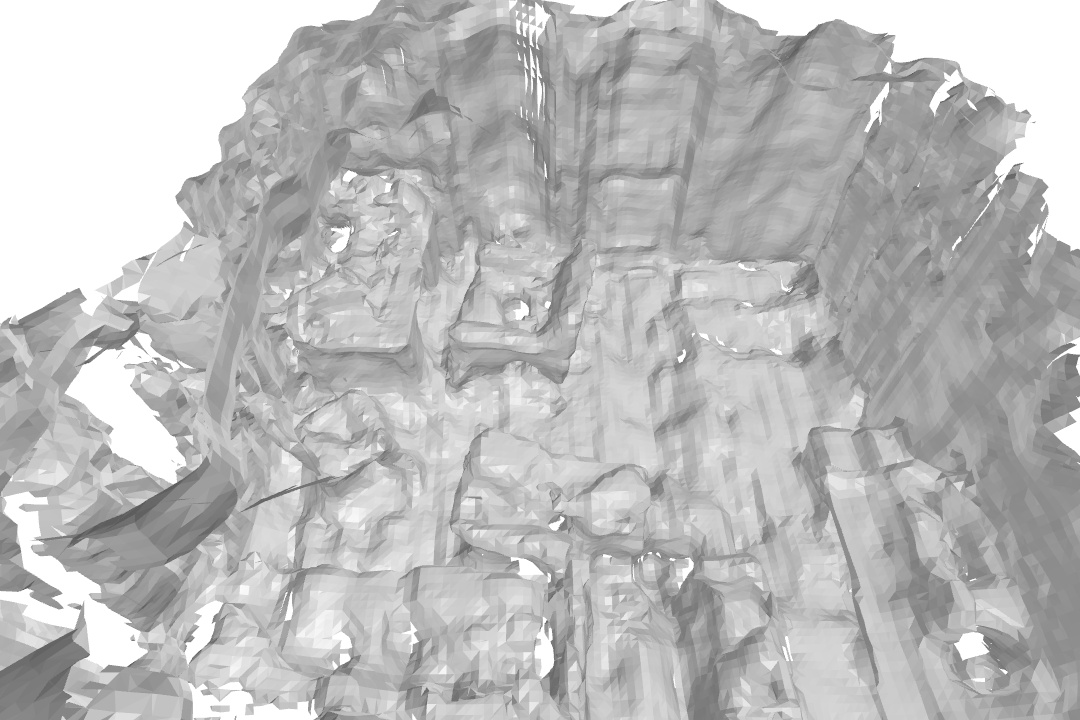}&
            \includegraphics[width=\mywidth]{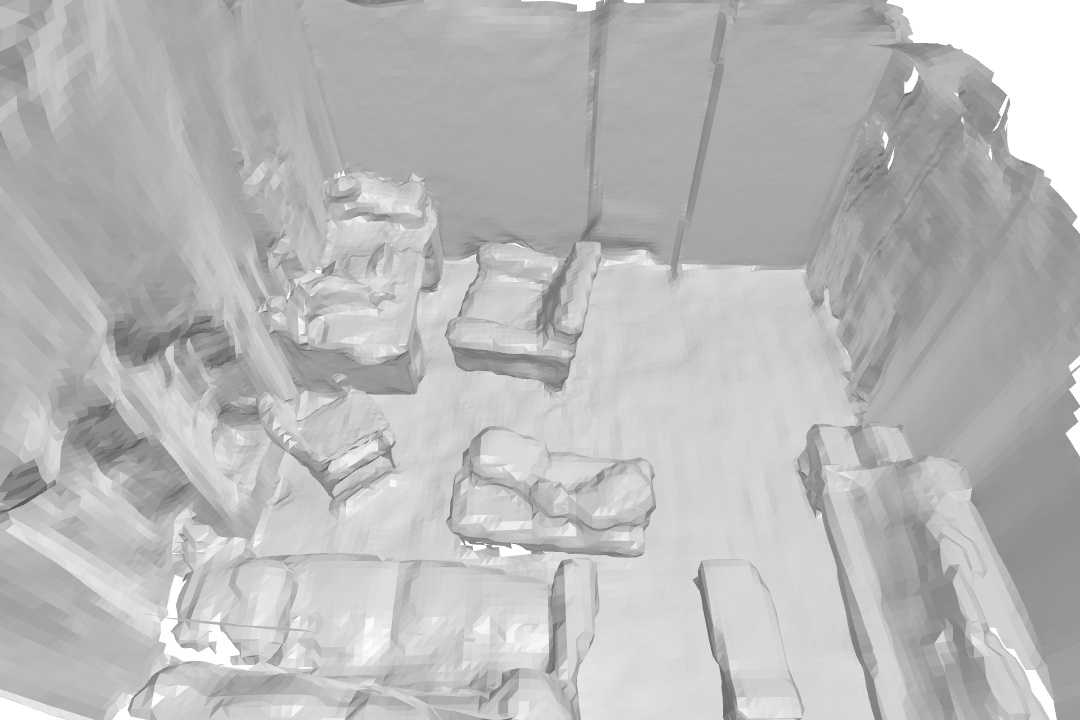}&
            \includegraphics[width=\mywidth]{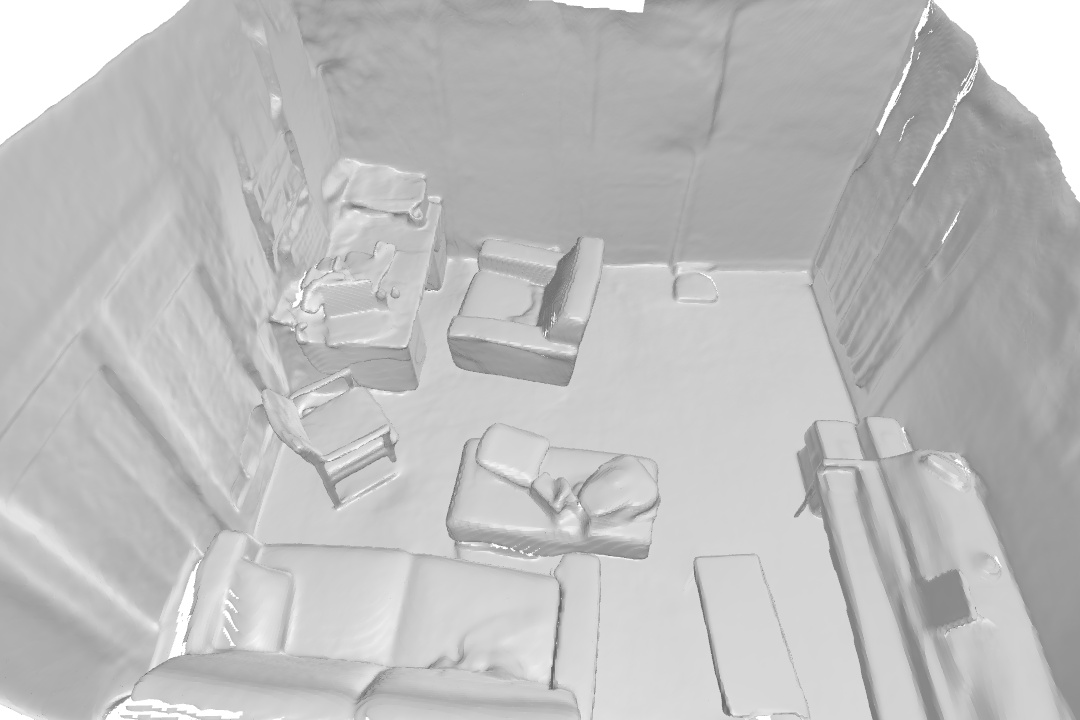}&
            \includegraphics[width=\mywidth]{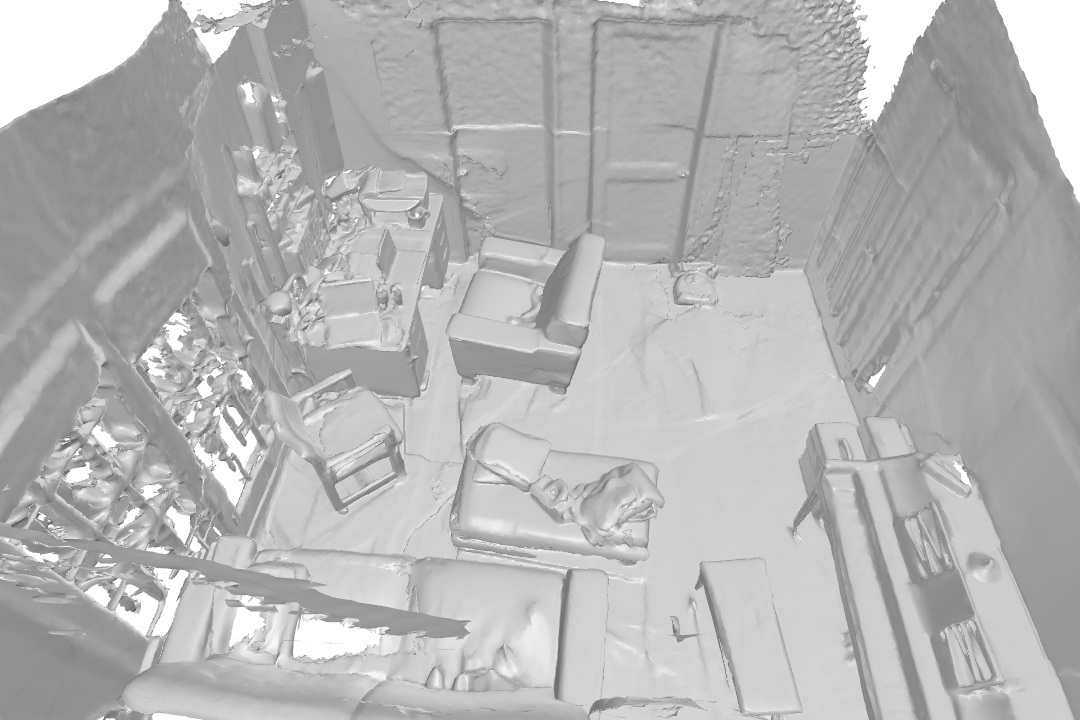}\\
            
            \includegraphics[width=\mywidth]{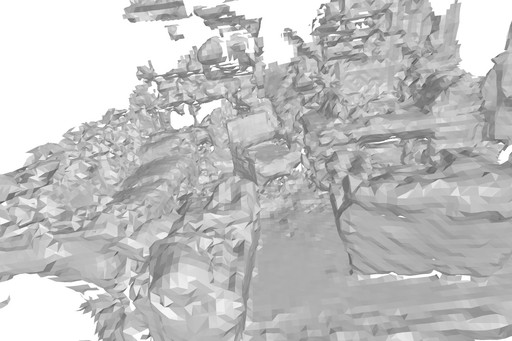}&
            \includegraphics[width=\mywidth]{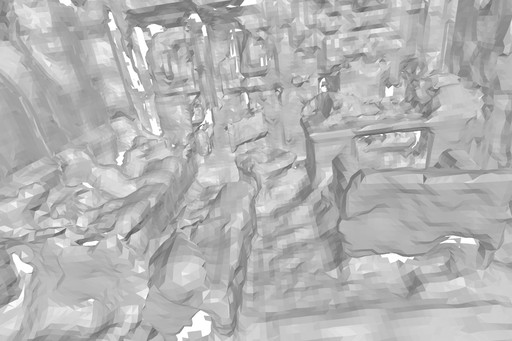}&
            \includegraphics[width=\mywidth]{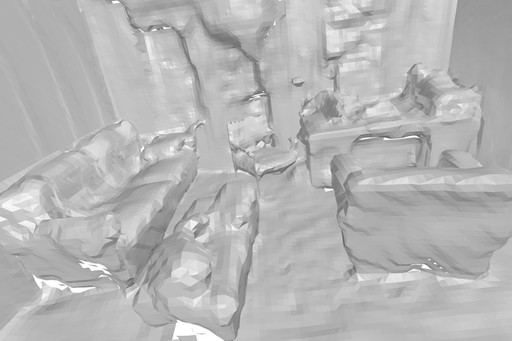}&
            \includegraphics[width=\mywidth]{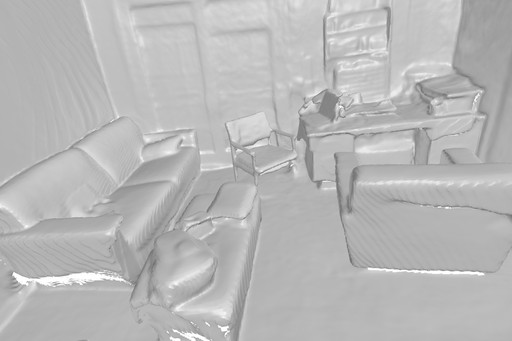}&
            \includegraphics[width=\mywidth]{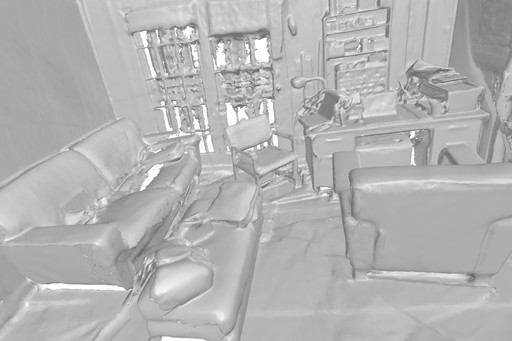}\\
            
            \includegraphics[width=\mywidth]{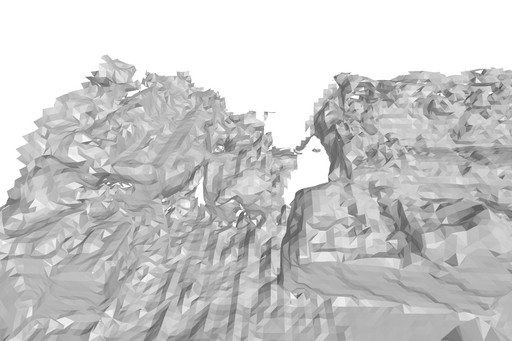}&            \includegraphics[width=\mywidth]{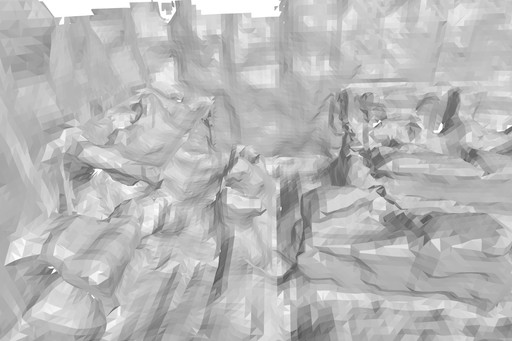}&
            \includegraphics[width=\mywidth]{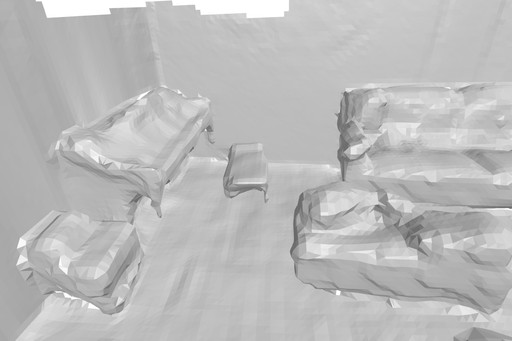}&
            \includegraphics[width=\mywidth]{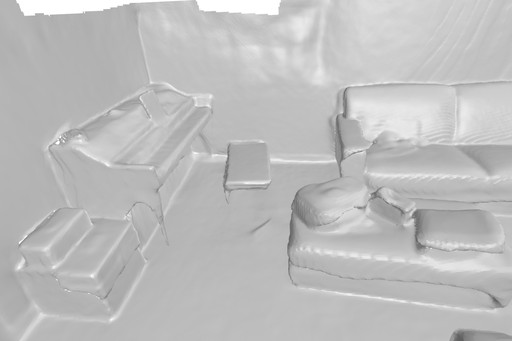}&
            \includegraphics[width=\mywidth]{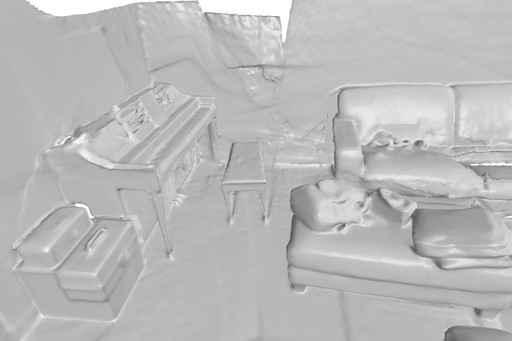}\\

            \includegraphics[width=\mywidth]{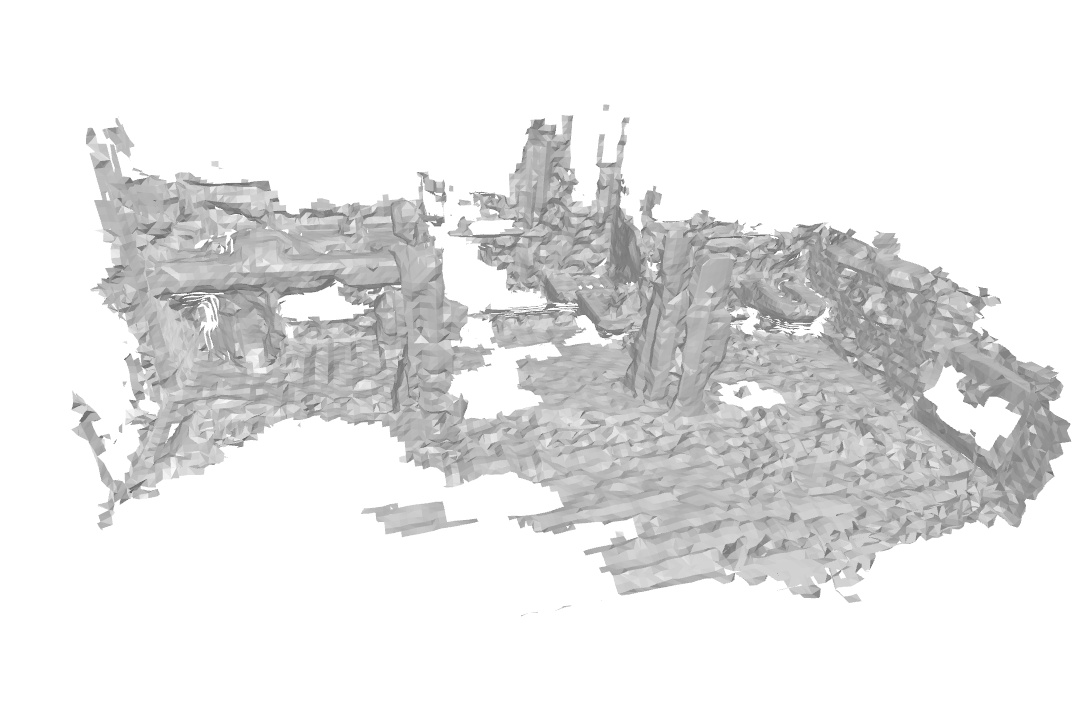}&
            \includegraphics[width=\mywidth]{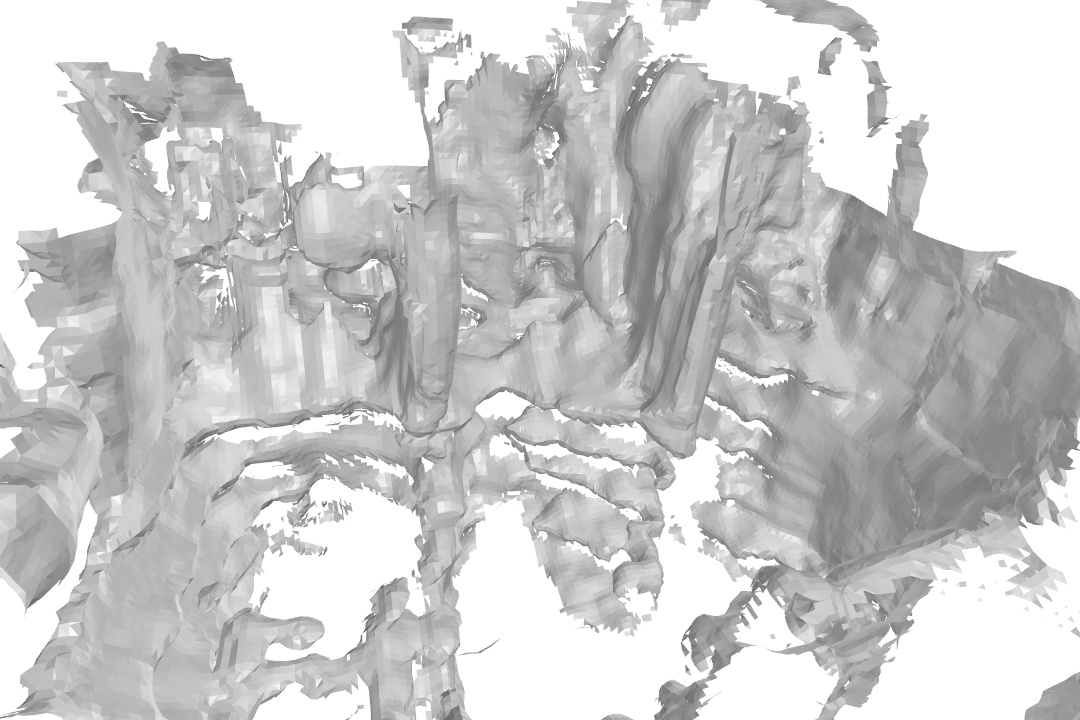}&
            \includegraphics[width=\mywidth]{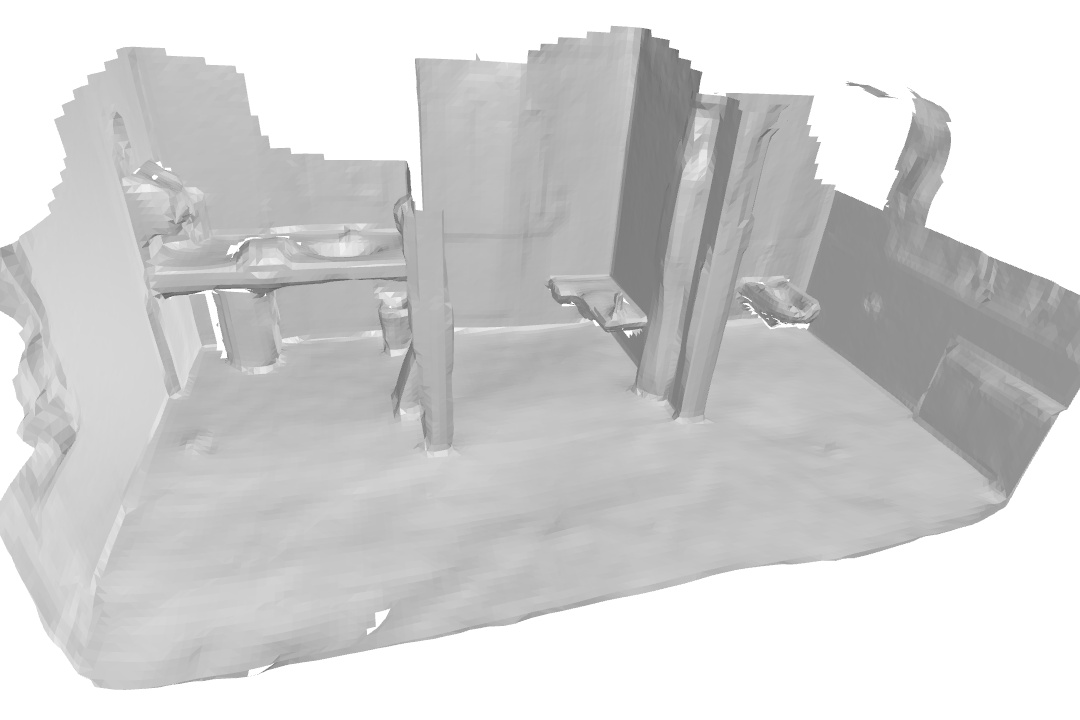}&
            \includegraphics[width=\mywidth]{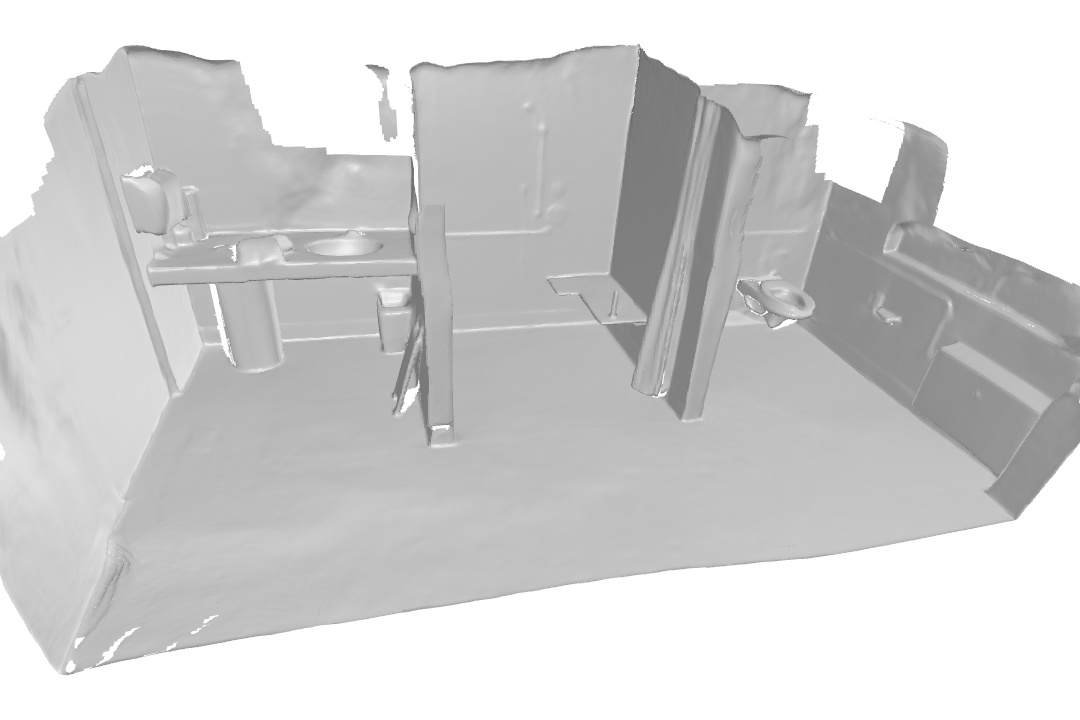}&
            \includegraphics[width=\mywidth]{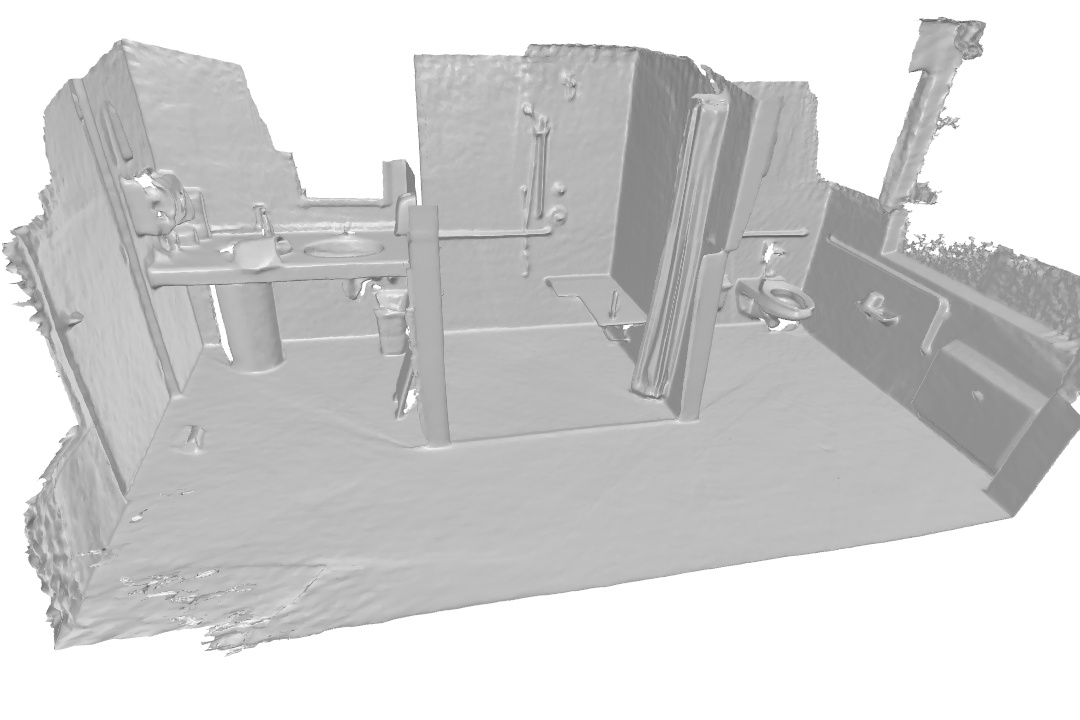}\\
            
            \includegraphics[width=\mywidth]{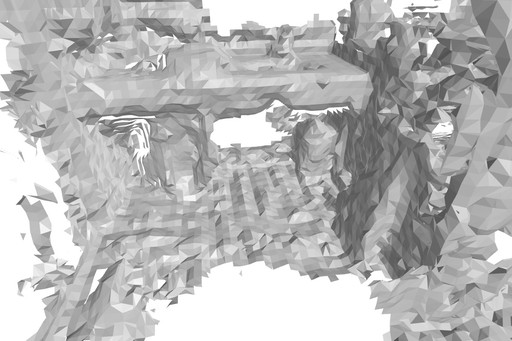}&
            \includegraphics[width=\mywidth]{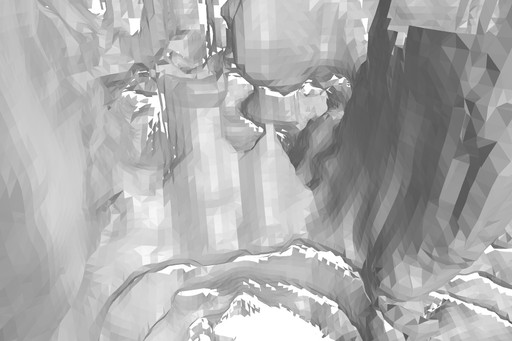}&
            \includegraphics[width=\mywidth]{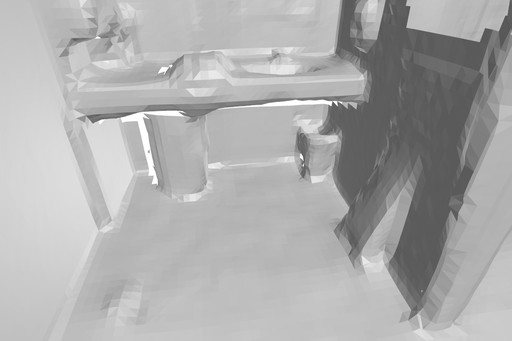}&
            \includegraphics[width=\mywidth]{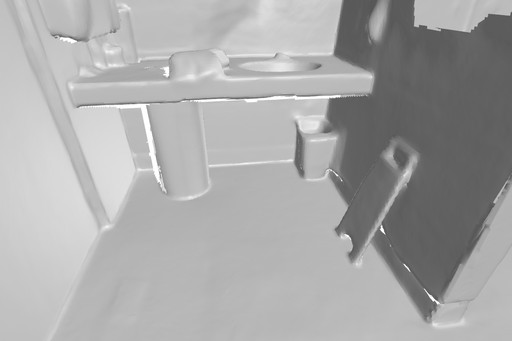}&
            \includegraphics[width=\mywidth]{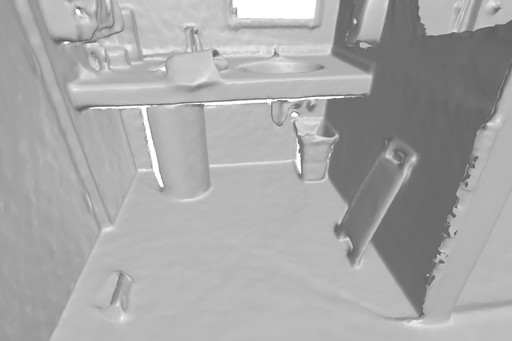}\\
            
            \includegraphics[width=\mywidth]{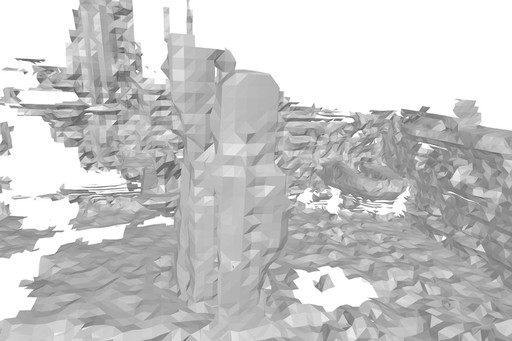}&
            \includegraphics[width=\mywidth]{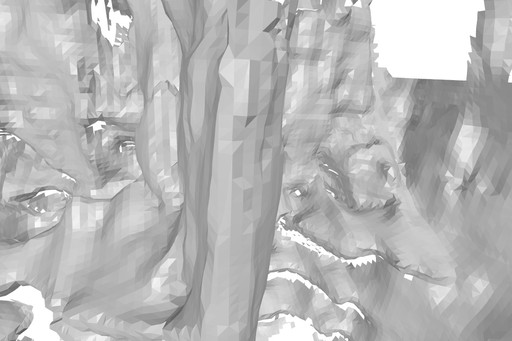}&
            \includegraphics[width=\mywidth]{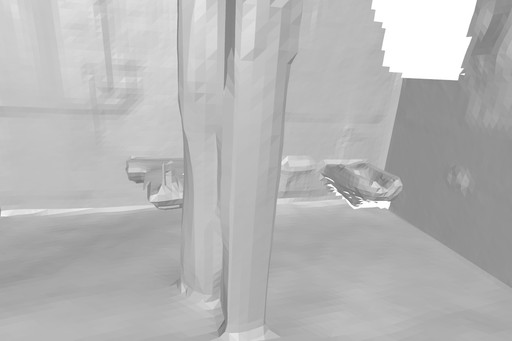}&
            \includegraphics[width=\mywidth]{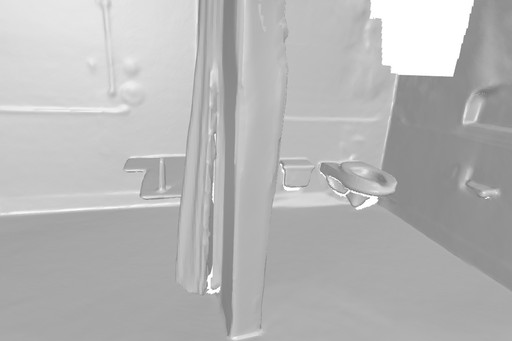}&
            \includegraphics[width=\mywidth]{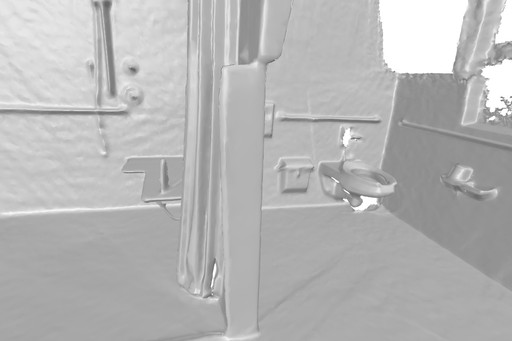}\\
            
            \includegraphics[width=\mywidth]{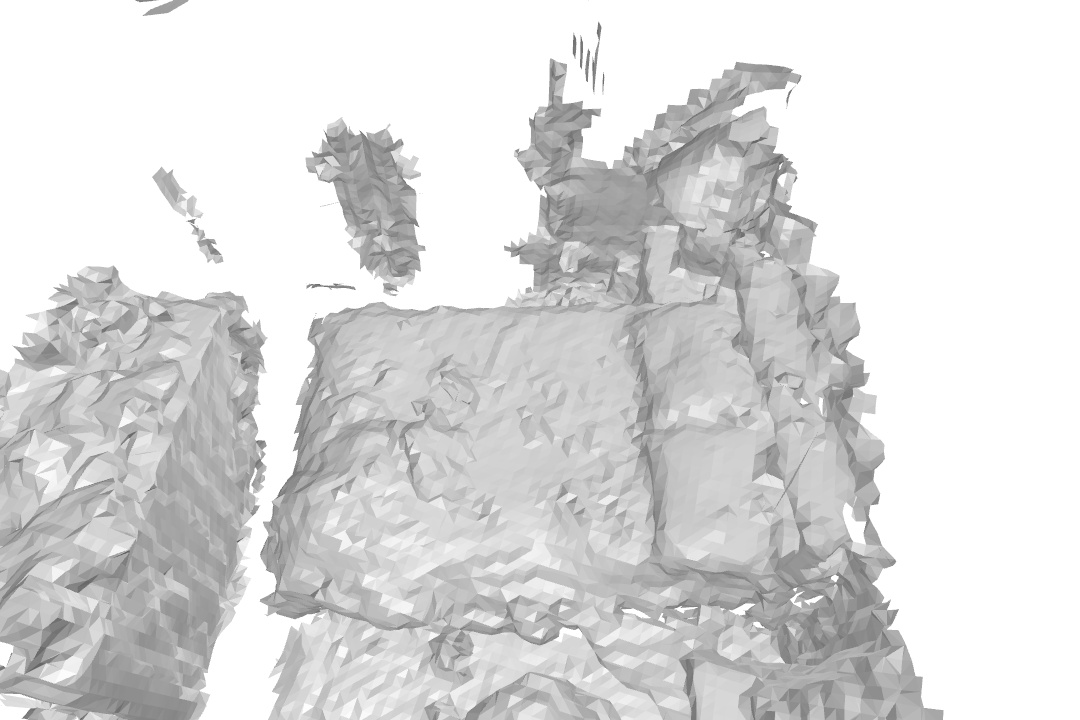}&
            \includegraphics[width=\mywidth]{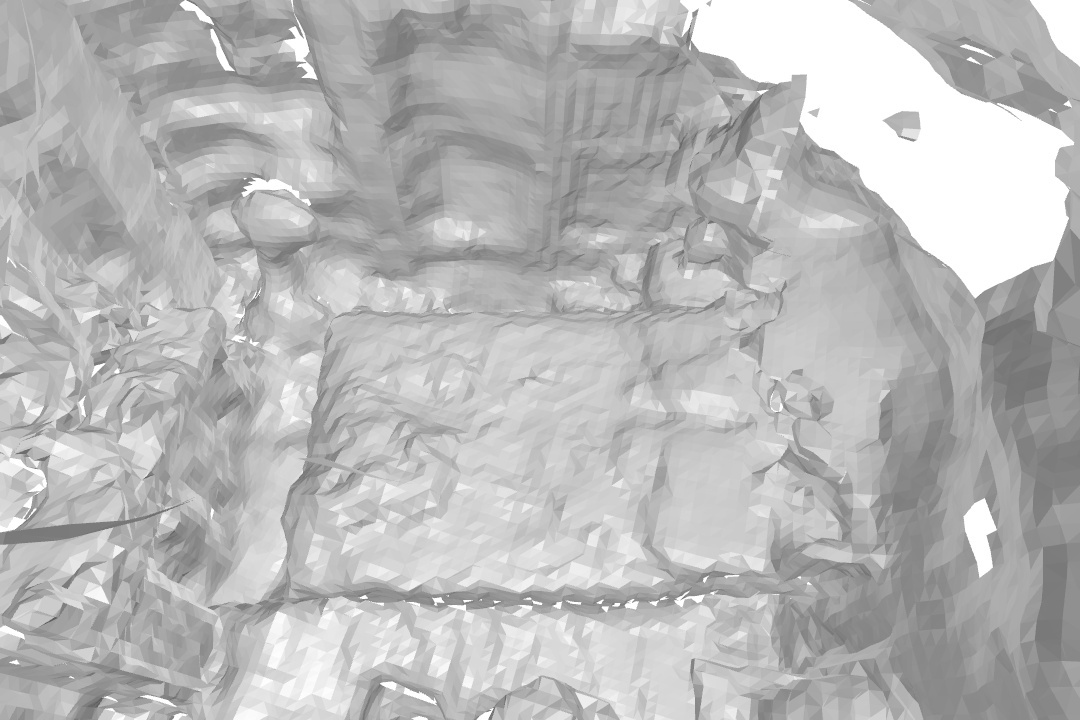}&
            \includegraphics[width=\mywidth]{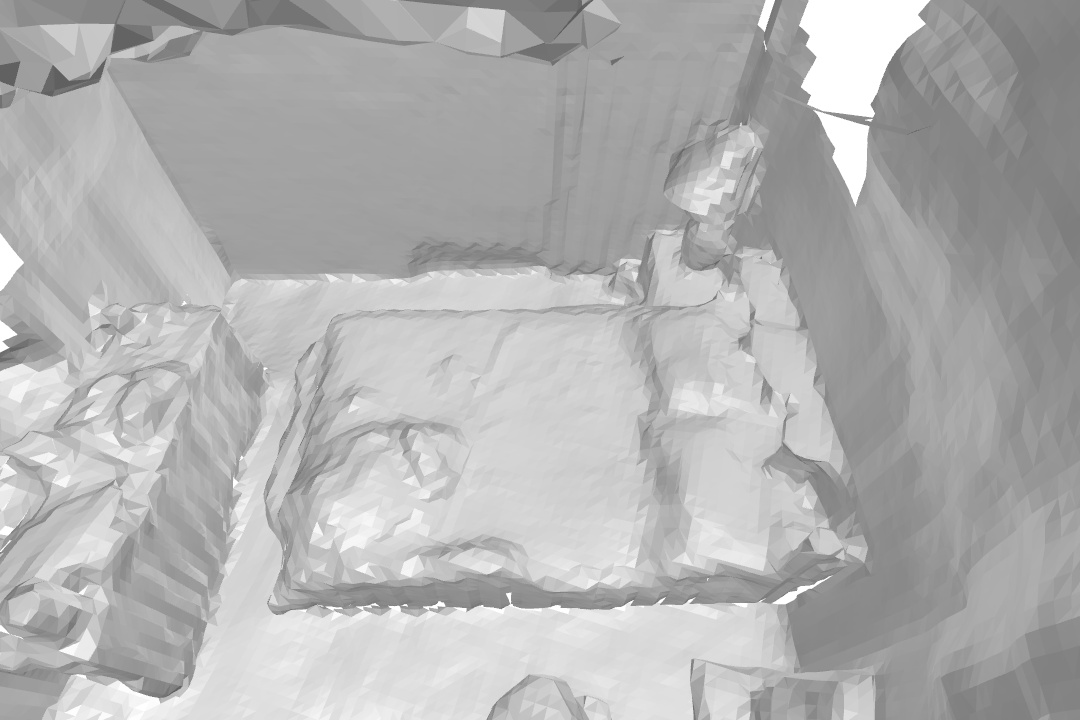}&
            \includegraphics[width=\mywidth]{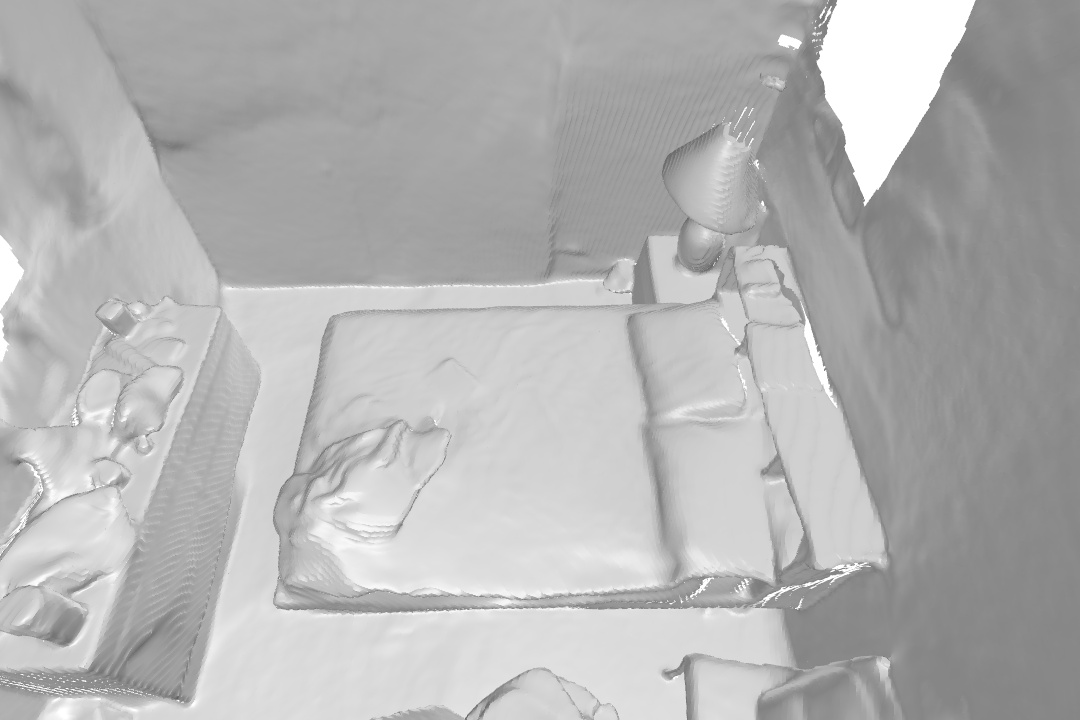}&
            \includegraphics[width=\mywidth]{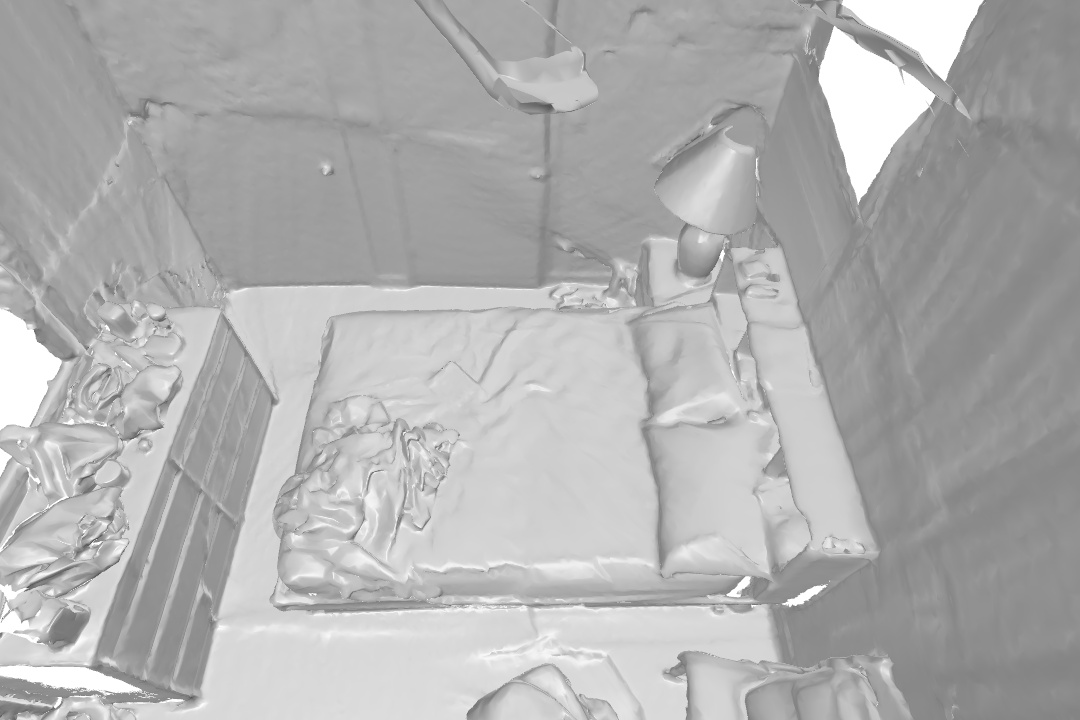}\\
    
            \includegraphics[width=\mywidth]{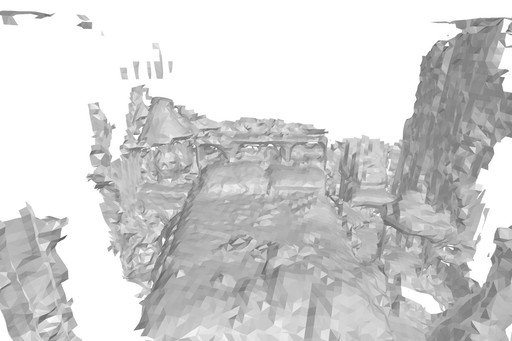}&
            \includegraphics[width=\mywidth]{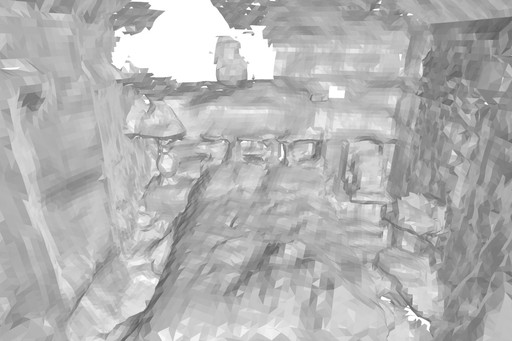}&
            \includegraphics[width=\mywidth]{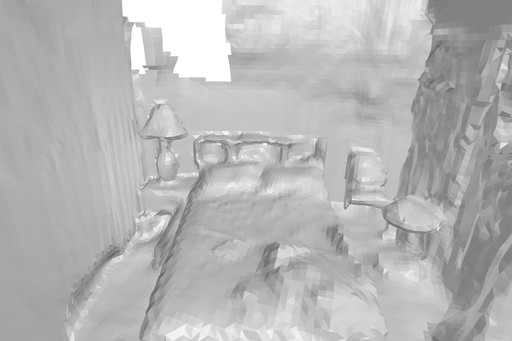}&
            \includegraphics[width=\mywidth]{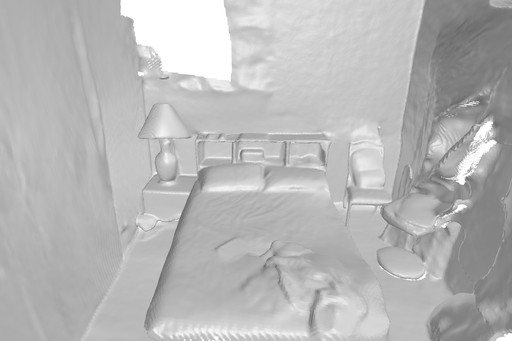}&
            \includegraphics[width=\mywidth]{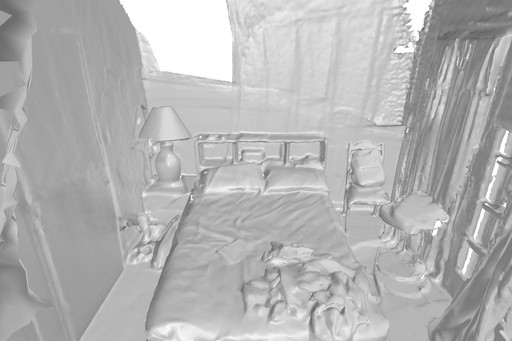}\\
    
            \includegraphics[width=\mywidth]{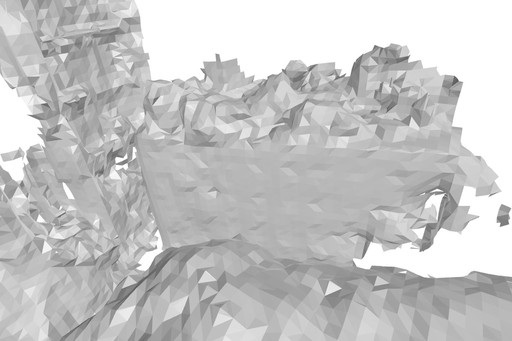}&
            \includegraphics[width=\mywidth]{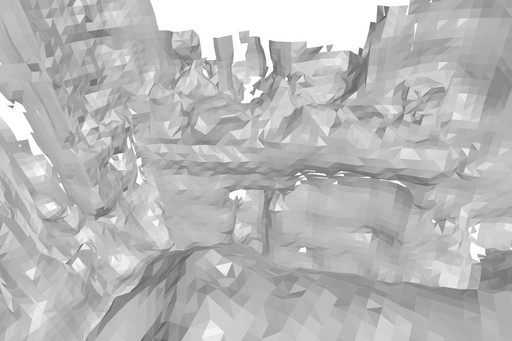}&
            \includegraphics[width=\mywidth]{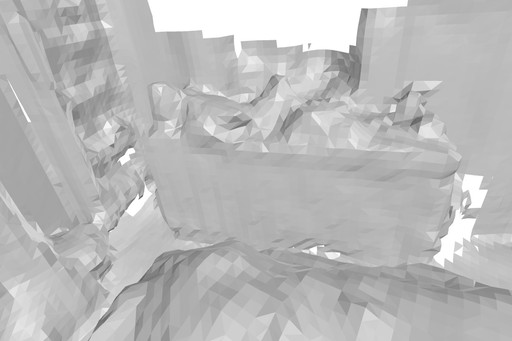}&
            \includegraphics[width=\mywidth]{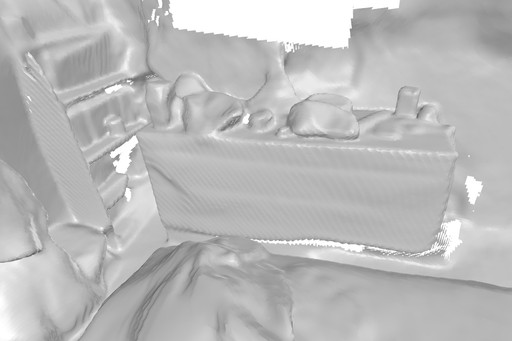}&
            \includegraphics[width=\mywidth]{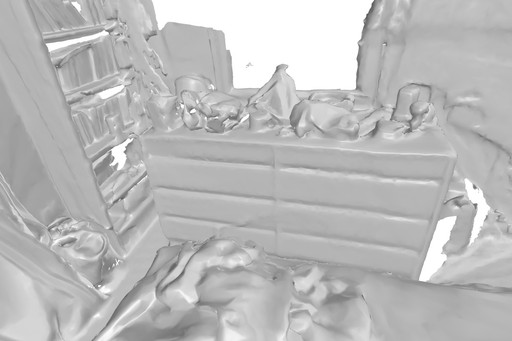}\\
    
            \includegraphics[width=\mywidth]{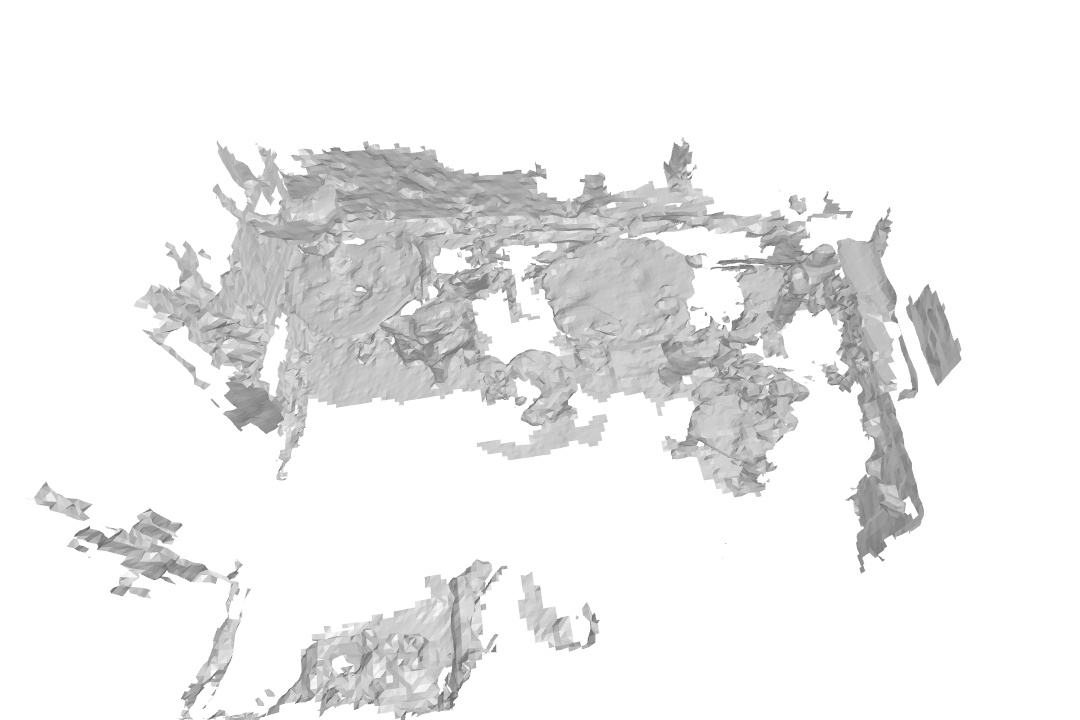}&
            \includegraphics[width=\mywidth]{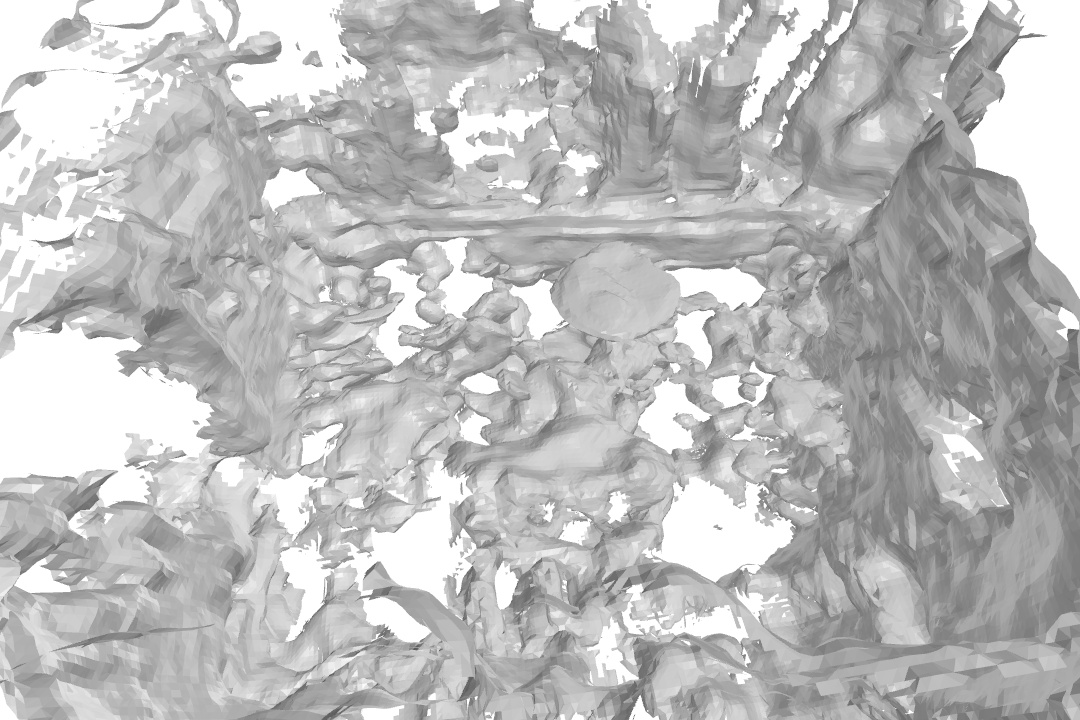}&
            \includegraphics[width=\mywidth]{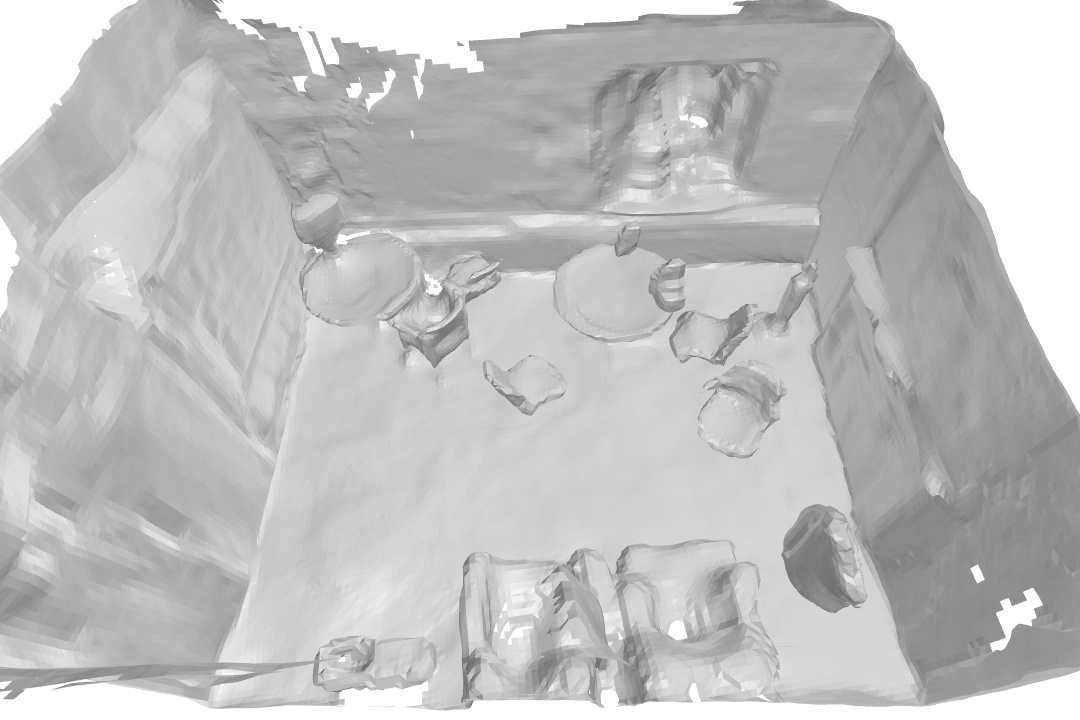}&
            \includegraphics[width=\mywidth]{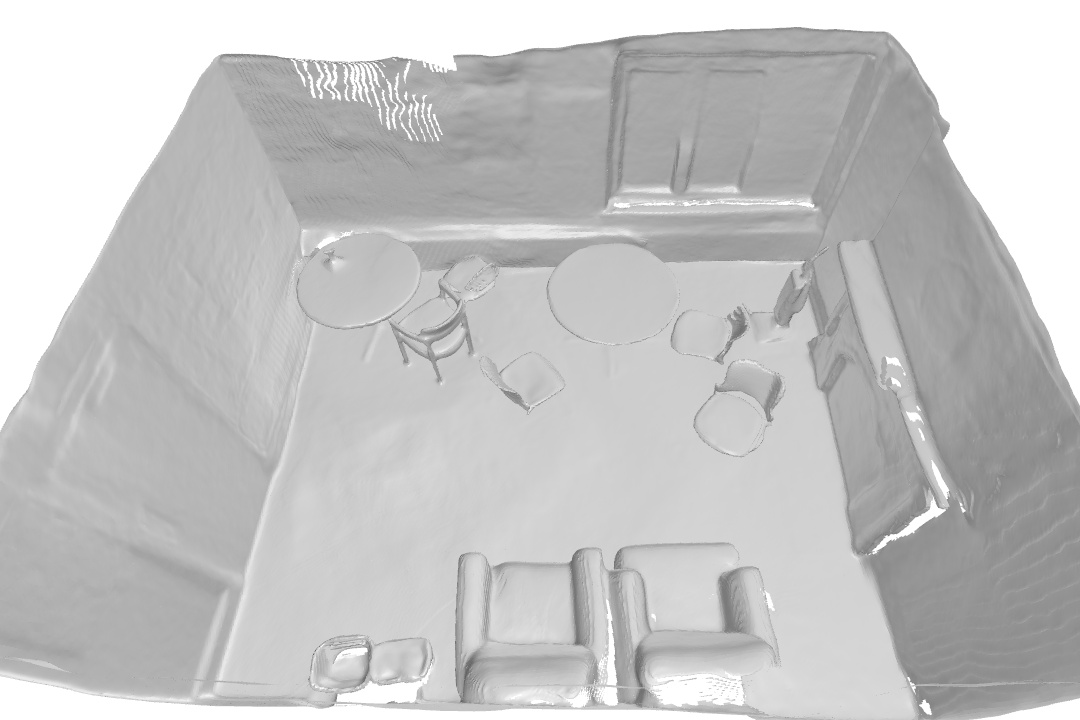}&
            \includegraphics[width=\mywidth]{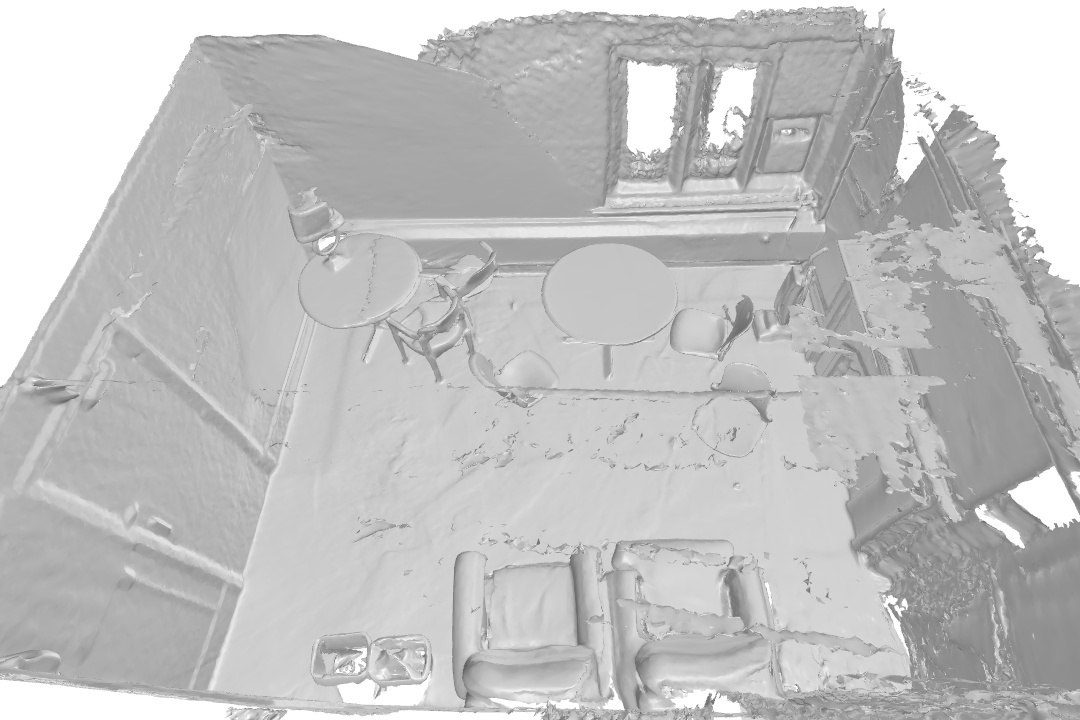}\\

            \includegraphics[width=\mywidth]{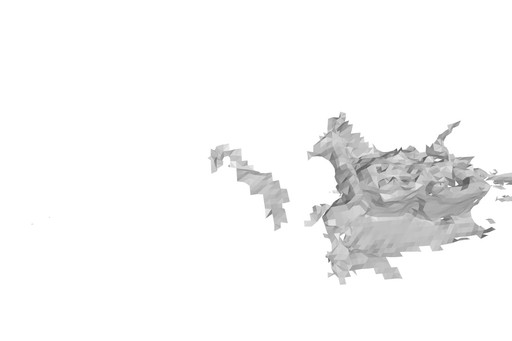}&
            \includegraphics[width=\mywidth]{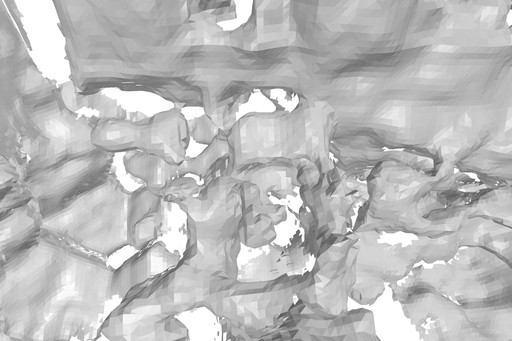}&
            \includegraphics[width=\mywidth]{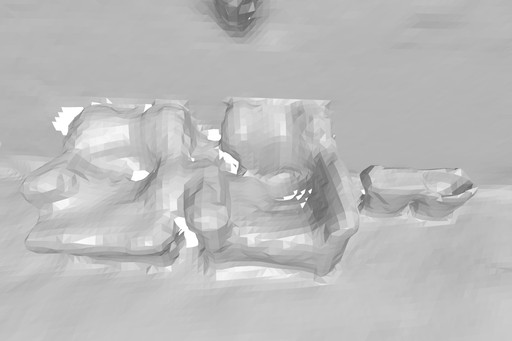}&
            \includegraphics[width=\mywidth]{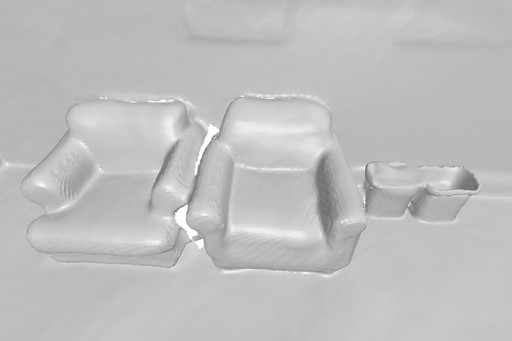}&
            \includegraphics[width=\mywidth]{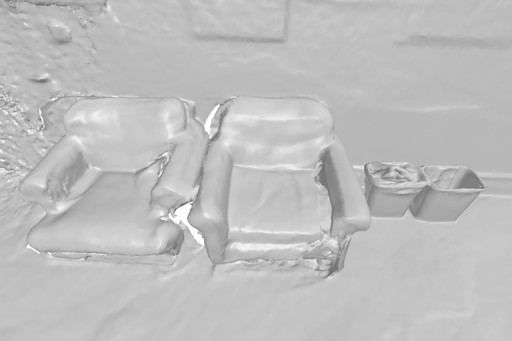}\\
            
             COLMAP \cite{Schoenberger2016ECCV}&
             VolSDF \cite{Mildenhall2020ECCV} &
             Manhattan-SDF \cite{guo2022manhattan}&
             \textbf{Ours} (MLP) & Ground Truth\\
        \end{tabular}
        \caption{
        \textbf{Qualitative Comparison on ScanNet.} We show different views for each scene. Our method leads to better results containing smooth surfaces and detailed reconstructions compared against state-of-the-art neural implicit methods.
        }
    \label{fig:scannet}
    \end{figure*}
}

\newcommand{\figureablationcuesjpg}{
\begin{figure*}[t]
        \centering
        \setlength{\tabcolsep}{0.1em}
        \renewcommand{\arraystretch}{0.7}
        \hfill{}\hspace*{-0.5em}
        \begin{tabular}{cccc}
            \includegraphics[width=\rwidth]{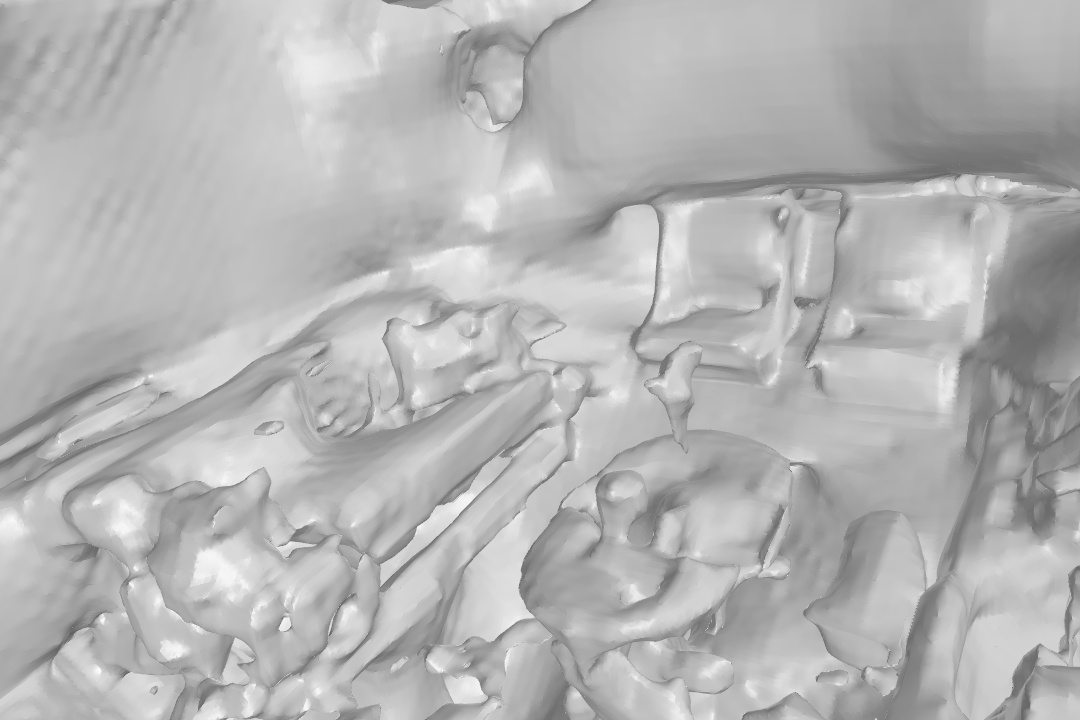}&
            \includegraphics[width=\rwidth]{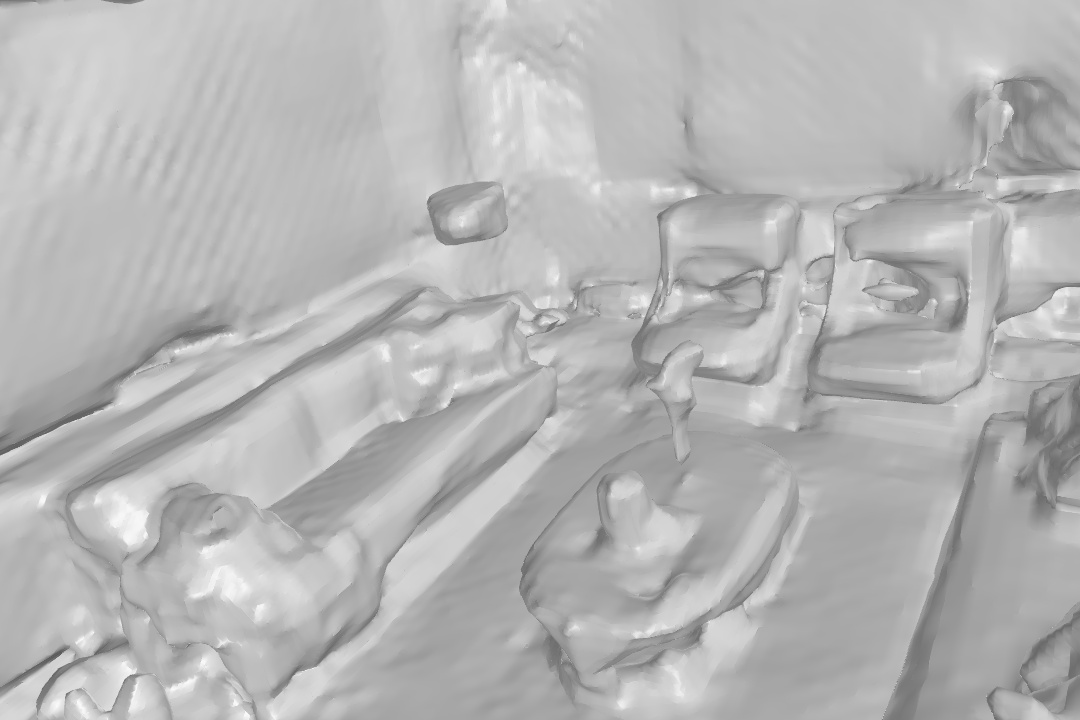}&
            \includegraphics[width=\rwidth]{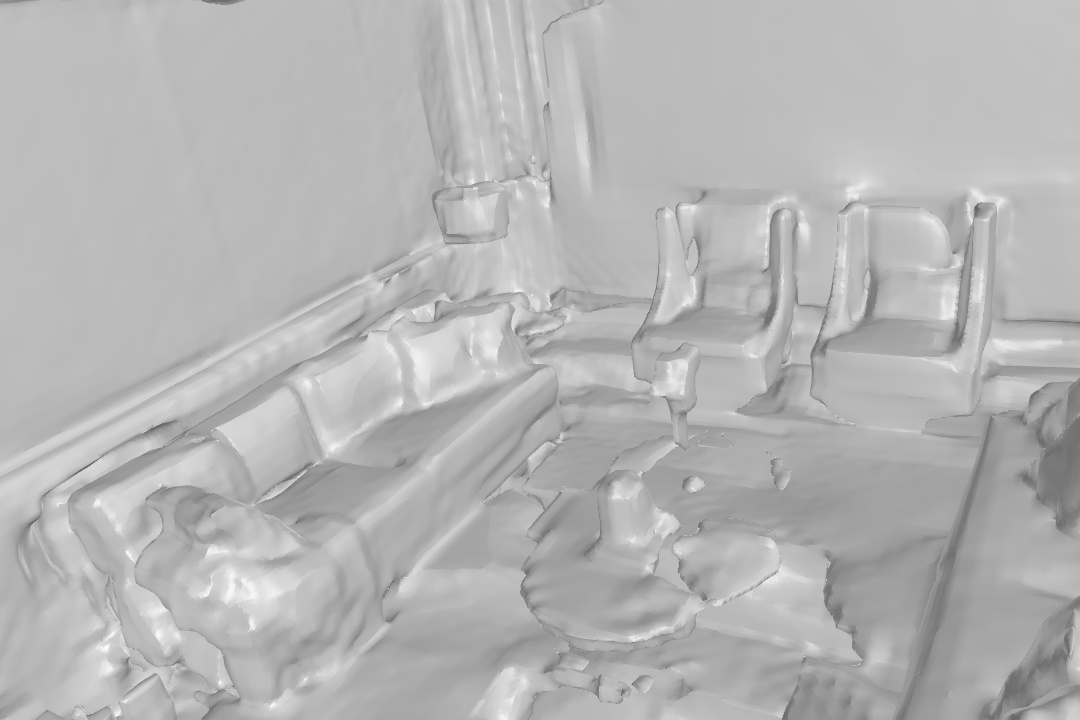}&
            \includegraphics[width=\rwidth]{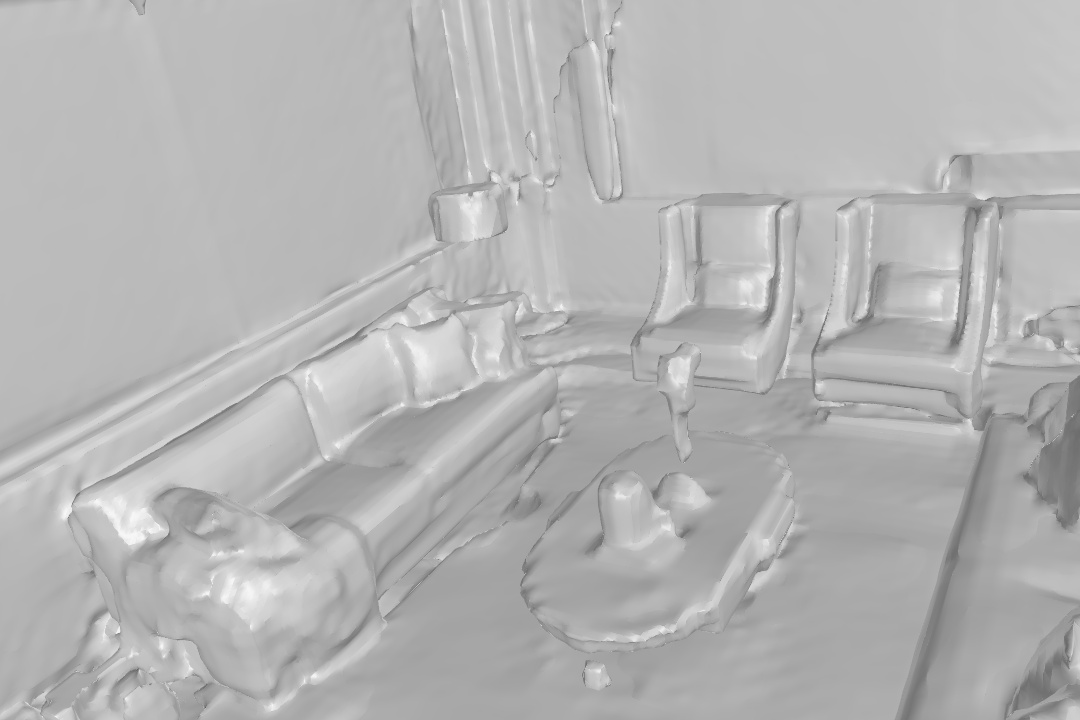}\\
            \includegraphics[width=\rwidth]{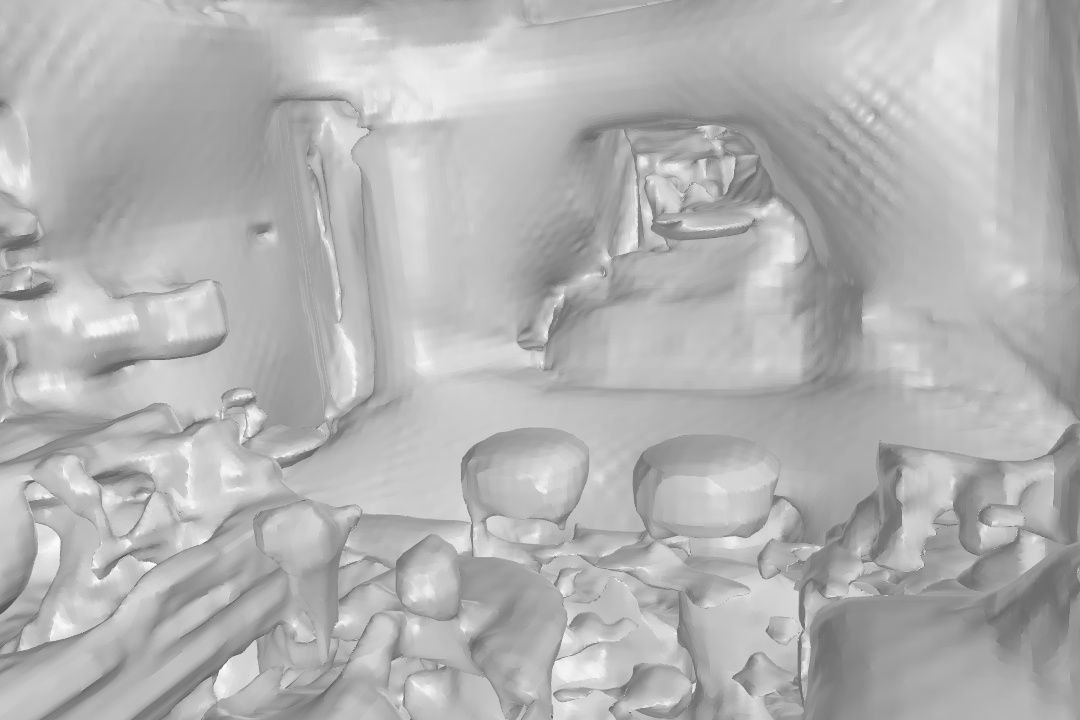}&
            \includegraphics[width=\rwidth]{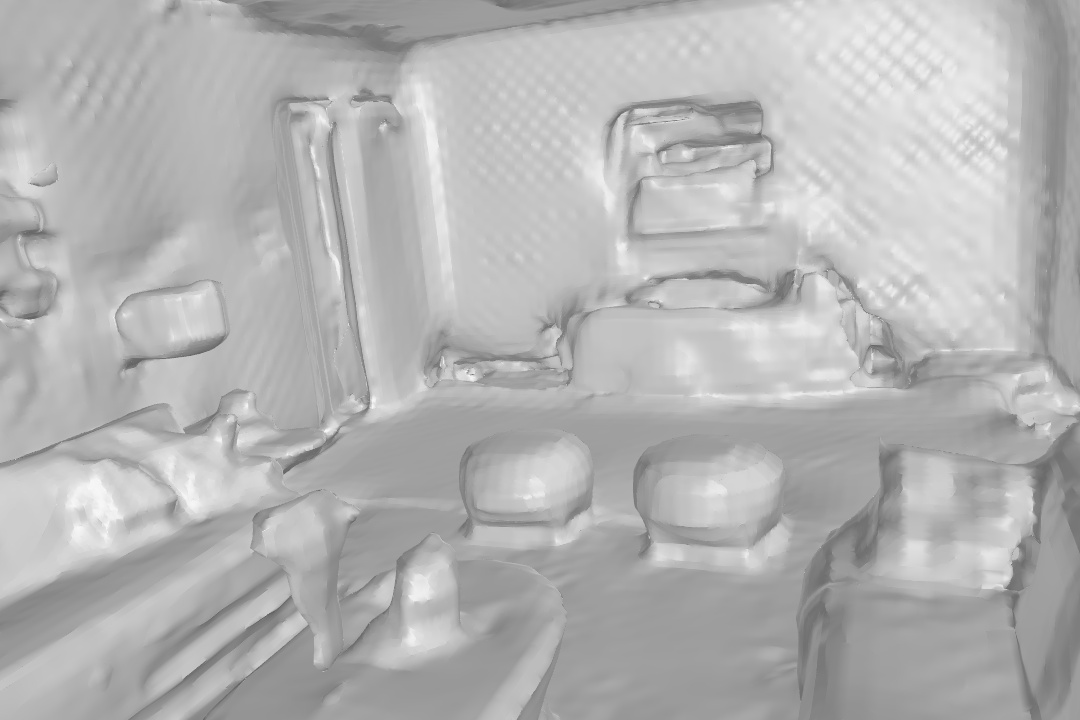}&
            \includegraphics[width=\rwidth]{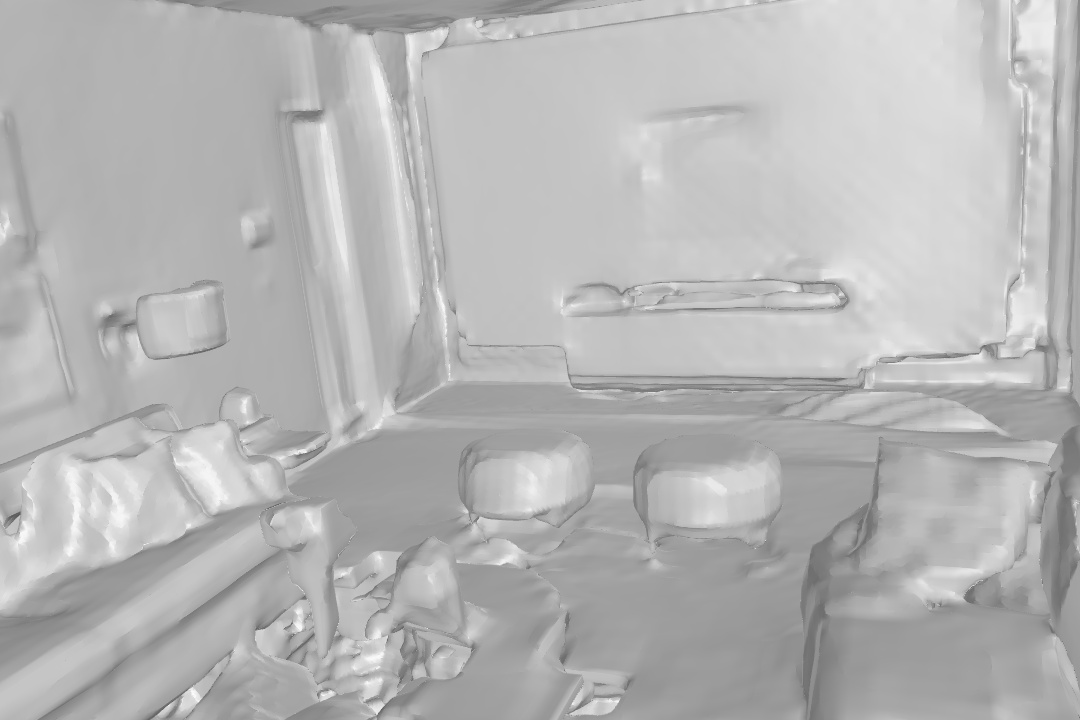}&
            \includegraphics[width=\rwidth]{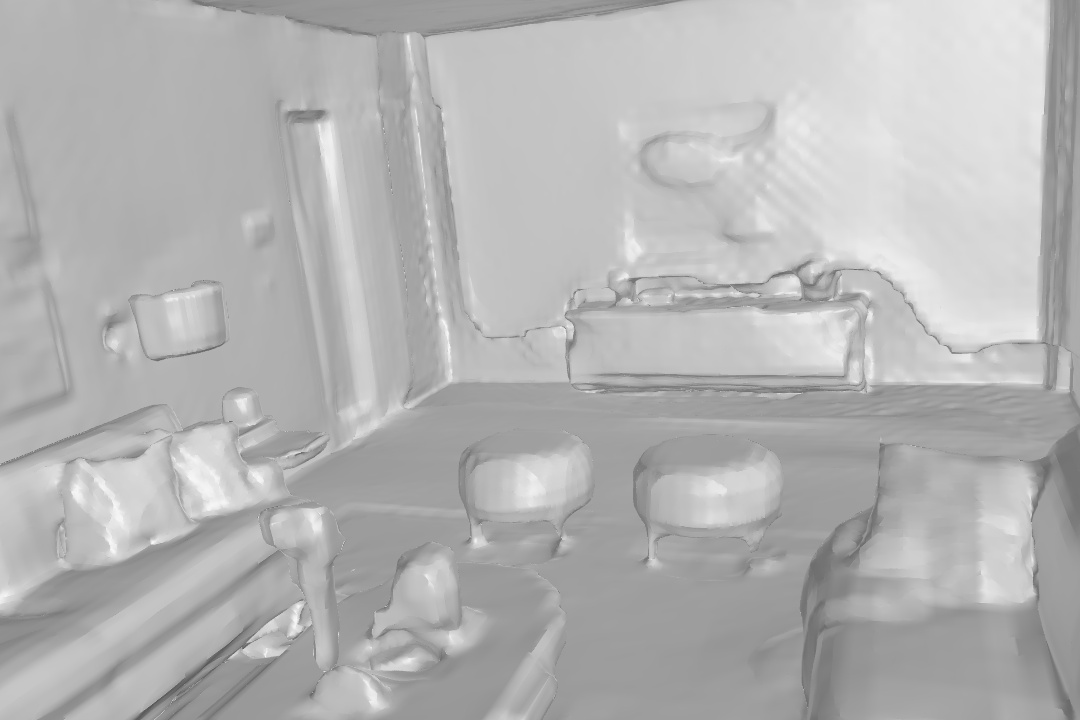}\\
             Dense SDF Grid &
             + Depth  &
             + Normal&
             + Both\\
            \includegraphics[width=\rwidth]{gfx/ablation_cues/room1_mlp/room1_mlp_no_cues.ply.jpg}&
            \includegraphics[width=\rwidth]{gfx/ablation_cues/room1_mlp/room1_mlp_depth.ply.jpg}&
            \includegraphics[width=\rwidth]{gfx/ablation_cues/room1_mlp/room1_mlp_normal.ply.jpg}&
            \includegraphics[width=\rwidth]{gfx/ablation_cues/room1_mlp/room1_mlp_both_cues.ply.jpg}\\
            \includegraphics[width=\rwidth]{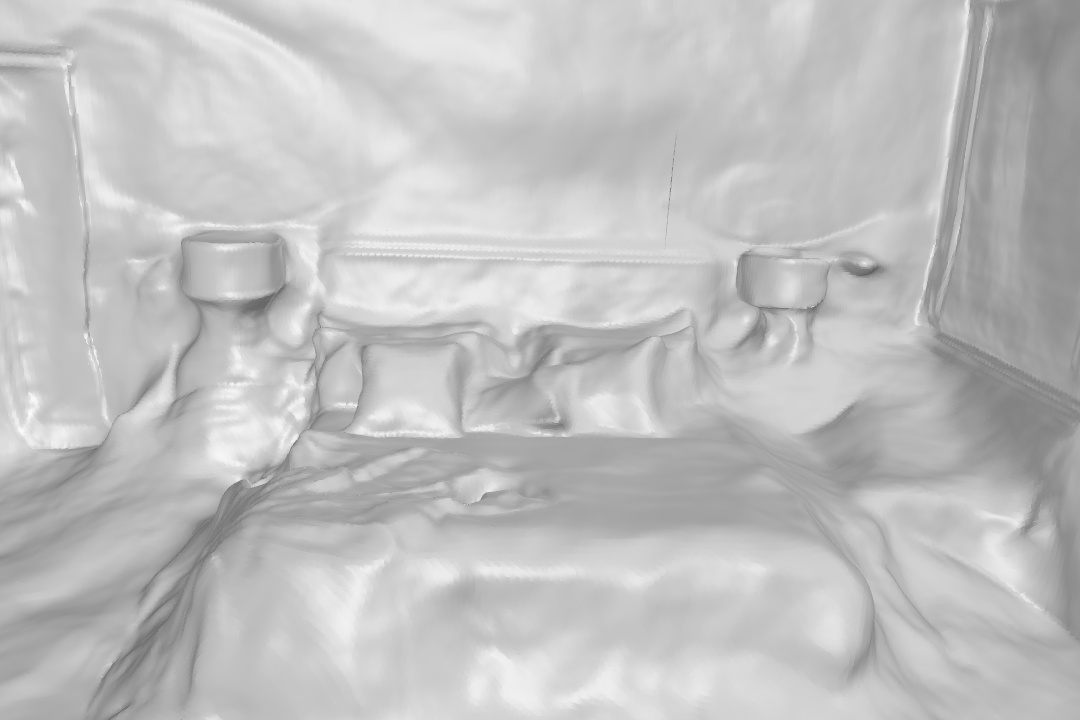}&
            \includegraphics[width=\rwidth]{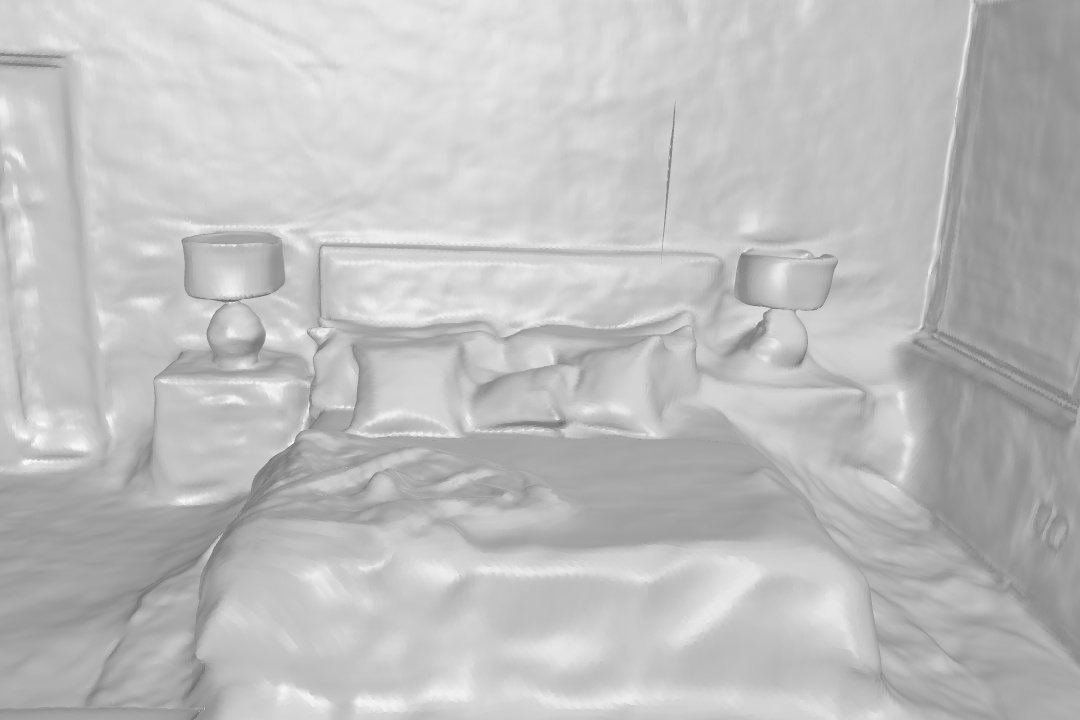}&
            \includegraphics[width=\rwidth]{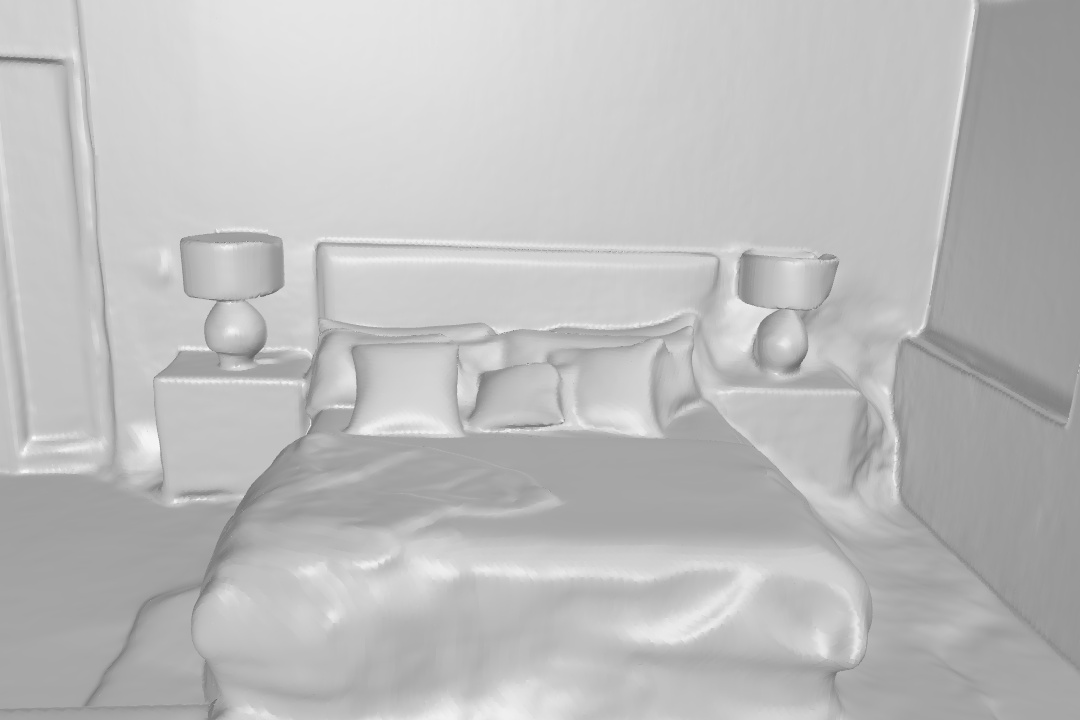}&
            \includegraphics[width=\rwidth]{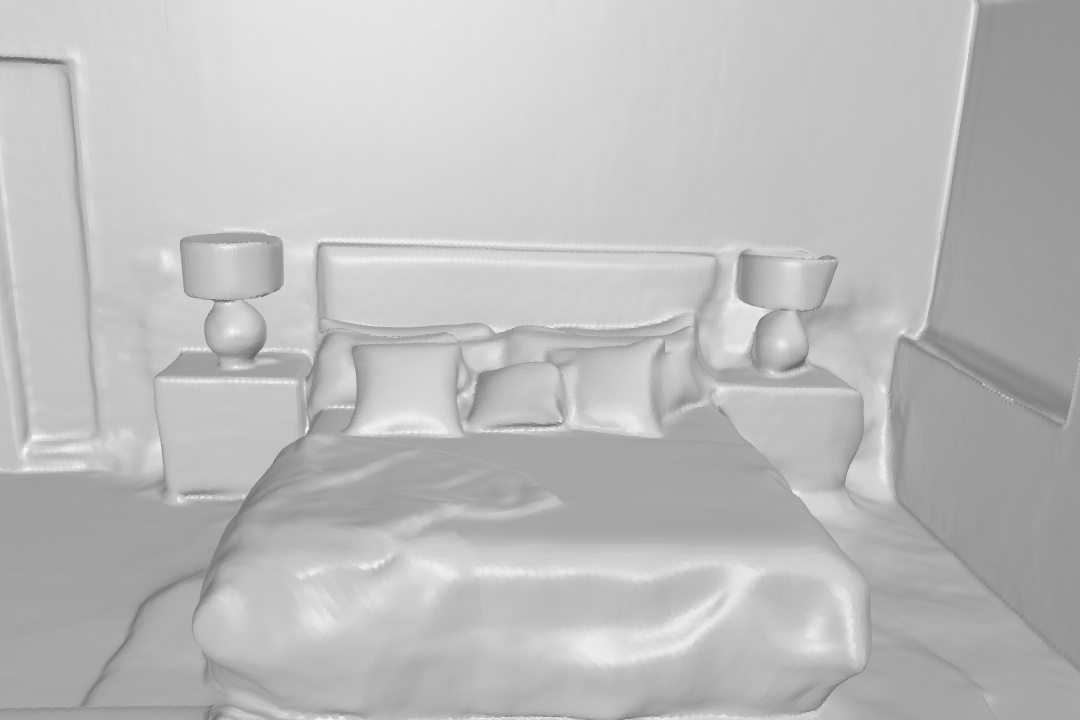}\\
             MLP &
             + Depth  &
             + Normal&
             + Both\\
            \includegraphics[width=\rwidth]{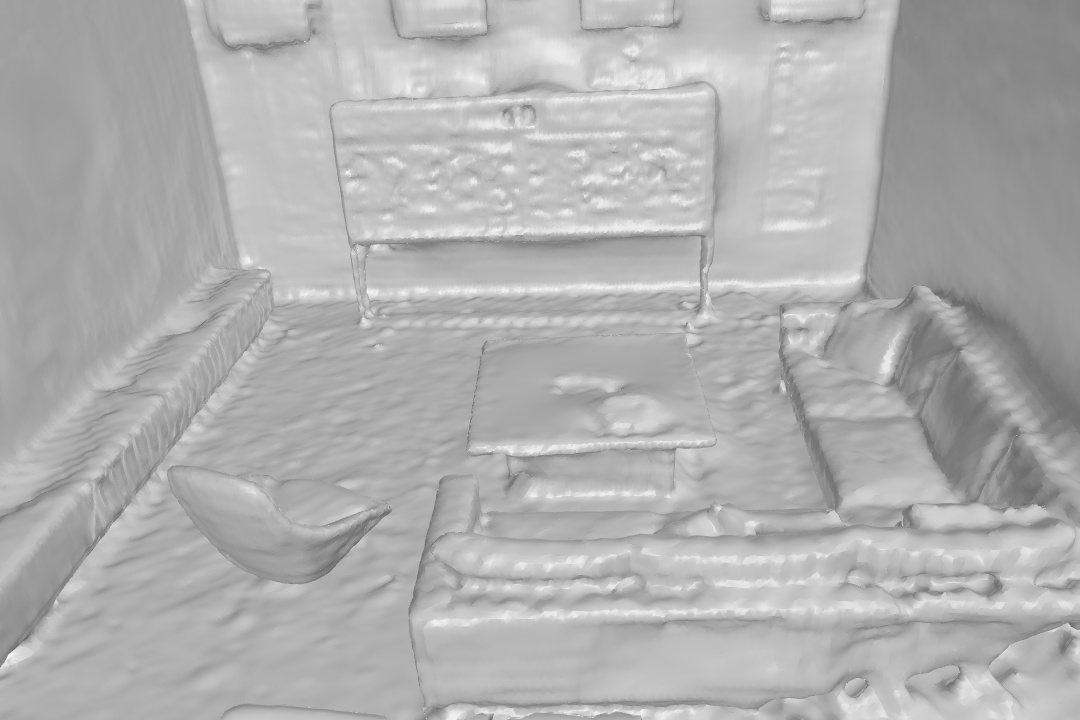}&
            \includegraphics[width=\rwidth]{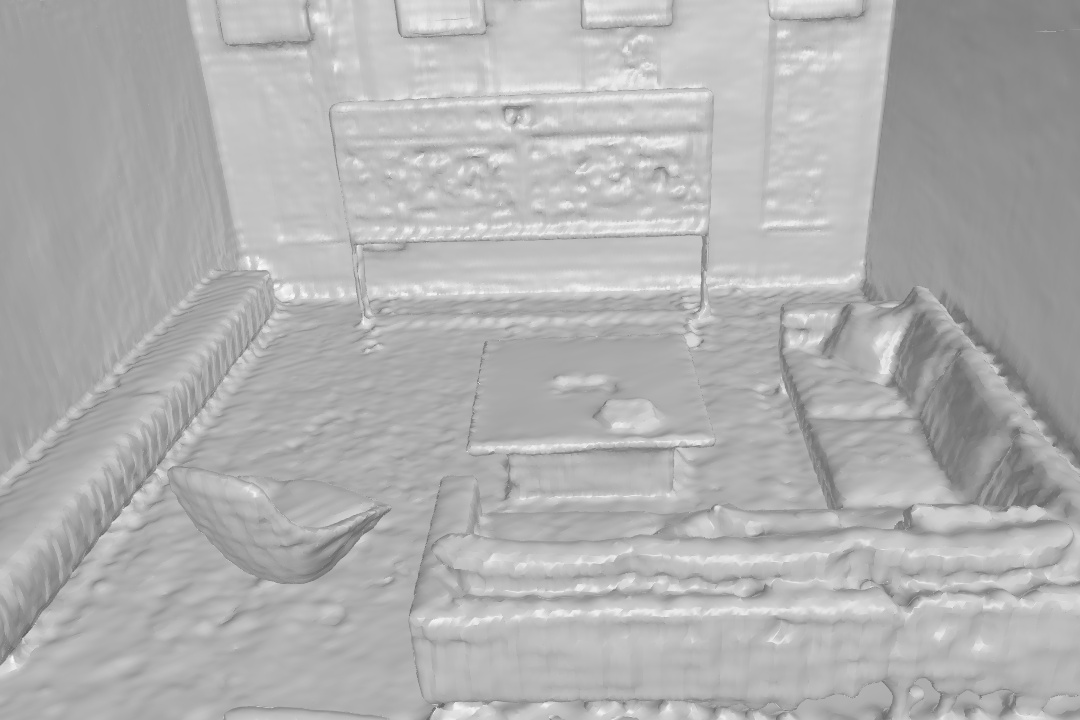}&
            \includegraphics[width=\rwidth]{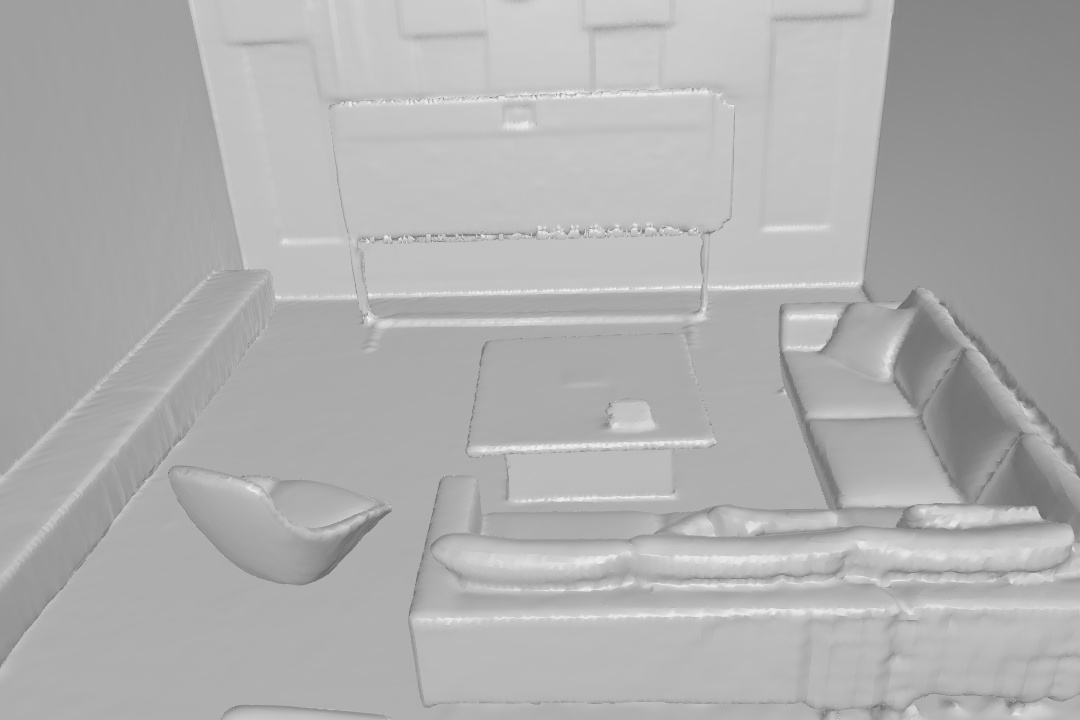}&
            \includegraphics[width=\rwidth]{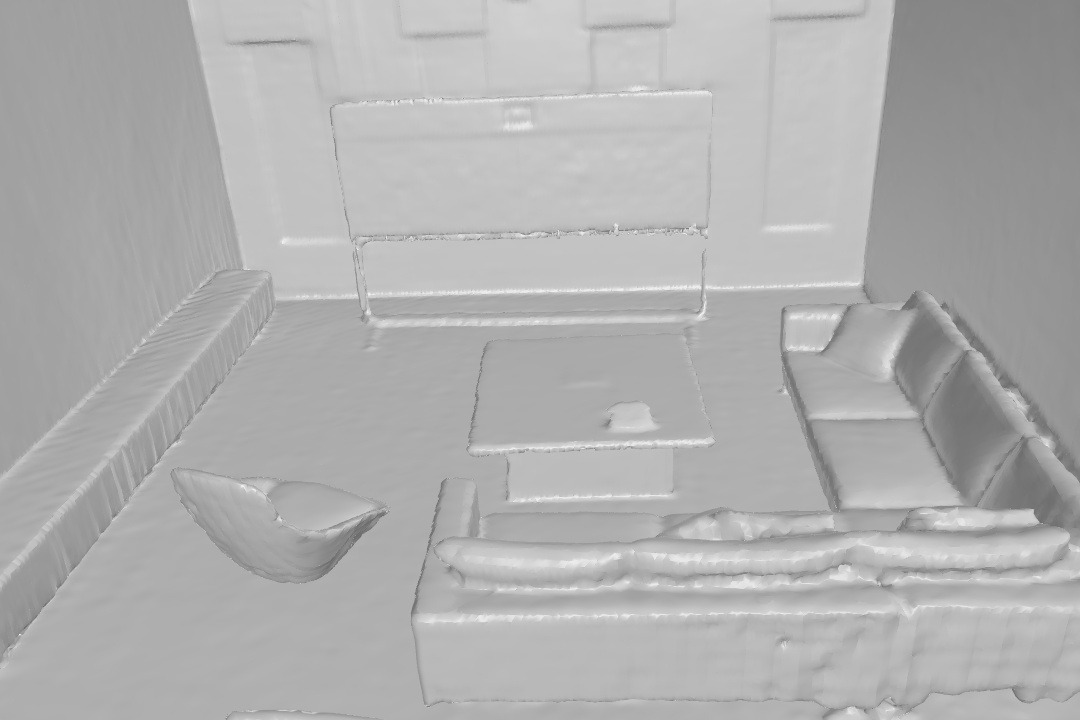}\\
            \includegraphics[width=\rwidth]{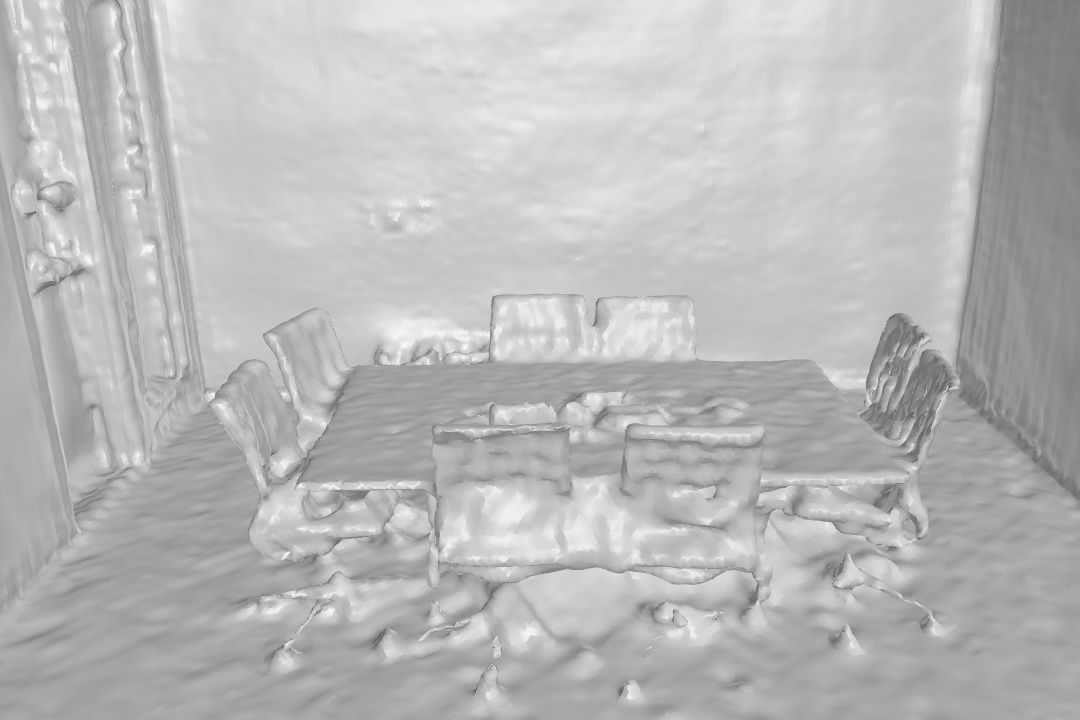}&
            \includegraphics[width=\rwidth]{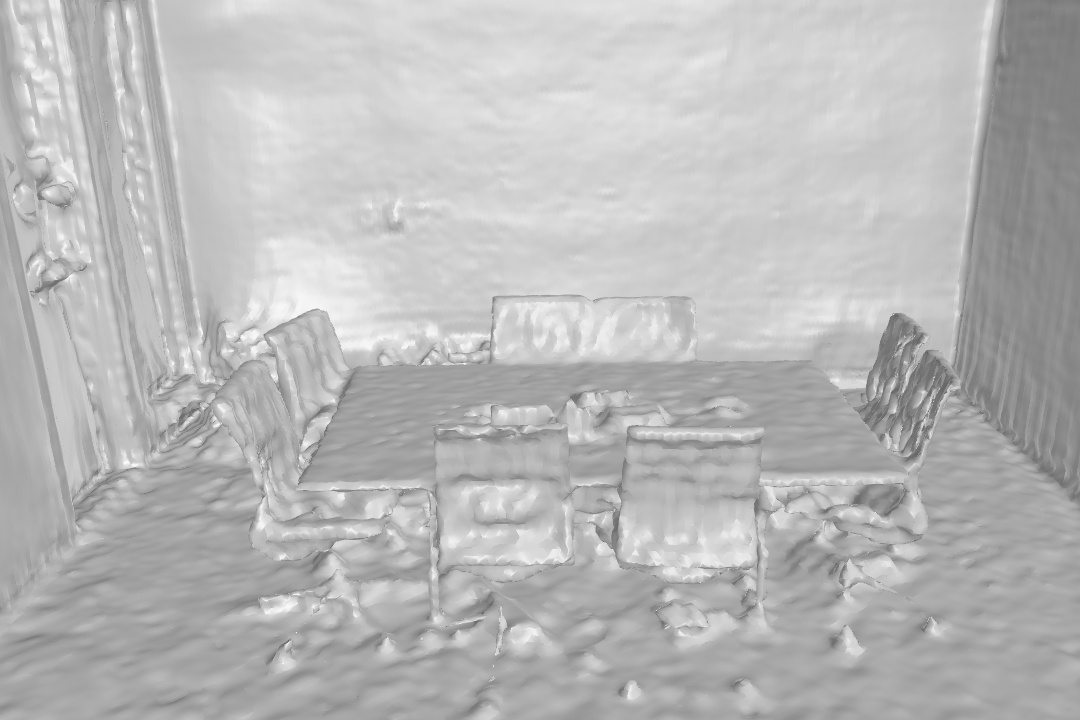}&
            \includegraphics[width=\rwidth]{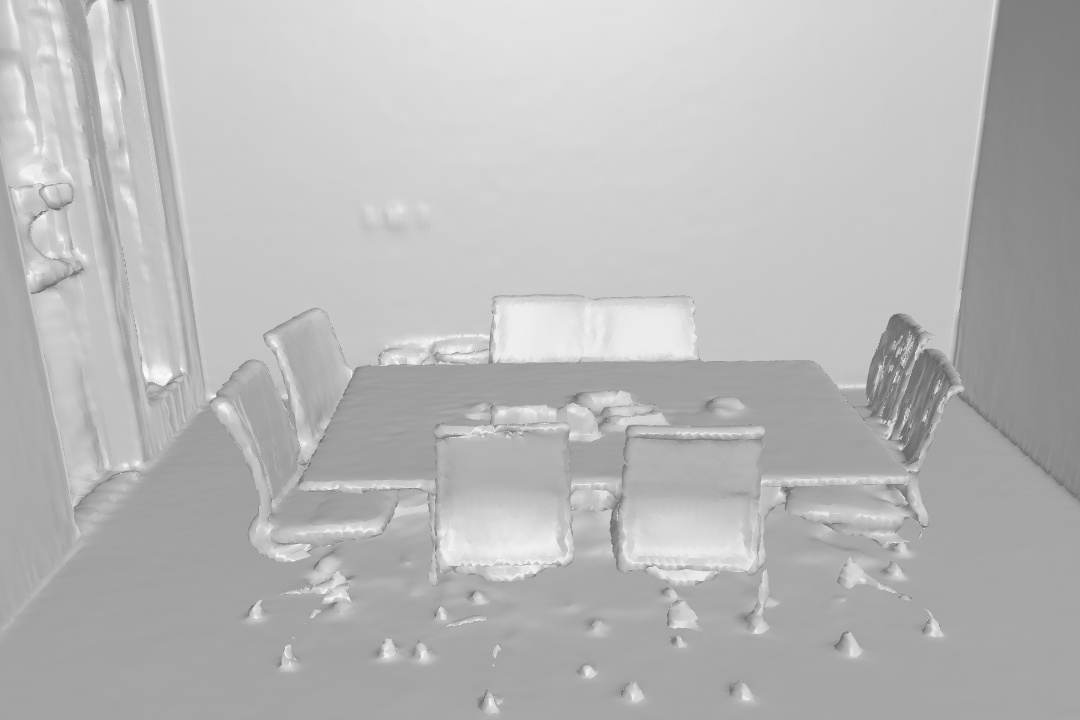}&
            \includegraphics[width=\rwidth]{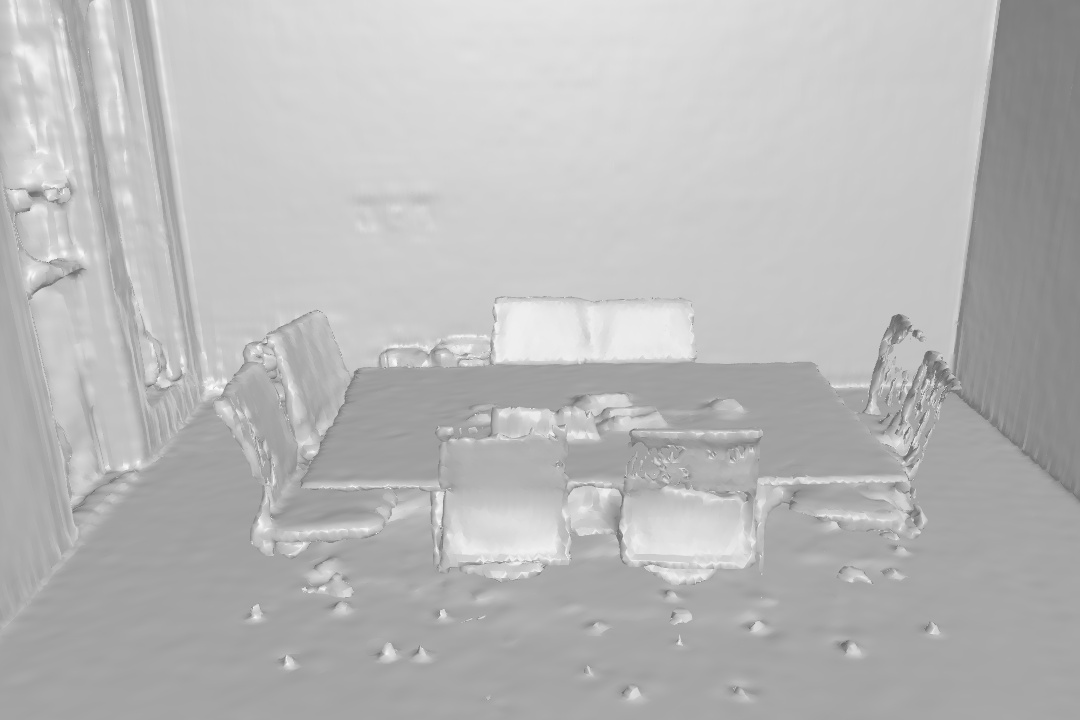}\\
             Single-Res. Grid&
             + Depth  &
             + Normal&
             + Both\\ 
            \includegraphics[width=\rwidth]{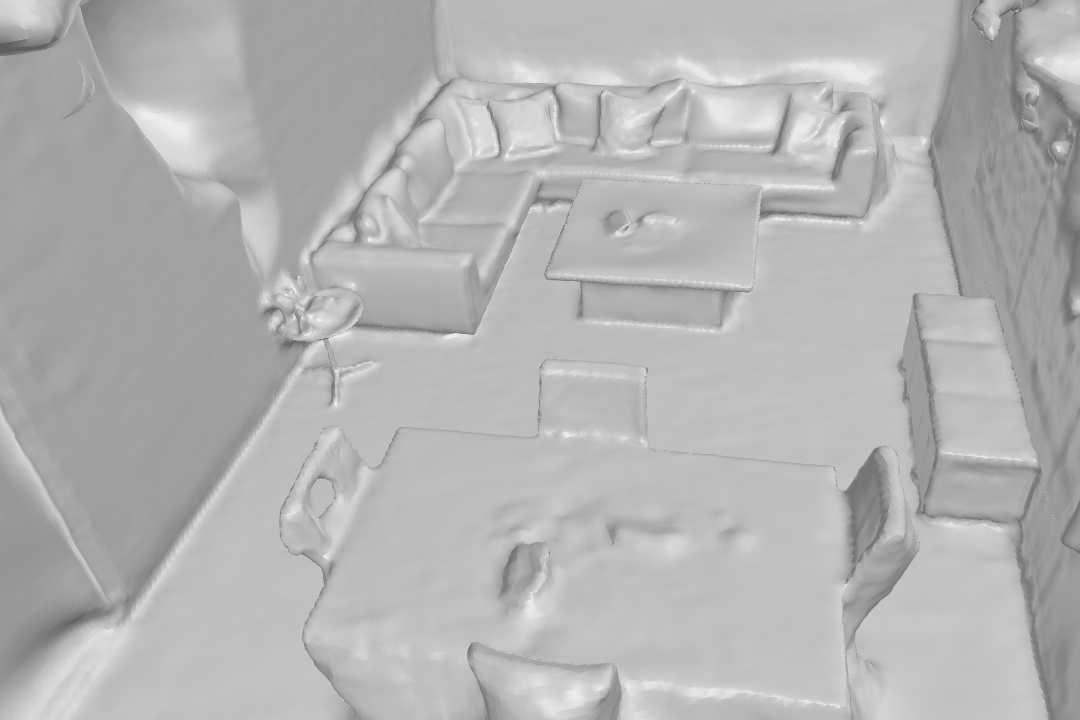}&
            \includegraphics[width=\rwidth]{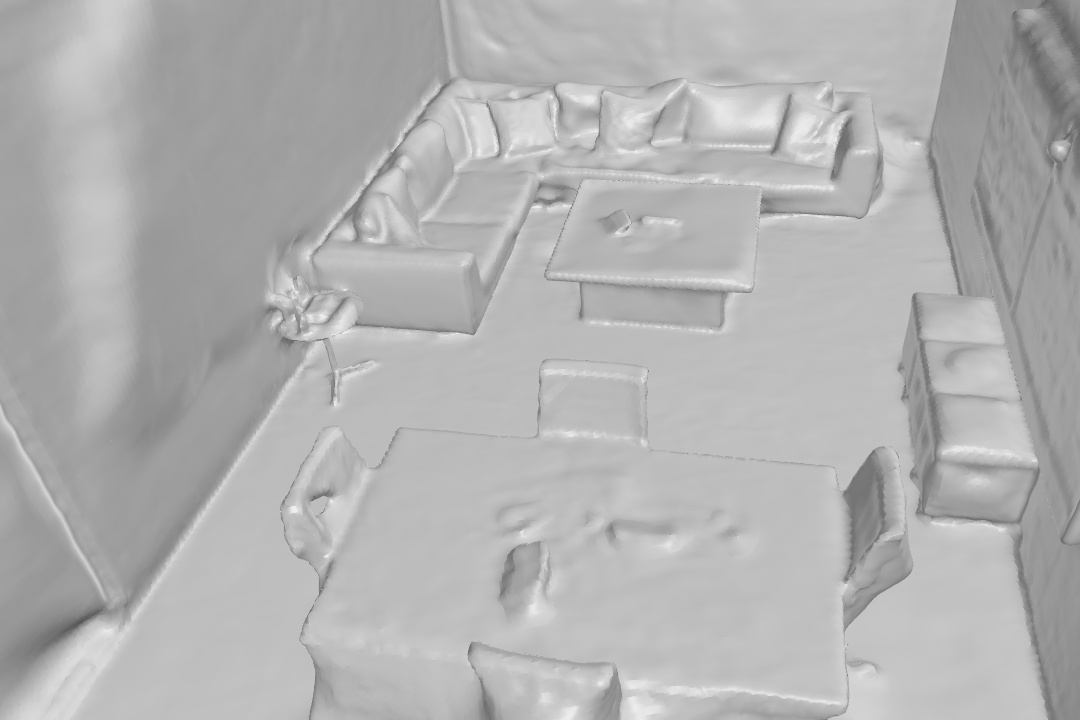}&
            \includegraphics[width=\rwidth]{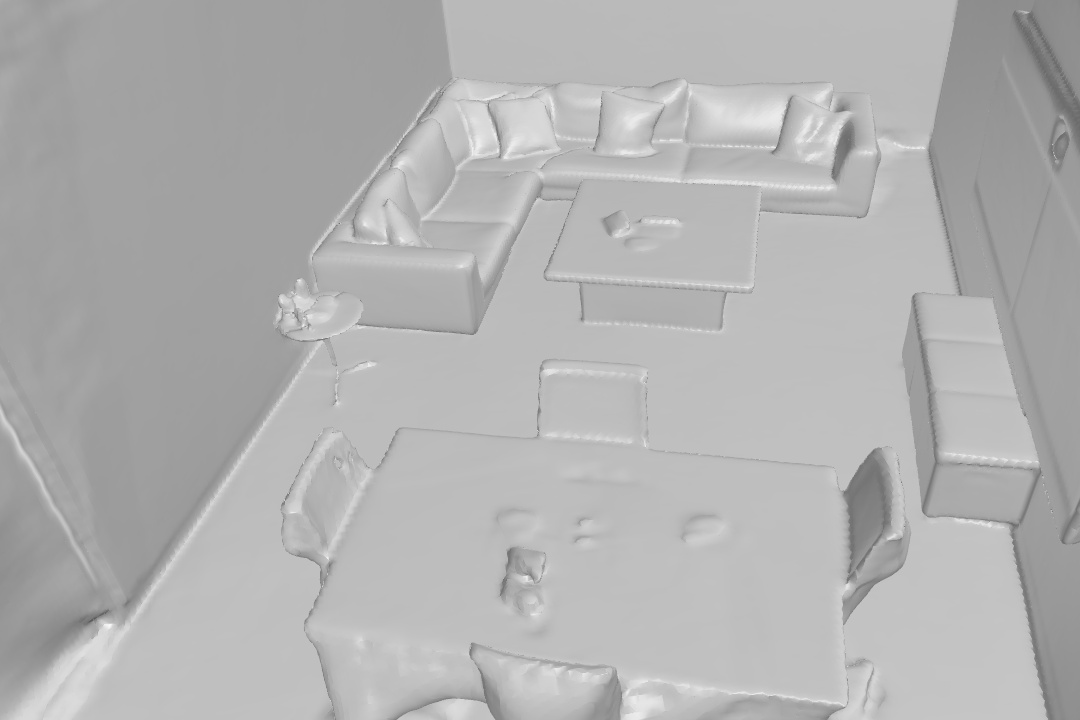}&
            \includegraphics[width=\rwidth]{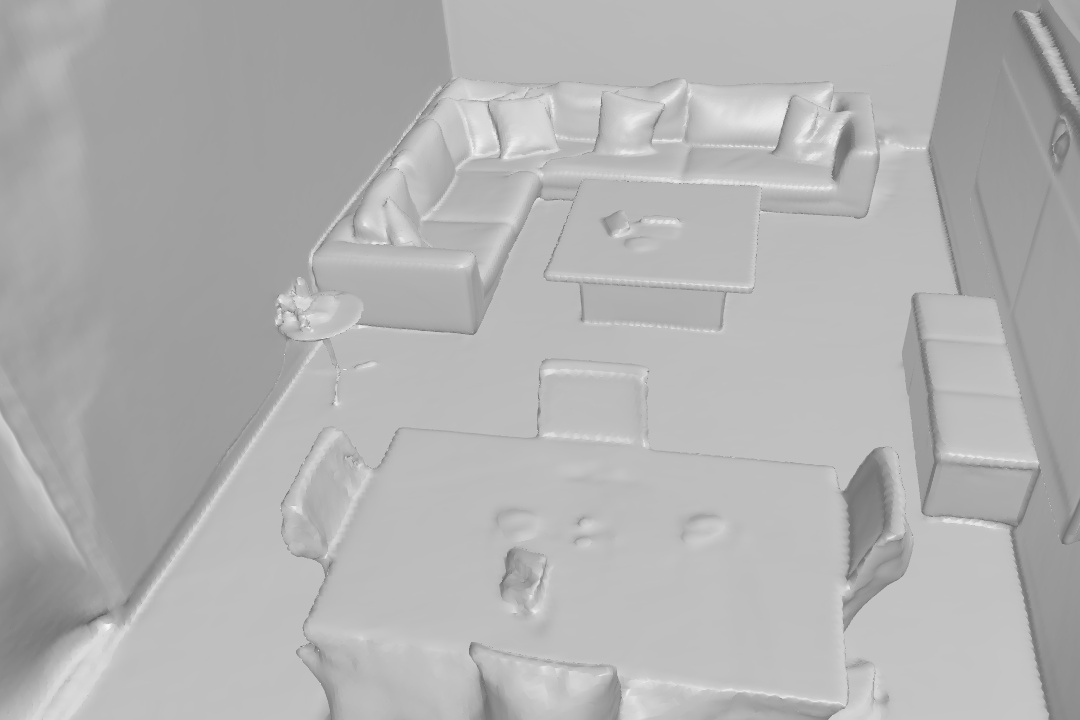}\\
            \includegraphics[width=\rwidth]{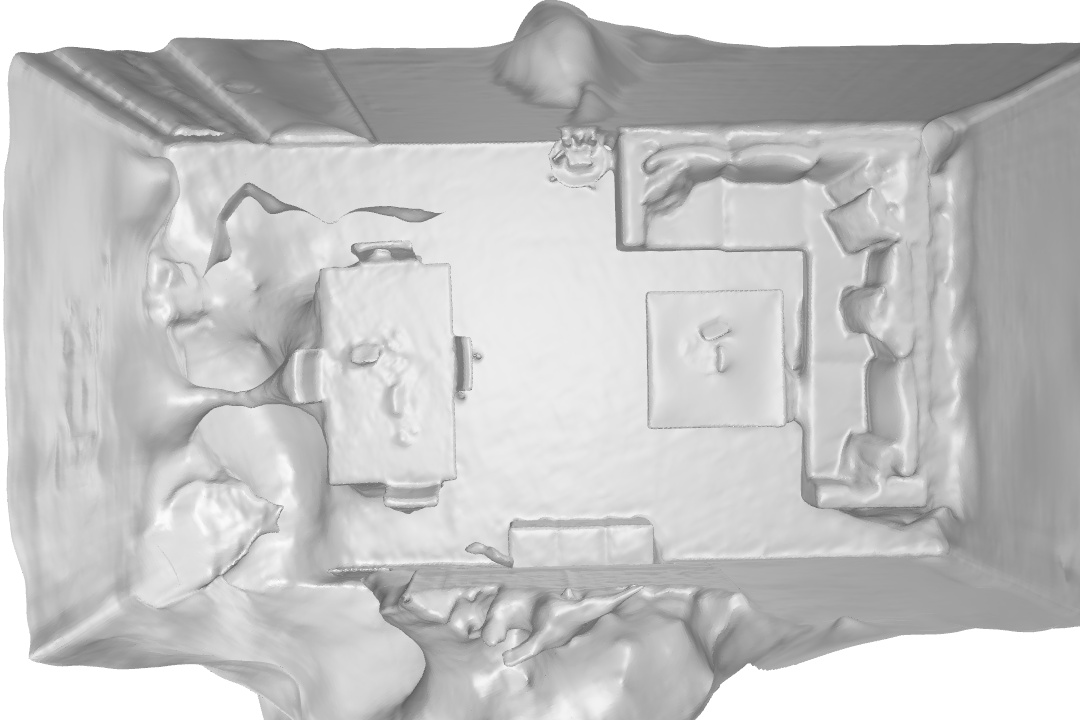}&
            \includegraphics[width=\rwidth]{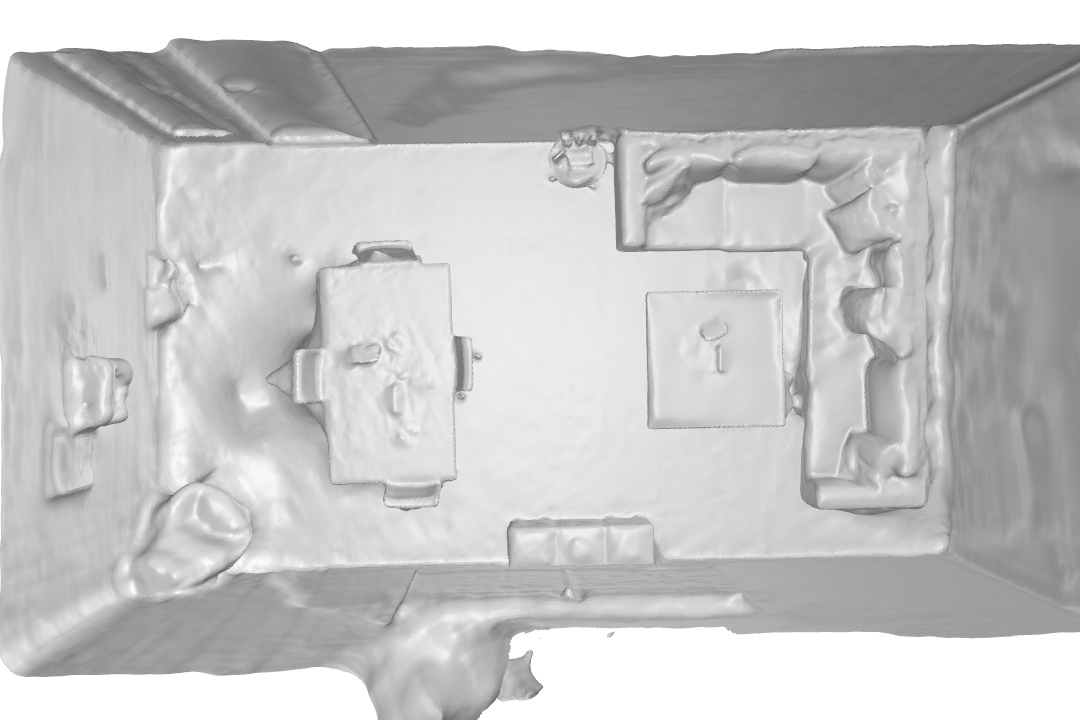}&
            \includegraphics[width=\rwidth]{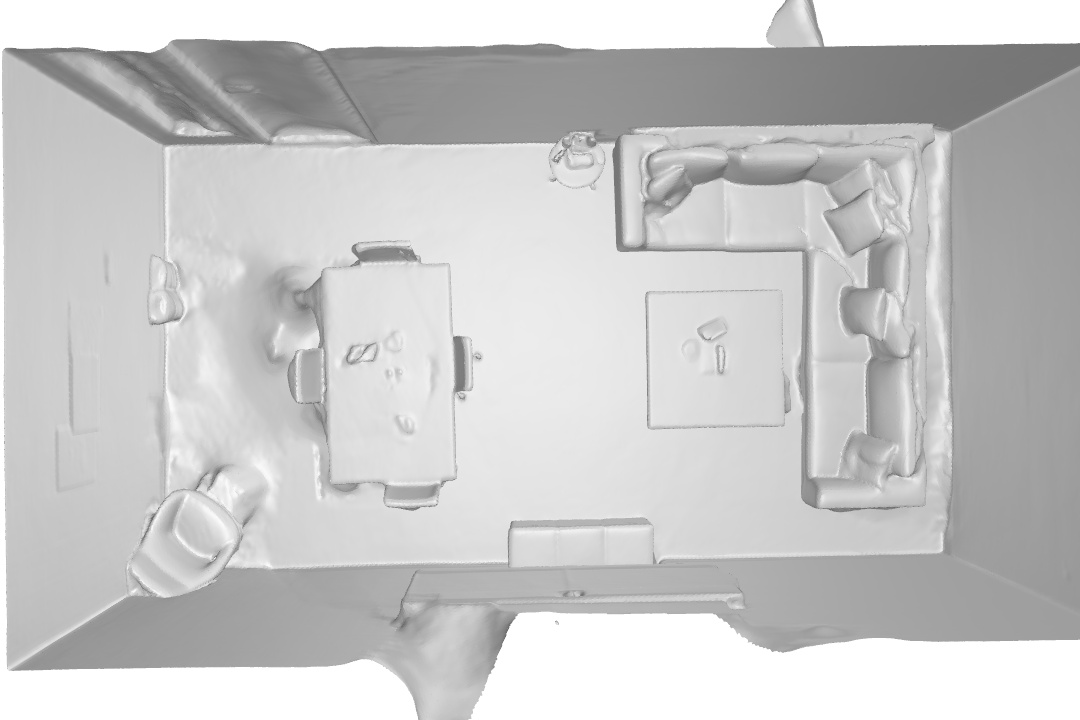}&
            \includegraphics[width=\rwidth]{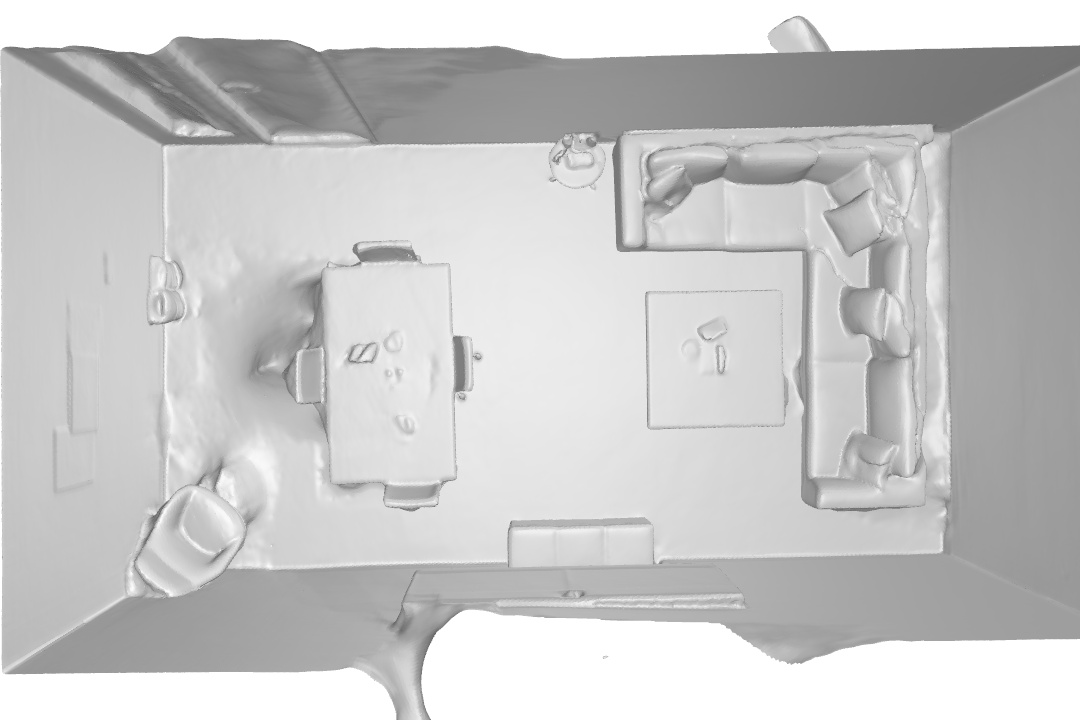}\\
             Multi-Res. Grids&
             + Depth  &
             + Normal&
             + Both\\
        \end{tabular}
        \caption{
        \textbf{Ablation of Monocular Geometric Cues on the Replica Dataset.}
        Monocular geometric cues significantly improve reconstruction quality for all architectures. 
        With monocular depth cues, the recovered geometry contains more details and a better overall structure. Similarly, with our normal cues, missing details are added and the results become smoother. Using both cues leads to the best performance. Zoom in for details.
        }
        \label{fig:ablation_cues_full}
\end{figure*}
}

\newcommand{\newwidtht}{0.248\textwidth}
\newcommand{\figuretnt}{
\begin{figure*}[t]
        \centering
        \setlength{\tabcolsep}{0.1em}
        \renewcommand{\arraystretch}{0.5}
        \footnotesize
        \begin{tabular}{cccc}
\includegraphics[width=\newwidtht]{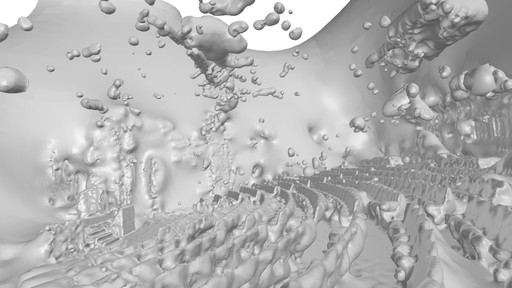}&
\includegraphics[width=\newwidtht]{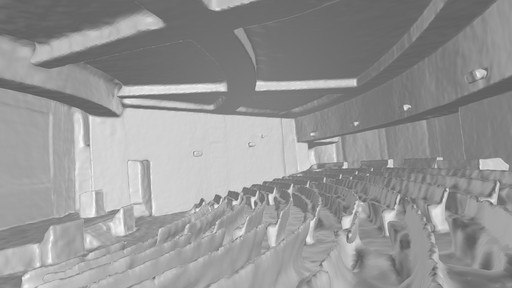}&
\includegraphics[width=\newwidtht]{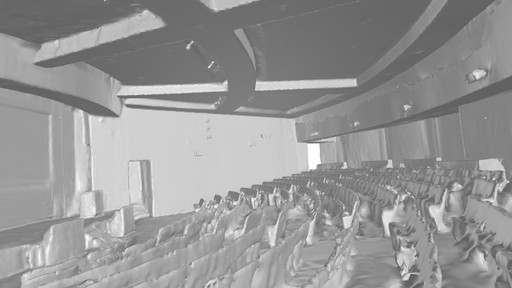}&
\includegraphics[width=\newwidtht]{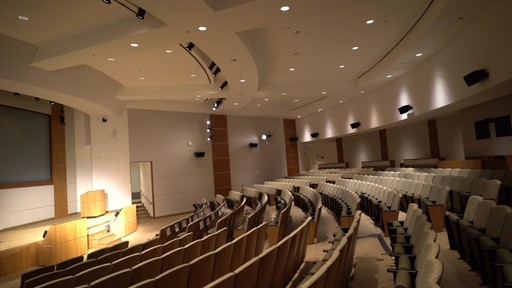}\\
\includegraphics[width=\newwidtht]{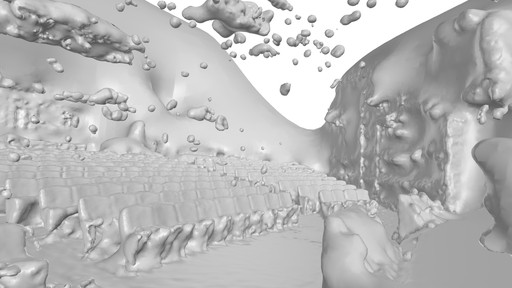}&
\includegraphics[width=\newwidtht]{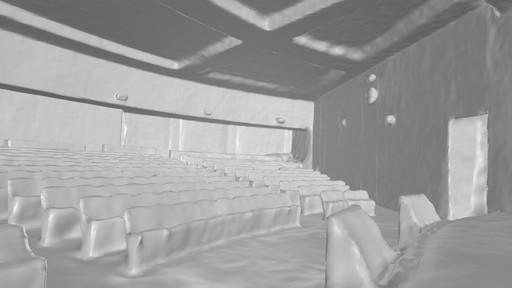}&
\includegraphics[width=\newwidtht]{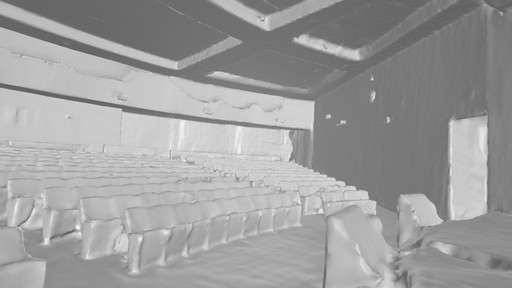}&
\includegraphics[width=\newwidtht]{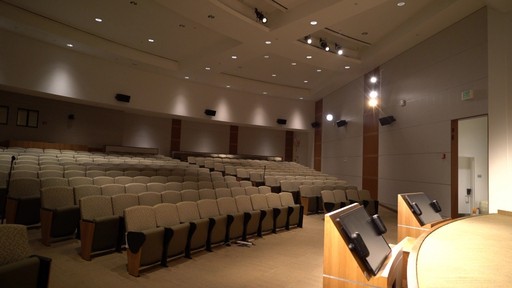}\\
\includegraphics[width=\newwidtht]{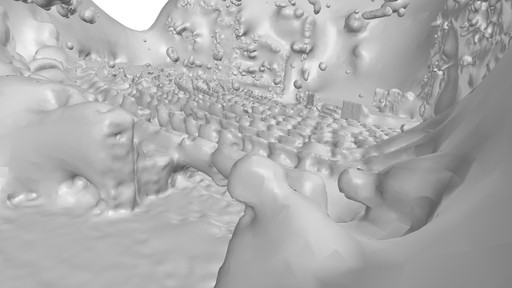}&
\includegraphics[width=\newwidtht]{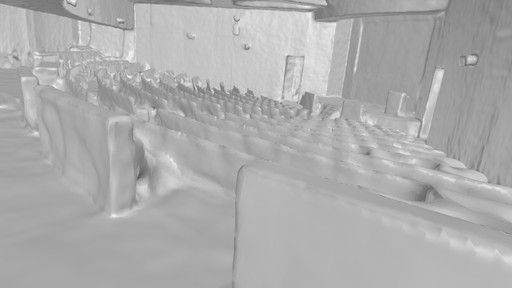}&
\includegraphics[width=\newwidtht]{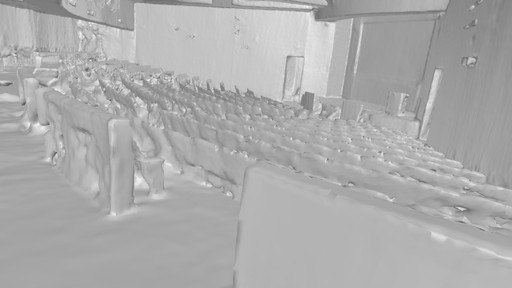}&
\includegraphics[width=\newwidtht]{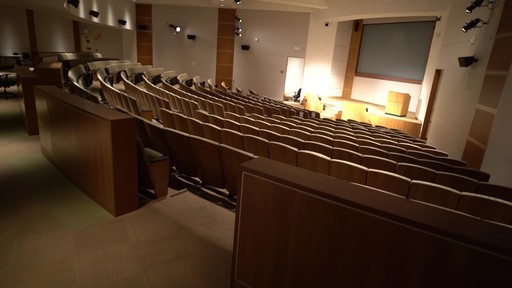}\\
\includegraphics[width=\newwidtht]{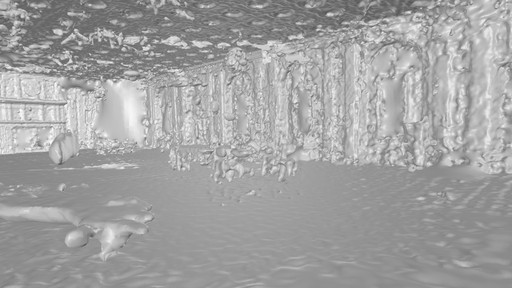}&
\includegraphics[width=\newwidtht]{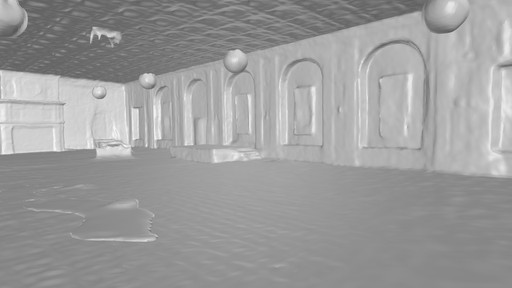}&
\includegraphics[width=\newwidtht]{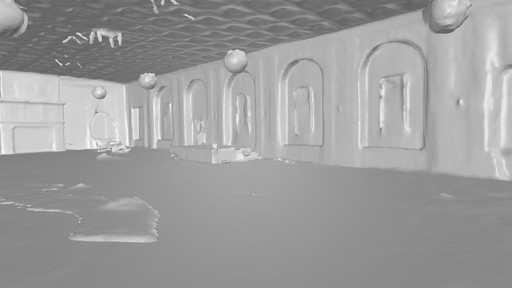}&
\includegraphics[width=\newwidtht]{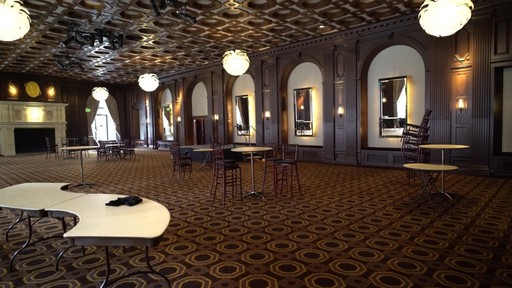}\\
\includegraphics[width=\newwidtht]{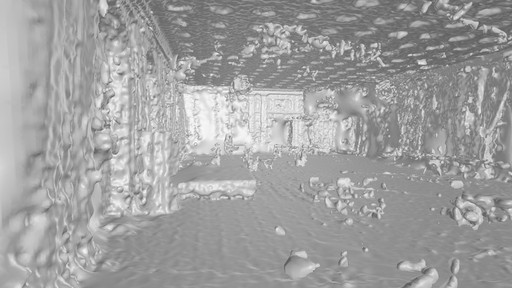}&
\includegraphics[width=\newwidtht]{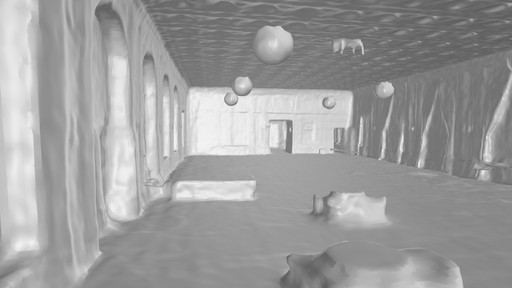}&
\includegraphics[width=\newwidtht]{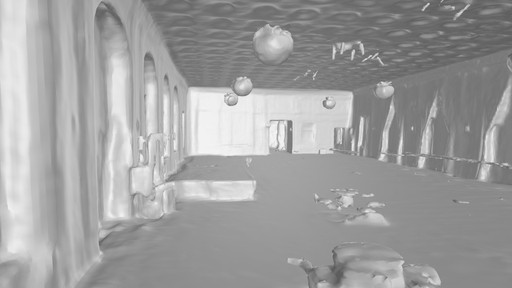}&
\includegraphics[width=\newwidtht]{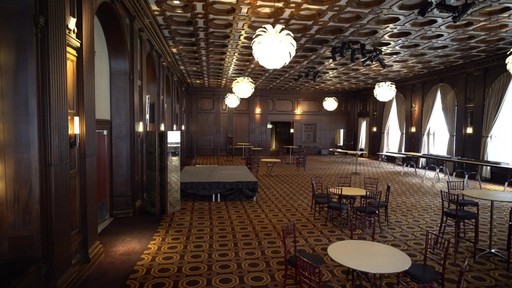}\\
\includegraphics[width=\newwidtht]{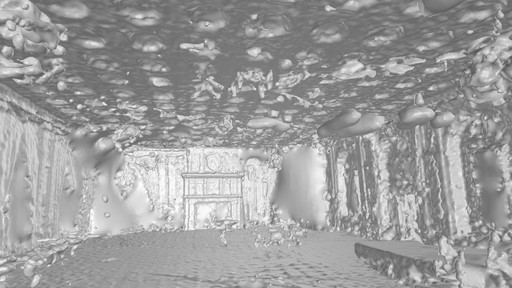}&
\includegraphics[width=\newwidtht]{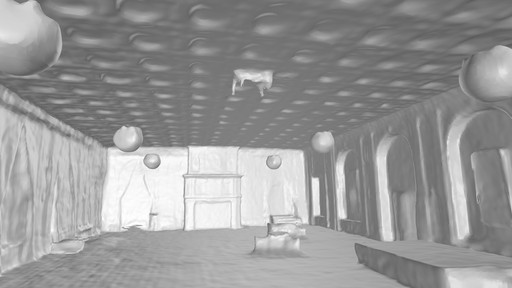}&
\includegraphics[width=\newwidtht]{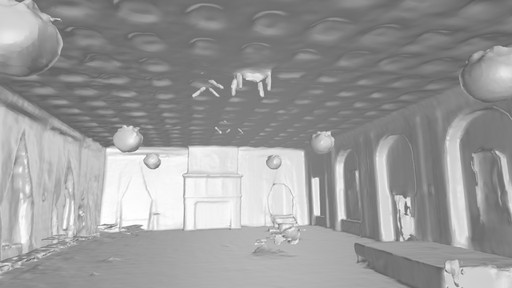}&
\includegraphics[width=\newwidtht]{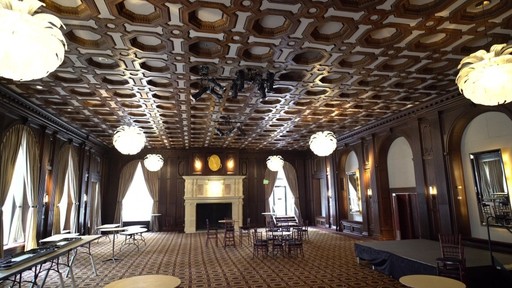}\\
\includegraphics[width=\newwidtht]{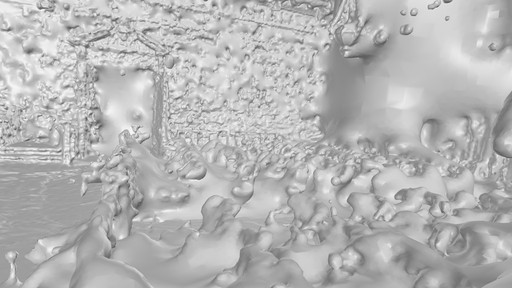}&
\includegraphics[width=\newwidtht]{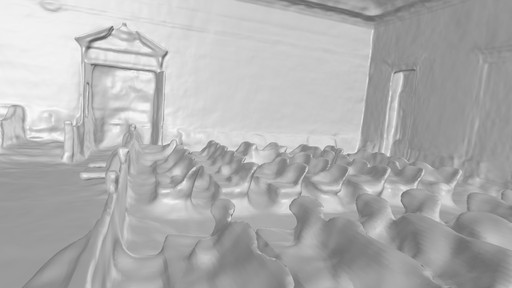}&
\includegraphics[width=\newwidtht]{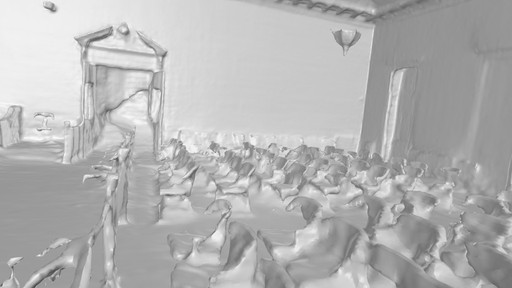}&
\includegraphics[width=\newwidtht]{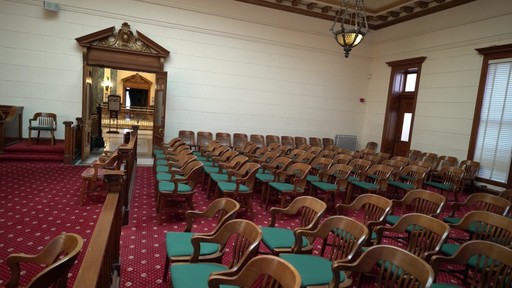}\\
\includegraphics[width=\newwidtht]{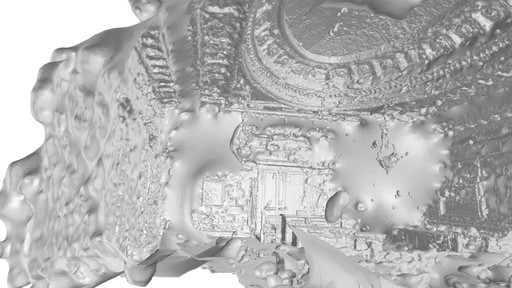}&
\includegraphics[width=\newwidtht]{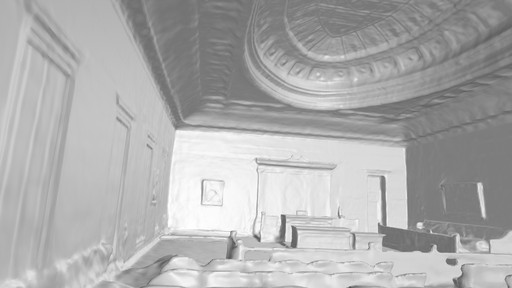}&
\includegraphics[width=\newwidtht]{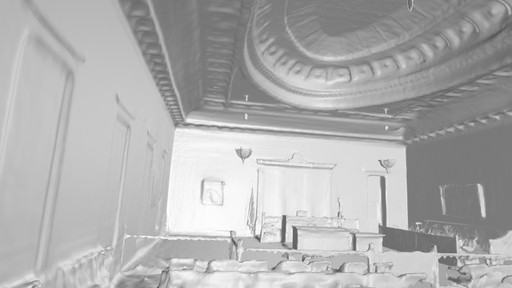}&
\includegraphics[width=\newwidtht]{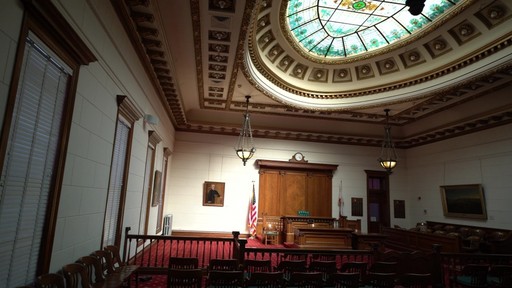}\\
\includegraphics[width=\newwidtht]{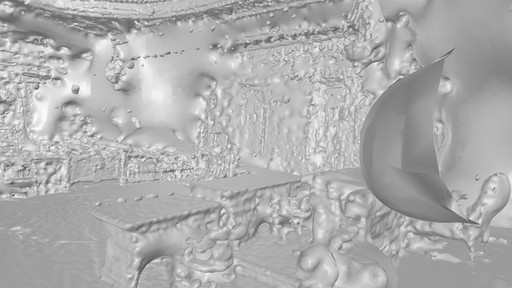}&
\includegraphics[width=\newwidtht]{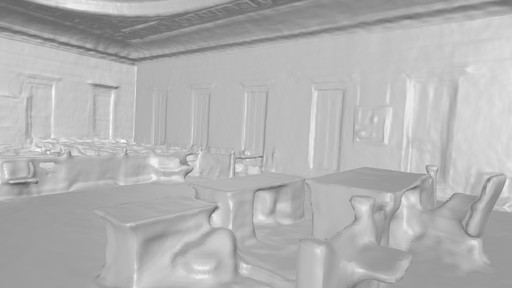}&
\includegraphics[width=\newwidtht]{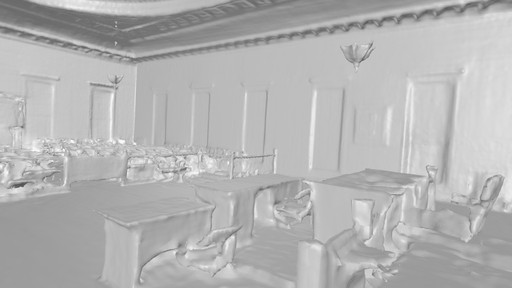}&
\includegraphics[width=\newwidtht]{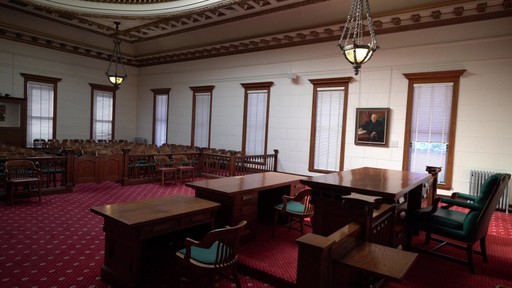}\\
\includegraphics[width=\newwidtht]{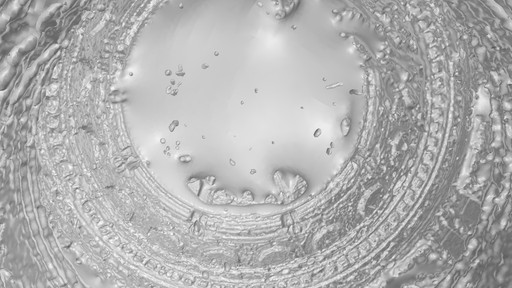}&
\includegraphics[width=\newwidtht]{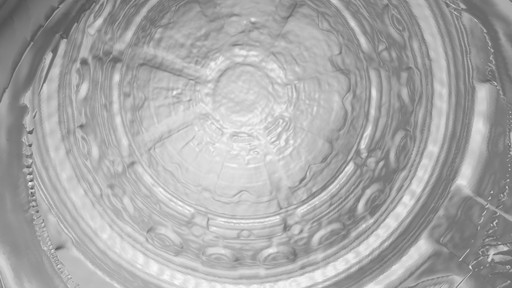}&
\includegraphics[width=\newwidtht]{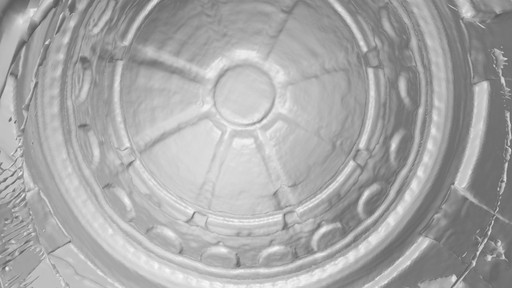}&
\includegraphics[width=\newwidtht]{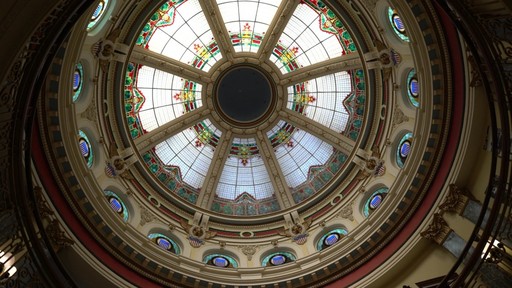}\\
\includegraphics[width=\newwidtht]{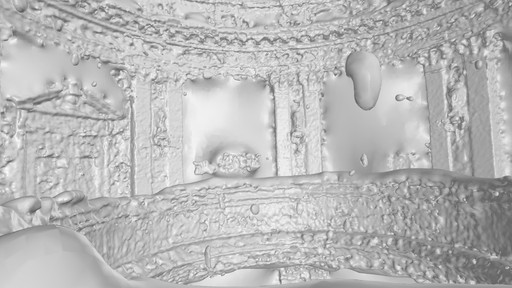}&
\includegraphics[width=\newwidtht]{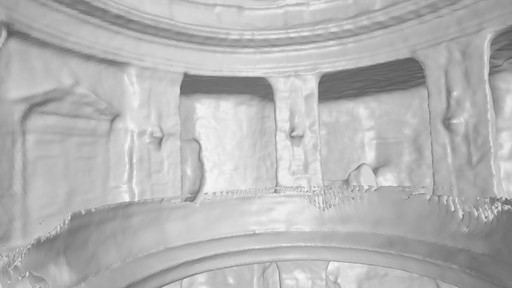}&
\includegraphics[width=\newwidtht]{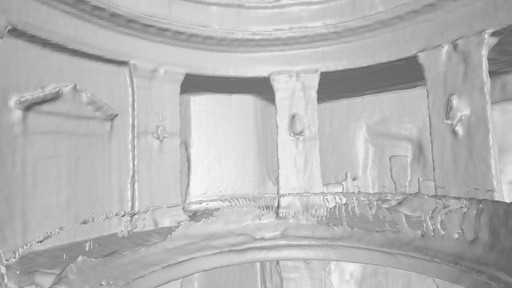}&
\includegraphics[width=\newwidtht]{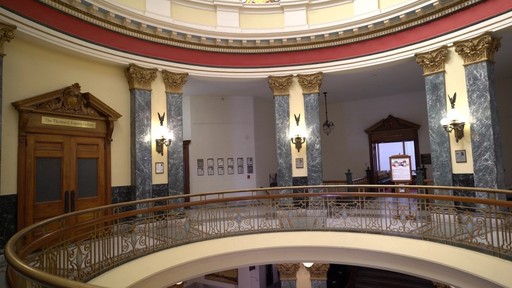}\\

             VisMVSNet~\cite{Zhang2020BMVC}& 
             Ours (MLP)&
             Ours (Multi-Res. Grids)&
             GT view\\
        \end{tabular}
    \vspace{-0.1cm}
    \caption{\textbf{Qualitative Comparison on Tanks \& Temples.}%
    }
    \label{fig:tnt}
\end{figure*}

}

\newcommand{\threewidth}{0.32\textwidth}
\newcommand{\figuretntcompareone}{
\begin{figure*}[t]
        \centering
        \setlength{\tabcolsep}{0.1em}
        \renewcommand{\arraystretch}{0.7}
        \footnotesize
        \begin{tabular}{ccc}
\includegraphics[width=\threewidth]{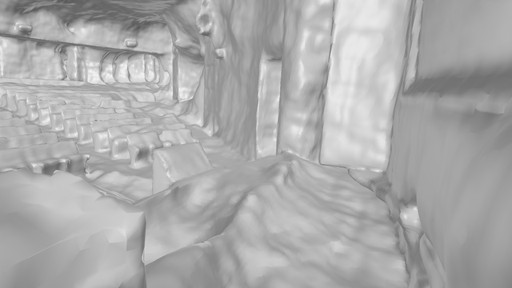}&
\includegraphics[width=\threewidth]{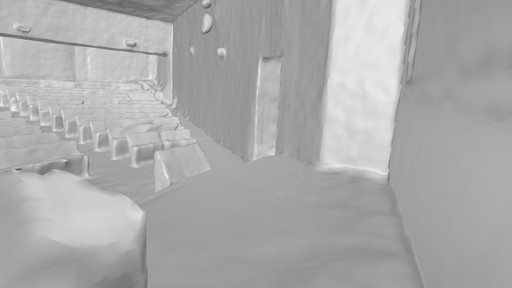}&
\includegraphics[width=\threewidth]{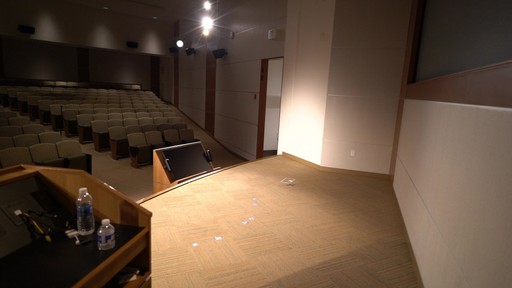}\\
\includegraphics[width=\threewidth]{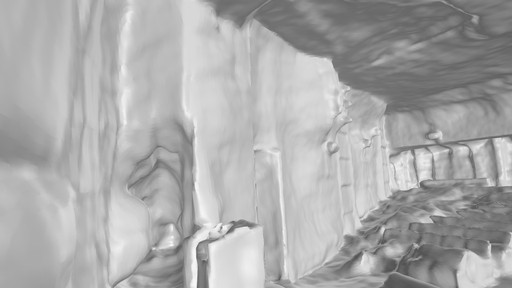}&
\includegraphics[width=\threewidth]{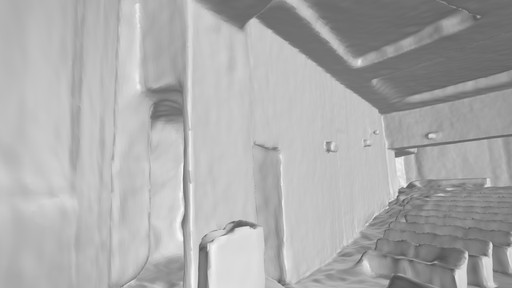}&
\includegraphics[width=\threewidth]{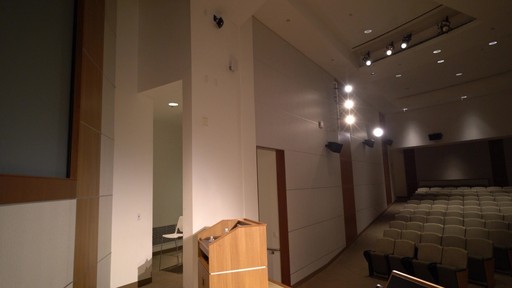}\\
\includegraphics[width=\threewidth]{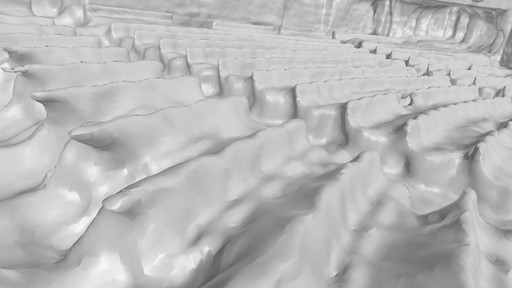}&
\includegraphics[width=\threewidth]{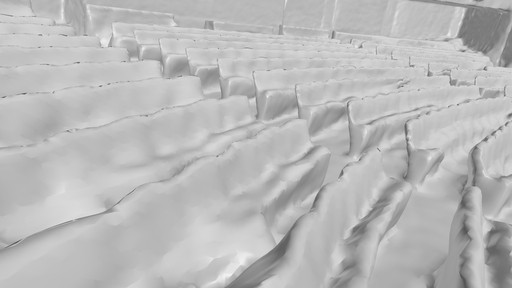}&
\includegraphics[width=\threewidth]{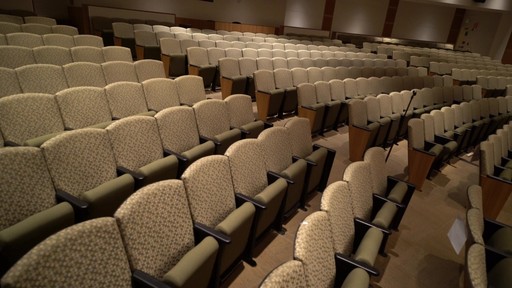}\\
\includegraphics[width=\threewidth]{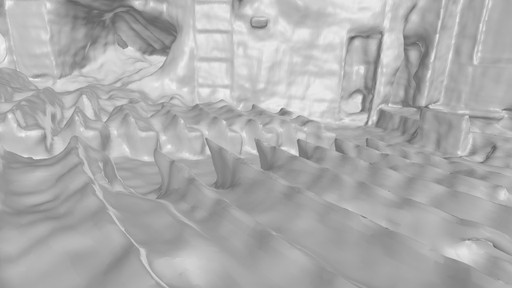}&
\includegraphics[width=\threewidth]{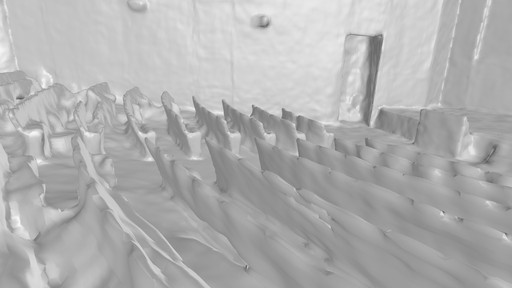}&
\includegraphics[width=\threewidth]{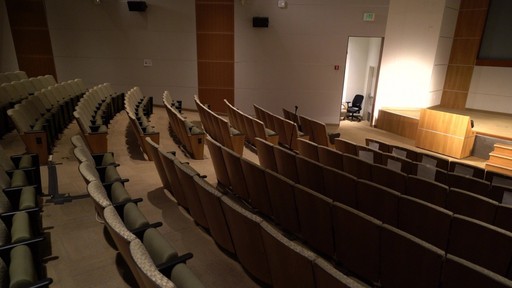}\\
\includegraphics[width=\threewidth]{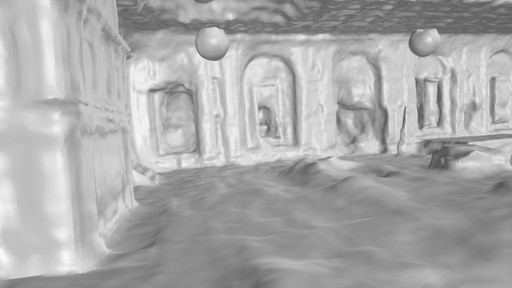}&
\includegraphics[width=\threewidth]{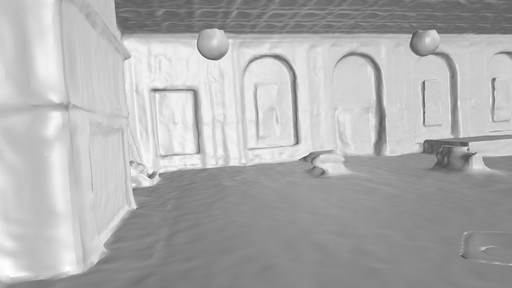}&
\includegraphics[width=\threewidth]{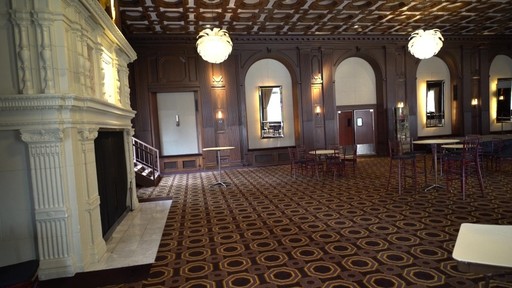}\\
\includegraphics[width=\threewidth]{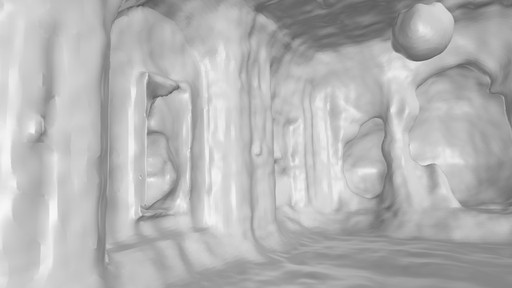}&
\includegraphics[width=\threewidth]{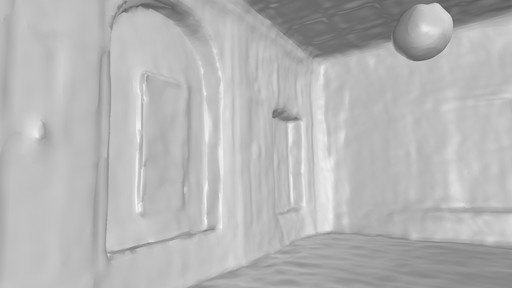}&
\includegraphics[width=\threewidth]{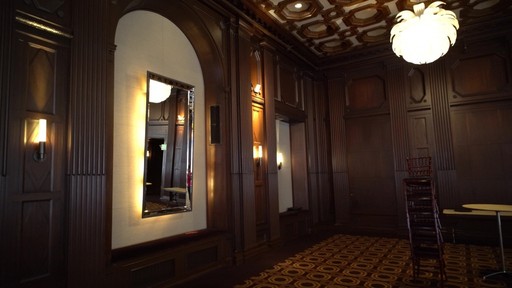}\\
\includegraphics[width=\threewidth]{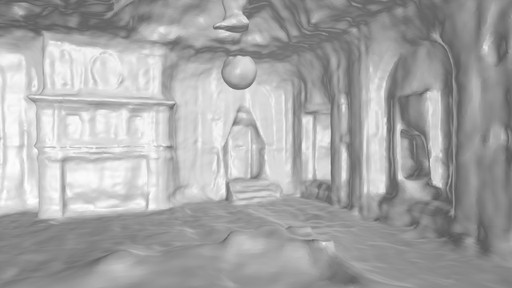}&
\includegraphics[width=\threewidth]{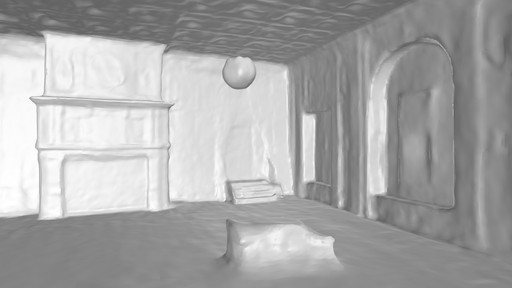}&
\includegraphics[width=\threewidth]{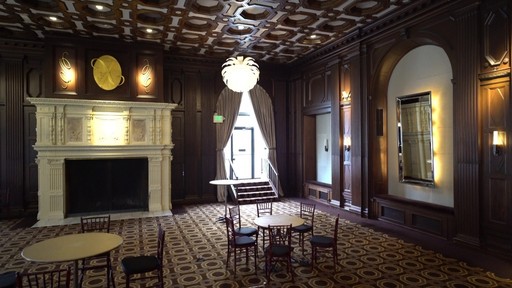}\\
\includegraphics[width=\threewidth]{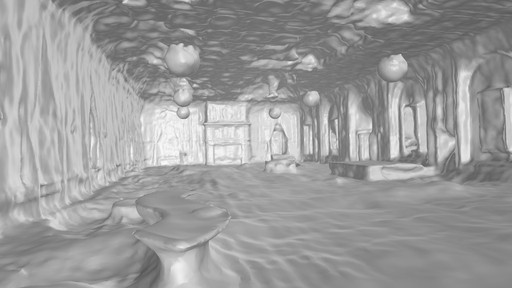}&
\includegraphics[width=\threewidth]{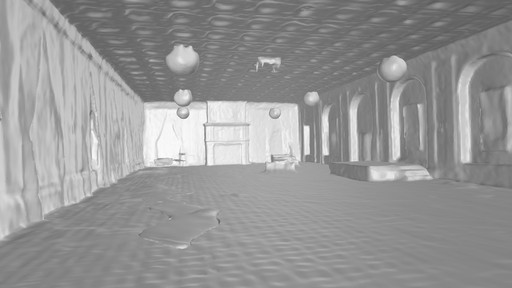}&
\includegraphics[width=\threewidth]{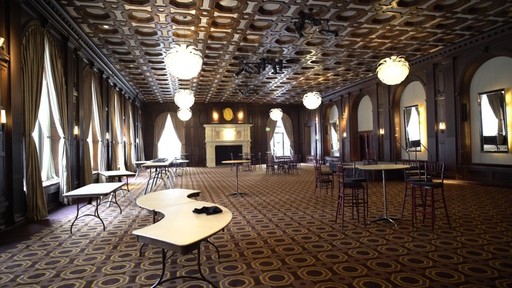}\\

             MLP~\cite{Yariv2021NEURIPS}& 
             MLP w/ cues&
             GT view\\
        \end{tabular}
    
    \caption{\textbf{Qualitative Comparison on Tanks \& Temples.}
    We use a single MLP as the scene geometry representation~\cite{Yariv2021NEURIPS} and compare the reconstruction when using monocular cues or not on Auditorium and Ballroom.
    }
    \label{fig:tnt_1}
\end{figure*}

}

\newcommand{\figuretntcomparetwo}{
\begin{figure*}[t]
        \centering
        \setlength{\tabcolsep}{0.1em}
        \renewcommand{\arraystretch}{0.7}
        \footnotesize
        \begin{tabular}{ccc}
        
\includegraphics[width=\threewidth]{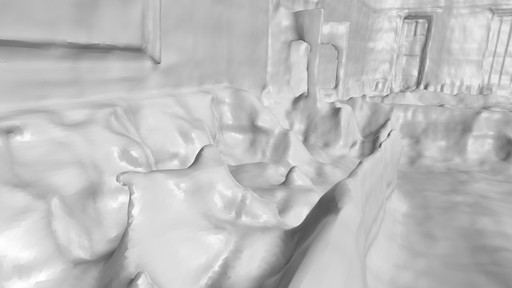}&
\includegraphics[width=\threewidth]{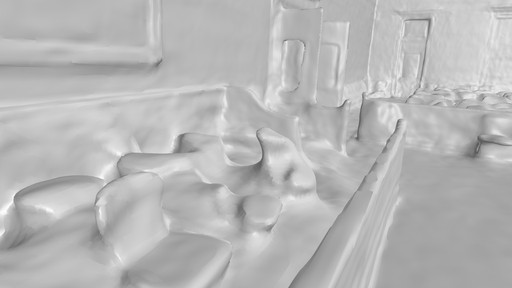}&
\includegraphics[width=\threewidth]{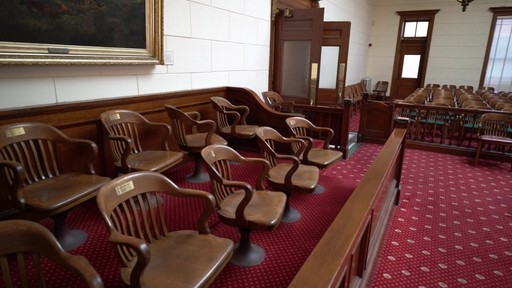}\\
\includegraphics[width=\threewidth]{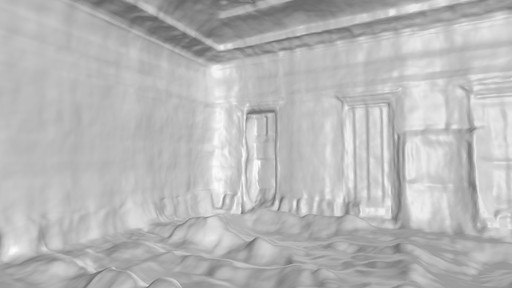}&
\includegraphics[width=\threewidth]{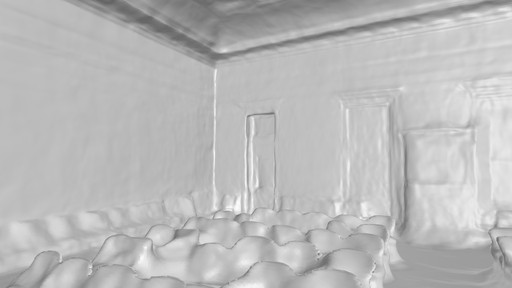}&
\includegraphics[width=\threewidth]{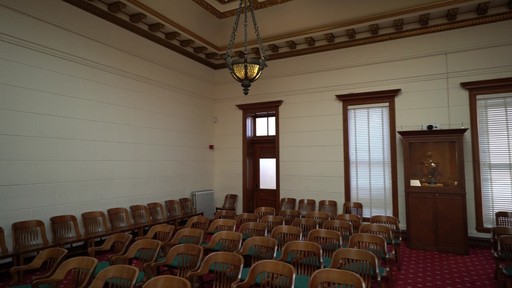}\\
\includegraphics[width=\threewidth]{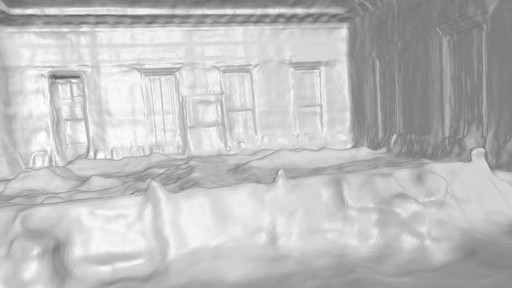}&
\includegraphics[width=\threewidth]{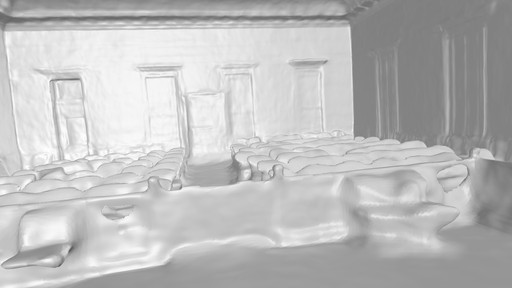}&
\includegraphics[width=\threewidth]{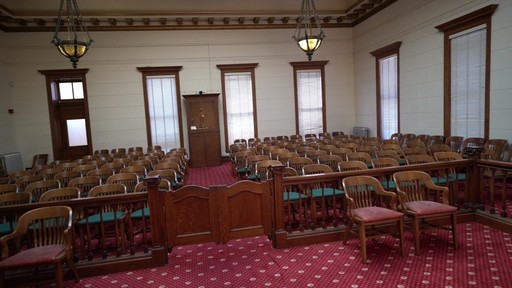}\\
\includegraphics[width=\threewidth]{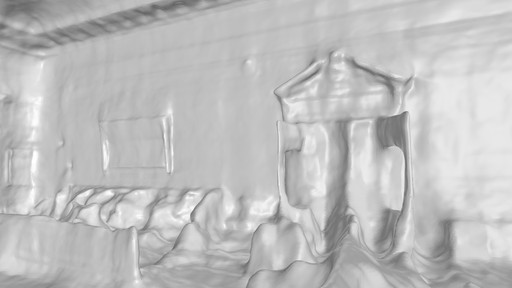}&
\includegraphics[width=\threewidth]{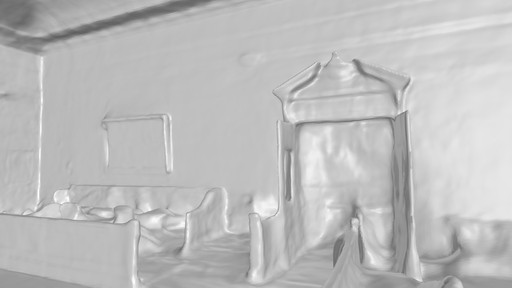}&
\includegraphics[width=\threewidth]{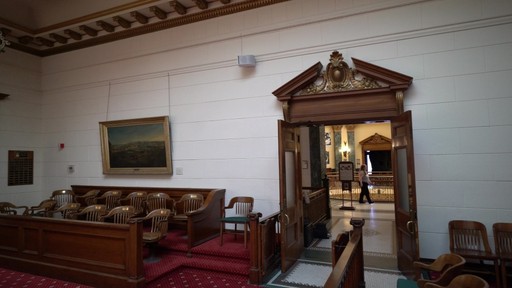}\\
\includegraphics[width=\threewidth]{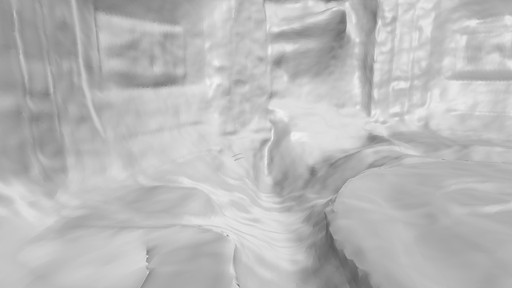}&
\includegraphics[width=\threewidth]{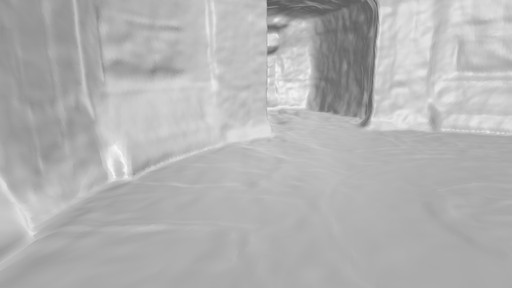}&
\includegraphics[width=\threewidth]{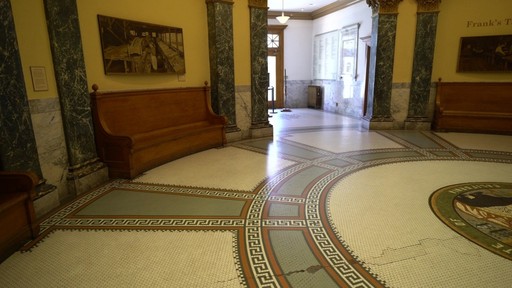}\\
\includegraphics[width=\threewidth]{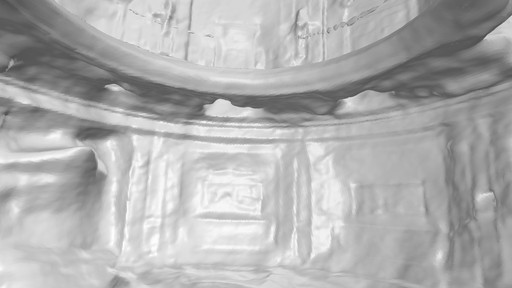}&
\includegraphics[width=\threewidth]{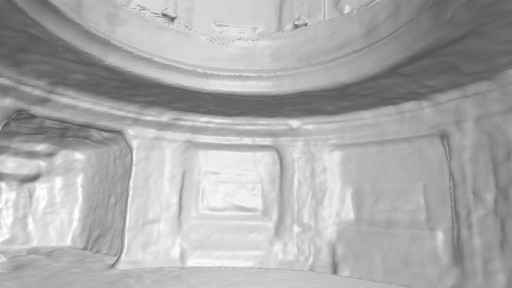}&
\includegraphics[width=\threewidth]{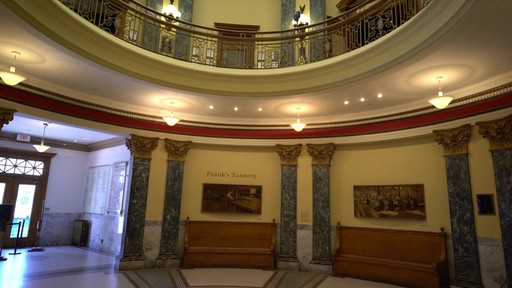}\\
\includegraphics[width=\threewidth]{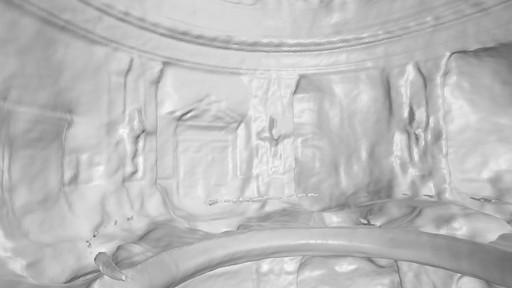}&
\includegraphics[width=\threewidth]{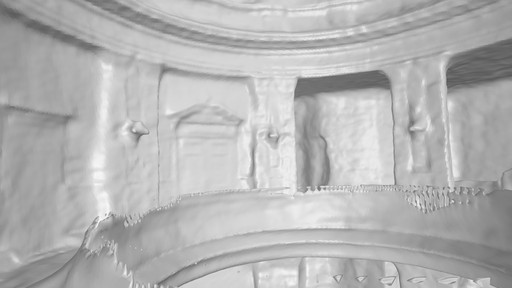}&
\includegraphics[width=\threewidth]{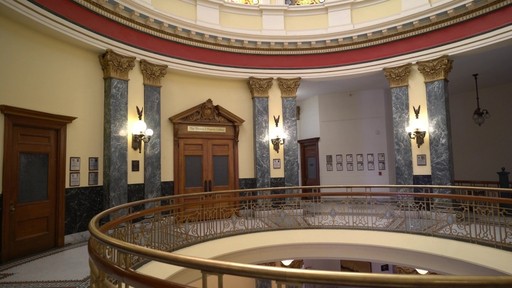}\\
\includegraphics[width=\threewidth]{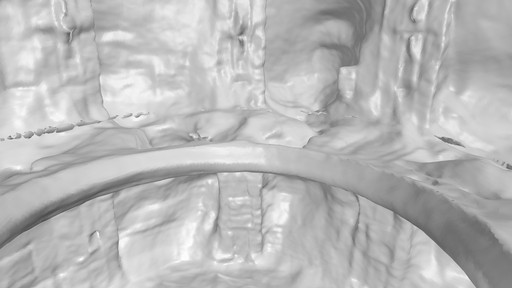}&
\includegraphics[width=\threewidth]{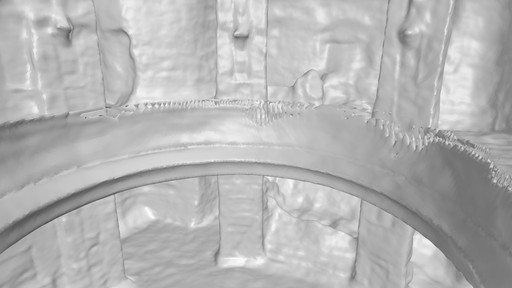}&
\includegraphics[width=\threewidth]{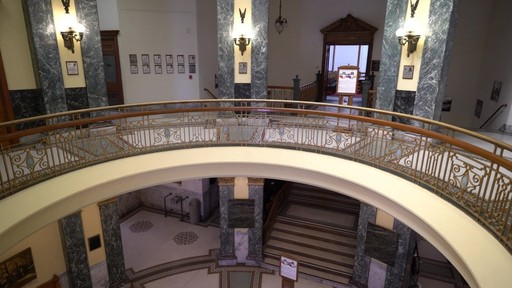}\\

             MLP~\cite{Yariv2021NEURIPS}& 
             MLP w/ cues&
             GT view\\
        \end{tabular}
    
    \caption{\textbf{Qualitative Comparison on Tanks \& Temples Dataset.} We use a single MLP as the scene geometry representation~\cite{Yariv2021NEURIPS} and compare the reconstruction quality when using monocular cues or not on Courtroom and Museum.
    }
    \label{fig:tnt_2}
\end{figure*}

}

\newcommand{\figuretntcomparethree}{
\begin{figure*}[t]
        \centering
        \setlength{\tabcolsep}{0.1em}
        \renewcommand{\arraystretch}{0.7}
        \footnotesize
        \begin{tabular}{ccc}
\includegraphics[width=\threewidth]{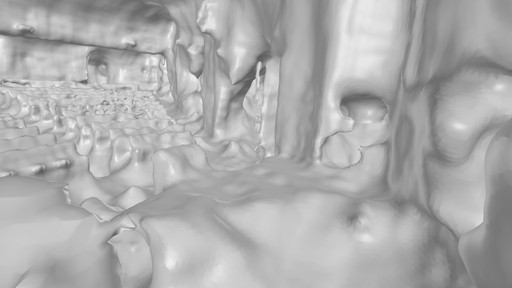}&
\includegraphics[width=\threewidth]{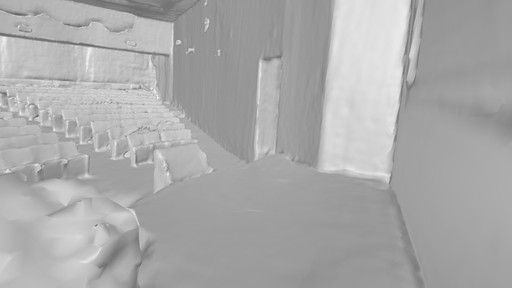}&
\includegraphics[width=\threewidth]{gfx/tnt2/rgb_Auditorium_000035.jpg}\\
\includegraphics[width=\threewidth]{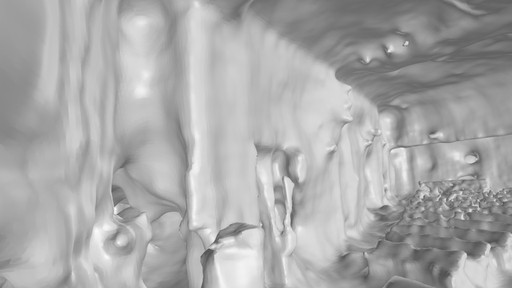}&
\includegraphics[width=\threewidth]{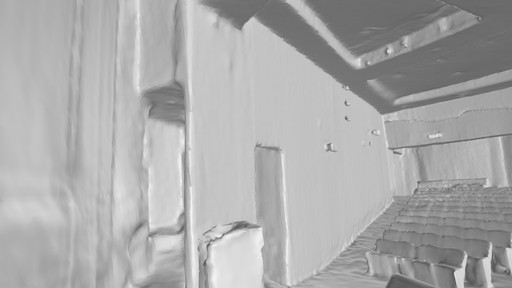}&
\includegraphics[width=\threewidth]{gfx/tnt2/rgb_Auditorium_000048.jpg}\\
\includegraphics[width=\threewidth]{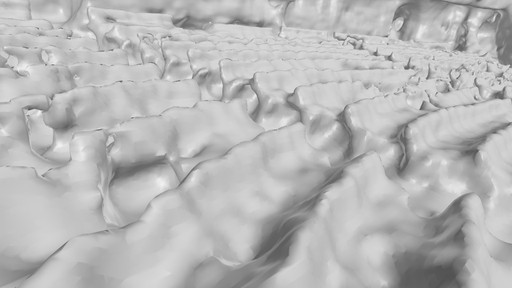}&
\includegraphics[width=\threewidth]{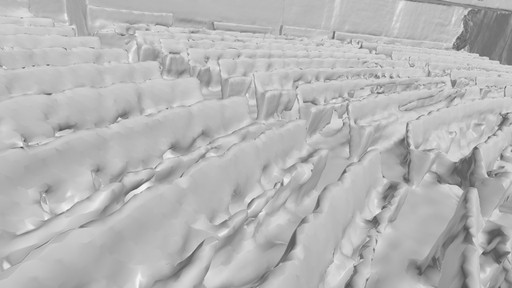}&
\includegraphics[width=\threewidth]{gfx/tnt2/rgb_Auditorium_000129.jpg}\\
\includegraphics[width=\threewidth]{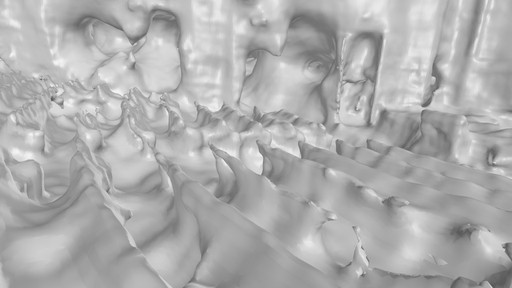}&
\includegraphics[width=\threewidth]{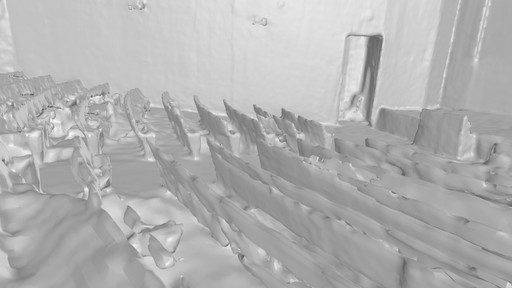}&
\includegraphics[width=\threewidth]{gfx/tnt2/rgb_Auditorium_000168.jpg}\\
\includegraphics[width=\threewidth]{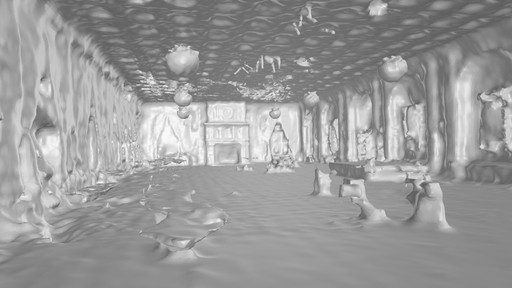}&
\includegraphics[width=\threewidth]{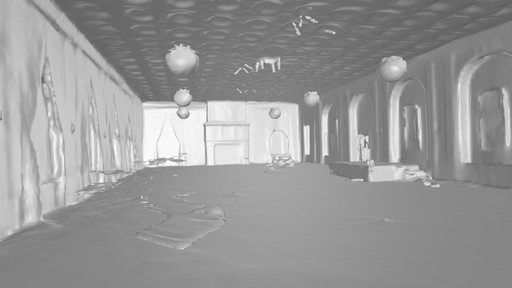}&
\includegraphics[width=\threewidth]{gfx/tnt2/rgb_Ballroom_000001.jpg}\\
\includegraphics[width=\threewidth]{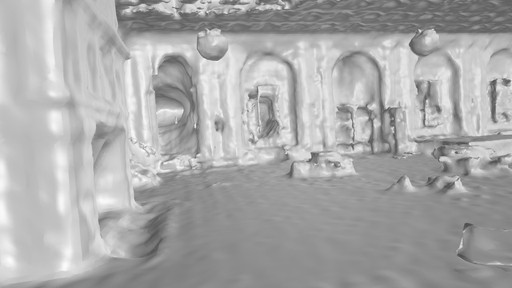}&
\includegraphics[width=\threewidth]{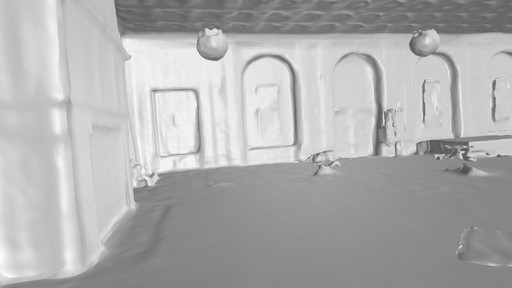}&
\includegraphics[width=\threewidth]{gfx/tnt2/rgb_Ballroom_000114.jpg}\\
\includegraphics[width=\threewidth]{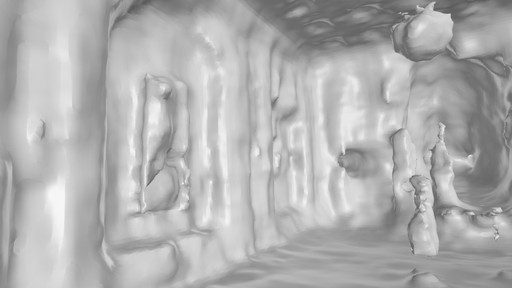}&
\includegraphics[width=\threewidth]{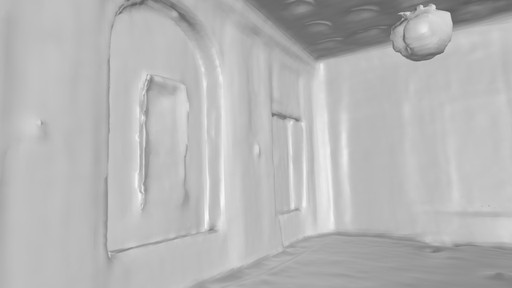}&
\includegraphics[width=\threewidth]{gfx/tnt2/rgb_Ballroom_000249.jpg}\\
\includegraphics[width=\threewidth]{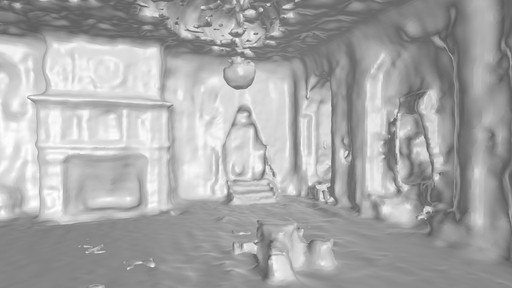}&
\includegraphics[width=\threewidth]{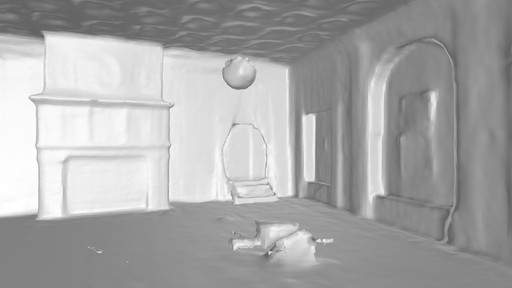}&
\includegraphics[width=\threewidth]{gfx/tnt2/rgb_Ballroom_000267.jpg}\\
        
             Multi-Res. Grids& 
             Multi-Res. Grids w/ cues&
             GT view\\
        \end{tabular}
    
    \caption{\textbf{Qualitative Comparison on Tanks \& Temples.} We use Multi-Res. Grids as the scene geometry representation and compare the reconstruction when using monocular cues or not on Auditorium and Ballroom.
    }
    \label{fig:tnt_3}
\end{figure*}

}

\newcommand{\figuretntcomparefour}{
\begin{figure*}[t]
        \centering
        \setlength{\tabcolsep}{0.1em}
        \renewcommand{\arraystretch}{0.7}
        \footnotesize
        \begin{tabular}{ccc}
        
\includegraphics[width=\threewidth]{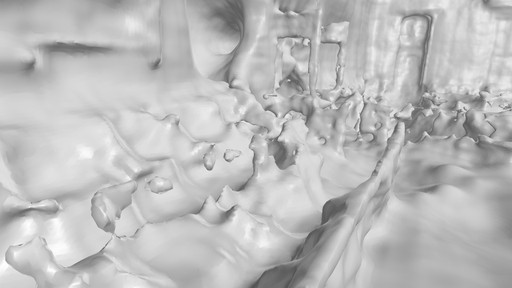}&
\includegraphics[width=\threewidth]{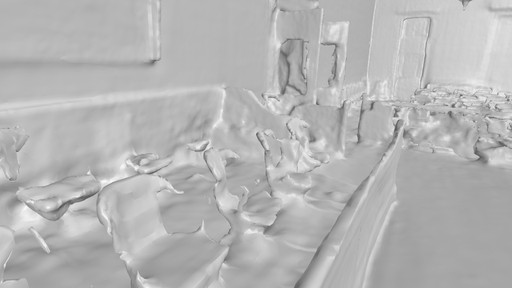}&
\includegraphics[width=\threewidth]{gfx/tnt2/rgb_Courtroom_000076.jpg}\\
\includegraphics[width=\threewidth]{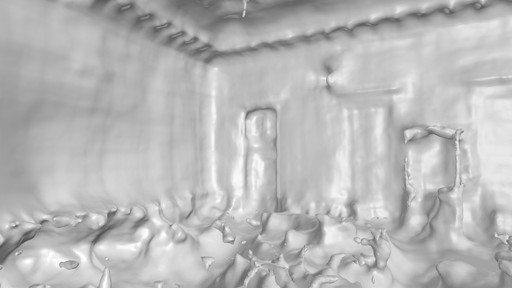}&
\includegraphics[width=\threewidth]{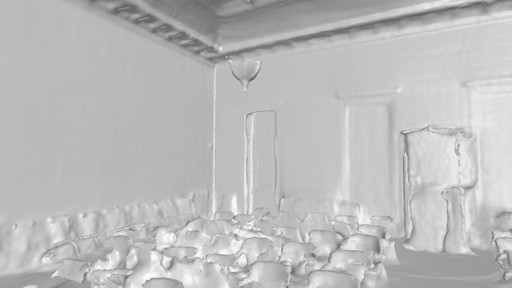}&
\includegraphics[width=\threewidth]{gfx/tnt2/rgb_Courtroom_000140.jpg}\\
\includegraphics[width=\threewidth]{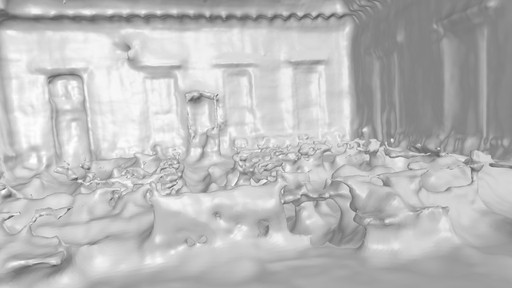}&
\includegraphics[width=\threewidth]{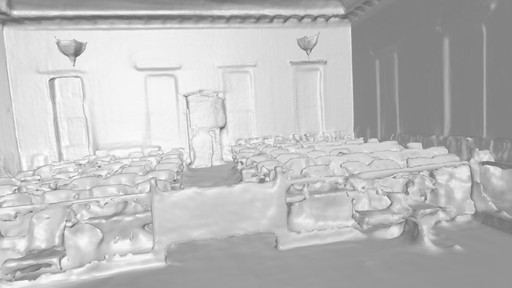}&
\includegraphics[width=\threewidth]{gfx/tnt2/rgb_Courtroom_000297.jpg}\\
\includegraphics[width=\threewidth]{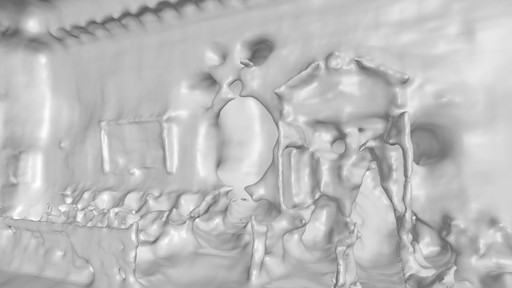}&
\includegraphics[width=\threewidth]{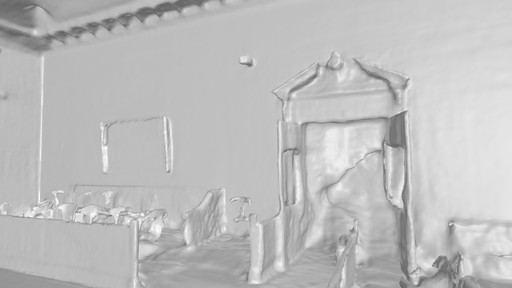}&
\includegraphics[width=\threewidth]{gfx/tnt2/rgb_Courtroom_000143.jpg}\\
\includegraphics[width=\threewidth]{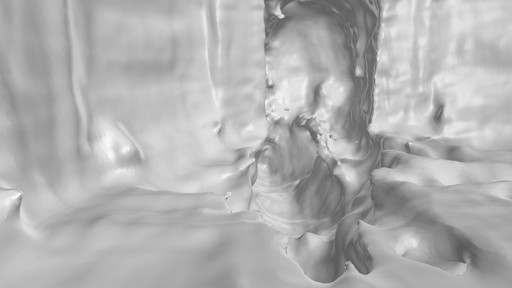}&
\includegraphics[width=\threewidth]{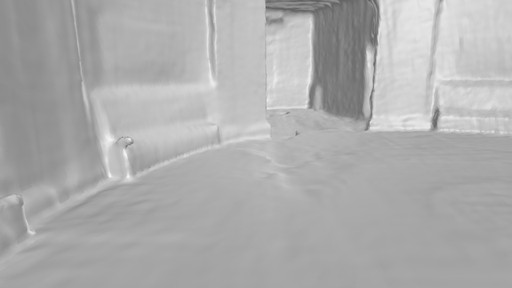}&
\includegraphics[width=\threewidth]{gfx/tnt2/rgb_Museum_000027.jpg}\\
\includegraphics[width=\threewidth]{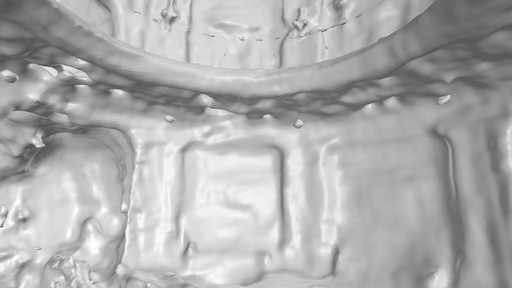}&
\includegraphics[width=\threewidth]{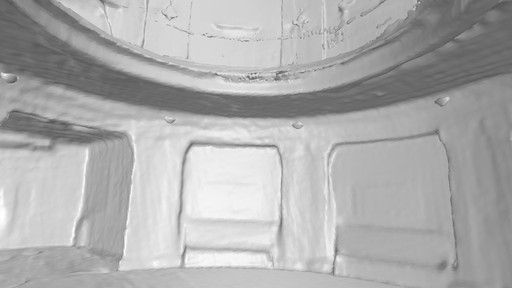}&
\includegraphics[width=\threewidth]{gfx/tnt2/rgb_Museum_000115.jpg}\\
\includegraphics[width=\threewidth]{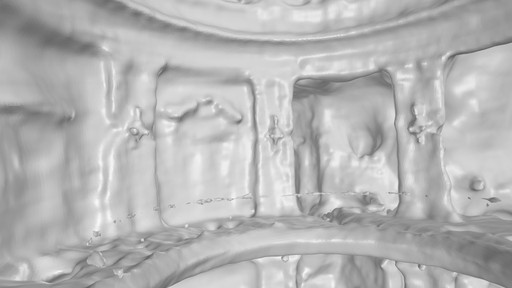}&
\includegraphics[width=\threewidth]{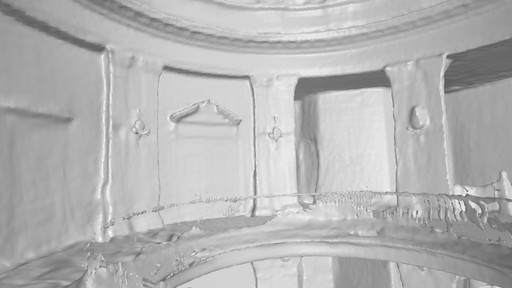}&
\includegraphics[width=\threewidth]{gfx/tnt2/rgb_Museum_000131.jpg}\\
\includegraphics[width=\threewidth]{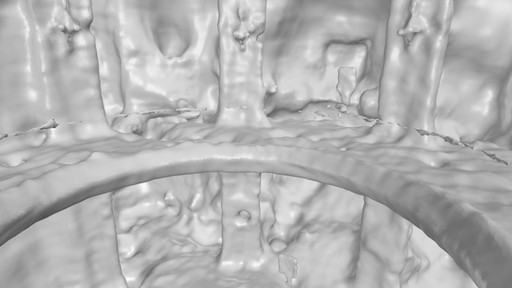}&
\includegraphics[width=\threewidth]{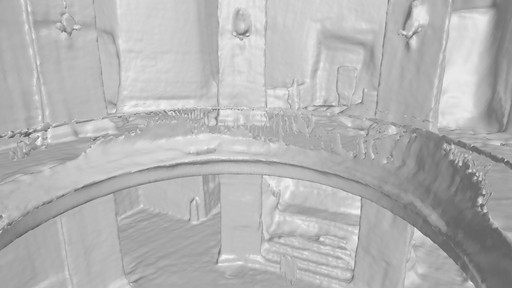}&
\includegraphics[width=\threewidth]{gfx/tnt2/rgb_Museum_000159.jpg}\\

             Multi-Res. Grids& 
             Multi-Res. Grids w/ cues&
             GT view\\
        \end{tabular}
    
    \caption{\textbf{Qualitative Comparison on Tanks \& Temples.} We use Multi-Res. Grids as the scene geometry representation and compare the reconstruction when using monocular cues or not on Courtroom and Museum.
    }
    \label{fig:tnt_4}
\end{figure*}

}

\newcommand{\dutheight}{3.1cm}
\newcommand{\fail}{
\begin{figure*}[t]
    \centering
    \setlength{\tabcolsep}{0.1em}
    
    \renewcommand{\arraystretch}{0.8}
    \hfill{}\hspace*{-0.5em}
    \footnotesize
    \begin{tabular}{cccc}
            \includegraphics[height=\dutheight]{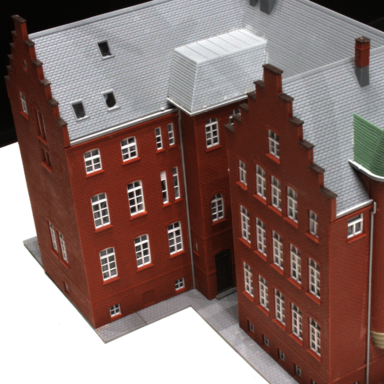} &
            \includegraphics[height=\dutheight]{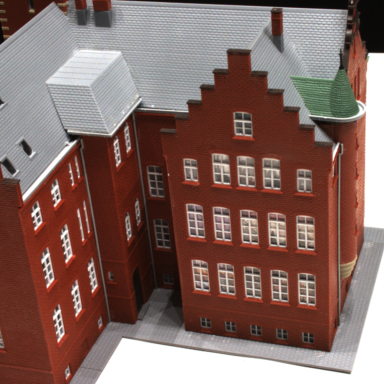} &
            \includegraphics[height=\dutheight]{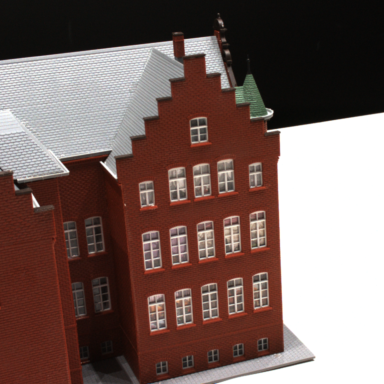} &
            \includegraphics[height=\dutheight]{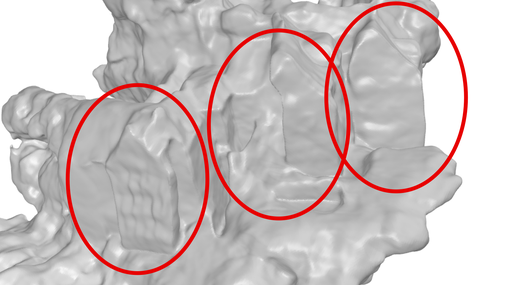}\\
            Input View 1 & Input View 2 & Input View 3 & Ours\\
    \end{tabular}
  \caption{\textbf{Failure Case on DTU Dataset with 3 Input Views.} The reconstructed mesh duplicate the object in front of each camera frustum.
  }
  \label{fig:fail}
\end{figure*}
}

\newcommand{\sixwidth}{0.161\textwidth}
\newcommand{\figuredtunvs}{
\begin{figure*}[t]
        \centering
        \setlength{\tabcolsep}{0.1em}
        \renewcommand{\arraystretch}{0.7}
        \footnotesize
        \begin{tabular}{cccccc}
            \includegraphics[width=\sixwidth]{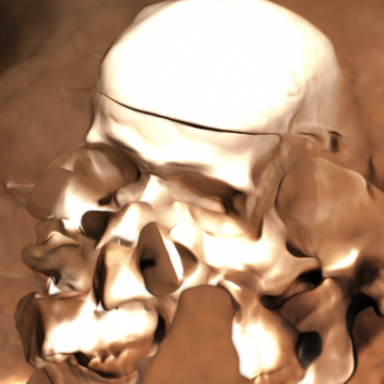}&
            \includegraphics[width=\sixwidth]{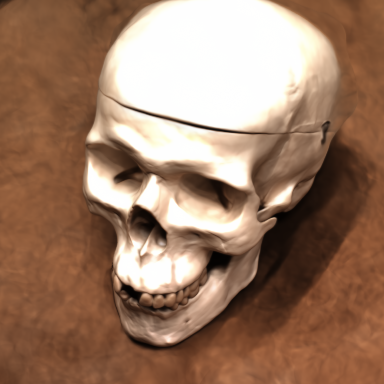}&
            \includegraphics[width=\sixwidth]{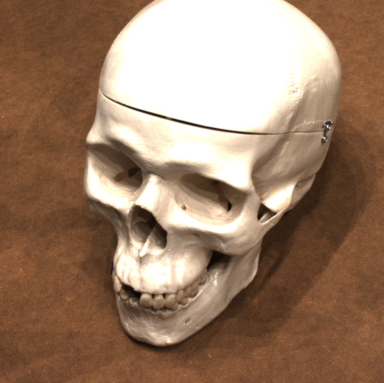}&
            \includegraphics[width=\sixwidth]{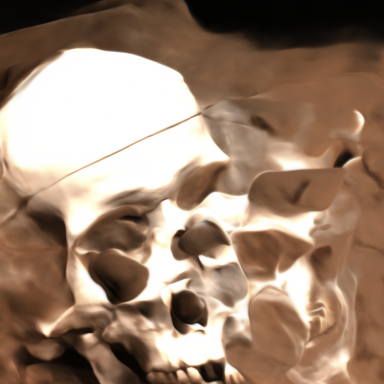}&
            \includegraphics[width=\sixwidth]{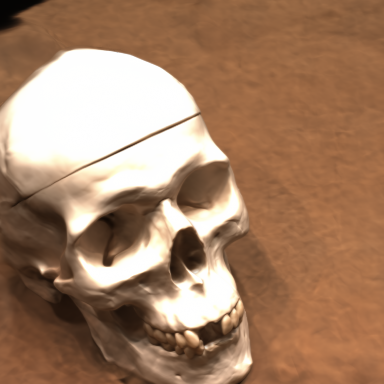}&
            \includegraphics[width=\sixwidth]{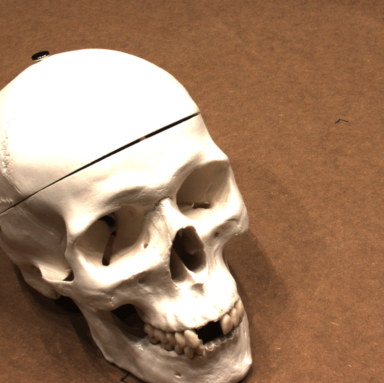}\\
            \includegraphics[width=\sixwidth]{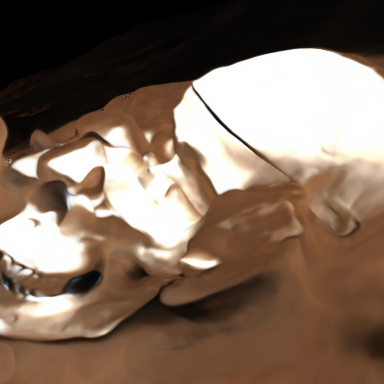}&
            \includegraphics[width=\sixwidth]{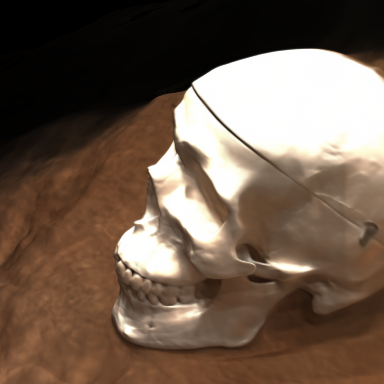}&
            \includegraphics[width=\sixwidth]{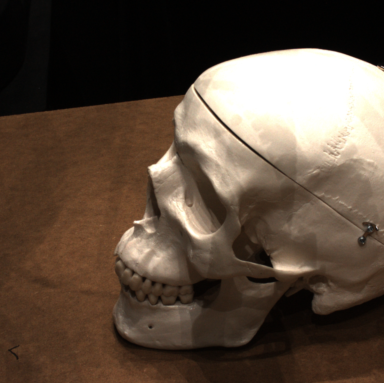}&
            \includegraphics[width=\sixwidth]{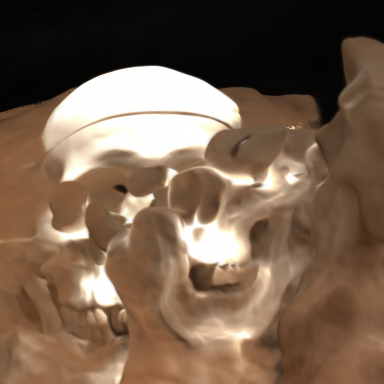}&
            \includegraphics[width=\sixwidth]{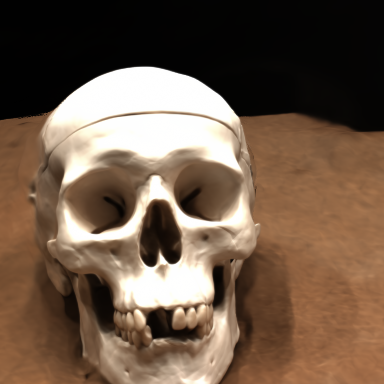}&
            \includegraphics[width=\sixwidth]{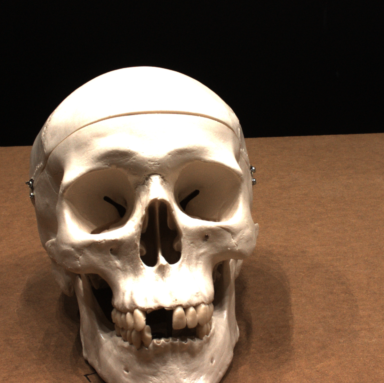}\\
            \includegraphics[width=\sixwidth]{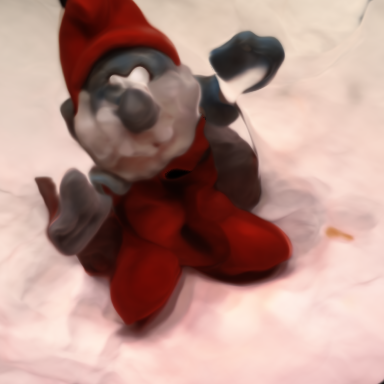}&
            \includegraphics[width=\sixwidth]{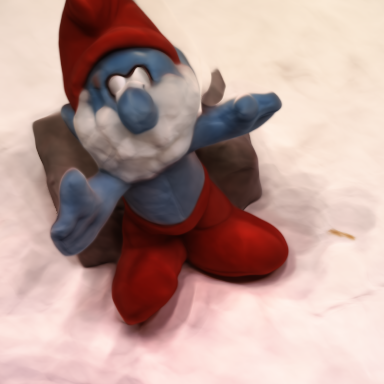}&
            \includegraphics[width=\sixwidth]{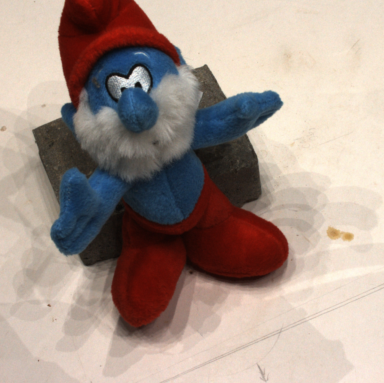}&
            \includegraphics[width=\sixwidth]{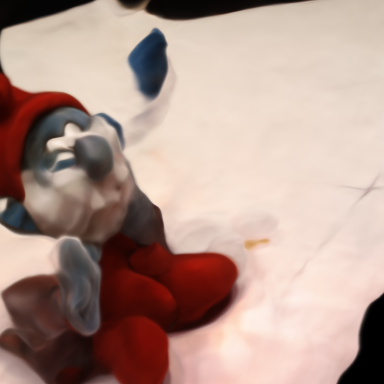}&
            \includegraphics[width=\sixwidth]{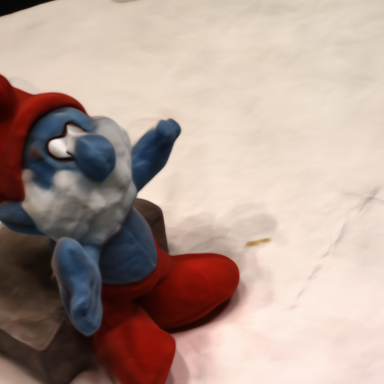}&
            \includegraphics[width=\sixwidth]{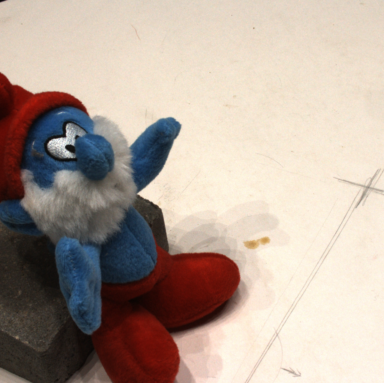}\\
            \includegraphics[width=\sixwidth]{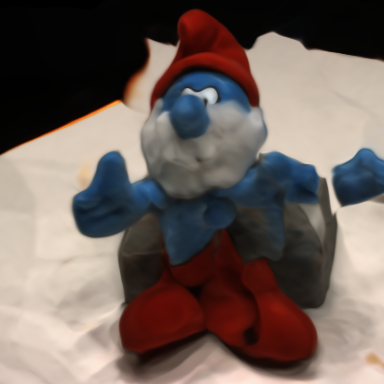}&
            \includegraphics[width=\sixwidth]{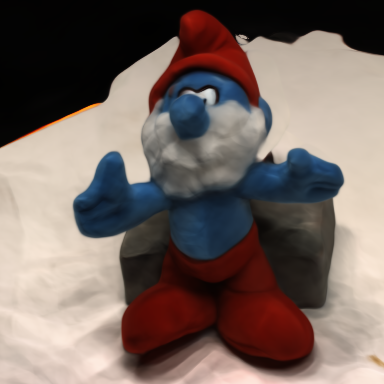}&
            \includegraphics[width=\sixwidth]{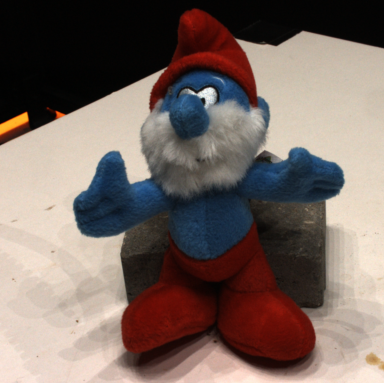}&
            \includegraphics[width=\sixwidth]{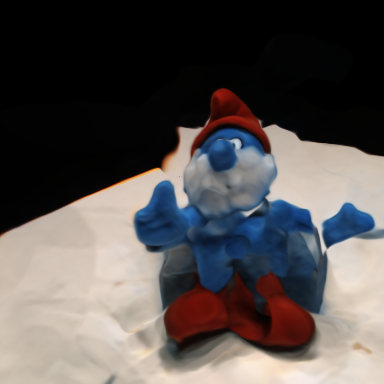}&
            \includegraphics[width=\sixwidth]{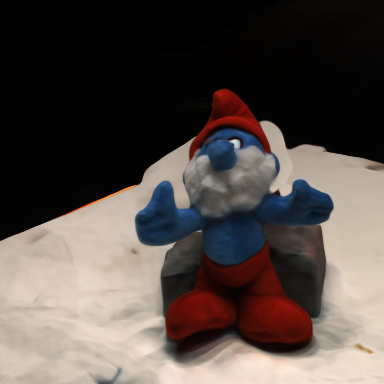}&
            \includegraphics[width=\sixwidth]{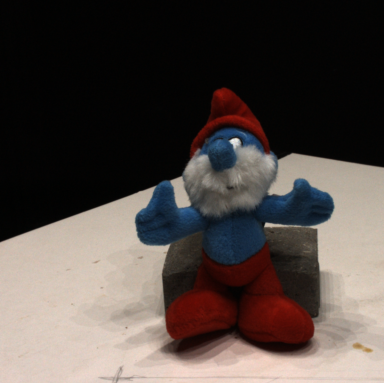}\\
            \includegraphics[width=\sixwidth]{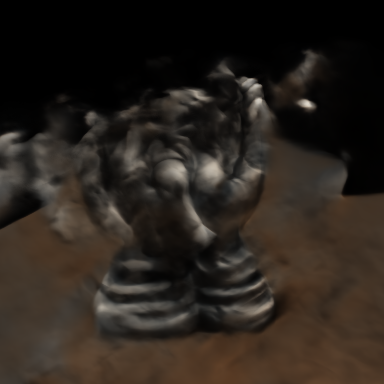}&
            \includegraphics[width=\sixwidth]{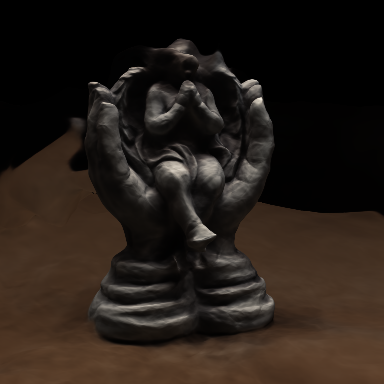}&
            \includegraphics[width=\sixwidth]{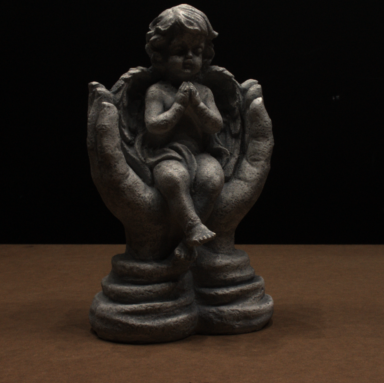}&
            \includegraphics[width=\sixwidth]{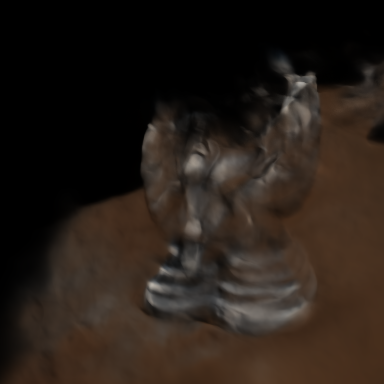}&
            \includegraphics[width=\sixwidth]{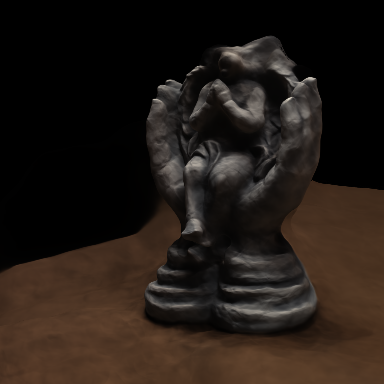}&
            \includegraphics[width=\sixwidth]{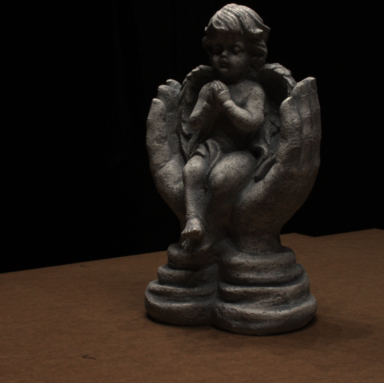}\\
            \includegraphics[width=\sixwidth]{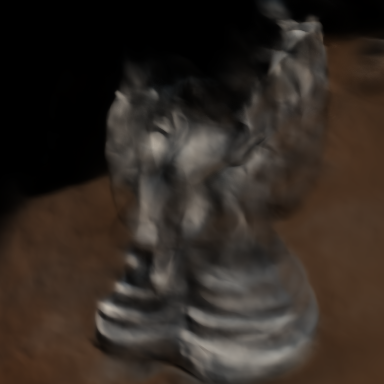}&
            \includegraphics[width=\sixwidth]{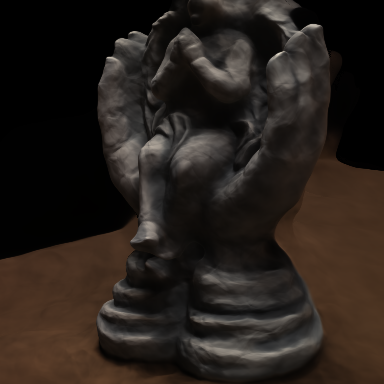}&
            \includegraphics[width=\sixwidth]{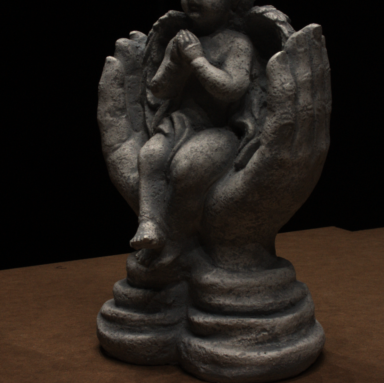}&
            \includegraphics[width=\sixwidth]{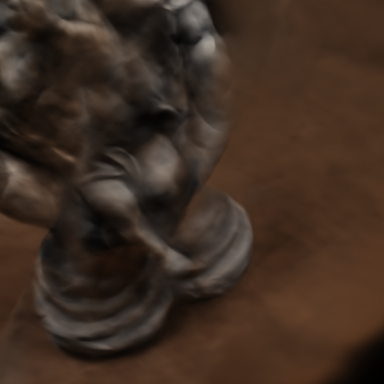}&
            \includegraphics[width=\sixwidth]{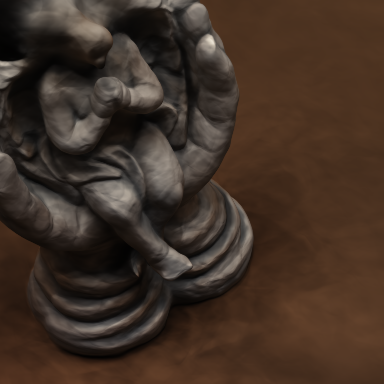}&
            \includegraphics[width=\sixwidth]{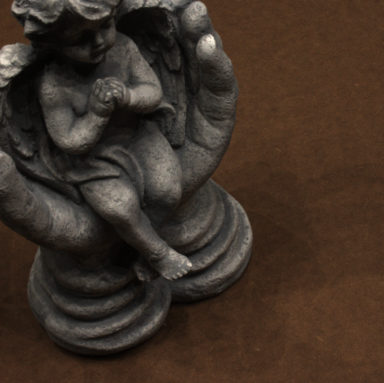}\\
            \includegraphics[width=\sixwidth]{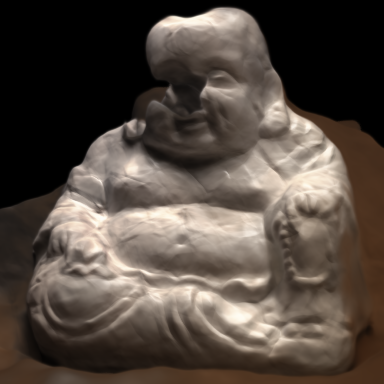}&
            \includegraphics[width=\sixwidth]{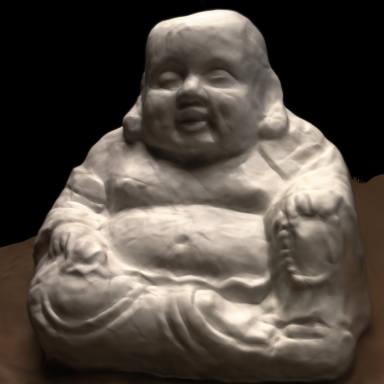}&
            \includegraphics[width=\sixwidth]{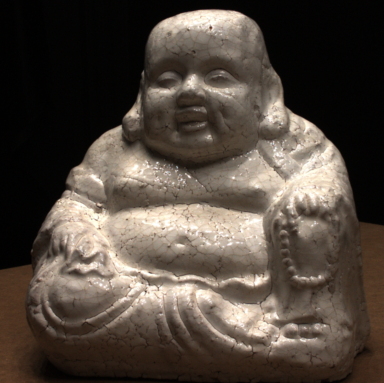}&
            \includegraphics[width=\sixwidth]{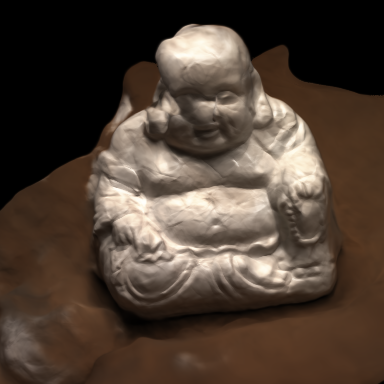}&
            \includegraphics[width=\sixwidth]{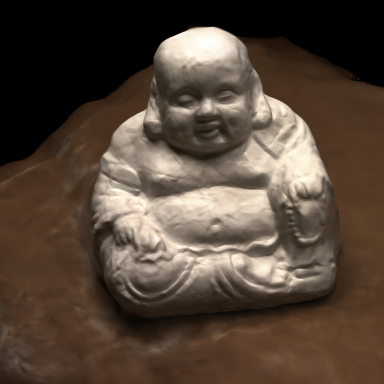}&
            \includegraphics[width=\sixwidth]{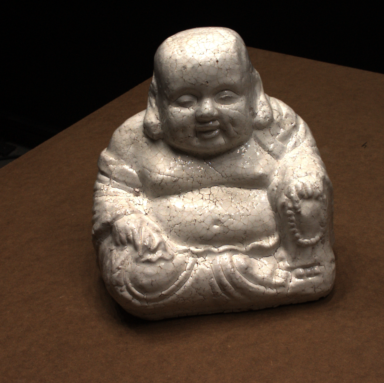}\\
            \includegraphics[width=\sixwidth]{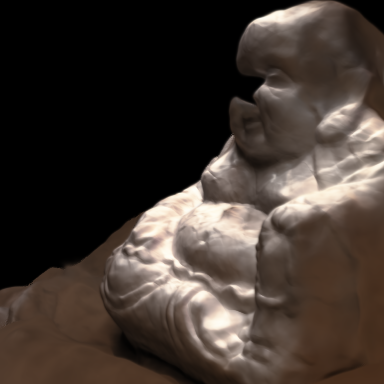}&
            \includegraphics[width=\sixwidth]{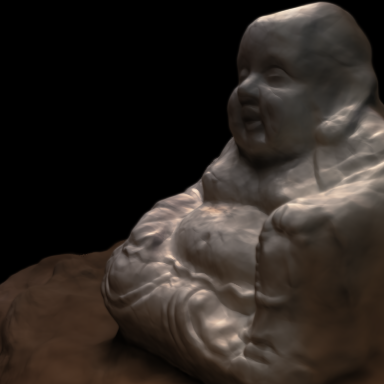}&
            \includegraphics[width=\sixwidth]{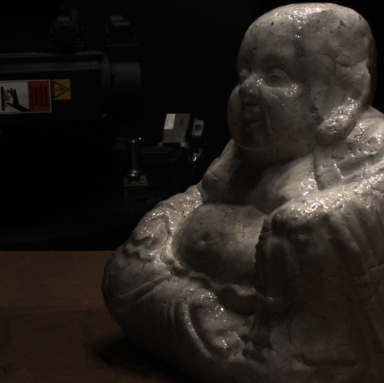}&
            \includegraphics[width=\sixwidth]{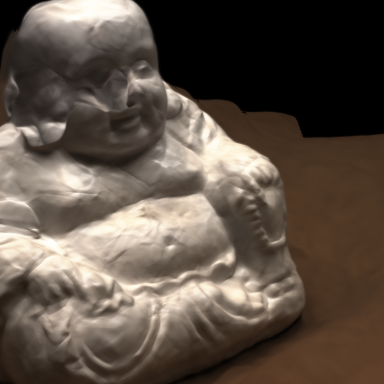}&
            \includegraphics[width=\sixwidth]{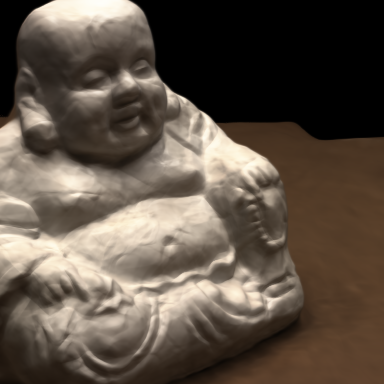}&
            \includegraphics[width=\sixwidth]{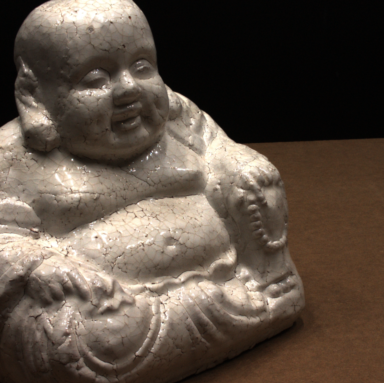}\\
            \includegraphics[width=\sixwidth]{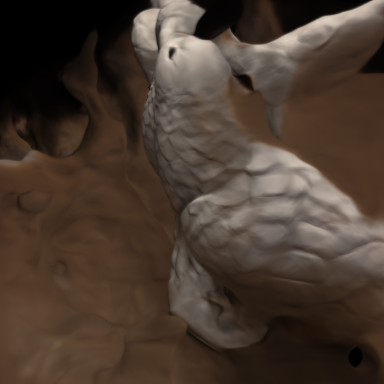}&
            \includegraphics[width=\sixwidth]{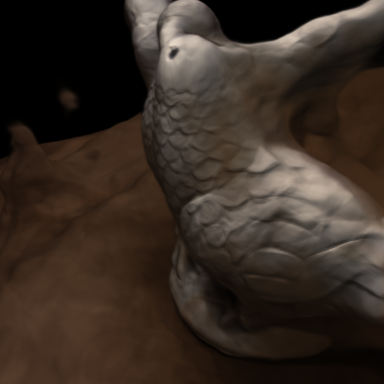}&
            \includegraphics[width=\sixwidth]{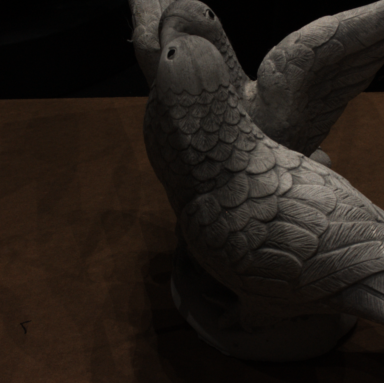}&
            \includegraphics[width=\sixwidth]{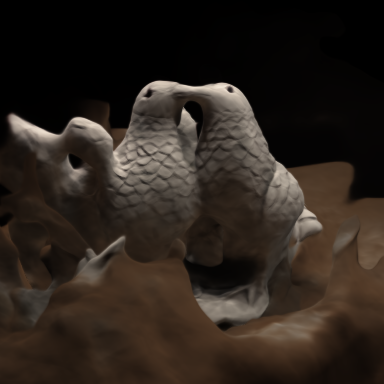}&
            \includegraphics[width=\sixwidth]{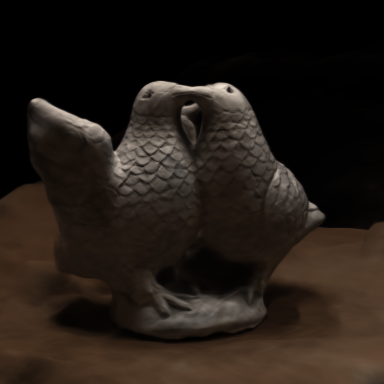}&
            \includegraphics[width=\sixwidth]{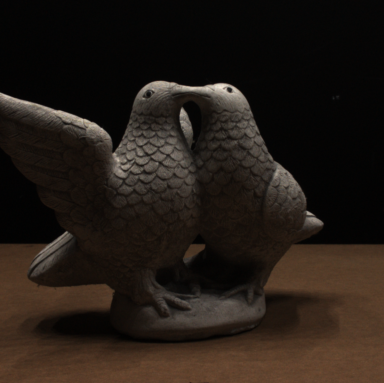}\\
             MLP~\cite{Yariv2021NEURIPS}& 
             MLP w/cues& 
             GT View& 
             MLP~\cite{Yariv2021NEURIPS}& 
             MLP w/cues& 
             GT View\\
        \end{tabular}
    \caption{\red{\textbf{Qualitative Comparison of Novel View Synthesis on the DTU Dataset with 3 Input Views.} Adding monocular geometric cues improves novel view synthesis quality.  }}
    \label{fig:dtu_nvs}
\end{figure*}
}

\newcommand{\figureablationdiffview}{
\begin{figure*}[t]
        \centering
        \setlength{\tabcolsep}{0.1em}
        \renewcommand{\arraystretch}{0.7}
        \hfill{}\hspace*{-0.5em}
        \begin{tabular}{cccccc}
             &10 Views &
             20 Views  &
             30 Views&
             40 Views &
             GT Views\\
            \rot{\; MLP}&
            \includegraphics[width=\fivewidth]{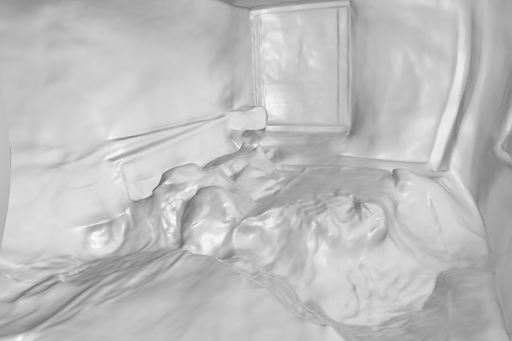}&
            \includegraphics[width=\fivewidth]{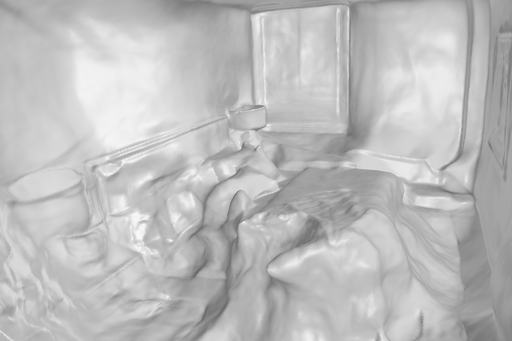}&
            \includegraphics[width=\fivewidth]{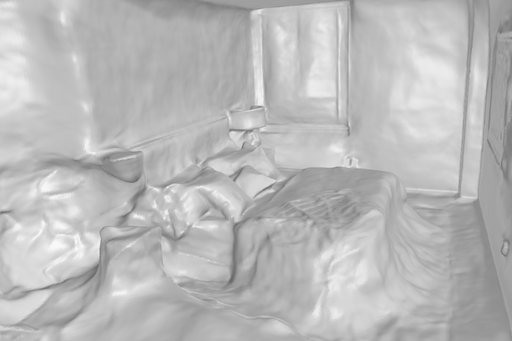}&
            \includegraphics[width=\fivewidth]{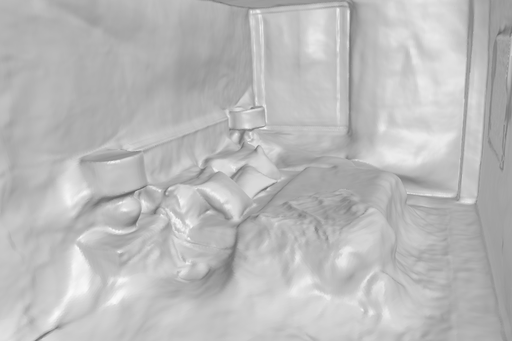}&
            \includegraphics[width=\fivewidth]{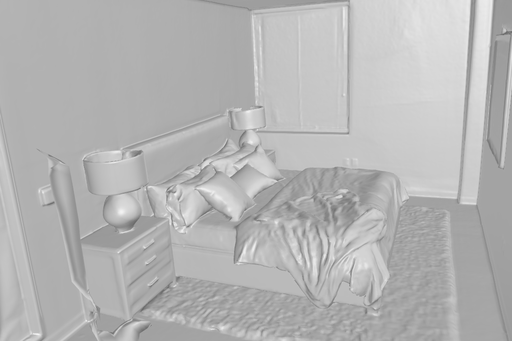}\\
            & 28.5 &
             44.6  &
             61.6&
             61.4\\
            \rot{ w/ Cues}&
            \includegraphics[width=\fivewidth]{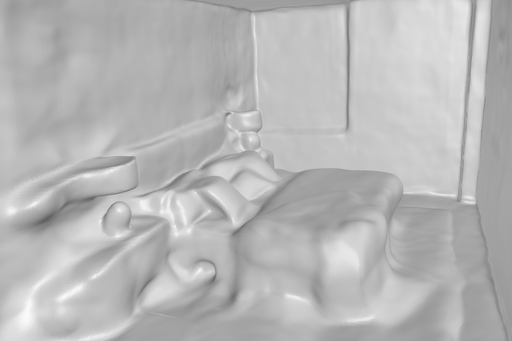}&
            \includegraphics[width=\fivewidth]{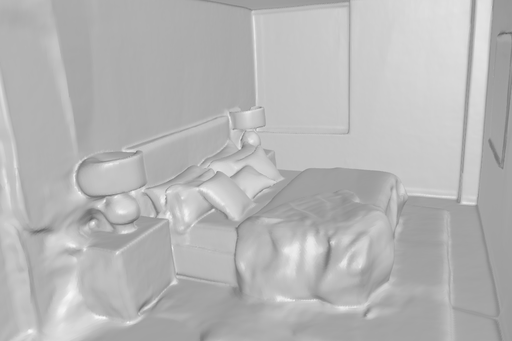}&
            \includegraphics[width=\fivewidth]{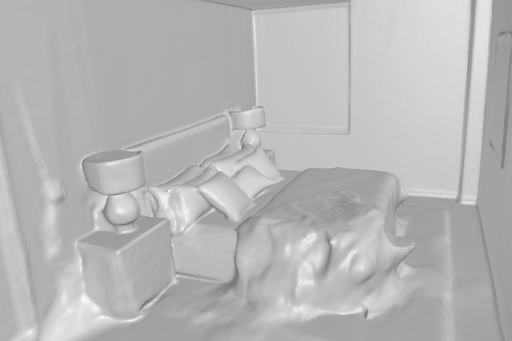}&
            \includegraphics[width=\fivewidth]{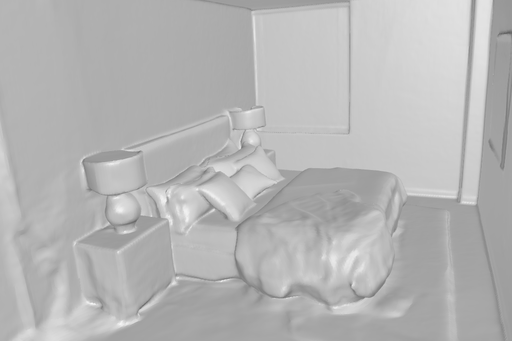}&
            \includegraphics[width=\fivewidth]{gfx/diffviews/diffview_room1_GT.ply.png}\\
             & 75.4 &
             89.8  &
             90.9&
             92.8\\
            
        \end{tabular}
        \caption{
        \red{\textbf{Ablation of Different Number of Input Views on the Replica Dataset.} We show F-score under each image. 
        We observe that using more input views for training improves reconstruction quality.
        Further, adding monocular geometric cues improves reconstruction quality. When using only 10 input views, the MLP fails to reconstruct reasonable results while using monocular geometric cues significantly improves results.
        }}
        \label{fig:ablation_diffviews}
\end{figure*}
}

\newcommand{\figureablationweightdecay}{
\begin{figure*}[t]
        \centering
        \setlength{\tabcolsep}{0.1em}
        \renewcommand{\arraystretch}{0.7}
        \hfill{}\hspace*{-0.5em}
        \begin{tabular}{ccc}
            \includegraphics[width=\threewidth]{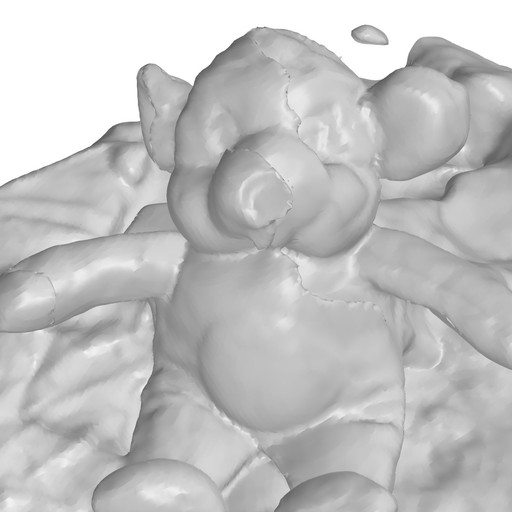}&
            \includegraphics[width=\threewidth]{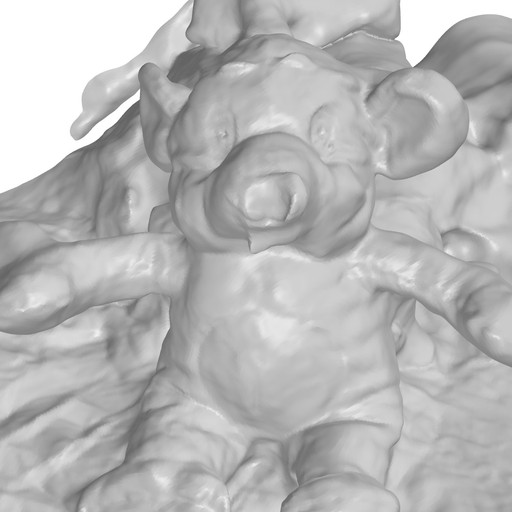}&
            \includegraphics[width=\threewidth]{gfx/DTU_vis/scan105_rgb.png}\\
            \includegraphics[width=\threewidth]{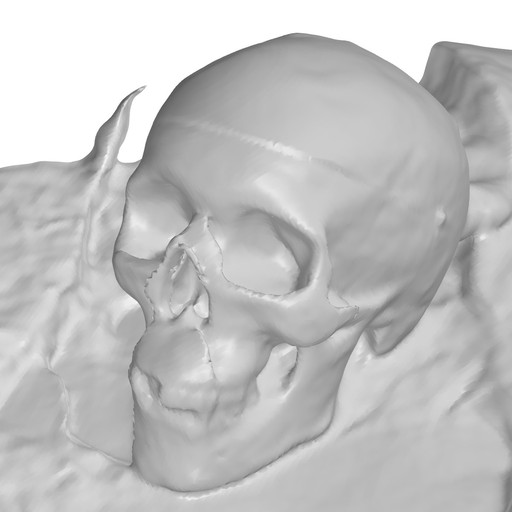}&
            \includegraphics[width=\threewidth]{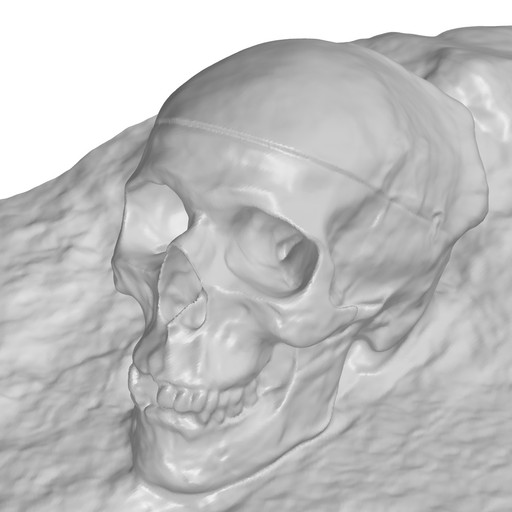}&
            \includegraphics[width=\threewidth]{gfx/DTU_vis/scan65_rgb.png}\\
             Without Weight Annealing &
             With Weight Annealing  &
             GT View\\
        \end{tabular}
        \caption{
        \red{\textbf{Ablation of Weight Annealing on the DTU Dataset with 3 Input Views. }Using weight schedule improves reconstruction quality.} 
        }
        \label{fig:ablation_weightdecay}
\end{figure*}
}

\newcommand{\figureablationmoreconfig}{
\begin{figure*}[t]
  \centering
  \includegraphics[width=0.9\textwidth]{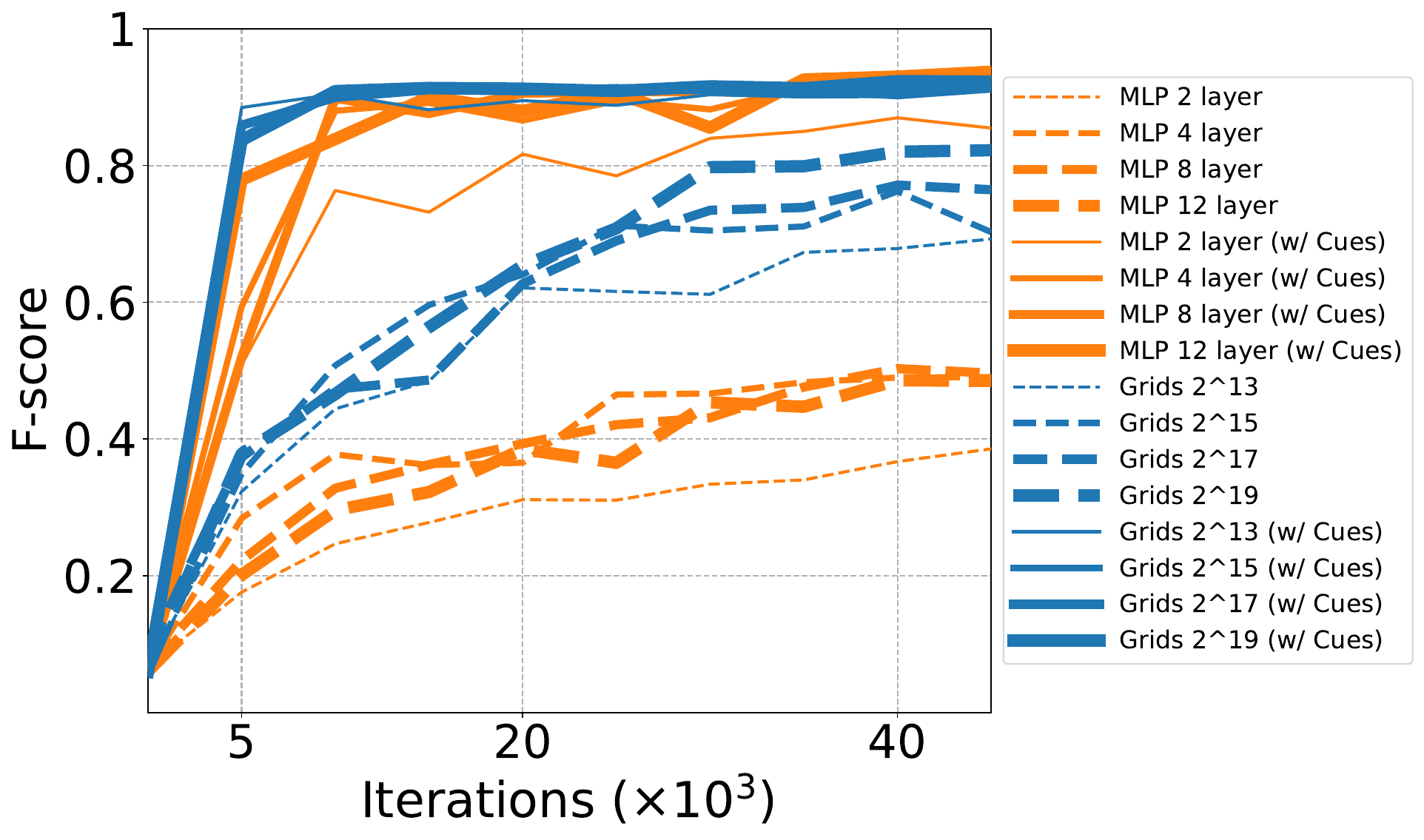}
  \caption{\red{\textbf{Optimization Processes Using Different Architecture Configurations.} Using monocular geometric cues improves reconstruction quality and convergence speed independent of the network configurations.}
  }
  \label{fig:ablation_moreconfig}
 \end{figure*}
}

\newcommand{\figurehighrescuesvis}{
\begin{figure*}\centering
    \begin{subfigure}{.98\linewidth}
     \includegraphics[width=1.0\linewidth]{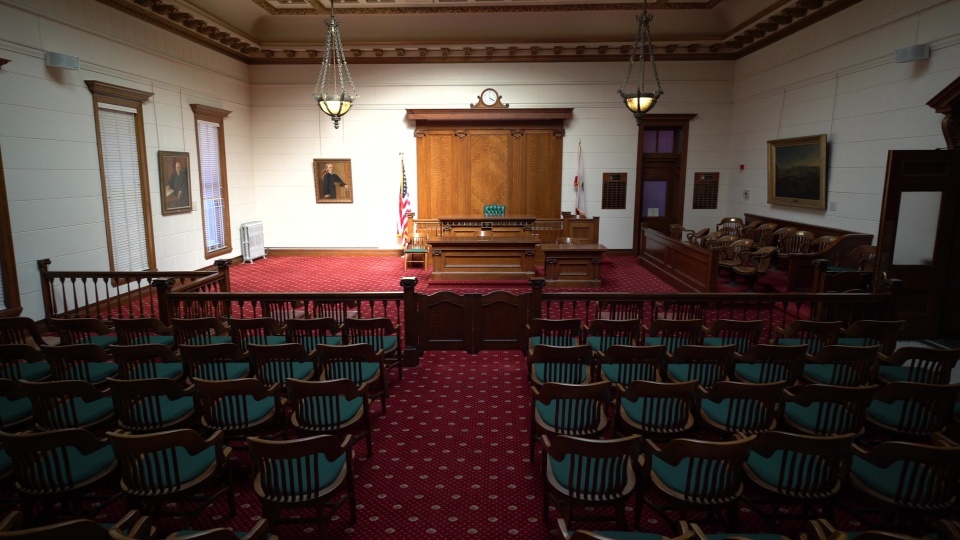}
        \caption{RGB Image.}
    \end{subfigure}
    
    \begin{subfigure}{.49\linewidth}
      \includegraphics[width=0.98\linewidth]{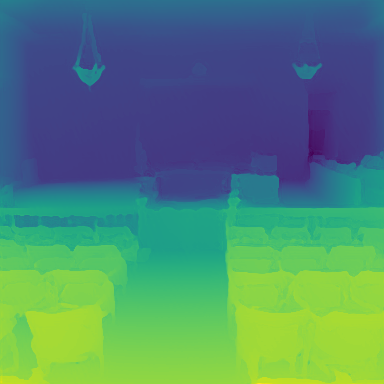}
        \caption{Low Resolution Depth Map.}
    \end{subfigure}
    \begin{subfigure}{.49\linewidth}
     \includegraphics[width=0.98\linewidth]{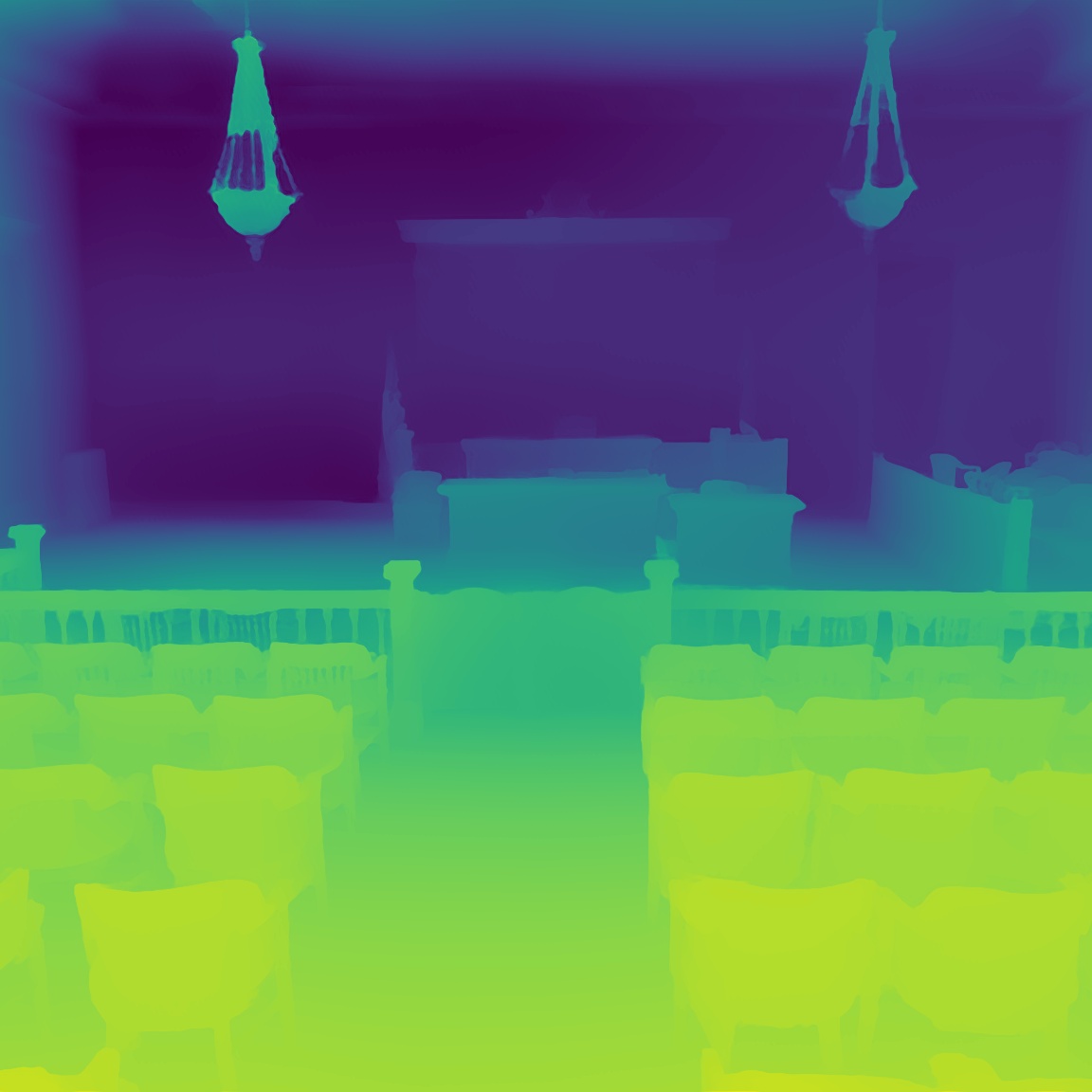}
        \caption{High Resolution Depth Map.}
    \end{subfigure}
    \begin{subfigure}{.49\linewidth}
     \includegraphics[width=0.98\linewidth]{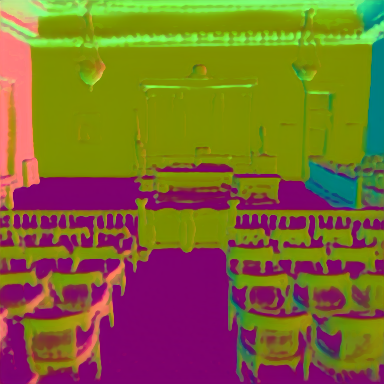}
        \caption{Low Resolution Normal Map.}
    \end{subfigure}
    \begin{subfigure}{.49\linewidth}
     \includegraphics[width=0.98\linewidth]{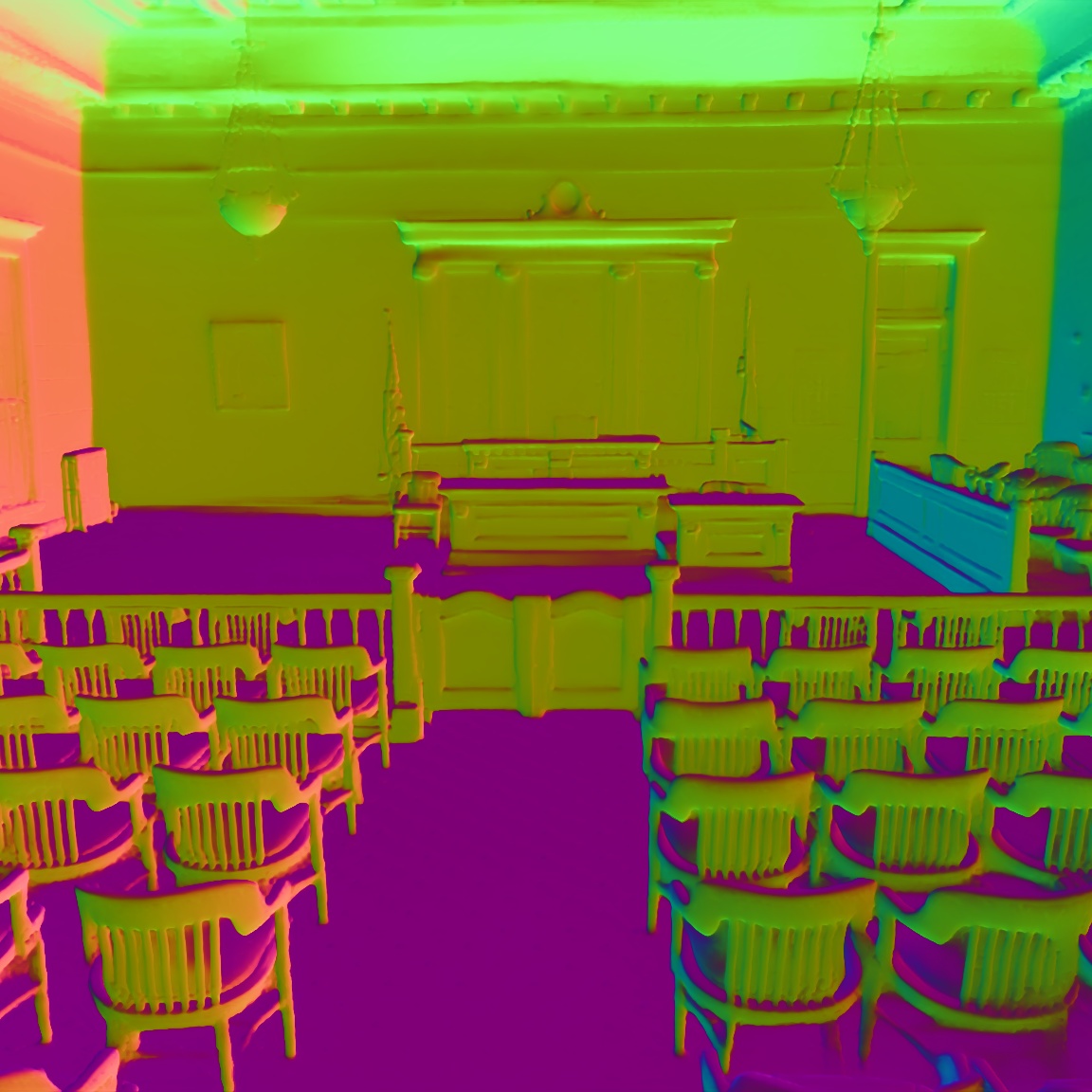}
        \caption{High Resolution Normal Map.}
    \end{subfigure}
    
\caption{\red{\textbf{Visual Comparison of Different Resolution Monocular Cues.}}}
 \label{fig:high_res_cues_vis}
\end{figure*}
}

\newcommand{\figuretnthighres}{
\begin{figure*}[t]
        \centering
        \setlength{\tabcolsep}{0.1em}
        \renewcommand{\arraystretch}{0.7}
        \footnotesize
        \begin{tabular}{ccc}
        
\includegraphics[width=\threewidth]{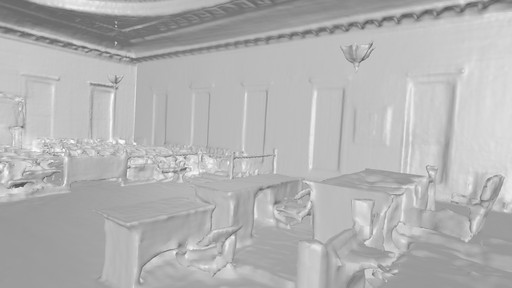}&
\includegraphics[width=\threewidth]{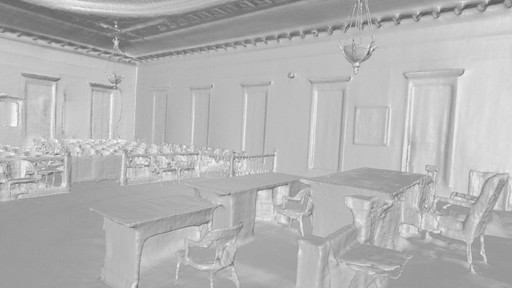}&
\includegraphics[width=\threewidth]{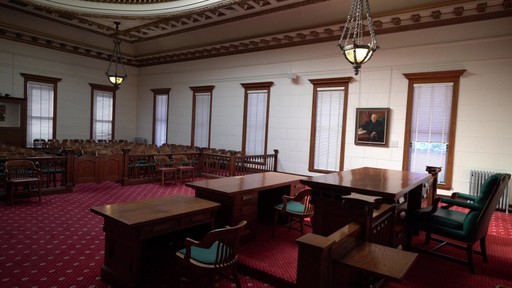}\\
\includegraphics[width=\threewidth]{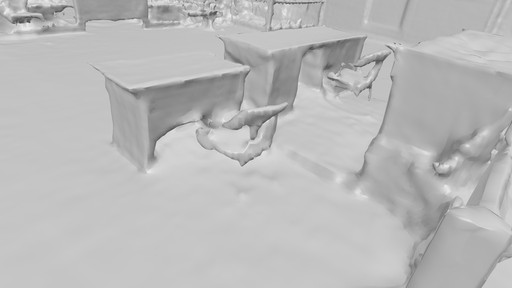}&
\includegraphics[width=\threewidth]{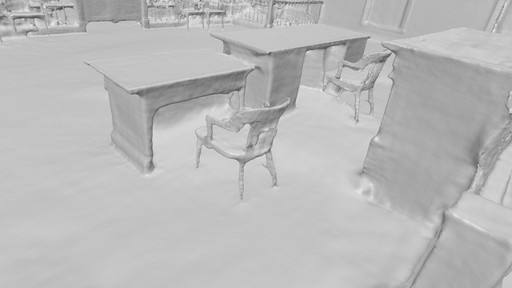}&
\includegraphics[width=\threewidth]{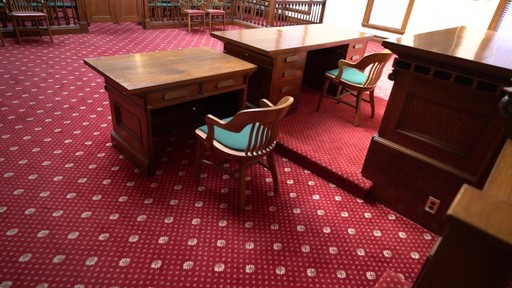}\\
\includegraphics[width=\threewidth]{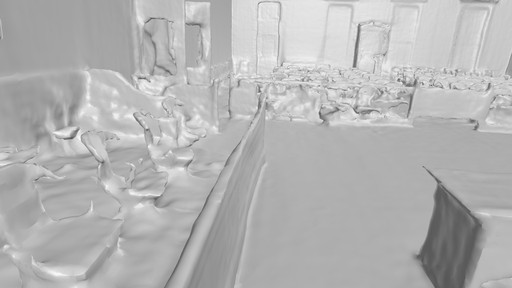}&
\includegraphics[width=\threewidth]{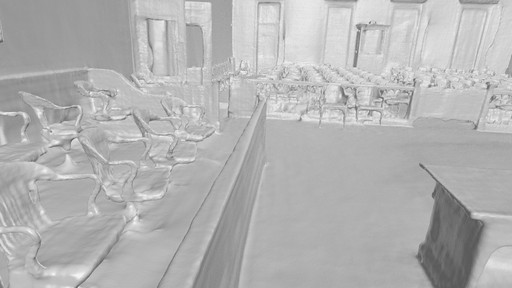}&
\includegraphics[width=\threewidth]{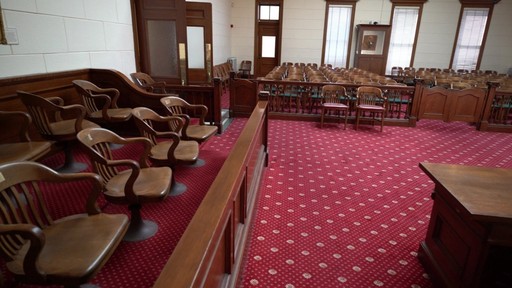}\\
\includegraphics[width=\threewidth]{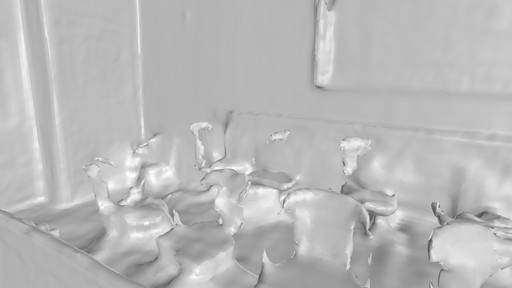}&
\includegraphics[width=\threewidth]{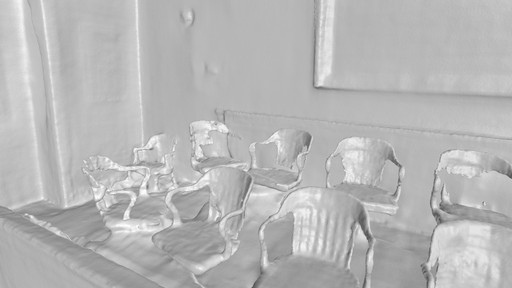}&
\includegraphics[width=\threewidth]{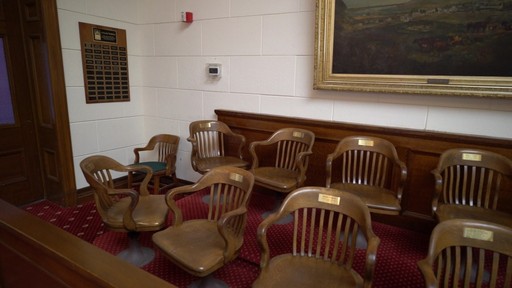}\\
\includegraphics[width=\threewidth]{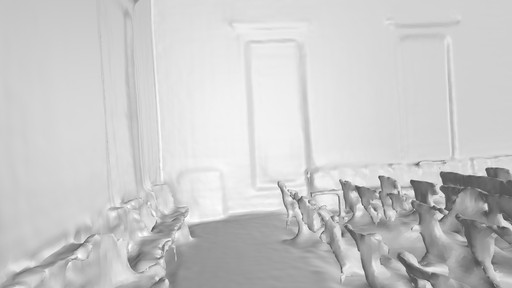}&
\includegraphics[width=\threewidth]{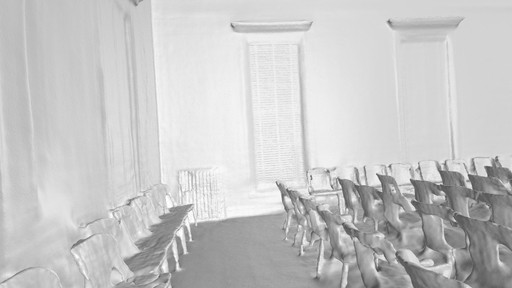}&
\includegraphics[width=\threewidth]{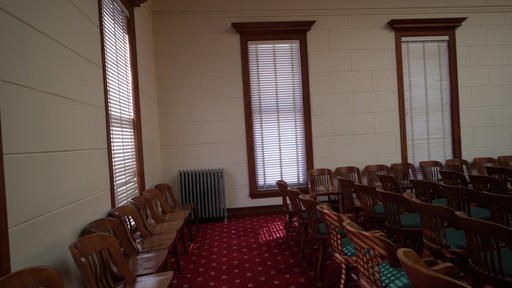}\\
\includegraphics[width=\threewidth]{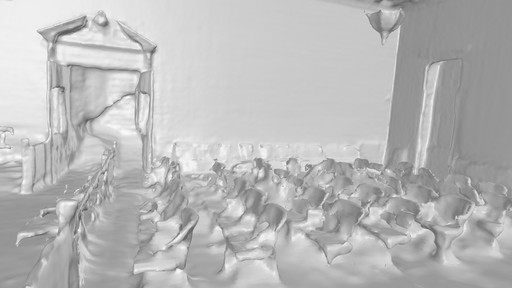}&
\includegraphics[width=\threewidth]{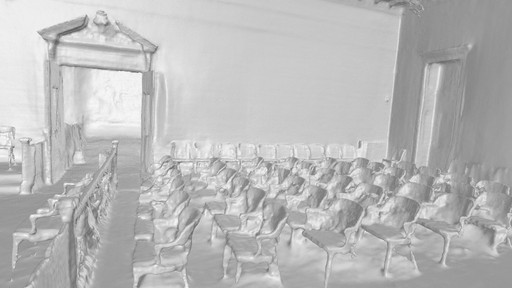}&
\includegraphics[width=\threewidth]{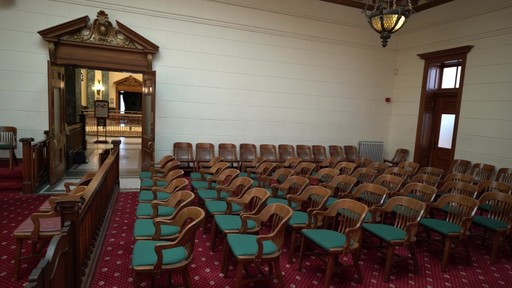}\\
\includegraphics[width=\threewidth]{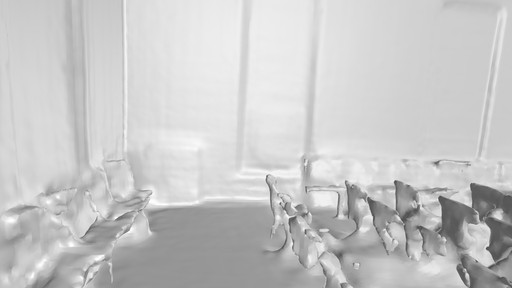}&
\includegraphics[width=\threewidth]{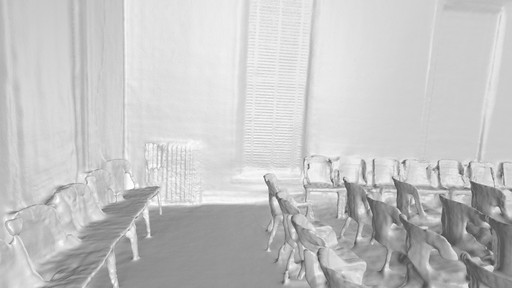}&
\includegraphics[width=\threewidth]{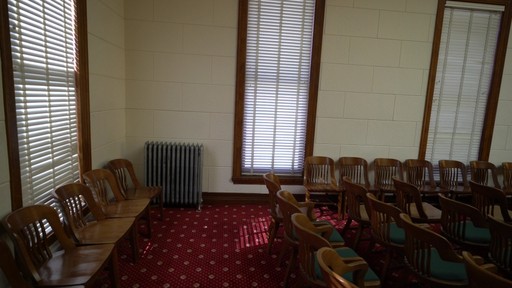}\\
\includegraphics[width=\threewidth]{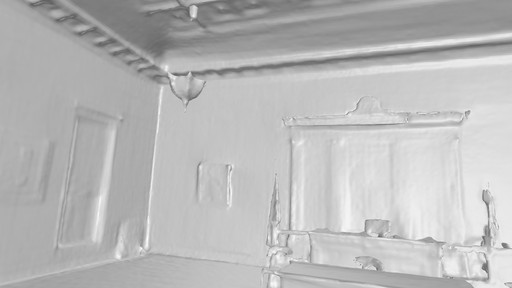}&
\includegraphics[width=\threewidth]{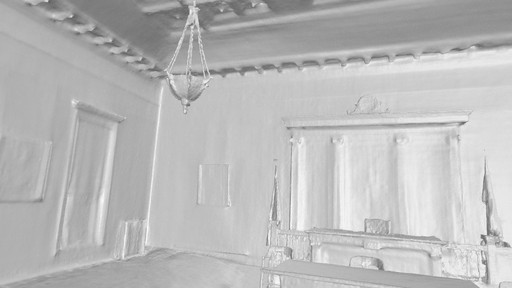}&
\includegraphics[width=\threewidth]{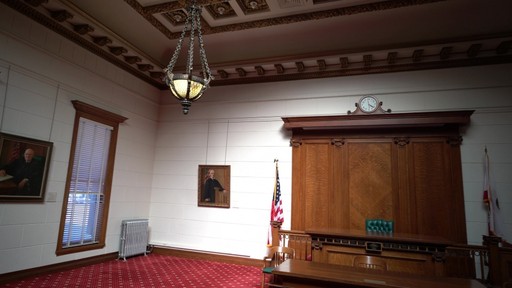}\\

             Low Resolution Cues& 
             High Resolution Cues&
             GT view\\
        \end{tabular}
    
    \caption{\red{\textbf{Qualitative Comparison of Low Resolution Cues and High Resolution cues on Tanks \& Temples.} We use Multi-Res. Grids as the scene geometry representation and compare the reconstruction when using different resolution of monocular cues.}
    }
    \label{fig:tnt_highres}
\end{figure*}

}

\newcommand{\figuredtuallview}{
\begin{figure*}[t]
        \centering
        \setlength{\tabcolsep}{0.1em}
        \renewcommand{\arraystretch}{0.7}
        \footnotesize
        \begin{tabular}{ccccc}

\includegraphics[width=\fivewidth]{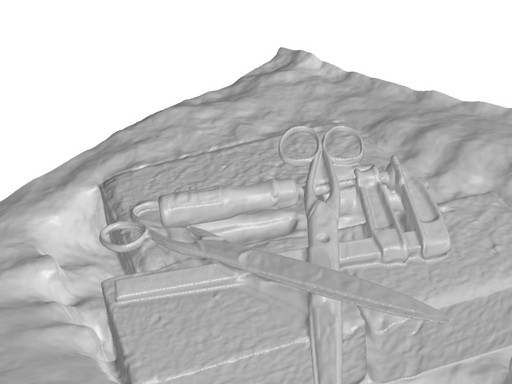}&
\includegraphics[width=\fivewidth]{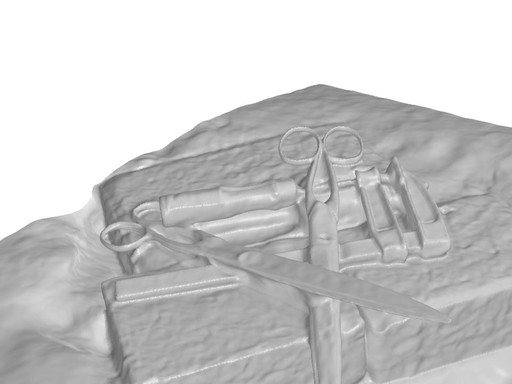}&
\includegraphics[width=\fivewidth]{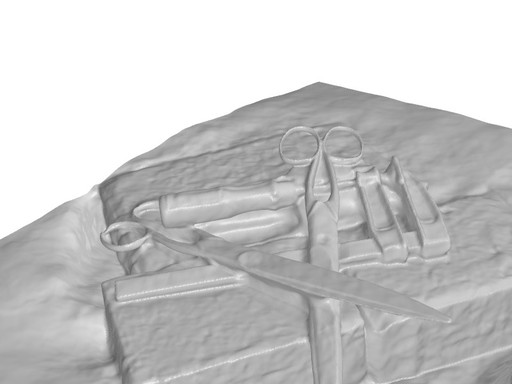}&
\includegraphics[width=\fivewidth]{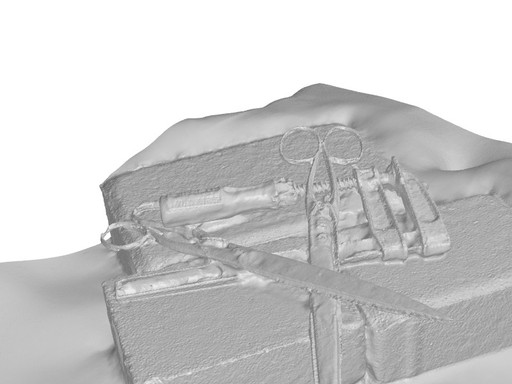}&
\includegraphics[width=\fivewidth]{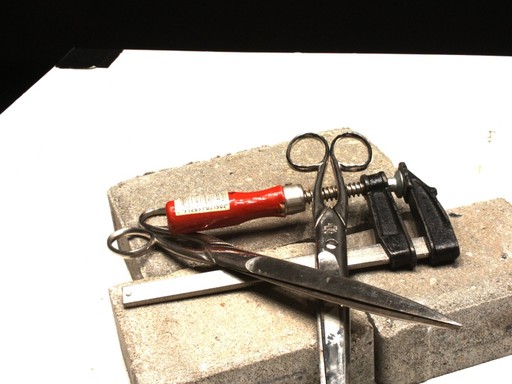}\\
\includegraphics[width=\fivewidth]{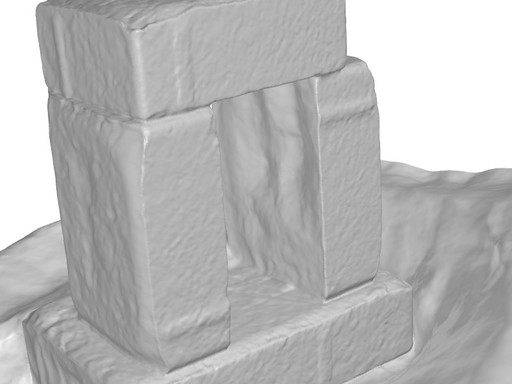}&
\includegraphics[width=\fivewidth]{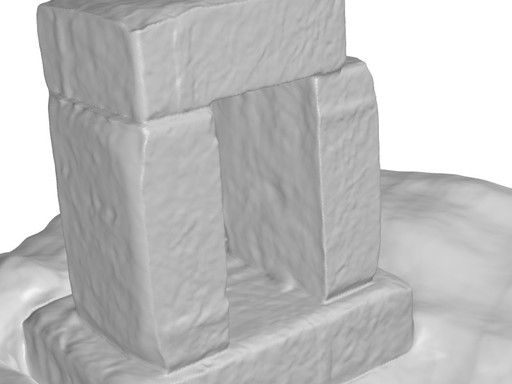}&
\includegraphics[width=\fivewidth]{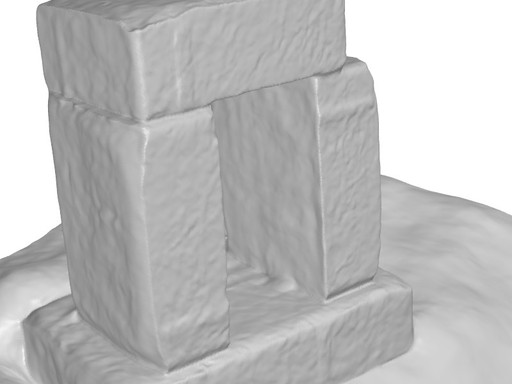}&
\includegraphics[width=\fivewidth]{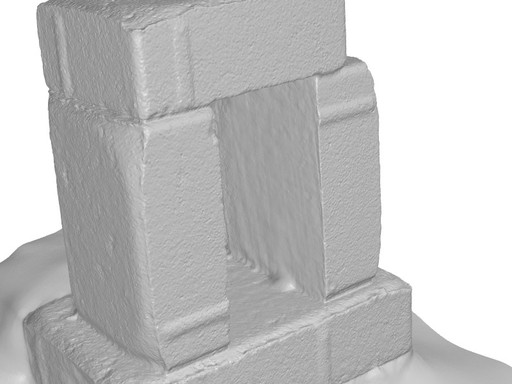}&
\includegraphics[width=\fivewidth]{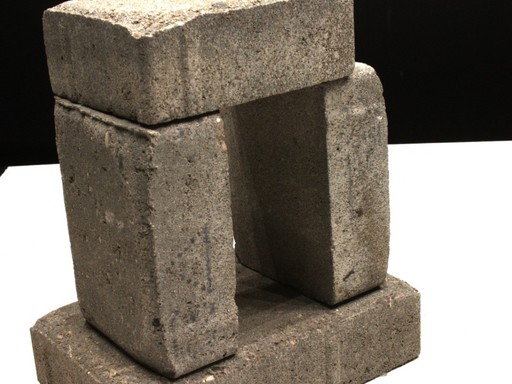}\\
\includegraphics[width=\fivewidth]{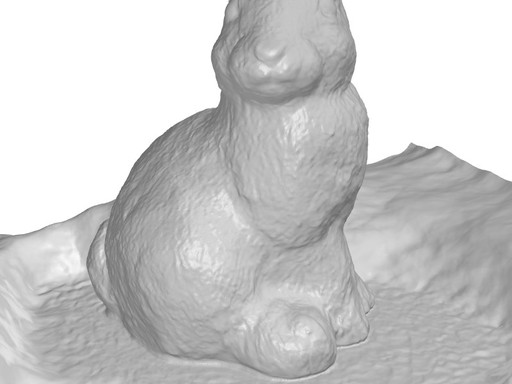}&
\includegraphics[width=\fivewidth]{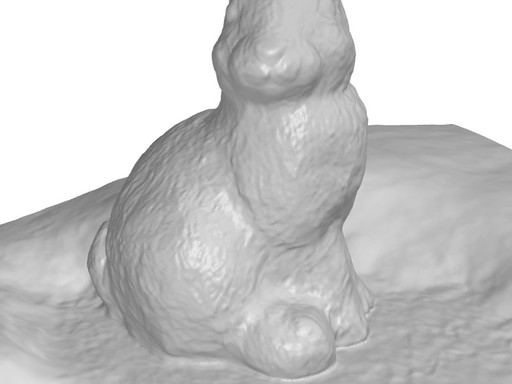}&
\includegraphics[width=\fivewidth]{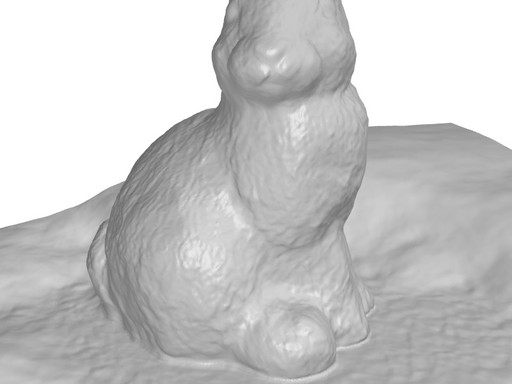}&
\includegraphics[width=\fivewidth]{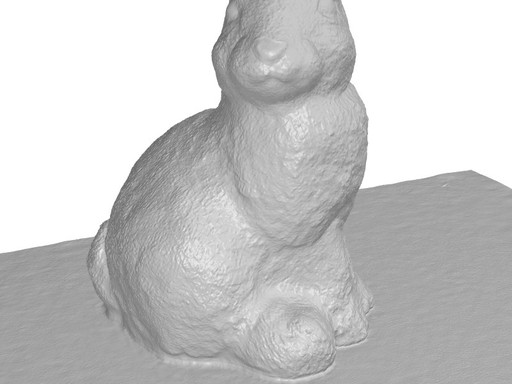}&
\includegraphics[width=\fivewidth]{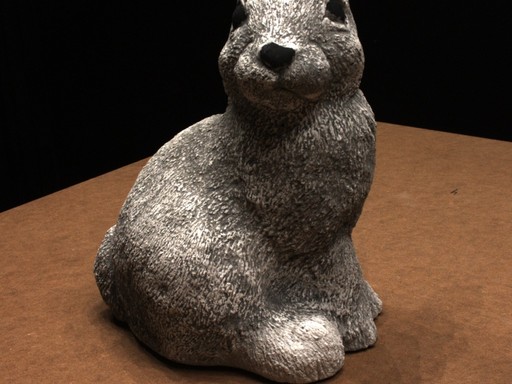}\\
\includegraphics[width=\fivewidth]{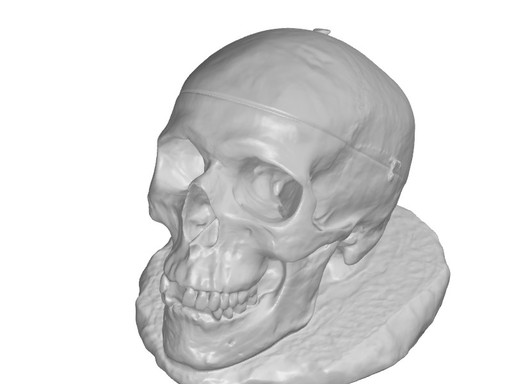}&
\includegraphics[width=\fivewidth]{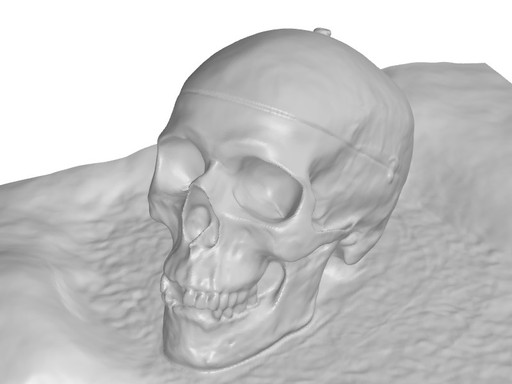}&
\includegraphics[width=\fivewidth]{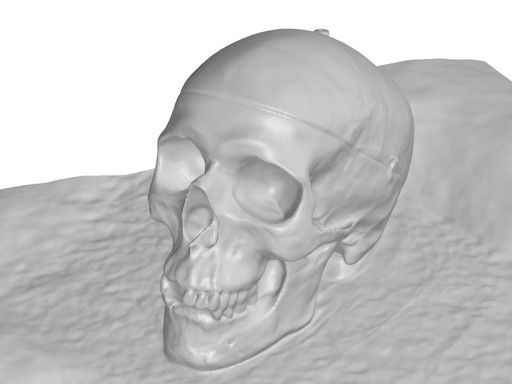}&
\includegraphics[width=\fivewidth]{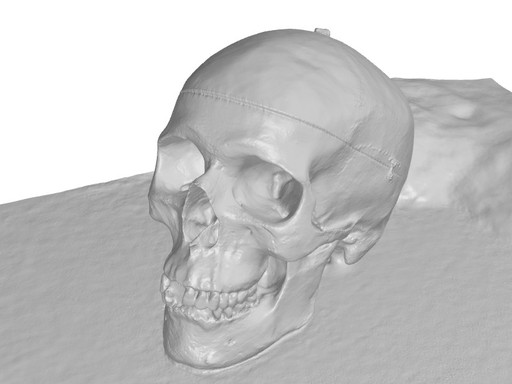}&
\includegraphics[width=\fivewidth]{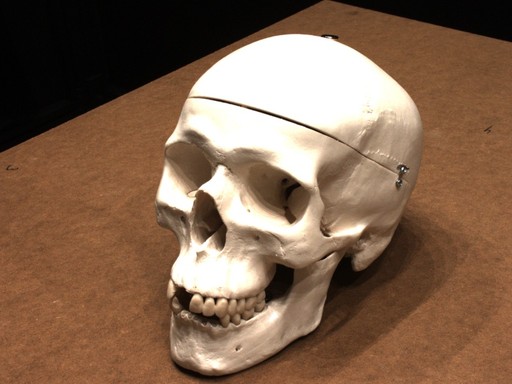}\\
\includegraphics[width=\fivewidth]{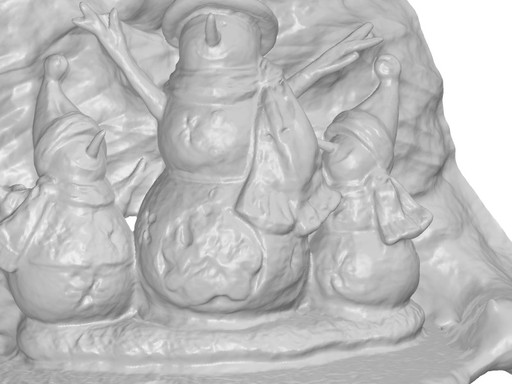}&
\includegraphics[width=\fivewidth]{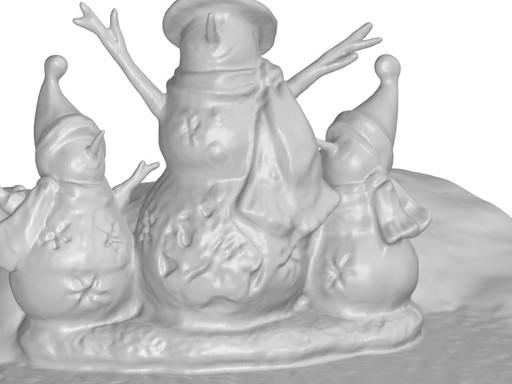}&
\includegraphics[width=\fivewidth]{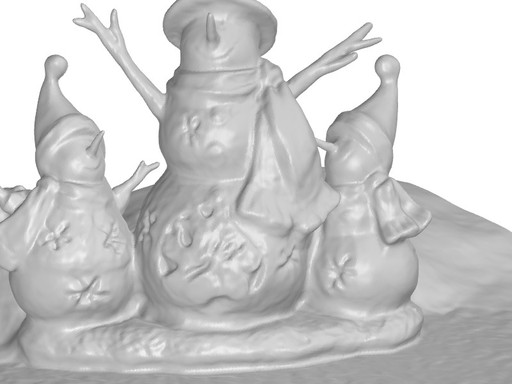}&
\includegraphics[width=\fivewidth]{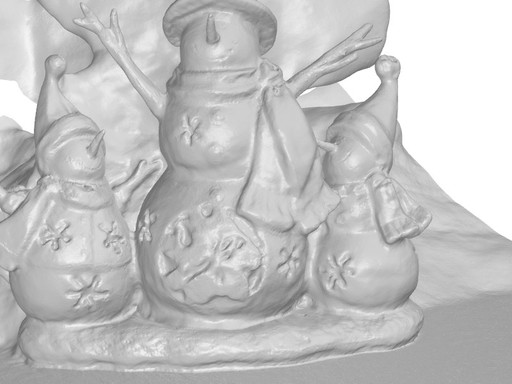}&
\includegraphics[width=\fivewidth]{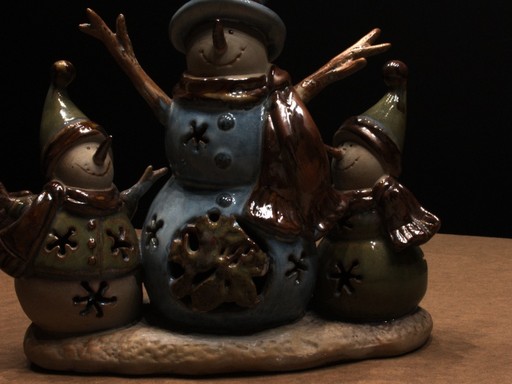}\\
\includegraphics[width=\fivewidth]{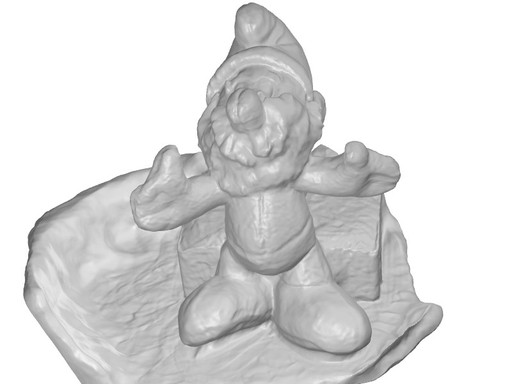}&
\includegraphics[width=\fivewidth]{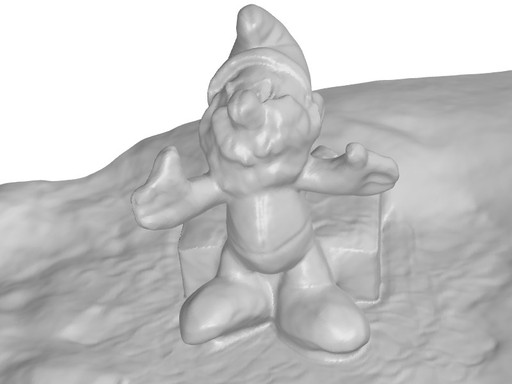}&
\includegraphics[width=\fivewidth]{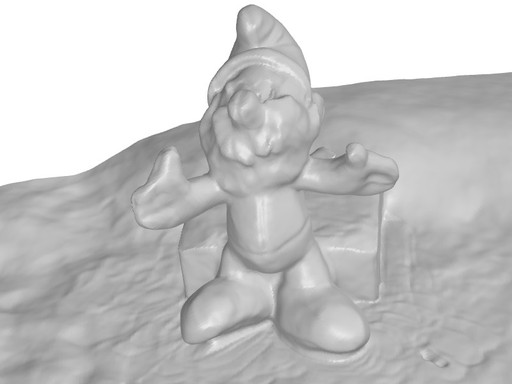}&
\includegraphics[width=\fivewidth]{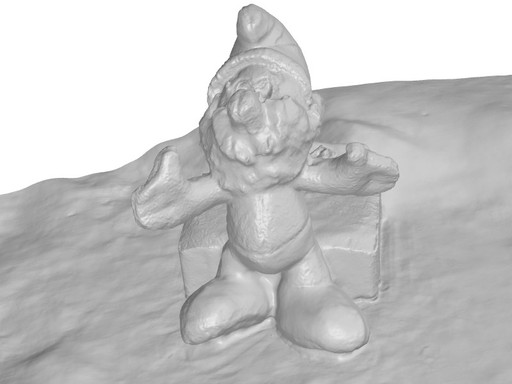}&
\includegraphics[width=\fivewidth]{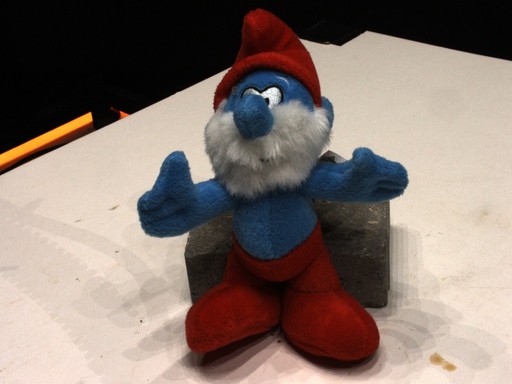}\\
\includegraphics[width=\fivewidth]{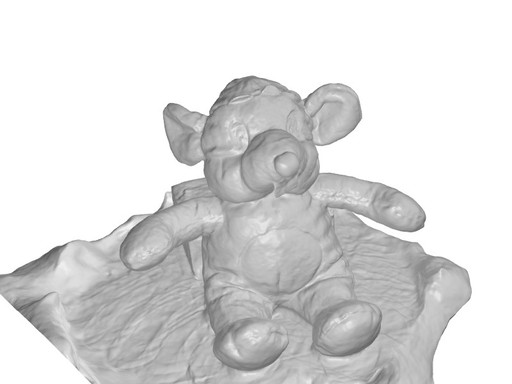}&
\includegraphics[width=\fivewidth]{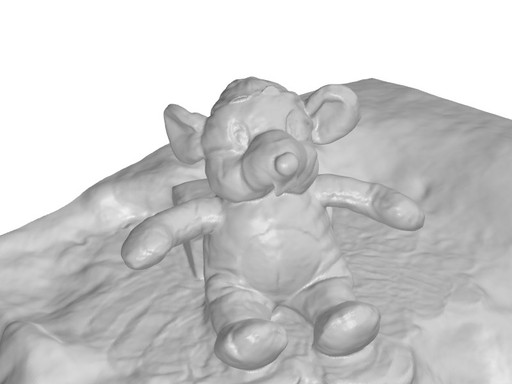}&
\includegraphics[width=\fivewidth]{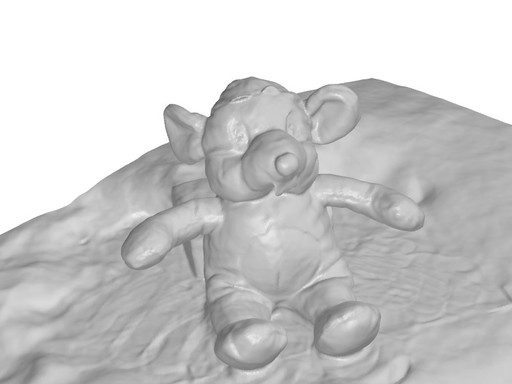}&
\includegraphics[width=\fivewidth]{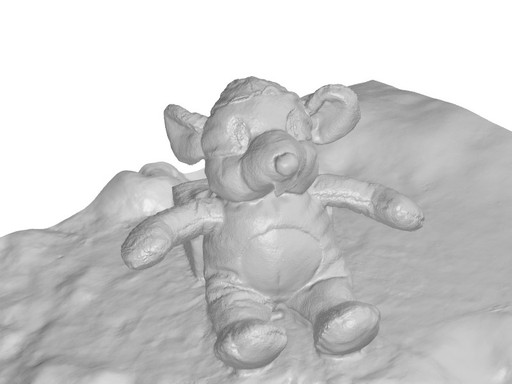}&
\includegraphics[width=\fivewidth]{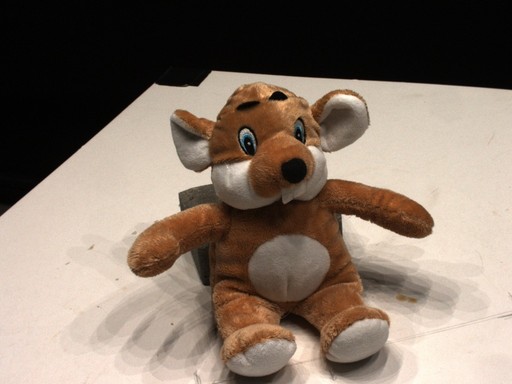}\\
\includegraphics[width=\fivewidth]{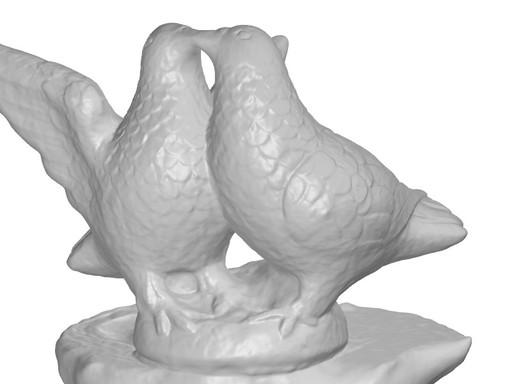}&
\includegraphics[width=\fivewidth]{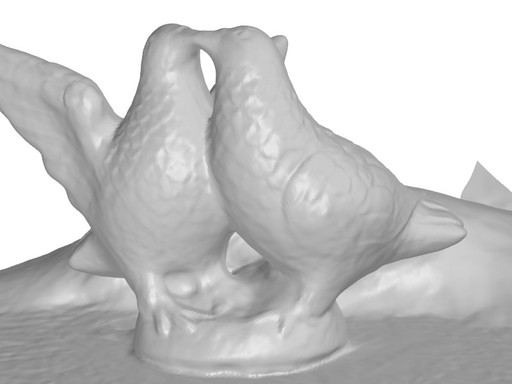}&
\includegraphics[width=\fivewidth]{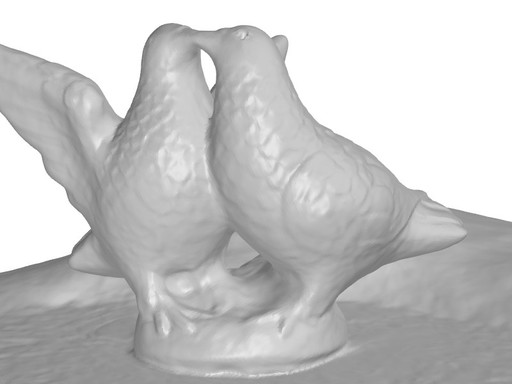}&
\includegraphics[width=\fivewidth]{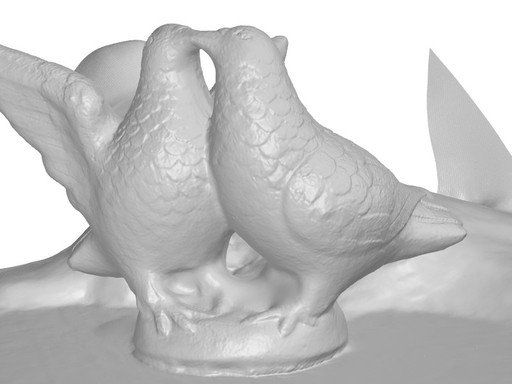}&
\includegraphics[width=\fivewidth]{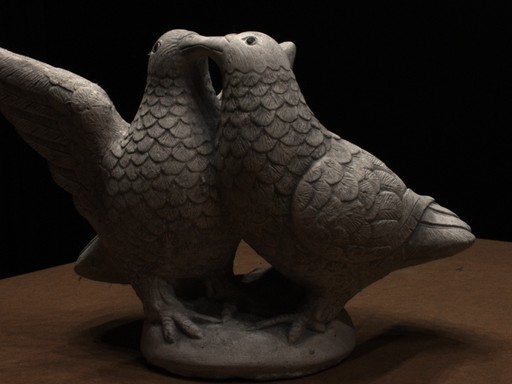}\\
\includegraphics[width=\fivewidth]{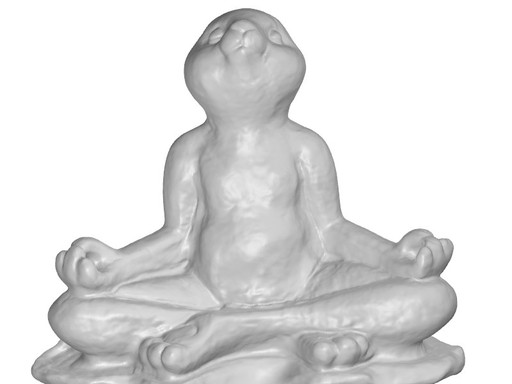}&
\includegraphics[width=\fivewidth]{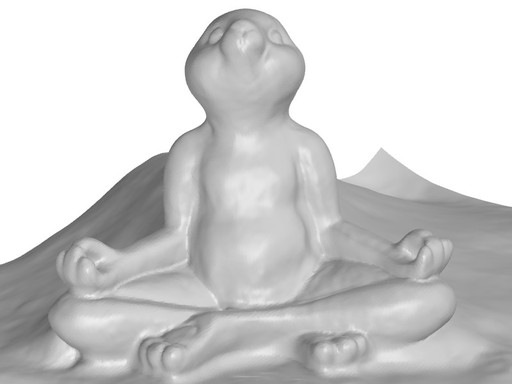}&
\includegraphics[width=\fivewidth]{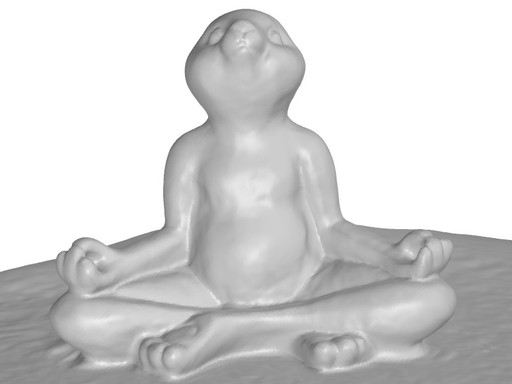}&
\includegraphics[width=\fivewidth]{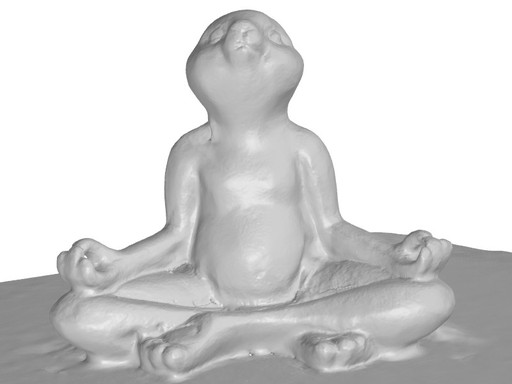}&
\includegraphics[width=\fivewidth]{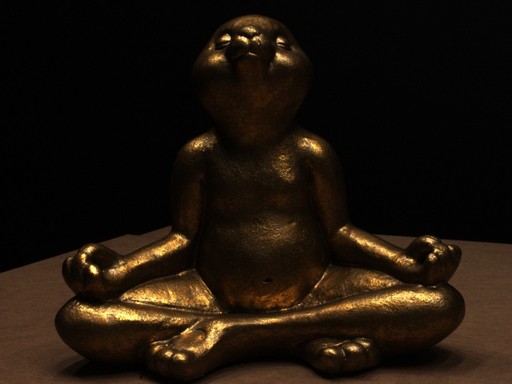}\\
\includegraphics[width=\fivewidth]{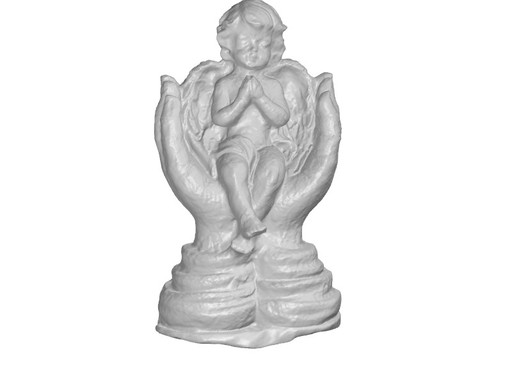}&
\includegraphics[width=\fivewidth]{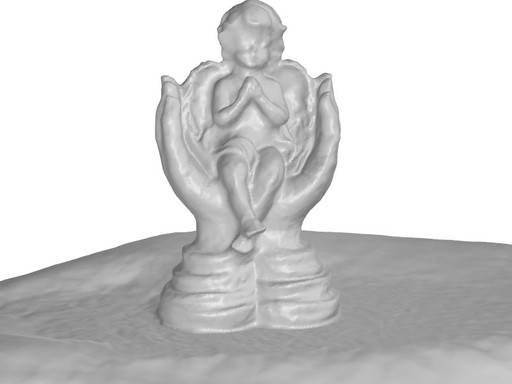}&
\includegraphics[width=\fivewidth]{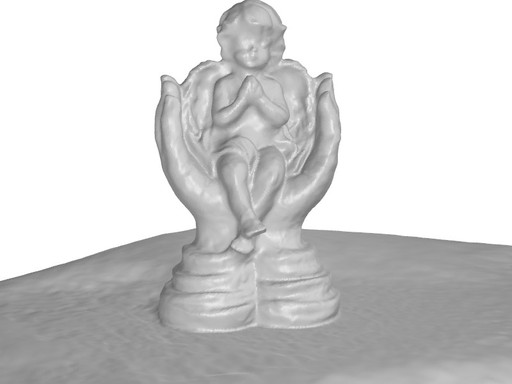}&
\includegraphics[width=\fivewidth]{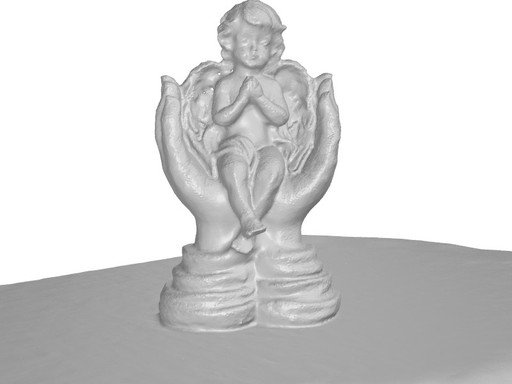}&
\includegraphics[width=\fivewidth]{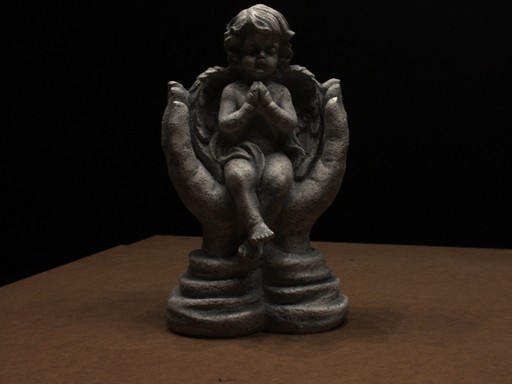}\\
        \end{tabular}
    
    \caption{\red{\textbf{Qualitative Comparison on DTU Dataset with all input views.}  Our approach with MLP achieves similar results with previous method, while our method with Multi-Res. Fea. Grids reconstruct more detailed surface.}}
    \label{fig:dtuallview_morevis}
\end{figure*}

}

\newcommand{\definition}{
\begin{table}[ht]
\centering
\setlength\extrarowheight{10pt}
    \begin{tabular}{lc}
    \toprule
    Metric & Definition \\
    \midrule
    Acc & $\underset{\bp \in P}{\mbox{mean}}\left( \underset{\bp^*\in P^*}{\mbox{min}} ||\bp-\bp^*||_1 \right)$ \\
    Comp & $\underset{\bp^* \in P^*}{\mbox{mean}}\left( \underset{\bp \in P}{\mbox{min}} ||\bp-\bp^*||_1 \right)$\\
    Chamfer & $\frac{\text{Acc} + \text{Comp}}{2} $ \\
    Precision & $\underset{\bp \in P}{\mbox{mean}}\left( \underset{\bp^*\in P^*}{\mbox{min}} ||\bp-\bp^*||_1 < 0.05 \right)$  \\
    Recall & $\underset{\bp^* \in P^*}{\mbox{mean}}\left( \underset{\bp \in P}{\mbox{min}} ||\bp-\bp^*||_1 < 0.05\right)$ \\
    F-score & $\frac{ 2 \cdot \text{Precision} \cdot \text{Recall} }{\text{Precision} + \text{Recall}}$ \\
    Normal-Acc  &  $\underset{\bp \in P}{\mbox{mean}}\left(  \bn_\bp^T\bn_{\bp^*}  \right) \,\, \text{s.t.} \, \, \bp^* = \underset{p^* \in P^*}{\argmin} ||\bp-\bp^*||_1 $  \\
    
    Normal-Comp &  $\underset{\bp^* \in P^*}{\mbox{mean}}\left( \bn_\bp^T \bn_{\bp^*} \right) \,\, \text{s.t.} \, \, \bp = \underset{p \in P}{\argmin} ||\bp-\bp^*||_1 $  \\
    Normal-Consistency & $ \frac{\text{Normal-Acc} + \text{Normal-Comp}}{2} $\\
    \bottomrule
    \end{tabular}
    \vspace{.1cm}
\caption{\textbf{Evaluation Metrics.} 
We show the evaluation metrics with their definitions that we use to measure reconstruction quality.
$P$ and $P^*$ are the point clouds sampled from the predicted and the ground truth mesh. $\bn_\bp$ is the normal vector at point $\bp$.
}
\label{tab:definition}
\end{table}
}

\newcommand{\scannetresultfull}{
\begin{table}[!t]
    \centering
    \footnotesize
    \begin{tabular}{lcccccc}
    \toprule
    {} &Acc$\downarrow$ & Comp$\downarrow$ & Chamfer-$L_1$ $\downarrow$ & Prec$\uparrow$ & Recall$\uparrow$ & F-score$\uparrow$\\
    \midrule
    COLMAP~\cite{Schoenberger2016ECCV}       & 0.047 & 0.235 & 0.141 & 0.711 & 0.441 & 0.537  \\
    UNISURF~\cite{Oechsle2021ICCV}           & 0.554 & 0.164 & 0.359 & 0.212 & 0.362 & 0.267  \\
    NeuS~\cite{Wang2021ARXIVb}               & 0.179 & 0.208 & 0.194 & 0.313 & 0.275 & 0.291  \\
    VolSDF~\cite{Yariv2021NEURIPS}           & 0.414 & 0.120 & 0.267 & 0.321 & 0.394 & 0.346  \\ 
    Manhattan-SDF~\cite{guo2022manhattan}    & 0.072 & 0.068 & 0.070 & 0.621 & 0.586 & 0.602  \\
    NeuRIS~\cite{wang2022neuris}             & 0.050 & 0.049 & 0.050 & 0.717 & 0.669 & 0.692  \\
    \midrule
    \textbf{Ours} (Multi-Res.\ Grids)       & 0.072& 0.057 & 0.064 & 0.660 & 0.601 & 0.626  \\
    \textbf{Ours} (MLP)                     & \textbf{0.035}& \textbf{0.048}& \textbf{0.042} & \textbf{0.799} & \textbf{0.681} & \textbf{0.733}  \\
    \bottomrule
    \end{tabular}
    \vspace{0.1cm}	
    \caption{
    \textbf{Scene-level 3D Reconstruction on ScanNet.} 
    We report reconstruction results for our methods and baselines on ScanNet (baselines from~\cite{guo2022manhattan}).
    We find that our approaches outperform previous state-of-the-art, highlighting the effectiveness of the use of monocular geometric priors. As ScanNet's RGB images contain motion blur and the camera poses are partially noisy, we further observe that the MLP architecture is more robust to this noise and achieves the best results. It's worth noting that we perform better than concurrent work~\cite{wang2022neuris} even though they have some filtering mechanism.
    }
    \label{tab:scannet}
\end{table}
}

\newcommand{\tntresult}{
\begin{table}[]
    \centering
    \begin{tabular}{ccccc}
    \toprule
    {} & Grid & Grid w/ cues & MLP~\cite{Yariv2021NEURIPS} & MLP w/ cues \\
    \midrule
    Auditorium  &  1.36 & \textbf{3.17}  & 1.60 & 3.09 \\
    Ballroom    &  2.67 & \textbf{3.70}  & 2.04 & 2.47 \\
    Courtroom   &  7.84 & \textbf{13.75} & 8.03 & 10.00\\
    Museum      &  4.12 & \textbf{5.68}  & 2.96 & 5.10 \\
    \midrule
    mean        &  4.00 & \textbf{6.58} & 3.66 & 5.165 \\
    \bottomrule
    \end{tabular}
    \vspace{0.1cm}	
    \caption{\textbf{Evaluation Results on the Tanks and Temples Dataset Advanced Set.} 
    We evaluate the reconstructed meshes using the official server and report the F-score with 10mm. 
    Our monocular geometric cues improve the reconstruction quality for all scenes.}
    \label{tab:tnt}
    \vspace{-0.2cm}	
\end{table}
}

\newcommand{\dturesultfull}{
\begin{table}[!t]
    \setlength{\tabcolsep}{2pt} %
    \centering
    \begin{tabular}{cccccccc}
        \toprule
        {} & TSDF~\cite{Curless1996SIGGRAPH} &  COLMAP &  RealityCapture & MLP~\cite{Yariv2021NEURIPS} & MLP & Multi-Res.  & Multi-Res. \\
        {} & {} &  {} &  {} & {} & w/ cues  & Grids & Grids w/ cues\\
        \midrule
        scan24  &  5.01 &  4.45 &      4.19 &    5.24 &  \textbf{3.47} & 6.46  & 5.24\\
        scan37  &  5.28 &  4.67 &      3.85 &    5.09 &  \textbf{3.61} & 8.30  & 6.37\\
        scan40  &  5.09 &  2.51 &      2.26 &    3.99 &  \textbf{2.10} & 7.03  & 2.52\\
        scan55  &  4.63 &  1.90 &      2.49 &    1.42 &  \textbf{1.05} & 5.87  & 1.95\\
        scan63  &  5.03 &  2.81 &      3.49 &    5.10 &  \textbf{2.37} & 6.92  & 6.64\\
        scan65  &  4.50 &  2.92 &      3.97 &    4.33 &  \textbf{1.38} & 3.09  & 2.05\\
        scan69  &  4.55 &  2.12 &      1.91 &    5.36 &  \textbf{1.41} & 5.34  & 4.25\\
        scan83  &  4.88 &  2.05 &      2.49 &    3.15 &  \textbf{1.85} & 6.03  & 1.81\\
        scan97  &  6.22 &  2.93 &      2.37 &    5.78 &  \textbf{1.74} & 6.93  & 5.27\\
        scan105 &  3.89 &  2.05 &      2.27 &    2.07 &  \textbf{1.10} & 6.01  & 2.54\\
        scan106 &  5.67 &  2.01 &      2.90 &    2.79 &  \textbf{1.46} & 6.14  & 3.85\\
        scan110 &  3.80 &  N/A  &      4.60 &    5.73 &  \textbf{2.28} & 7.62  & 3.89\\
        scan114 &  4.67 &  1.10 &      1.38 &    \textbf{1.20} &  1.25 & 6.27  & 1.90\\
        scan118 &  4.51 &  2.72 &      2.57 &    5.64 &  \textbf{1.44} & 7.59  & 3.12\\
        scan122 &  4.35 &  1.64 &      1.76 &    6.20 &  \textbf{1.45} & 6.47  & 3.84\\
        \midrule
        mean    &  4.80 &  2.56 &      2.84 &     4.21 &  \textbf{1.86 }& 6.47 & 3.68 \\
        \bottomrule
        \end{tabular}
    \vspace{0.1cm}	
    \caption{\textbf{Evaluation Results on the DTU Dataset with 3 Input Views.} 
    Note the COLMAP fails on scan110 so we take the average over the remaining 14 scenes. We find that without geometric cues, neither Grids nor MLP works well with only 3 input views. When incorporating the monocular geometric cues, the results for both representations are significantly improved. Interestingly, the grid-based representations perform inferior to a single MLP as they are updated only locally and do not have an inductive smoothness bias compared to a monolithic MLP representation.}
    \label{tab:dtu_full}
    \vspace{-0.1cm}	
\end{table}
}

\newcommand{\replicaablationfull}{
\begin{table}[!t]
	\centering
	\footnotesize
	\setlength{\tabcolsep}{2pt}
	    \begin{tabular}{clcccccc}
            \toprule
            {}&{}&{}& Test Split & {}&{}&Train Split \\
            \multicolumn{2}{c}{} &Normal C.$\uparrow$& Chamfer-$L_1$~$\downarrow$ & F-score~$\uparrow$ &Normal C.$\uparrow$& Chamfer-$L_1$~$\downarrow$ & F-score~$\uparrow$\\
            \hline
            {}                            & No Cues     & 57.30 & 26.68 & 15.50 & 60.86 & 17.34 & 26.34\\
            \textbf{Dense SDF}            & Only Depth  & 71.81 & 12.60 & 30.09 & 73.15& 13.09& 30.30\\
            \textbf{Grid}                 & Only Normal & 73.95 & 13.62 & 33.34 & 77.80& 11.30& \textbf{42.45}\\
            {}                            & Both Cues   & \textbf{76.47} & \textbf{11.39} & \textbf{37.27} & \textbf{80.05} & \textbf{10.09} & 41.57\\
            \hline
            \multirow{4}{*}{\textbf{MLP}} & No Cues     & 86.48& 6.75& 66.88 & 86.69& 7.48& 63.24\\
            {}                            & Only Depth  & 90.56& 4.26& 76.42 & 91.80& 3.59&85.67\\
            {}                            & Only Normal & 91.35& 3.19& 85.84 & 92.85& 4.23&85.58\\
            {}                            & Both Cues   & \textbf{92.11}& \textbf{2.94}& \textbf{86.18} & \textbf{93.86}& \textbf{2.63}&\textbf{92.12}\\
            \hline
            {}                            & No Cues     & 86.41 & 6.28 & 64.22 &86.54&6.63&67.26\\
            \textbf{Single-Res.}          & Only Depth  & 90.50 & 3.94 & 78.42 &91.3 & 3.29&86.34\\
            \textbf{Grids}                & Only Normal & 89.60 & 4.07 & 76.47 &\textbf{91.87}&3.13&85.96\\
            {}                            & Both Cues   & \textbf{90.59} & \textbf{3.56} & \textbf{83.34} &\textbf{91.87}&\textbf{2.98}&\textbf{88.23}\\
            \hline
            {}                            & No Cues     & 87.95& 5.03&78.38 &87.15&5.83&72.13\\
            \textbf{Multi-Res.}           & Only Depth  & 90.87& 3.75&80.32 &91.25&3.41&\textbf{87.04}\\
            \textbf{Grids}                & Only Normal & 89.90& 3.61&81.28 &91.11&3.59&84.02\\
            {}                            & Both Cues   & \textbf{90.93}& \textbf{3.23} &\textbf{85.91} &\textbf{91.41}&\textbf{3.14} &86.87\\
            \bottomrule
        \end{tabular}
    \vspace{0.1cm}	
    \caption{ \textbf{Ablation of Monocular Geometric Cues on Replica.} Our monocular geometric cues significantly improve reconstruction quality across all architectures. 
    }
    \label{tab:ablation_cues_full}
\end{table}
}

\newcommand{\moreconfigurations}{
\begin{table}[]
    \centering
    \begin{tabular}{cccc}
        \toprule
        Model configuration & Num. Params \\
        \midrule
        MLP (2 layers) & 0.15M \\
        MLP (4 layers) & 0.26M \\
        MLP (8 layers) & 0.53M \\
        MLP (12 layers) & 0.8M \\
        \midrule
        Multi-res. Feature Grids (hash table size $2^{13}$) & 0.41M \\
        Multi-res. Feature Grids (hash table size $2^{15}$) & 1.11M \\
        Multi-res. Feature Grids (hash table size $2^{17}$) & 3.67M \\
        Multi-res. Feature Grids (hash table size $2^{19}$) & 12.67M \\
        \bottomrule
    \end{tabular}
    \caption{\red{\textbf{Number of Learnable Parameters Using Different Architecture Configurations.}}}
    \label{tab:more_config}
\end{table}
}

\newcommand{\differentpredictors}{
\begin{table}[!t]
	\centering
	\setlength{\tabcolsep}{2pt}
    \begin{tabular}{ccc}
    	{
    	    \begin{tabular}{cc}
            \toprule
            Method                               & F-score \\
            \midrule
            MLP                                  & 64.2\\
            w/ MiDaS ~\cite{Ranftl2020PAMI}      & 68.6 \\
            w/ LeReS ~\cite{Wei2021CVPR}         & 72.6 \\
            w/ Omnidata ~\cite{Eftekhar2021ICCV} & 86.7\\
            \bottomrule
            \end{tabular}
    	}
    	&
    	{
    	    \begin{tabular}{cc}
            \toprule
            Method                               & F-score \\
            \midrule
            MLP                                  & 64.2\\
            w/ Tilted~\cite{Do2020SurfaceNormal} & 45.6\\
            w/ Omnidata~\cite{Eftekhar2021ICCV}  & 92.2\\
            \bottomrule
            \end{tabular}
    	}&
    	{
    	    \begin{tabular}{cc}
            \toprule
            Method                               & F-score \\
            \midrule
            MLP                                  & 64.2\\
            w/ Self-supervised~\cite{IndoorSfMLearner,structdepth} & 45.6\\
            w/ Omnidata~\cite{Eftekhar2021ICCV}                     & 86.7\\
            \bottomrule
            \end{tabular}
    	}
    	
    	\\
    (a) \small Different Depth & \small (b) Different Normal & (c) Self-supervised Depth\\
    \end{tabular}
    \caption{
    \red{
    \textbf{Ablation of Different Monocular Cues Predictors.}
    a.) Adding monocular depth improves performance over a single MLP without cues. Unsurprisingly, better depth predictors lead to better performance, with the state-of-the-art Omnidata model giving the best results. 
    b.) Adding monocular normal improve the results. Similarly, using normals predicted by the state-of-the-art Omnidata model leads to the best performance.
    c.) Using self-supervised depth estimator degrades performance. We hypothesize that this is due to the weaker performance of the self-supervised model which is also trained with an RGB loss and hence suffers from the under-constrained problem of recovering geometry from multi-view images.
    }
    }
    \label{tab:ablation_different_predictors}
\end{table}
}

\newcommand{\dtunvstable}{
\begin{wraptable}[8]{r}{0.2186\linewidth}
    \centering
    \setlength{\tabcolsep}{2pt}

    \begin{tabular}{lc}
            \toprule
             & PSNR \\
            \midrule
            MLP~\cite{Yariv2021NEURIPS} & 17.65\\
            MLP w/ cues & \textbf{23.64}\\
            \bottomrule
    \end{tabular}
    
    \caption{\textbf{Novel view synthesis results on DTU (3 Views).}
    }
    \vspace{0.0cm}
    \label{tab:dtu_nvs}
\end{wraptable}
}

\begin{document}

\maketitle

\begin{abstract}

In recent years, neural implicit surface reconstruction methods have become popular for multi-view 3D reconstruction.
In contrast to traditional multi-view stereo methods, these approaches tend to
produce smoother and more complete reconstructions
due to the inductive smoothness bias of neural networks. State-of-the-art neural implicit methods allow for high-quality reconstructions of simple scenes from many input views. Yet, their performance drops significantly for larger and more complex scenes and scenes captured from sparse viewpoints.
This is caused primarily by the inherent ambiguity in the RGB reconstruction loss that does not provide enough constraints, in particular in less-observed and textureless areas. 
Motivated by recent advances in the area of monocular geometry prediction, we systematically explore the utility these cues provide for improving neural implicit surface reconstruction. 
We demonstrate that %
depth and normal cues, predicted by general-purpose monocular estimators, significantly improve reconstruction quality and optimization time. %
Further, we analyse and investigate multiple design choices for representing neural implicit surfaces, ranging from monolithic MLP models over single-grid to multi-resolution grid representations. 
We observe that geometric monocular priors improve performance
both for small-scale single-object as well as large-scale multi-object scenes, independent of the choice of representation. %
\end{abstract}
\section{Introduction}

3D reconstruction from multiple RGB images is a fundamental problem in computer vision %
with various applications in robotics, graphics, animation, virtual reality, and more.
Recently, coordinate-based neural networks have emerged as a powerful tool for representing 3D geometry and appearance.
The key idea is to use compact, memory efficient multi-layer perceptrons (MLPs) to parameterize implicit shape representations such as occupancy or signed distance fields.
While early works~\cite{Mescheder2019CVPR,Park2019CVPR,chen2018implicit_decoder} relied on 3D supervision, several recent works~\cite{Niemeyer2019ICCV,Sitzmann2019NIPS,Yariv2020NIPS} use %
differentiable surface rendering %
to reconstruct scenes from multi-view images. 
At the same time, neural radiance fields (NeRFs)~\cite{Mildenhall2020ECCV} achieved %
impressive novel view synthesis results with volume rendering techniques.
~\cite{Oechsle2021ICCV,Wang2021ARXIVb,Yariv2021NEURIPS} combine %
surface and volume rendering %
for the task of 3D reconstruction by %
expressing volume density as a function of the underlying 3D surface, which in turn improves scene geometry.

\teaser

Current neural implicit-based surface reconstruction approaches achieve %
impressive reconstruction results for simple scenes with dense viewpoint sampling. Yet, as shown in the first row of \figref{fig:teaser}, they struggle in the presence of limited input views (DTU with 3 views)
or for scenes that contain large textureless regions (walls in ScanNet or Tanks \& Temples). A key reason for this behavior is that these model are optimized using a per-pixel RGB reconstruction loss. Using only RGB images as input leads to an underconstrained problem as there exist an infinite number of photo-consistent explanations~%
\cite{barron2022mipnerf360,kaizhang2020nerfplusplus}. 
Previous works address this problem by incorporating priors on the structure of the scene into the optimization process, e.g., %
depth smoothness~\cite{Niemeyer2022CVPR}, surface smoothness~\cite{Oechsle2021ICCV,Zhang2021TOGNerfFactor}, semantic similarity~\cite{Jain2021ICCV}, or %
Manhattan world assumptions~%
\cite{guo2022manhattan}. In this paper, we explore monocular geometric priors as they are readily available and efficient to compute. We show that using such priors significantly improves 3D reconstruction quality in challenging scenarios (see second row of \figref{fig:teaser}). %

Estimating geometric cues such as depth and normals from a single image has been an active research area for decades. 
The seminal work by Eigen et al.~\cite{Eigen2014NIPS,Eigen2015ICCV} showed that learned models based on deep convolutional neural networks (CNNs)
significantly improved over early work in this area
~\cite{Saxena2006NIPS,Saxena2008IJCV,Saxena2009PAMI,Hoiem2008IJCV,Hoiem2005ICCV,Hoiem2007IJCV,Hoiem2005SIGGRAPH}. %
Recent work~\cite{Ranftl2020PAMI,Ranftl2021DPT,Eftekhar2021ICCV}, in particular Omnidata~\cite{Eftekhar2021ICCV}, has made significant headway in terms of prediction quality and generalization to new scenes using very large datasets for training.
These strong results on individual images, and the fact that monocular geometric cues can be computed %
efficiently, naturally lead to the question whether such models are able to provide the additional constraints required by implicit neural surface reconstruction approaches to handle more challenging settings.

This paper describes a framework, called MonoSDF, for integrating monocular geometric priors into neural implicit surface reconstruction methods: 
given multi-view images, we infer %
depth and surface normals for each image, and use them as additional supervision signals during optimization together with the RGB image reconstruction loss.
We observe that these priors %
lead to significant gains %
in reconstruction quality, especially in textureless and less-observed areas as shown in~\figref{fig:teaser}. %
This is due to the fact that the photometric consistency cues used by surface reconstruction methods and the recognition cues used by monocular networks are complementary: while photometric consistency fails in texturless regions such as %
walls, surface normals can be predicted reliably in these areas due to the structured 3D scene layout. Conversely, photoconsistency cues allow for establishing globally accurate 3D geometry in textured regions, while normal %
and (relative) depth cues only provide local geometric information.

Apart from incorporating monocular geometric cues, we
provide a systematic study and analysis of state-of-the-art design choices for coordinate-based neural representations in the context of implicit surface reconstruction.
More specifically, we investigate the following architectures:
a single, large MLP~\cite{Yariv2021NEURIPS,Yariv2020NIPS,Oechsle2021ICCV,Wang2021ARXIVb}, a dense SDF grid~\cite{Jiang2020CVPRa}, a single feature grid~\cite{Peng2020ECCV,Huang2021CVPR,Liu2020NEURIPS,Peng2021NEURIPS} and multi-resolution feature grids~\cite{Mueller2022ARXIV,Zhu2022CVPR,Takikawa2021CVPR,chibane20ifnet,hadadan2021neural}.
We observe that MLPs act globally and exhibit an inductive smoothness bias while being computationally expensive to optimize and evaluate. In contrast, grid-based representations benefit from locality during optimization and evaluation, hence they are computationally more efficient. However, reconstructions are noisier for sparse views or less-observed areas. 
Including monocular geometric priors improves neural implicit reconstruction results across different settings with faster convergence times and independent of the underlying representation.

In summary, we make the following contributions:
\vspace{-.2cm}\begin{itemize}[leftmargin=*]
    \item We introduce \textit{MonoSDF}, a novel framework %
    which exploits monocular geometric cues to improve multi-view 3D reconstruction quality, efficiency, and scalability for neural implicit surface models. %
    \item We provide a systematic comparison and detailed analysis of design choices of neural implicit surface representations, including vanilla MLP and grid-based approaches.
    \item We conduct extensive experiments on multiple challenging datasets, ranging from object-level reconstruction on the DTU dataset\cite{Aanes2016IJCV}, over room-level reconstruction on Replica~\cite{Straub2019ARXIV} and ScanNet~\cite{Dai2017CVPR}, to large-scale indoor scene reconstruction on Tanks and Temples~\cite{Knapitsch2017SIGGRAPH}. 
\end{itemize}

\section{Related Work}

\boldparagraph{Architectures for Neural Implicit Scene Representations.}
Neural implicit scene representations or neural fields~\cite{Xie2021EUROGRAPHICS} have recently gained popularity for representing 3D geometry due to their expressiveness and low memory footprint.
Seminal works~\cite{Mescheder2019CVPR,Park2019CVPR,chen2018implicit_decoder} use a single MLP as the scene representation and show impressive object-level reconstruction quality, but they do not scale to more complicated or large-scale scenes due to the limited model capacity. 
Follow-up works~\cite{Peng2020ECCV,Zhu2022CVPR,Takikawa2021CVPR,Martel2021SIGGRAPH,Mueller2022ARXIV,chibane20ifnet,hadadan2021neural} combine an MLP decoder with one or multi-level voxel grids of low-dimensional features.
Such hybrid representations are able to better represent fine geometric details and can be evaluated fast. 
However, they lead to a larger memory footprint with increasing scene size. %
In this paper we provide a systematic comparison of four architectural design choices for \textit{implicit surface reconstruction}.

\boldparagraph{3D Reconstruction from Multi-view Images.}
Reconstructing the underlying 3D geometry from multi-view images is a long-standing goal of computer vision.
Classic multi-view stereo (MVS) methods~\cite{Agrawal2001CVPR,Bleyer2011BMVC,Bonet1999ICCV,Kutulakos2000IJCV,Broadhurst2001ICCV,Kutulakos2000IJCV,Schoenberger2016ECCV,Seitz1997CVPR,Seitz2006CVPR} consider either feature matching for depth estimation~\cite{Bleyer2011BMVC,Schoenberger2016ECCV} or represent shapes with voxels~\cite{Agrawal2001CVPR,Bonet1999ICCV,Broadhurst2001ICCV,Seitz1997CVPR,Kutulakos2000IJCV,Paschalidou2018CVPR,Tulsiani2017CVPR,Ulusoy2015THREEDV}. %
Learning-based MVS methods usually replace some parts of the classic MVS pipeline, \eg, feature matching~\cite{Hartmann2017ICCV,Leroy2018ECCV,Luo2016CVPR,Ummenhofer2017CVPRa,Zagoruyko2015CVPR}, depth fusion~\cite{Donne2019CVPR,Riegler2017THREEDV}, or inferring depth from multi-view images~\cite{Huang2018CVPRDeepMVS,Yao2018ECCV,Yao2019CVPR,Yu2020fastmvsnet}. 
In contrast to the explicit scene representations used by classic MVS algorithms, recent neural approaches~\cite{Niemeyer2020CVPR,Yariv2020NIPS,Liu2020CVPR} %
represent surfaces via a single MLP with continuous outputs. 
Learned purely from posed 2D images, they show 
appealing reconstruction results and do not suffer from discretization. However, accurate object masks are required. 
Inspired by the density-based volume rendering in NeRF~\cite{Mildenhall2020ECCV}, which demonstrated impressive view synthesis without object masks, several works~\cite{Oechsle2021ICCV,Yariv2021NEURIPS,Wang2021ARXIVb} use volume rendering for neural implicit surface reconstruction without masks. %
However, 
these methods lead to poor results in large-scale scenes with textureless regions.
In this work, we show that incorporating monocular priors allows these approaches to obtain significantly more detailed reconstructions and to scale to larger and more challenging scenes. %

\boldparagraph{Incorporating Priors into Neural Scene Representations.}
Several researchers proposed to incorporate priors such as depth smoothness~\cite{Niemeyer2022CVPR}, %
semantic similarity~\cite{Jain2021ICCV}, or sparse MVS point clouds~\cite{Roessle2022CVPR} for the task of \textit{novel view synthesis} from sparse inputs. In contrast, in this work, our focus is on implicit 3D surface reconstruction. %
Concurrently, Manhattan-SDF~\cite{guo2022manhattan} uses dense MVS depth maps %
from COLMAP~\cite{Schoenberger2016CVPR} as supervision and adopts Manhattan world priors~\cite{Coughlan1999ICCV} to handle low-textured planar regions corresponding to walls, floors, etc. 
Our approach is based on the observation that
data-driven monocular depth and normal predictions%
~\cite{Eftekhar2021ICCV} provide high-quality priors for the full scene. 
Incorporating these priors into the optimization of neural implicit surfaces not only removes the
Manhattan world assumption~\cite{Coughlan1999ICCV} but also results in improved reconstruction quality and a simpler pipeline.\footnote{Manhattan-SDF~\cite{guo2022manhattan} requires semantic segmentation to determine where to enforce the assumption.} 
\red{Compared to NeuRIS~\cite{wang2022neuris}, a concurrent work that %
proposes to use normal priors for indoor scene reconstruction, we integrate monocular depth cues and further demonstrate the effectiveness of monocular cues on various neural scene representations, ranging from MLP to multi-resolution feature grids.}

\section{Method}
Our goal is to recover the underlying scene geometry from multiple posed images while utilizing monocular geometric cues to guide the optimization process.
To this end, we first review %
neural implicit scene representations and various design choices in~\secref{sec:representation} and discuss how to perform volume rendering of these representations in~\secref{sec:rendering}.
Next, we introduce the monocular geometric cues we investigate in our study in~\secref{sec:mono_cue} and discuss loss functions and the overall optimization process in~\secref{sec:optimization}.
An overview of our framework is provided in~\figref{fig:overview}.

\method

\subsection{Implicit Scene Representations}\label{sec:representation}

We represent scene geometry as a signed distance function (SDF). A signed distance function is a continuous function $f$ that, for a given 3D point, %
returns the point’s distance to the closest surface:
\begin{equation}
f: \nR^3 \to \nR \quad\quad\bx \mapsto s = \text{SDF}(\bx) \enspace .
\label{eq:f}
\end{equation}
Here, $\bx$ is the 3D point %
and  $s$ denotes the corresponding SDF value. %
In this work, we parameterize the SDF function with learnable parameters $\theta$ and investigate several different design choices for representing the function:
explicit as a dense grid of learnable SDF values, implicit as a single MLP, or hybrid using an MLP in combination with single- or multi-resolution feature grids.

\boldparagraph{Dense SDF Grid.}
The most straightforward way of parameterizing an SDF is to directly store SDF values in each cell of a discretized volume $\mathcal{G}_\theta$ with resolution of $R_H \times R_W \times R_D$~\cite{Jiang2020CVPRa}.
To query the SDF value $\hat{s}$ for an arbitrary point $\bx$ from the dense SDF grid, we can use any interpolation operation:
\begin{equation}
\hat{s} = \texttt{interp}(\bx, \mathcal{G}_\theta) \enspace .
\label{eq:dense}
\end{equation}
In our experiments, we implement $\texttt{interp}$ as trilinear interpolation.

\boldparagraph{Single MLP.} 
The SDF function can also be parameterized by a single MLP~\cite{Park2019CVPR} $f_\theta$:
\begin{equation}
\hat{s} = f_\theta(\gamma(\bx)) \enspace ,
\label{eq:f_mlp}
\end{equation}
where $\hat{s}$ is the predicted SDF value and $\gamma$ corresponds to a fixed positional encoding~\cite{Mildenhall2020ECCV, Tancik2020NEURIPS} mapping $\bx$ to a higher dimensional space. After their introduction to novel view synthesis~\cite{Mildenhall2020ECCV}, positional encoding functions are now widely used for neural implicit surface reconstruction~\cite{Yariv2020NIPS,Yariv2021NEURIPS,Wang2021ARXIVb,Oechsle2021ICCV} as they 
increase the expressiveness of coordinate-based networks~\cite{Tancik2020NEURIPS}.

\boldparagraph{Single-Resolution Feature Grid with MLP Decoder.} 
We can also combine both %
parameterizations and use a feature-conditioned MLP $f_\theta$ together with a feature grid $\Phi_\theta$ with a resolution of $R^3$,  where each cell of the grid stores a feature vector~\cite{Peng2020ECCV,Takikawa2021CVPR,Liu2020NEURIPS,Huang2021CVPR} instead of directly storing SDF values:
\begin{equation}
\hat{s} = f_\theta(\gamma(\bx), \texttt{interp}(\bx, \Phi_\theta)) \enspace .
\label{eq:single_grid}
\end{equation}
Note that the MLP $f_\theta$ is conditioned on the interpolated local feature vector from the feature grid $\Phi_\theta$.

\boldparagraph{Multi-Resolution Feature Grids with MLP Decoder.}\label{subsec:multi_res_grid}
Instead of using a single feature grid $\Phi_\theta$, one can also employ multi-resolution feature grids $\{\Phi_\theta^l\}_{l=1}^L$ with resolutions $R_l$~\cite{Mueller2022ARXIV,Zhu2022CVPR,Takikawa2021CVPR,chibane20ifnet,hadadan2021neural}. %
The resolutions are sampled in geometric space~\cite{Mueller2022ARXIV} to combine features at different frequencies: %
\begin{equation}
    R_l := \;\lfloor R_\text{min} b^l \rfloor \hspace{1cm}
    b := \;\text{exp}\left(\frac{\ln R_\text{max} - \ln R_\text{min}}{L-1}\right) \enspace ,
    \label{eq:res}
\end{equation}

where $R_\text{min}, R_\text{max}$ are the coarsest and finest resolution, respectively.
Similarly, we extract the interpolated features at each level and concatenate them together:
\begin{equation}
	\hat{s} = f_\theta(\gamma(\bx), \{\texttt{interp}(\bx, \Phi_\theta^l)\}_l)) \enspace .
	\label{eq:f_full}
\end{equation}
As the total number of grid cells grows cubically, we use a fixed number of parameters to store the feature grids and use a spatial hash function to index the feature vector at finer levels~\cite{Mueller2022ARXIV} (see supplementary for details).

\boldparagraph{Color Prediction.} 
In addition to the 3D geometry, we also predict color values such that our model can be optimized with a reconstruction loss.
Following~\cite{Yariv2020NIPS}, we %
therefore define a second function $\bc_\theta$ 
\begin{equation}\label{eq:color-mlp}
    \hat{\bc} = \bc_\theta(\bx, \bv, \hat{\bn}, \hat{\bz})
\end{equation}
that predicts a RGB color value $\hat{\bc}$ for a 3D point $\bx$ and a viewing direction $\bv$. The 3D unit normal $\hat{\bn}$ is the analytical gradient of our SDF function. The feature vector $\hat{\bz}$ is the output of a second linear head of the SDF network as in~\cite{Yariv2020NIPS}. We parameterize $\bc_\theta$ with a two-layer MLP with network weights $\theta$. In case of the dense grid SDF parameterization, we similarly optimize a dense feature grid and obtain the feature vector $\hat{\bz}$ via the interpolation function $\texttt{interp}$.

\subsection{Volume Rendering of Implicit Surfaces}\label{sec:rendering}
Following recent work~\cite{Yariv2020NIPS,Yariv2021NEURIPS,Oechsle2021ICCV,Wang2021ARXIVb}, we optimize the implicit representations described in~\secref{sec:representation} via an image-based reconstruction loss
using differentiable volume rendering. 
More specifically, to render a pixel, we cast a ray $\br$ from the camera center $\bo$ through the pixel along its view direction $\bv$. 
We sample $M$ points $\bx_\br^i = \bo + t_\br^i\bv$ along the ray and predict their SDF %
$\hat{s}_\br^i$ and color values $\hat{\bc}^i_\br$. 
We follow~\cite{Yariv2021NEURIPS} to transform the SDF values $\hat{s}_\br^i$ to density values $\sigma_\br^i$ for volume rendering: 
\begin{equation}\label{e:laplace}
\sigma_\beta(s) = \begin{cases} \frac{1}{2\beta} \exp\left( \frac{s}{\beta} \right) & \text{if } s\leq 0 \\
\frac{1}{\beta}\left( 1-\frac{1}{2}\exp\left ( -\frac{s}{\beta} \right ) \right) & \text{if } s>0
\end{cases} \enspace ,
\end{equation}
where $\beta$ is a learnable parameter. 
Following NeRF~\cite{Mildenhall2020ECCV}, the color $\hat{C}(\br)$ for the current ray $\br$ is computed %
via numerical integration: 
\begin{equation}
\hat{C}(\br) = \sum_{i=1}^M \, T_\br^i \, \alpha_\br^i \, \hat{\bc}_\br^i \hspace{1cm}
T_\br^i = \prod_{j=1}^{i-1}\left(1-\alpha_\br^j\right)\hspace{1cm}
\alpha_\br^i = 1-\exp\left(-\sigma_\br^i\delta_\br^i\right) \enspace ,
\label{eq:volume_render}
\end{equation}
where $T_\br^i$ and $\alpha_\br^i$ denote the transmittance and alpha value of sample point $i$ along ray $\br$, respectively,
and $\delta_\br^i$ is the distance between neighboring sample points. 
Similarly, we compute %
the depth $\hat{D}(\br)$ and normal $\hat{N}(\br)$ of the surface intersecting the current ray as:
\begin{equation}
\hat{D}(\br) = \sum_{i=1}^M \, T_\br^i \, \alpha_\br^i \, t_\br^i \hspace{1cm}
\hat{N}(\br) = \sum_{i=1}^M \, T_\br^i \, \alpha_\br^i \, \hat{\bn}_\br^i \enspace .
\label{eq:volume_render_dn}
\end{equation}

\subsection{Exploiting Monocular Geometric Cues}\label{sec:mono_cue}
Unifying volume rendering with implicit surfaces leads to impressive 3D reconstruction results. 
Yet, this approach struggles
with more complex scenes especially in textureless and sparsely covered regions.
To overcome this limitation,
we use %
readily available, 
efficient-to-compute
monocular geometric priors thereby improving neural implicit surface methods.

\boldparagraph{Monocular Depth Cues.}
One common monocular geometric cue is a monocular depth map, which can be easily obtained via an off-the-shelf monocular depth predictor.
More specifically, we use a pretrained Omnidata model~\cite{Eftekhar2021ICCV} to predict a depth map $\bar{D}$ for each input RGB image.
Note that the absolute scale is difficult to estimate in general scenes, so $\bar{D}$ must be considered as a relative cue. However, this relative depth information is provided also over larger distances in the image.

\boldparagraph{Monocular Normal Cues.}
Another geometric cue we use is the surface normal. 
Similar to the depth cues, we apply the same pretrained Omnidata model to acquire a normal map $\bar{N}$ for each RGB image. 
Unlike depth cues that provide semi-local relative information, normal cues are local and capture geometric detail.
We hence expect that surface normals and depth are complementary to each other.

\subsection{Optimization}\label{sec:optimization}
\boldparagraph{Reconstruction Loss.} \eqnref{eq:volume_render} provides a linkage from the 3D scene representation to 2D observations. We can therefore optimize the scene representation with a simple RGB reconstruction loss:
\begin{equation}
\mathcal{L}_\text{rgb} = \sum_{\br \in \cR} {\lVert \hat{C}(\br) - C(\br) \rVert}_1 \enspace . %
\end{equation}
Here $\mathcal{R}$ denotes the set of pixels/rays in the minibatch and $C(\br)$ is the observed pixel color.

\boldparagraph{Eikonal Loss.} Following common practice, we also add an Eikonal term~\cite{Gropp2020ICML} on the sampled points to regularize SDF values in 3D space
\begin{equation}
\mathcal{L}_\text{eikonal} = \sum_{\bx \in \cX} ({\lVert \nabla f_\theta(\bx) \rVert}_2 - 1)^2 \enspace ,
\end{equation} 
where $\cX$
are a set of uniformly sampled points together with near-surface points~\cite{Yariv2021NEURIPS}.

\boldparagraph{Depth Consistency Loss.} Besides $\mathcal{L}_{\text{rgb}}$ and $\mathcal{L}_{\text{eikonal}}$, we also enforce consistency between our rendered expected depth $\hat{D}$ and the monocular depth $\bar{D}$: %
\begin{equation}
\mathcal{L}_\text{depth} = \sum_{\br \in \cR} {\lVert (w \hat{D}(\br) + q) - \bar{D}(\br) \rVert}^2 \enspace , \label{eq:depth_prior}
\end{equation}
where $w$ and $q$ are the scale and shift used to align $\hat{D}$ and $\bar{D}$ since $\bar{D}$ is defined only up to scale. 
Note that these factors have to be estimated individually per batch as the depth maps predicted for different batches can differ in scale and shift.
Specifically, we solve for $w$ and $q$ with a least-squares criterion~\cite{Eigen2014NIPS,Ranftl2020PAMI} which has a closed-form solution (see supplementary for details).

\boldparagraph{Normal Consistency Loss.} Similarly, we impose consistency on the volume-rendered normal $\hat{N}$ and the predicted monocular normals $\bar{N}$ transformed to the same coordinate system with angular and L1 losses~\cite{Eftekhar2021ICCV}:
\begin{equation}
\mathcal{L}_\text{normal} = \sum_{\br \in \cR} {\lVert \hat{N}(\br) - \bar{N}(\br) \rVert}_1 +  {\lVert  1 - \hat{N}(\br)^\top  \bar{N}(\br) \rVert}_1 \enspace .
\end{equation}
The overall loss we use to optimize our implicit surfaces jointly with the appearance network is:
\begin{equation}
    \mathcal{L} = \mathcal{L}_{\text{rgb}} + \lambda_1 \mathcal{L}_\text{eikonal} + \lambda_2\mathcal{L}_{\text{depth}} + \lambda_3\mathcal{L}_{\text{normal}} \enspace .
\end{equation}

\boldparagraph{Implementation Details.}
We implement our method in PyTorch~\cite{Pytorch2019NIPS} and use the Adam optimizer~\cite{Kingma2015ICML} with a learning rate of 5e-4 for neural networks and 1e-2 for feature grids and dense SDF grids. We set $\lambda_1$, $\lambda_2$, $\lambda_3$ to 0.1, 0.1, 0.05, respectively.
We sample 1024 rays per iteration and apply the error-bounded sampling strategy introduced by~\cite{Yariv2021NEURIPS} to sample points along each ray.
For MLPs and feature grids, we adapt the architecture and initialization scheme from~\cite{Yariv2021NEURIPS} and~\cite{Mueller2022ARXIV}, respectively.
For obtaining monocular cues, we first resize each image and center crop it to $384\times384$, which we then feed as input to the pretrained Omnidata model~\cite{Eftekhar2021ICCV}. See supplementary for more details.

\section{Experiments}
We first analyze different architectural design choices and perform ablation studies \wrt monocular cues and optimization time on a room-level dataset (Replica) with %
perfect ground truth. %
Next, we provide qualitative and quantitative comparisons against state-of-the-art baselines on real-world indoor scenes.
Finally, we evaluate %
our method on object-level reconstruction for both sparse input and dense input scenarios.

\boldparagraph{Datasets.}
While previous neural implicit-based reconstruction methods mainly focused on single-object scenes with many input views, in this work, we investigate the importance of monocular geometric cues for scaling to more complex scenes. Thus we consider:
a) Real-world indoor scans: Replica~\cite{Straub2019ARXIV} and ScanNet~\cite{Dai2017CVPR};
b) Real-world large-scale indoor scenes: Tanks and Temples~\cite{Knapitsch2017SIGGRAPH} advanced scenes;
c) Object-level scenes: DTU~\cite{Aanes2016IJCV} in the sparse 3-view setting from~\cite{Niemeyer2022CVPR,Yu2021CVPR}.

\boldparagraph{Baselines.}
We compare against
a) state-of-the-art neural implicit surfaces methods: UNISURF~\cite{Oechsle2021ICCV}, VolSDF~\cite{Yariv2021NEURIPS}, NeuS~\cite{Wang2021ARXIVb}, and Manhattan-SDF~\cite{guo2022manhattan}.
b) Classic MVS methods: COLMAP~\cite{Schoenberger2016ECCV} and a state-of-the-art commercial software (RealityCapture\footnote{\url{https://www.capturingreality.com/}}).
c) TSDF-Fusion~\cite{Curless1996SIGGRAPH} with predicted monocular depth cues, where GT depth maps are used to recover the scale and shift values (cf. Eq.~\eqref{eq:depth_prior}). This baseline shows %
the reconstruction quality if only monocular depth cues and no implicit surface model is used. %

\boldparagraph{Evaluation Metrics.}
For DTU, we follow the official evaluation protocol and report the Chamfer distance.
For Replica and ScanNet,  following~\cite{Sucar2021ICCV,Zhu2022CVPR,Mescheder2019CVPR,guo2022manhattan,Peng2020ECCV,Peng2021NEURIPS}, we report the Chamfer Distance, the F-score with a threshold of 5cm, as well as a Normal Consistency measure.

\subsection{Ablation Study}
We first analyze 
different scene representation choices on the Replica dataset. %
Next, we ablate the %
impact of our geometric cues on reconstruction quality and convergence time. %

\figureablationgrid

\replicatable
\boldparagraph{Architecture Choices for Scene Representations.} 
We compare the four %
different 
scene geometry representations introduced in~\secref{sec:representation} and report %
metrics averaged over %
the Replica dataset in~\tabref{tab:replica_architecture}. 
Note that no monocular geometric cues are used here. %
We first observe that using a single MLP as the scene geometry representation leads to decent results, but the reconstruction tends to be over-smooth (see~\tabref{tab:replica_architecture} and ~\figref{fig:ablation_architecture}).
For grid-based representations, optimizing a dense SDF grid leads to a significantly worse performance compared to all other neural implicit scene representations, even with careful parameter tuning.
The reason is the lack of a smoothness bias: 
The SDF values in grid cells are all stored and optimized independently of each other, hence there is no local or global smoothness bias. %
In contrast, the Single-Res.\ Fea.\ Grid replaces the SDF value in each grid cell with a low-dimensional latent code, and uses a shallow MLP conditioned on these features to read out SDF values of arbitrary 3D points.
This modification leads to a notable boost in reconstruction quality over the dense grid, performing similarly well as the single MLP.
Using a %
Multi-Res.\ Fea.\ Grids as in~\cite{Mueller2022ARXIV} further increases performance. %
We observe that the Multi-Res.\ Fea.\ Grids is the best-performing grid-based model, and from now on we report results for the single MLP and the Multi-Res.\ Feature Grids.
For simplicity, we will refer to the multi-resolution feature grids as 
\textit{Multi-Res. Grids} or \textit{Grids} in the following.

\boldparagraph{Ablation of Different Cues.} 
We now investigate the effectiveness of different monocular geometric cues for the two chosen representations.
\tabref{tab:ablation_cues} (a) and~\figref{fig:ablation_cues} show that, for both representations, using either one or both monocular cues significantly boosts reconstruction quality.
We also find both cues to be complementary, %
with the best performance being achieved when using both. 
Similar behavior can be observed for the other two representations (\cf supplementary material).
It is worth noting that the differences between the two representations become negligible when using monocular cues, indicating that those serve as a general drop-in to improve reconstruction quality. 
\figureablationcues

\replicaablation
\boldparagraph{Optimization Time.} %
\tabref{tab:ablation_cues} (b) shows %
optimization time for the two scene representations with and without cues.
We see that the Multi-Res.\ Grids converge faster than the single MLP model. %
Further, adding the monocular cues significantly speeds up the convergence process.
After only 10K iterations, both representations perform better than the converged models without monocular cues. 
Note that the overhead required for incorporating the monocular cues into the optimization process is small and can be neglected. An extended version of \tabref{tab:ablation_cues} (b) can be found in the supplementary materials.

\subsection{Real-world Large-scale Scene Reconstruction}
To show the effectiveness of our method for large-scale scene reconstruction, we compare against various baselines on two challenging large-scale indoor datasets.

\boldparagraph{ScanNet.}
On ScanNet, we use the test split from~\cite{guo2022manhattan}
and also follow their evaluation protocol in which depth maps are rendered from input camera poses and then re-fused using TSDF Fusion~\cite{Curless1996SIGGRAPH} to evaluate only observed areas.
We observe in~\tabref{tab:scannet_fused} that our MLP variant outperforms all baselines achieving smoother reconstructions with more fine details. \red{Note that we outperform concurrent work~\cite{wang2022neuris}.} %
Further, we find that the MLP variant performs significantly better than using Multi-Res. Grids.
ScanNet's RGB images contain motion blur and the camera poses are also noisy. %
This can be harmful to the local geometry updates in grid-based representations, while MLPs are more robust to this noise due to their smoothness bias.

\scannetfused
\boldparagraph{Tanks \& Temples.}
To further investigate the scalability of our method to larger-scale scenes, we conduct experiments on the Tanks and Temples advanced sets.
The qualitative results in~\figref{fig:teaser} show that the monocular cues significantly boost the performance of VolSDF~\cite{Yariv2021NEURIPS}, making MonoSDF the first neural implicit model achieving reasonable results on such a large-scale indoor scene.
See the supplementary material for more visual comparisons and discussions.

\subsection{Object-level Reconstruction from Sparse Views}

We now evaluate our method on another challenging task: reconstructing single objects from sparse input views.
We adopt the test split from~\cite{Yariv2020NIPS, Yariv2021NEURIPS} on DTU
and choose \textit{three} input views following~\cite{Niemeyer2022CVPR}.

\dtufused

We first observe in~\tabref{tab:dtu_fused} and~\figref{fig:teaser} that without the usage of the monocular geometric cues, neither the MLP (VolSDF~\cite{Yariv2021NEURIPS}) nor the Multi-Res. Grids work well with only 3 input views.
When incorporating the cues, the results for both representations are significantly improved. %
Interestingly, the grid-based representations perform inferior to a single MLP
as they are updated locally and do not benefit from the inductive bias of a monolithic MLP representation.

Comparing against TSDF Fusion~\cite{Curless1996SIGGRAPH} that fused predicted depth cues from all views into a TSDF volume without any optimization, we observe that this baseline has difficulties in reconstructing meaningful details due to inconsistencies in the monocular depth cues.
Note that this baseline uses the GT depth maps from~\cite{Donne2019CVPR} to compute scale and shift for the depth cues. %
Classic MVS methods perform well quantitatively, but they heavily rely on dense matching, and in case of three input images, this inevitably leads to incomplete reconstructions (see supplementary material). 
In contrast, our approach combines neural implicit surface representations with the benefits from monocular geometric cues that are more robust to less-observed regions. %

\subsection{Object-level Reconstruction from Dense Views}
\dtuallviewfused

\red{To further investigate the effectiveness and flexibility of our method, we evaluate our approach on the DTU dataset with all input views, which is a common setting in recent work~\cite{Oechsle2021ICCV,Wang2021NEURIPS,Yariv2021NEURIPS}. In this experiment, we simply resize the low-resolution monocular cues to full resolution (from $384 \times 384$ to $1200 \times 1200$ pixels) while keeping the image ratio. As the original image is of size $1200 \times 1600$, the monocular cues are missing in the left and right part of the image. Therefore, we only use the monocular cues where they are available. 

As shown in~\tabref{tab:dtuallview_fused}, our approach with MLP architecture achieves reconstruction quality similar to state-of-the-art methods~\cite{Oechsle2021ICCV,Wang2021NEURIPS,Yariv2021NEURIPS}. This is reasonable as the dense input views provide enough constraints and the prior information from monocular cues is negligible. However, our method with multi-resolution feature grid architecture outperforms previous work by a large margin. We attribute this to the expressiveness of multi-resolution feature grids where monocular cues are still effective to suppress noise and therefore can reconstruct smooth and detailed surfaces. We kindly refer the reader to the supplementary material for additional visual comparisons.
}
\section{Conclusion}
We have presented MonoSDF, a novel framework that systematically explores how monocular geometric cues can be incorporated into the optimization of neural implicit surfaces from multi-view images.
We show that such easy-to-obtain monocular cues can significantly improve 3D reconstruction quality, efficiency, and scalability for a variety of neural implicit representations.
When using monocular cues, a simple MLP architecture performs best overall, demonstrating that MLPs in principle are able to represent complex scenes, albeit being slower to converge compared to grid-based representations.
Multi-resolution feature grids in general can converge fast and capture details, but are less robust to noise and ambiguities in the input images.

\boldparagraph{Limitations.} 
The performance of our model depends on the quality of the monocular cues.
Filtering strategies to handle failures of the monocular predictor
are thus a promising direction to further improve reconstruction quality. We kindly refer the reader
to the supplementary material for additional analysis.
\red{While we demonstrated that integrating depth and normal cues significantly improves reconstruction, exploring other cues such as occlusion edges, plane, or curvature~\cite{YuZLZG19,Eftekhar2021ICCV} is an interesting future direction.}
We are currently limited by the low-resolution ($384\times384$ pixels) output of the Omnidata model~\cite{Eftekhar2021ICCV} and
plan to explore different ways of using higher-resolution cues. %
\red{ We provide some preliminary results of using high-resolution cues in the supplementary.}
Joint optimization of scene representations and camera parameters~\cite{Azinovic2022CVPR,Zhu2022CVPR} is another interesting direction, especially for multi-resolution grids, in order to better handle noisy camera poses. %

\begin{ack}
This work was supported by an NVIDIA research gift. 
We thank the Max Planck ETH Center for Learning Systems (CLS) for supporting SP and the International Max Planck Research School for Intelligent Systems (IMPRS-IS) for supporting MN. 
ZY is supported by BMWi in the project KI Delta Learning (project number 19A19013O). 
AG is supported by the ERC Starting Grant LEGO-3D (850533) and DFG EXC number 2064/1 - project number 390727645.
TS is supported by the EU Horizon 2020 project RICAIP (grant agreeement No.857306), and the European Regional Development Fund under project IMPACT (No. CZ.02.1.01/0.0/0.0/15$\_$003/0000468).
We thank the authors of Manhattan-SDF and NeuRIS for sharing results on ScanNet. \red{We also thank Christian Reiser and Zijian Dong for proofreading.%
}

\end{ack}

\bibliographystyle{ieee}
\bibliography{bibliography_long,bibliography_custom,bibliography}
\section*{Checklist}

\begin{enumerate}

\item For all authors...
\begin{enumerate}
  \item Do the main claims made in the abstract and introduction accurately reflect the paper's contributions and scope?
    \answerYes{}
  \item Did you describe the limitations of your work?
    \answerYes{}
  \item Did you discuss any potential negative societal impacts of your work?
    \answerYes{We discuss potential negative societal impacts in our supplementary material.}
  \item Have you read the ethics review guidelines and ensured that your paper conforms to them?
    \answerYes{}
\end{enumerate}

\item If you are including theoretical results...
\begin{enumerate}
  \item Did you state the full set of assumptions of all theoretical results?
     \answerNA{}
        \item Did you include complete proofs of all theoretical results?
     \answerNA{}
\end{enumerate}

\item If you ran experiments...
\begin{enumerate}
  \item Did you include the code, data, and instructions needed to reproduce the main experimental results (either in the supplemental material or as a URL)?
    \answerYes{Code and data are released.}
  \item Did you specify all the training details (e.g., data splits, hyperparameters, how they were chosen)?
    \answerYes{}
        \item Did you report error bars (e.g., with respect to the random seed after running experiments multiple times)?
    \answerNo{}
        \item Did you include the total amount of compute and the type of resources used (e.g., type of GPUs, internal cluster, or cloud provider)?
    \answerYes{We describe details of our computational resources in supplementary material.}
\end{enumerate}

\item If you are using existing assets (e.g., code, data, models) or curating/releasing new assets...
\begin{enumerate}
  \item If your work uses existing assets, did you cite the creators?
    \answerYes{}
  \item Did you mention the license of the assets?
     \answerNA{}
  \item Did you include any new assets either in the supplemental material or as a URL?
     \answerNA{}
  \item Did you discuss whether and how consent was obtained from people whose data you're using/curating?
     \answerNA{}
  \item Did you discuss whether the data you are using/curating contains personally identifiable information or offensive content?
     \answerNA{}
\end{enumerate}

\item If you used crowdsourcing or conducted research with human subjects...
\begin{enumerate}
  \item Did you include the full text of instructions given to participants and screenshots, if applicable?
     \answerNA{}
  \item Did you describe any potential participant risks, with links to Institutional Review Board (IRB) approvals, if applicable?
     \answerNA{}
  \item Did you include the estimated hourly wage paid to participants and the total amount spent on participant compensation?
     \answerNA{}
\end{enumerate}

\end{enumerate}

\clearpage
\appendix

\section*{\LARGE \centering Supplementary Material for\\MonoSDF: Exploring Monocular Geometric Cues \\for Neural Implicit Surface Reconstruction}
\vspace{12mm}

\maketitle

In this \textbf{supplementary document}, we first discuss architectural and implementation details in~\secref{sec:implementation}. Next, we provide additional ablation studies of our monocular geometric cues for four different scene representations in~\secref{sec:ablation} and report additional quantitative and qualitative results in~\secref{sec:result}. Finally, we discuss potential negative impact of this work in~\secref{sec:impact}.

\section{Implementation Details} \label{sec:implementation}
In this section, we first present an overview of 4 different architectures for neural implicit scene representations and details of Multi-Res. Grids in~\secref{sec:architecture} and provide details of the depth loss computation in~\secref{sec:depth}. Next, we describe additional details regarding our parameterizations and optimization in~\secref{sec:details} and discuss evaluation metrics in~\secref{sec:evaluation}.

\subsection{Architectures} \label{sec:architecture}

In the main paper, we investigate four different architectures as our scene representation: \textit{Dense SDF Grid}, \textit{Single MLP}, \textit{Single-Res. Grid}, and \textit{Multi-Res. Grids} . See~\figref{fig:architecture} for an overview over the architectures. In the following, we provide details for Multi-Res.\ Feature Grids.

\boldparagraph{Multi-Res. Grids.}
Following Instant-NGP~\cite{Mueller2022ARXIV}, we use $L$ levels of feature grids with resolutions sampled in geometric space to combine features at different frequencies: 
\begin{equation}
    R_l := \;\lfloor R_\text{min} b^l \rfloor \hspace{1cm}
    b := \;\text{exp}\left(\frac{\ln R_\text{max} - \ln R_\text{min}}{L-1}\right) \enspace ,
\end{equation}
where $R_\text{min}, R_\text{max}$ are the coarsest and finest resolutions, respectively.
As the total number of grid cells grows cubically, we use a fixed number of parameters to store the feature grids and use a spatial hash function to index the feature vector at finer levels. More specifically, each grid contains up to $T$ feature vectors with dimensionality $F$. At the coarse level where $R_l^3 \leq T$, %
the feature grid is stored densely. At the finer level where $R_l^3 > T$, a spatial hash function~\cite{Matthias2003SpatialHash} is used to index the corresponding feature vector:
\begin{equation}
    h(\bx) = \left(\bigoplus_{i=1}^{3} \bx_i \pi_i \right) \, \text{mod} \, T \enspace ,
\end{equation}
where $\bigoplus$ is the bit-wise XOR operation and $\pi_i$ are unique, large prime numbers. We use the default values $R_\text{min} = 16$, $R_\text{max} = 2048$, $L=16$, $F=2$, and $T = 2^{19}$ similar to~\cite{Mueller2022ARXIV} in all experiments.

\subsection{Depth Consistency Loss} \label{sec:depth}
We enforce consistency between our rendered expected depth $\hat{D}$ and the monocular depth $\bar{D}$ with a scale invariant loss function: %
\begin{equation}
\mathcal{L}_\text{depth} = \sum_{\br \in \cR} {\lVert (w \hat{D}(\br) + q) - \bar{D}(\br) \rVert}^2 \enspace , 
\end{equation}
where $w$ and $q$ are the scale and shift used to align $\hat{D}$ and $\bar{D}$ since $\bar{D}$ is given only up to scale. Specifically, we solve $w$ and $q$ with a least-squares criterion~\cite{Eigen2014NIPS,Ranftl2020PAMI}:
\begin{equation}
(w, q) = \operatorname*{arg\,min}_{w, q} \sum_{\br \in \cR} \left(w \hat{D}(\br) + q - \bar{D}(\br)\right)^2. \label{eq:least_square}
\end{equation}
$w$ and $q$ can be efficiently computed as follows: Let $\bh = (w, q)^T$ and $\bd_\br = (\hat{D}(\br), 1)^T$, then ~\eqnref{eq:least_square} can be rewrite as:
\begin{equation}
\bh^{\text{opt}} = \operatorname*{arg\,min}_{\bh} \sum_{\br \in \cR} \left( \bd_\br^T \bh - \bar{D}(\br)\right)^2. \label{eq:least_square_rewrite}
\end{equation}
which has the closed-form solution:
\begin{equation}
\bh = \left(\sum_\br \bd_\br \bd_\br^T\right)^{-1}\left(\sum_\br \bd_\br \bar{D}(\br)\right) \label{eq:close_form} \enspace .
\end{equation}
\red{Note that we estimate $w$ and $q$ individually at each iteration for a batch of randomly sampled rays within a single image because depth maps predicted by the monocular depth predictor can differ in scale and shift and the underlying scene geometry changes at each iteration.}

\subsection{Additional Details} \label{sec:details}
For our single MLP architecture, we use an 8-layer MLP with hidden dimension 256. We use a two-layer MLP with hidden dimension 256 for the SDF prediction for both, Single-Res. Grid and Multi-Res. Grids. We implement the color network with a two-layer MLP with hidden dimension 256 and use it for all architectures. We use Softplus activation for geometric network and use ReLU activation for the color network. We explicitly initialize the SDF grid with a sphere and use the geometric initialization from~\cite{Atzmon2020CVPR} for other architectures. For obtaining monocular cues, we first resize each image and center crop it to $384\times384$, which we then feed as input to the pretrained Omnidata model~\cite{Eftekhar2021ICCV}.
The output depth and normal maps have the same resolution of $384\times384$. 
As a result, we use the same resolution for RGB images, depth cues and normal cues and adjust camera intrinsics accordingly for all experiments.
We optimize our model for 200k iterations which takes about 6 hours and 11 hours for our Multi-Res.\ Grids and MLP, respectively, on a single NVIDIA RTX3090 GPU.

\architecture

\subsection{Evaluation Metrics} \label{sec:evaluation}

For the DTU dataset~\cite{Aanes2016IJCV}, we follow the official evaluation protocol and report the reconstruction quality with: \textit{Accuracy}, \textit{Completeness} and \textit{Chamfer Distance}. \textit{Accuracy} measures how close the reconstructed points are to the ground truth and is defined as the mean distance of the reconstructed points to the ground truth. 
\textit{Completeness} measures to what extent the ground truth points are recovered and is defined as the mean distance of the ground truth points to the reconstructed points. \textit{Chamfer Distance} is the mean of \textit{Accuracy} and \textit{Completeness}. It measures the overall reconstruction quality. For efficiency, we use the Python script\footnote{\url{https://github.com/jzhangbs/DTUeval-python}} to compute these evaluation metrics. %

\definition

For Replica~\cite{Straub2019ARXIV} and ScanNet~\cite{Dai2017CVPR}, we report \textit{Accuracy}, \textit{Completeness}, \textit{Chamfer Distance}, \textit{Precision}, \textit{Recall}, and \textit{F-score} with a threshold of 5cm following~\cite{Sucar2021ICCV,Zhu2022CVPR,guo2022manhattan}. We further report \textit{Normal Consistency}  for the Replica dataset following~\cite{Sucar2021ICCV,Zhu2022CVPR,Mescheder2019CVPR,guo2022manhattan,Peng2020ECCV,Peng2021NEURIPS} as near-perfect ground truth is available. These metrics are defined in ~\tabref{tab:definition}. 

For the Tanks and Temples dataset~\cite{Knapitsch2017SIGGRAPH}, we submit our reconstruction results to the official evaluation server\footnote{\url{https://www.tanksandtemples.org/}} and report the provided F-score.

\section{Ablation} \label{sec:ablation}
\red{In this section, we first conduct several ablation studies to verify the effectiveness of our method, including using geometric cues with different scene representations in~\secref{sec:ab_cues}, different architecture configurations in~\secref{sec:ab_archi}, different number of input views in~\secref{sec:ab_nview}, different cues predictors in~\secref{sec:ab_predictors}.} Next, we analyze the optimization time of our framework in~\secref{sec:optimization_supp}.

\subsection{Ablation of Different Cues} \label{sec:ab_cues}
\replicaablationfull

\figureablationdiffview

\figureablationmoreconfig

\figureablationcuesjpg

To evaluate the effectiveness of our monocular geometric cues for different scene representations, we conduct ablation studies on the Replica dataset with our four different scene representations. Note that as the Replica dataset is part of the training set of Omnidata (making up 0.46\% of the entire training data)~\cite{Eftekhar2021ICCV}, we split the evaluation into the train/test split of Omnidata~\cite{Eftekhar2021ICCV}.

As shown in~\tabref{tab:ablation_cues} and~\figref{fig:ablation_cues_full}, our geometric cues improve reconstruction quality significantly independent of the underlying scene representations. 
We observe that using both, depth cues and normal cues, leads to the best results, indicating the complementary nature of the different cues. We further observe that the reconstruction quality as well as the improvements from adding geometric cues are similar for the train and test split of Omnidata, showing that the monocular predictor did not overfit to the training data.

\red{
\subsection{Ablation of Different Architecture Configurations}\label{sec:ab_archi}
\moreconfigurations

In order to evaluate the performance with different model capacities, we consider MLPs with a different number of layers and Multi-res. Feature Grids with different sizes of the hash table. We list the number of learnable parameters using different architecture configurations in the Table~\ref{tab:more_config}, and show their performance over the optimization processes in ~\figref{fig:ablation_moreconfig}. Our experiments show that using monocular geometric cues improves reconstruction quality and convergence speed independent of the network configuration.

\subsection{Ablation of Different Numbers of Input Views}\label{sec:ab_nview}
We ran experiments with a different number of input images and monocular geometric cues. As shown in ~\figref{fig:ablation_diffviews}, adding the monocular geometric cues leads to consistent improvements across different numbers of input views.

\subsection{Ablation of Different Monocular Cues Predictors}\label{sec:ab_predictors}
\differentpredictors
To further analyze the robustness of our approach to monocular geometric cues of different levels of quality, we further tested our model with different supervised depth predictors~\cite{Ranftl2020PAMI,Wei2021CVPR}, normal predictors~\cite{Do2020SurfaceNormal}, and self-supervised depth predictors~\cite{IndoorSfMLearner,structdepth}. The result is shown in ~\tabref{tab:ablation_different_predictors}. We found that using the state-of-the-art Omnidata model leads to the best results, indicating that the development of better geometric cues will further improve the performance of our approach.
}

\subsection{Optimization Time}\label{sec:optimization_supp}
Adding monocular geometric cues to the optimization introduces a small overhead to our overall optimization pipeline. First, predicting these cues with a pretrained Omnidata model is very efficient (36 FPS with an NVIDIA RTX3090 GPU). For example, it takes less than 26 seconds to predict both depth maps and normal maps for 464 images for one of the ScanNet scene. Note that this only needs to be done once and that we measure FPS with a batch size of one; using a larger batch size will result in a speed up.
Second, we volume render depth and normals during optimization in order to apply a loss against these monocular cues.  This overhead is also small and can be neglected since the most expensive part \wrt compute is the inference of the network. For our MLP variant, the additional flops for volume rendering depth and normal is only 0.0002\% of the MLP inference time.  While adding monocular geometric cues introduce a small overhead, the improvements in terms of reconstruction quality and converge speed are significant. As shown in Table 2 (b) in the main paper, with only 5k iterations, our Multi-Res. Grids representation with cues performs better than the converged models without geometric cues, which implies a $40\times$ speed up (5k vs. 200k).
\section{Additional Results} \label{sec:result}
In this section, we provide more qualitative and quantitative results for three datasets: ScanNet (~\secref{sec:scannet}), Tanks and Temples (~\secref{sec:tnt}), and DTU (~\secref{sec:dtu}).

\subsection{ScanNet}\label{sec:scannet}
\scannetresultfull

\figurescannetjpg

We report quantitative results with all metrics for ScanNet in~\tabref{tab:scannet} and show more visualizations in~\figref{fig:scannet}.
Compared to state-of-the-art methods, our approach with MLP architecture produces significantly better reconstructions both visually as well as quantitatively. It's worth noting that we perform better than concurrent work~\cite{wang2022neuris} even though they have some filtering mechanism.

\subsection{Tanks and Temples}\label{sec:tnt}
\tntresult

\figuretntcompareone

\figuretntcomparetwo

\figuretntcomparethree

\figuretntcomparefour

\figuretnt

We show quantitative results for Tanks and Temples in~\tabref{tab:tnt}. Qualitative comparisons of with or without monocular cues of our MLP variant are shown in~\figref{fig:tnt_1} and~\figref{fig:tnt_2}. ~\figref{fig:tnt_3} and~\figref{fig:tnt_4} show qualitative comparison of our Mulit-Res. Grids. Our monocular geometric cues significantly improve the reconstruction quality. 

We further show an additional comparison against state-of-the-art MVS methods in~\figref{fig:tnt}. We use a pretrained Vis-MVSNet~\cite{Zhang2020BMVC} to predict depth maps for the input images and fuse them to point clouds follow the official code.\footnote{Available at \url{https://github.com/jzhangbs/Vis-MVSNet}}
Next, we use Meshlab's screened Poisson reconstruction~\cite{Kazhdan2013SIGGRAPH} to reconstruct a mesh from point clouds with default parameters. 
We observe that our reconstructions are more complete which is useful for many applications.
Further, reconstructing a mesh from point clouds involves lossy post-processing, leading to floating artifacts and bloated areas in less-observed areas. 

\subsection{Preliminary Results of Using High-resolution Monocular Cues}
\figurehighrescuesvis

\figuretnthighres

\red{In the main paper, we center-crop each image and resize it to $384 \times 384$. Then, we use a pretrained Omnidata model to predict depth maps and normal maps which are also of size $384 \times 384$. While we have shown that training at a resolution of $384 \times 384$ produces impressive results, we believe that exploring different ways to generate and integrate higher resolution cues could further improve reconstruction quality. Here, we provide a proof-of-concept experiment for generating higher resolution monocular cues and integrating them into our model. We use a divide-and-conquer method for generating high-resolution cues. First, we partition a high-resolution image to multiple overlapping sub-images, and we predict monocular depth and normal for each sub-image. Next, we merge these predictions. We use Eq.~\ref{eq:close_form} to align the depth maps and solve the rotation for the normal maps. An example of the resulting high-resolution monocular cues is shown in~\figref{fig:high_res_cues_vis}. We found that our high-resolution cues contain more fine details compared to low-resolution cues. Note that using other methods for generating high-resolution depth maps is also possible, e.g.,~\cite{Miangoleh2021Boosting}. We then use the high-resolution cues to train our model, and the results are shown in~\figref{fig:tnt_highres}. We observe significant improvements when using high-resolution monocular cues.
}

\subsection{DTU}\label{sec:dtu}
\dturesultfull

\figuredturgb

\figuredtunvs

\boldparagraph{Geometry.} We show per-scene quantitative results on the DTU dataset \red{with 3 input-views} in~\tabref{tab:dtu_full} and more qualitative results in~\figref{fig:dtu}. We find that without the monocular geometric cues, both MLP and Multi-Res.\ Grids fail to produce satisfying reconstructions, while with our monocular cues, both methods are improved and are able to reconstruct high-quality meshes. %
\red{We further show more visualizations on the DTU dataset using all input views in ~\figref{fig:dtuallview_morevis}. Compared to state-of-the-art methods, our approach with multi-resolution feature grids produces more accurate reconstructions.}

\figuredtuallview

\dtunvstable

\boldparagraph{Novel View Synthesis.}
\red{We further compare our novel view synthesis results on the DTU dataset with three input views. As shown in~\tabref{tab:dtu_nvs} and ~\figref{fig:dtu_nvs}, using monocular geometric cues improves novel view synthesis results significantly. }

\boldparagraph{Weight Annealing.}
\red{As the monocular depth and normal predictor is not perfect, we exponentially anneal the loss weight for the monocular depth consistency and normal consistency loss, $\lambda_2$ and $\lambda_3$, to $0$ during the first 200 epochs of optimization. Qualitative comparison in ~\figref{fig:ablation_weightdecay} verifies the importance of weight annealing.}

\figureablationweightdecay

\boldparagraph{Failure cases.} We show a failure case on DTU with 3 input views in~\figref{fig:fail}. 
The reconstructed mesh duplicates the object in front of each camera frustum. %
One reason is that the monocular depth cues that we use are only up to scale so they do not guarantee multi-view consistency. 
Therefore, the optimization is still underconstrained since the input RGB images and monocular cues can be explained by individual objects in front of the image plane. 
One possible solution would be incorporating explicit multi-view constraints such as using sparse point clouds from COLMAP~\cite{Schoenberger2016ECCV} as an additional supervision~\cite{Deng2022CVPR}. 
\fail

\section{Societal Impact}\label{sec:impact}
Our method can faithfully reconstruct a 3D scene which can be used for application ranging from virtual reality to robotics. However, it can also have potential negative societal impact. First, our method relies on a general purpose monocular geometric predictor that needs to be trained on large amounts of data and with large computational resources, which potentially has a negative impact on global climate change. Second, accurate reconstruction of a scene may raise privacy concerns that need to be addressed carefully. Finally, accurate geometry reconstructed by our method can potentially be used for malicious purposes.

\end{document}